\documentclass{article}
\usepackage[utf8]{inputenc}
\usepackage{main}
\usepackage{microtype}
\usepackage{graphicx}
\usepackage{subfig}
\usepackage{times}
\usepackage{latexsym}
\usepackage{amsmath}
\usepackage{float}
\usepackage{footnote}
\usepackage{bm}
\usepackage{arydshln}
\usepackage{booktabs}
\usepackage{wrapfig}
\usepackage{multicol}
\usepackage{multirow}
\usepackage{color}
\usepackage{xcolor}     
\usepackage{colortbl}
\usepackage{bbding}
\usepackage{makecell}
\usepackage{mathtools}
\usepackage{imakeidx}
\makeindex
\usepackage{xcolor}
\usepackage{lipsum}
\usepackage{natbib}
\usepackage[toc]{multitoc}
\usepackage[edges]{forest}
\usepackage[normalem]{ulem}
\usepackage{afterpage}
\definecolor{mydarkblue}{rgb}{0,0.08,0.45}
\usepackage[colorlinks=true,linkcolor=mydarkblue,citecolor=mydarkblue,filecolor=mydarkblue,urlcolor=mydarkblue]{hyperref}
\usepackage{caption}
\usepackage{CJKutf8}
\usepackage{awesomebox} 
\usepackage[most]{tcolorbox}

\definecolor{myblue}{rgb}{0.9, 0.1, 0.94}
\definecolor{tablegreen}{rgb}{0.82, 0.94, 0.75}
\definecolor{mygreen}{rgb}{0.64, 0.56, 0.88}
\definecolor{myyellow}{rgb}{0.98, 0.94, 0.75}
\definecolor{mygreen}{rgb}{0.68, 0.9, 0.6}

\definecolor{color4}{rgb}{0.91, 0.63, 0.82}
\definecolor{color3}{rgb}{0.33, 0.73, 0.78}
\definecolor{color2}{rgb}{0.54, 0.84, 0.56}
\definecolor{color5}{rgb}{0.85, 0.74, 0.95}
\definecolor{color6}{rgb}{0.84, 0.99, 0.37}

\usepackage{fancyhdr} 
\usepackage{blindtext} 

\definecolor{plain_text_color}{rgb}{0.9677975592919913, 0.44127456009157356, 0.5358103155058701}
\definecolor{rotten_tomatoes_color}{rgb}{0.8616090647292522, 0.536495730113334, 0.19548899031476086}
\definecolor{daily_mail_color}{rgb}{0.6804189127793346, 0.6151497514677574, 0.19405452111445337}
\definecolor{wikidata_color}{rgb}{0.46810256823426105, 0.6699492535792404, 0.1928958739904499}
\definecolor{wikihow_color}{rgb}{0.20125317221201128, 0.6907920815379025, 0.47966761189275336}
\definecolor{wikipedia_color}{rgb}{0.21044753832183283, 0.6773105080456748, 0.6433941168468681}
\definecolor{wordnet_color}{rgb}{0.2197995660828324, 0.6625157876850336, 0.7732093159317209}
\definecolor{qa_datasets_color}{rgb}{0.433280341176423, 0.6065273407962815, 0.9585467098271748}
\definecolor{arxiv_color}{rgb}{0.8004936186423958, 0.47703363533737203, 0.9579547196007522}
\definecolor{pwc_color}{rgb}{0.962272393509669, 0.3976451968965351, 0.8008274363432775}

\usepackage{enumerate}
\newenvironment{itemize*}%
 {\leftmargini=10pt\begin{itemize}%
  \setlength{\itemsep}{0pt}%
  \setlength{\parskip}{0pt}%
  }%
 {\end{itemize}}
\newenvironment{enumerate*}%
 {\begin{enumerate}%
  \setlength{\itemsep}{0pt}%
  \setlength{\parskip}{0pt}}%
 {\end{enumerate}}

\usepackage{pifont}
\newcommand{\cmark}{\ding{51}}%
\newcommand{\xmark}{\ding{55}}%

\makeatletter
\newenvironment{chapquote}[2][2em]
  {\setlength{\@tempdima}{#1}%
   \def\chapquote@author{#2}%
   \parshape 1 \@tempdima \dimexpr\textwidth-2\@tempdima\relax%
   \itshape}
  {\par\normalfont\hfill--\ \chapquote@author\hspace*{\@tempdima}\par\bigskip}
\makeatother

\newtcolorbox{myboxi}[1][]{
  breakable,
  title=#1,
  colback=red!5,
  colbacktitle=red!5,
  coltitle=black,
  fonttitle=\bfseries,
  bottomrule=0pt,
  toprule=0pt,
  leftrule=2pt,
  rightrule=2pt,
  titlerule=0pt,
  arc=0pt,
  outer arc=0pt,
  colframe=red,
}

\usepackage[edges]{forest}

\tikzset{%
    parent/.style =          {align=center,text width=1.4cm,rounded corners=3pt, line width=0.3mm, fill=gray!10,draw=gray!80},
    child/.style =           {align=center,text width=2cm,rounded corners=3pt, fill=blue!10,draw=blue!80,line width=0.3mm},
    grandchild/.style =      {align=center,text width=1.5cm,rounded corners=3pt},
    greatgrandchild/.style = {align=center,text width=1.5cm,rounded corners=3pt},
    greatgrandchild2/.style = {align=center,text width=1.5cm,rounded corners=3pt},    
    referenceblock/.style =  {align=center,text width=1.5cm,rounded corners=2pt},
    plaintext/.style =           {align=center,text width=3cm,rounded corners=3pt, fill=plain_text_color!10,draw=plain_text_color!80,line width=0.3mm},   
    plaintext_work/.style =           {align=center, text width=6cm,rounded corners=3pt, fill=plain_text_color!10,draw=plain_text_color!0,line width=0.3mm},  
    pretrain/.style =           {align=center,text width=3cm,rounded corners=3pt, fill=rotten_tomatoes_color!10,draw=rotten_tomatoes_color!80,line width=0.3mm},   
    pretrain_work/.style =           {align=center, text width=6cm,rounded corners=3pt, fill=rotten_tomatoes_color!10,draw=rotten_tomatoes_color!0,line width=0.3mm},  
    template/.style =           {align=center,text width=3cm,rounded corners=3pt, fill=daily_mail_color!10,draw=daily_mail_color!80,line width=0.3mm},   
    template_work/.style =           {align=center,text width=6cm,rounded corners=3pt, fill=daily_mail_color!10,draw=daily_mail_color!0,line width=0.3mm},    
    answer/.style =           {align=center,text width=3cm,rounded corners=3pt, fill= wikidata_color!10,draw= wikidata_color!80,line width=0.3mm},   
    answer_work/.style =           {align=center,text width=6cm,rounded corners=3pt, fill= wikidata_color!10,draw= wikidata_color!0,line width=0.3mm},      
    multiple/.style =           {align=center,text width=3cm,rounded corners=3pt, fill= wikihow_color!15,draw= wikihow_color!85,line width=0.3mm},   
    multiple_work/.style =           {align=center,text width=6cm,rounded corners=3pt, fill= wikihow_color!15,draw= wikihow_color!0,line width=0.3mm},     
    tuning/.style =           {align=center,text width=3cm,rounded corners=3pt, fill= wikipedia_color!10,draw= wikipedia_color!80,line width=0.3mm},   
    tuning_work/.style =           {align=center,text width=6cm,rounded corners=3pt, fill= wikipedia_color!10,draw= wikipedia_color!0,line width=0.3mm},        
    wordnet/.style =           {align=center,text width=3cm,rounded corners=3pt, fill= wordnet_color!20,draw= wordnet_color!100,line width=0.3mm},   
    wordnet_example/.style =           {align=center,text width=6cm,rounded corners=3pt, fill= wordnet_color!20,draw= wordnet_color!0,line width=0.3mm},
    qa/.style =           {align=center,text width=3cm,rounded corners=3pt, fill= qa_datasets_color!15,draw= qa_datasets_color!85,line width=0.3mm},   
    qa_example/.style =           {align=center,text width=6cm,rounded corners=3pt, fill= qa_datasets_color!15,draw= qa_datasets_color!0,line width=0.3mm},
    arxiv/.style =           {align=center,text width=3cm,rounded corners=3pt, fill= arxiv_color!15,draw= arxiv_color!85,line width=0.3mm},   
    arxiv_example/.style =           {align=center,text width=6cm,rounded corners=3pt, fill= arxiv_color!15,draw= arxiv_color!0,line width=0.3mm},
    pwc/.style =           {align=center,text width=3cm,rounded corners=3pt, fill= pwc_color!20,draw= pwc_color!100,line width=0.3mm},   
    pwc_example/.style =           {align=center,text width=6cm,rounded corners=3pt, fill= pwc_color!20,draw= pwc_color!0,line width=0.3mm},
}

\title{\textit{re}Structured Pre-training}

\begin{document}

\author{
Weizhe Yuan
\thanks{Work done with Pengfei at \href{http://expressai.co/}{X-Lab}.} 
\\
  CMU \\
  \texttt{bellyapplerian@gmail.com} \\
  \And
  Pengfei Liu \thanks{Corresponding author.} \\ %
   CMU\\
\texttt{stefanpengfei@gmail.com}
  }
  
\maketitle

\begin{abstract}

In this work, we try to decipher the internal connection of NLP technology development in the past decades, searching for essence, which rewards us with a (potential) new learning paradigm for NLP tasks, dubbed as \textit{\textbf{r}e\textbf{S}tructured Pre-\textbf{t}raining} (RST).
In such a paradigm, the role of \textbf{\textit{data}} will be re-emphasized, and model pre-training and fine-tuning of downstream tasks are viewed as a process of data storing and accessing. 
Based on that, we operationalize the simple principle that \textit{a good storage mechanism should not only have the ability to cache a large amount of data but also consider the ease of access}. We achieve this by pre-training models over \textit{restructured} data that consist of a variety of valuable information instead of raw data after overcoming several engineering challenges.
Experimentally, RST models not only \textbf{surpass strong competitors (e.g., T0) on 52/55 popular datasets} from a variety of NLP tasks (e.g., classification, information extraction, fact retrieval, text generation, etc.) without fine-tuning on downstream tasks, but also achieve superior performance in National College Entrance Examination - English (Gaokao-English), the most authoritative examination in China, which millions of students will attend every year.
Specifically, the proposed system Qin ( \includegraphics[scale=0.05]{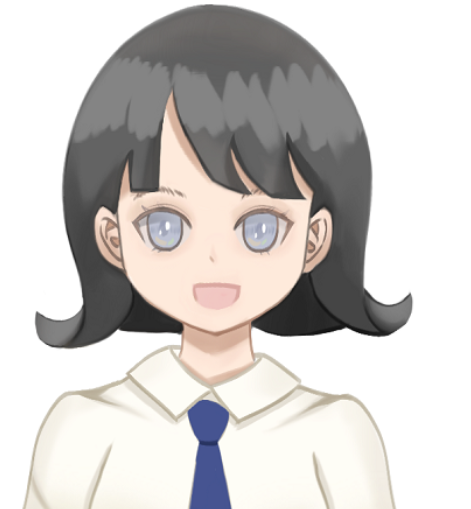}) \textbf{achieves 40 points higher than the average
scores made by students and 15 points higher than GPT3 with 1/16 parameters. In particular, Qin gets a high score of 138.5 (the full mark is 150) in the 2018 English exam (national paper III)}.
We have released the {Gaokao Benchmark} with an online submission platform that contains ten annotated English papers from 2018-2021 so far (and will be expanded annually), which allows more AI models to attend Gaokao, establishing a relatively fair test bed for human and AI competition and helping us better understand \textit{where we are}.

We test our model in the 2022 College Entrance Examination English that happened a few days ago (2022.06.08), and \textbf{it gets a total score of 134 (v.s. GPT3's 108).}
We released all \textcolor{plain_text_color}{data}\footnote{\url{https://github.com/ExpressAI/reStructured-Pretraining}} and \textcolor{plain_text_color}{models}.\footnote{\url{https://huggingface.co/XLab/rst-all-11b}}

\end{abstract}

\paragraph{}
\paragraph{}

\begin{chapquote}[50pt]{Clifford Stoll}
``Data is not information''\footnote{This paper claims that pre-training over \textit{information}, i.e., restructured data, will be more effective than simply pre-training on raw data. See \S\ref{easter:vi} for future work.}
\end{chapquote}

 \begin{figure}[ht]
    \centering
    \includegraphics[width=0.55\linewidth]{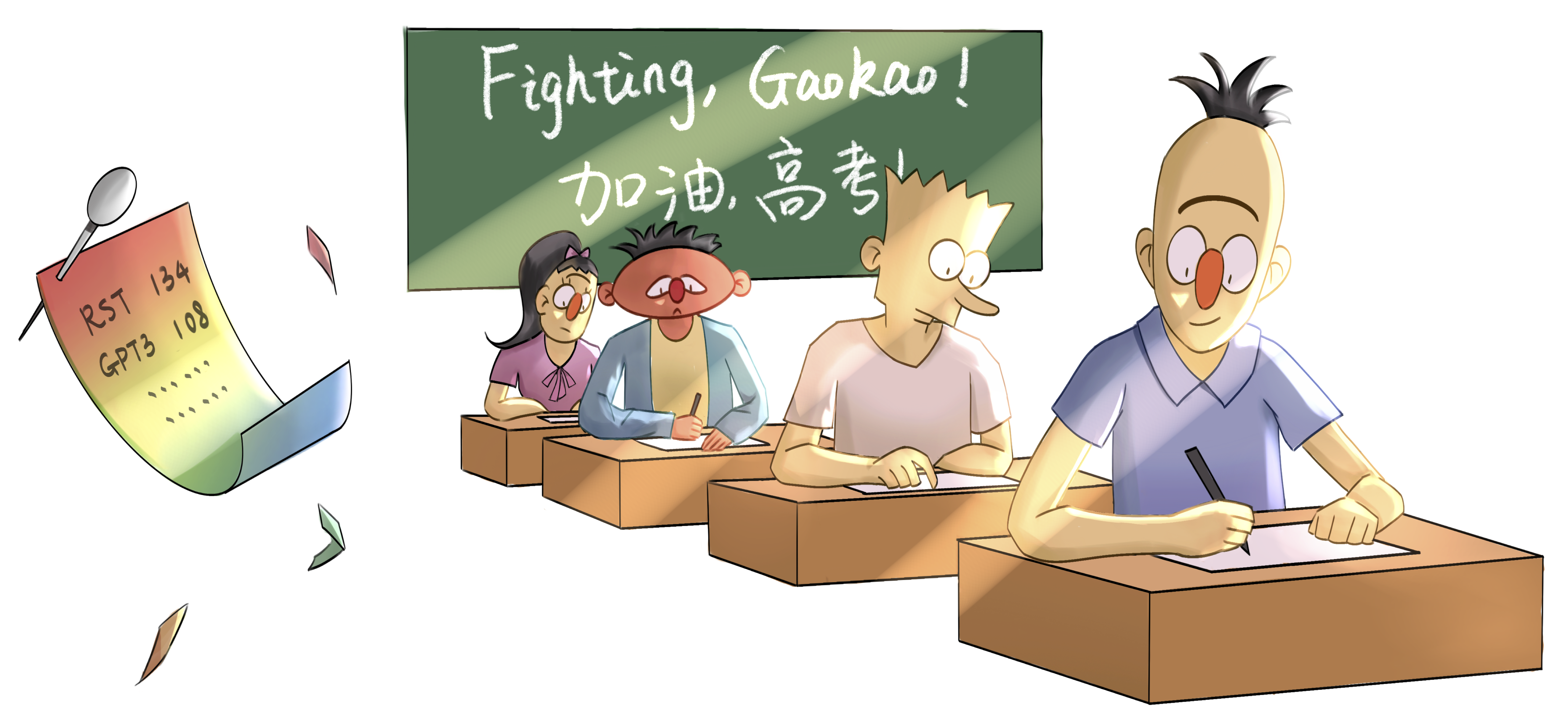}
  \caption{\href{https://explainaboard.inspiredco.ai/benchmark}{Gaokao Benchmark for AI} }
\end{figure}

\section*{Summary of Contributions} \label{sec:contribution}
This work is part position, part new methodology, part new resources, and part unification, which we believe will contribute to communities from multiple aspects:

\begin{figure}[!th]
    \centering
    \includegraphics[width=0.38\linewidth]{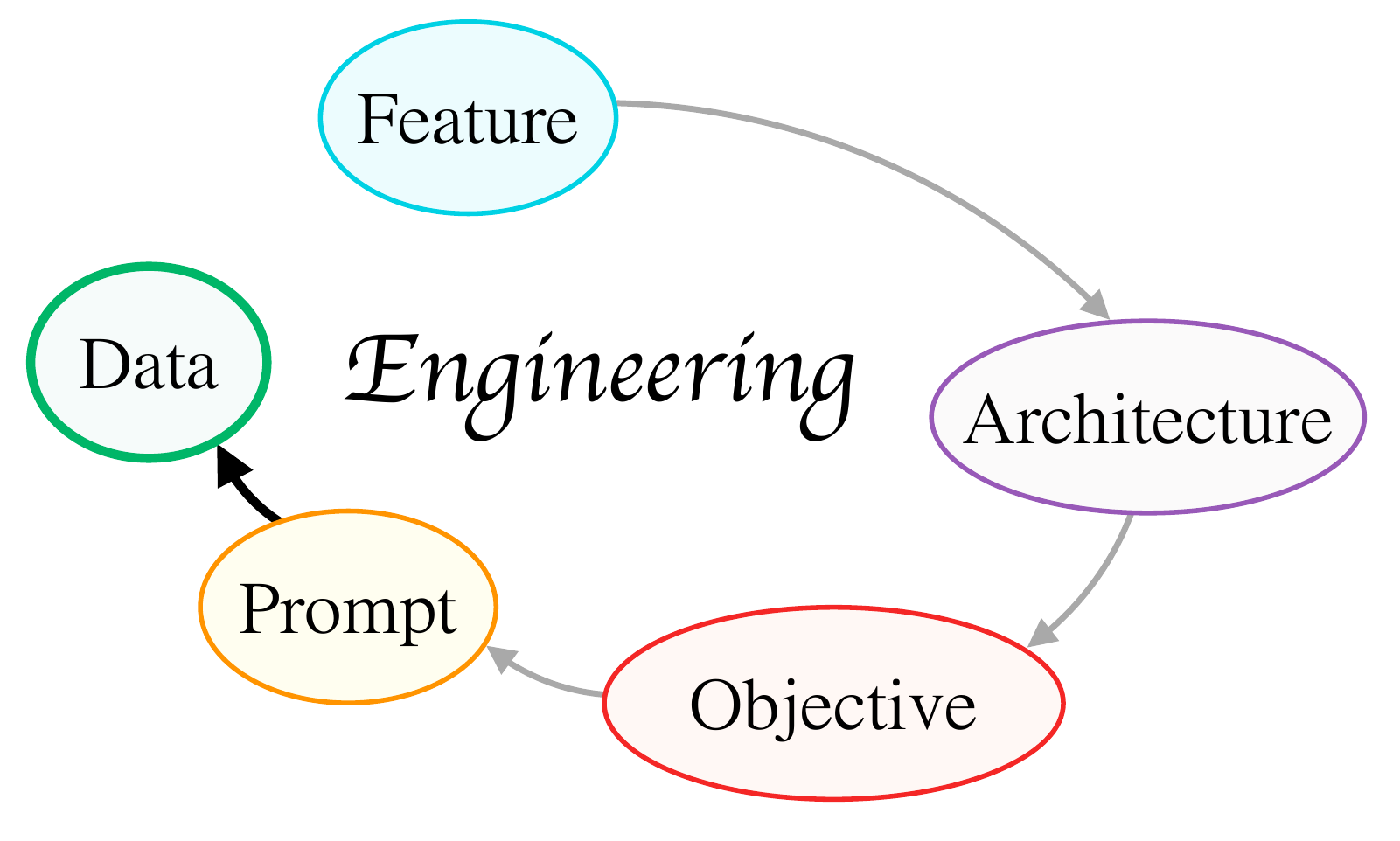}
    \caption{Hypothesis of NLP technique evolution.}
    \label{fig:evolution-summary}
\end{figure}

\noindent \textbf{I: Evolution Hypothesis}
This paper attempts to establish a ``Hypothesis of NLP Technique Evolution'' (\S\ref{sec:evolution}) from a global perspective by exploring the internal connection between the development of modern NLP technology (\S\ref{sec:paradigm_shift}).
Briefly, the core idea reflected in the hypothesis goes: \textit{the iteration of technology always moves along the direction that system developers can design a better and more general system by doing fewer things}. So far, technological evolution has gone through multiple waves of iterations as illustrated in Fig.\ref{fig:evolution-summary}: \textit{feature} engineering$\rightarrow$ \textit{architecture} engineering $\rightarrow$ \textit{objective} engineering $\rightarrow$ \textit{prompt} engineering, and we are moving towards a more practically effective data-centric engineering.
We hope to inspire more researchers to think about this critically in the future so as to grasp the core driving force of technological progress, find a ``gradient ascent'' path for academic development, and do more scientifically-meaningful works.

\paragraph{II: New Paradigm}
With the above hypothesis in mind, we propose a new paradigm for modeling NLP: \textit{reStructured Pre-training} (\S\ref{sec:rst_engineering}). This paradigm regards model pre-training/tuning as a data storing/accessing process and claims that \textbf{a good storage mechanism should make expected data easily accessible}.
With such a new paradigm, we are able to unify 26 different types of signals (e.g., entities of a sentence) in the world from 10 data sources (\S\ref{signal_restructure}) (e.g., Wikipedia), and the ``generalist'' model trained on that has achieved strong generalization ability (\S\ref{sec:exp_setup_nlp_tasks}) on various tasks, including 55 datasets of NLP.

\paragraph{III: AI for Gaokao}
Under this paradigm, we are able to develop an AI system \textsc{Qin} for the Gaokao-English examination (\S\ref{subsec:qin}), which, as far as we know, \textbf{is the first deep learning-based AI system in the world} for Gaokao-English.
Qin has achieved remarkable results in the Gaokao examination over the years: \textbf{40 points higher than the average human and 15 points higher than GPT3 with 1/16 parameters}. In particular, Qin gets a high score of \textbf{138.5} (the full mark is 150) in the $2018$ English exam (national paper III), in which it gets full marks in both listening and reading comprehension parts.

\paragraph{V: Rich Resources}
(1) We release the Gaokao Benchmark (\S\ref{sec:gaokao}) to track how well we make progress towards human-level intelligence. It can not only provide a comprehensive evaluation of different tasks and domains that are practically useful in a real-world scenario, but also provide rich human performance so that AI systems can be directly compared with humans over time.
(2) We set up an interactive leaderboard \footnote{\url{https://explainaboard.inspiredco.ai/benchmark}} using ExplainaBoard \cite{liu-etal-2021-explainaboard} for Gaokao Benchmark\footnote{\url{https://github.com/expressai/AI-Gaokao}} so that more AI systems can easily participate in Gaokao and automatically get the score.
(3) All resources can be found on GitHub.\footnote{\url{https://github.com/expressai/reStructured-Pretraining}}

\paragraph{IV: Inspiring Evidence}
 
(1) The success of AI in English for Gaokao Examination has provided us with much new thinking: \textbf{AI technology can empower education and help solve a series of problems} in education and teaching. For example, (a) helping teachers grade papers automatically, (b) helping students answer questions about their homework with detail explanation, and (c), more importantly, promoting education equity so that most families can receive education services of the same quality.
(2) 
For the first time, this work integrates \textbf{26} different signals in the world in a unified way, \textbf{not trying to distinguish between supervised and unsupervised data but being concerned with how much we can use the information that nature gives us and how}.\footnote{We argue that blindly sticking with supervised or unsupervised, pre-training or fine-tuning, few-shot, or zero-shot makes little sense. In practice, all it matters is how we make the best use of the information from data that we can get from the world.} The impressive performance on more than \textbf{50} datasets from varieties of NLP tasks shows the value of data-centric pre-training and inspires more future exploration.

\newpage
\textcolor{white}{\textbf{Human: Hey PLM guys, valuable information is like the gem of a mine, distributed in all kinds of data, and when you have the eyes to find that treasure, you will make the world better again.
Let's go on a treasure hunt...}}

\begin{figure*}[t]
\centering
\includegraphics[width=16cm]{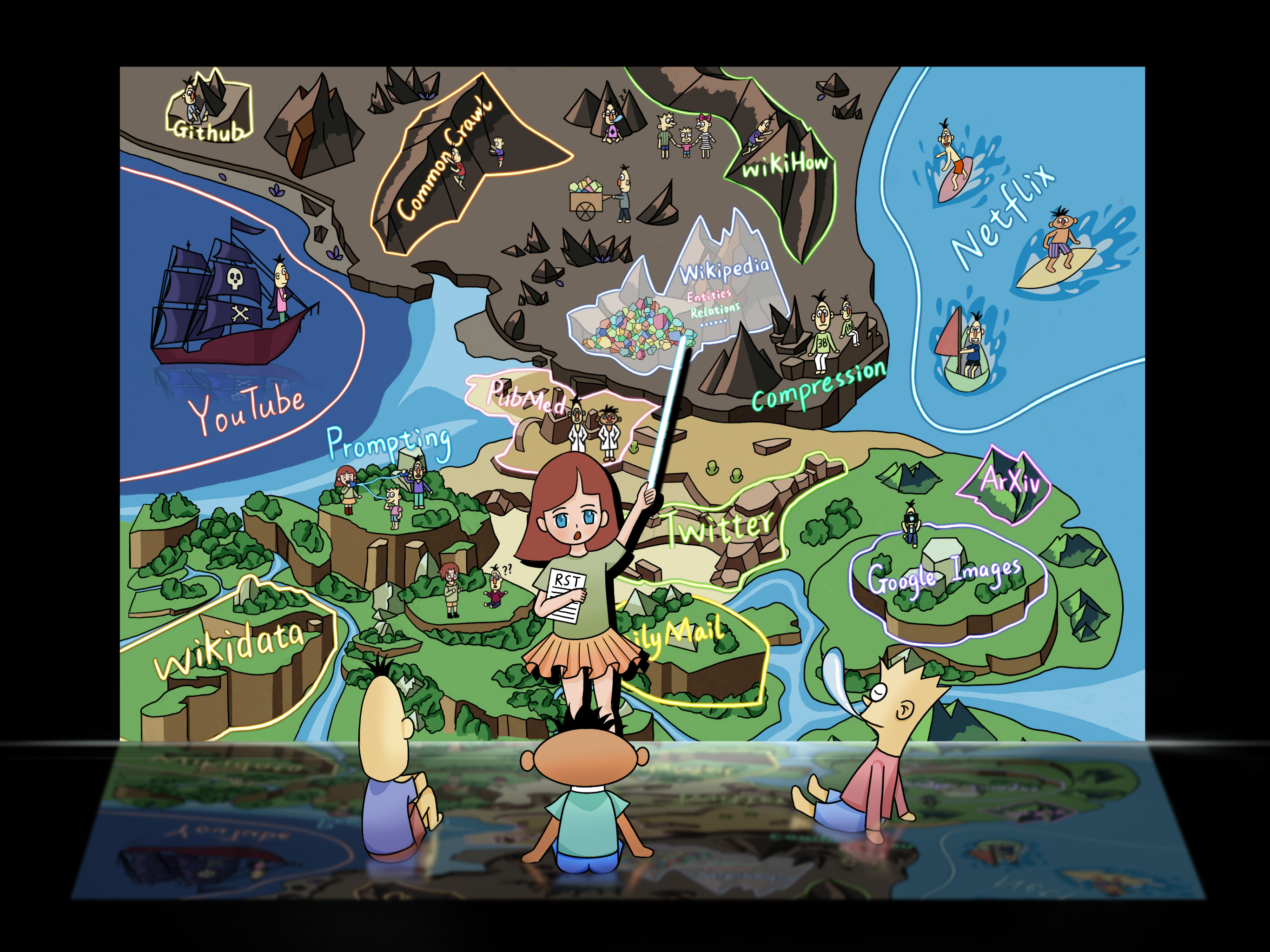}
\label{fig:essence}
\end{figure*}
\pagecolor{black} \afterpage{\nopagecolor}

\clearpage

\tableofcontents

\clearpage

\section{Introduction}

\textbf{The way we store data is evolving, from biological neural networks to artificial counterparts.}
The most common scenario is the use of our brains to cache the data \cite{posner1990attention,xia2013brainnet} that will be further distilled into knowledge or expertise.
With the ever-growing size of data available nowadays, people seek to store data with different external devices such as hard drives \cite{white1980disk} or cloud storage \cite{armbrust2010view}.
Amid a rise in deep learning techniques~\cite{hinton2006reducing,hinton2012deep}, another promising storage technology has emerged, which uses artificial neural networks \cite{mikolov2013distributed,devlin-etal-2019-bert,brown2020language} to store information from data.

\begin{figure*}[!th]
\centering
\subfloat[Biological neural networks. ]{{\includegraphics[height=0.17\linewidth]{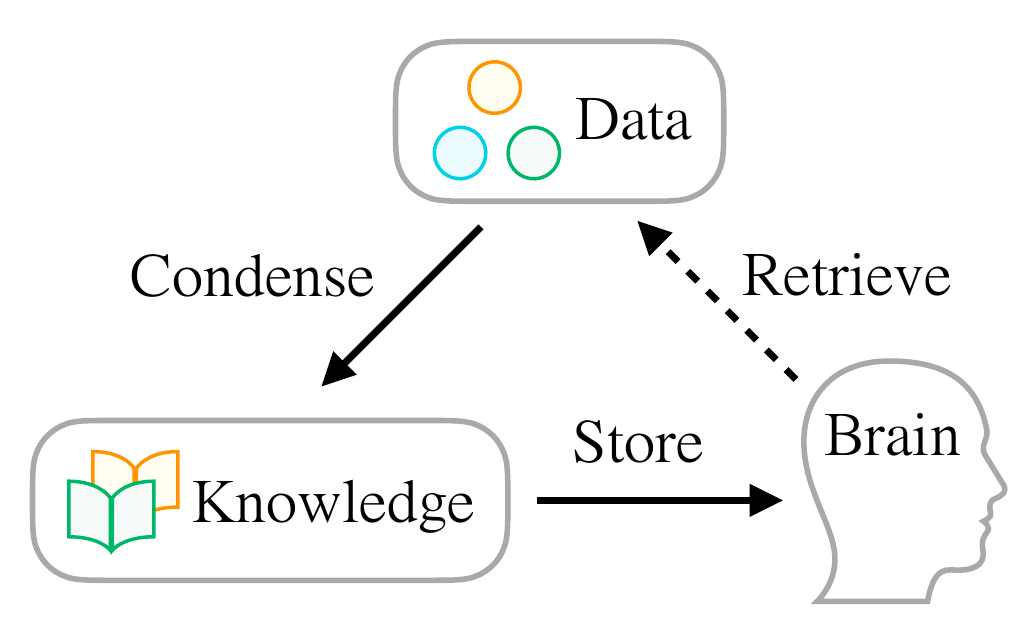} }}%
\hspace{10px}
\subfloat[Disk/Cloud storage.]{{\includegraphics[height=0.17\linewidth]{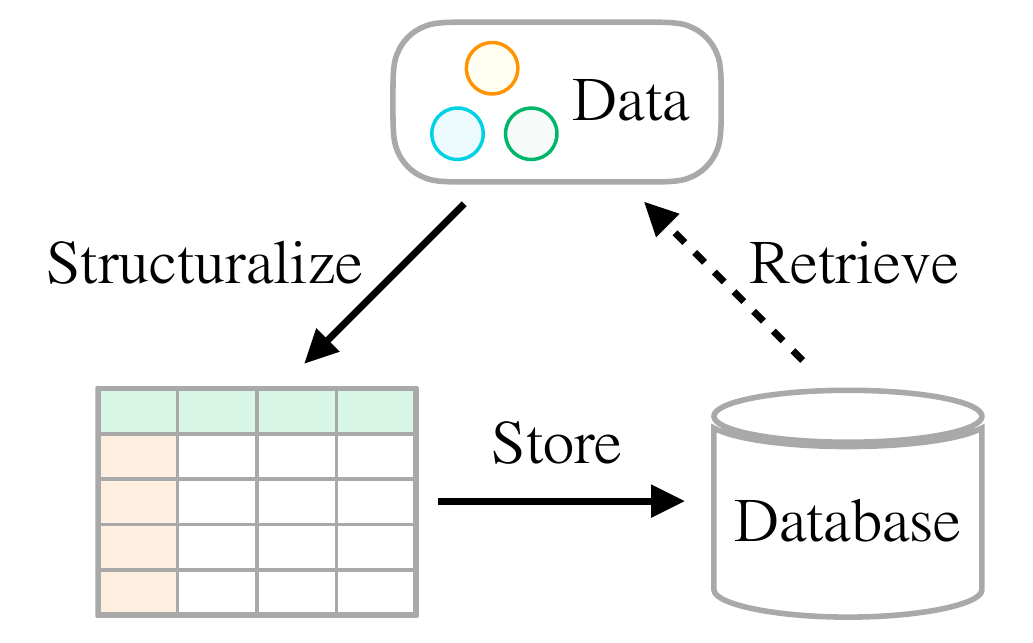} }}%
\hspace{10px}
\subfloat[Artificial neural netwokrs.]{{\includegraphics[height=0.17\linewidth]{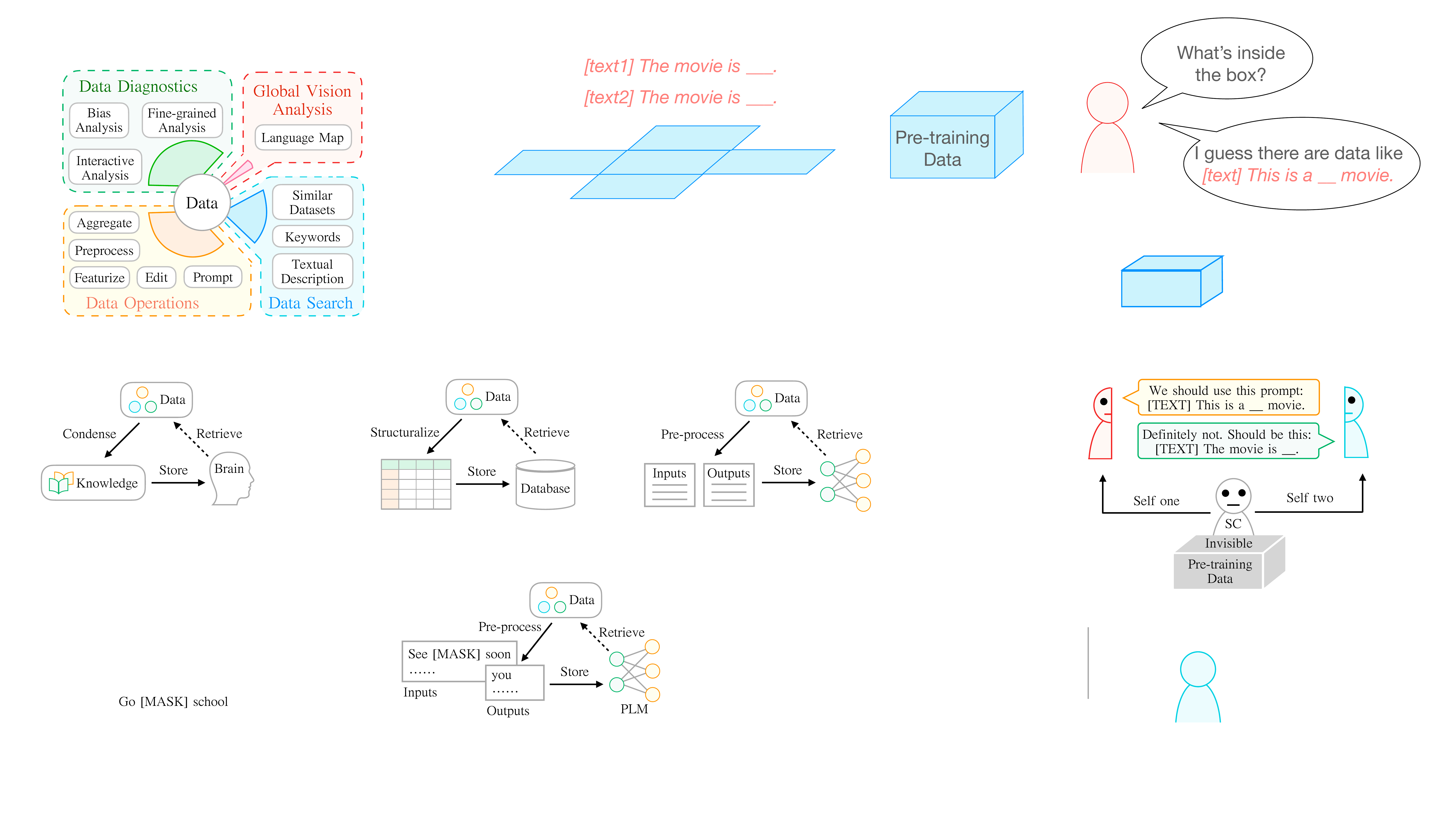} }}%
\caption{Three ways of storing/retrieving data. The solid arrows represent storing data while the dotted arrows represent retrieving data.}%
\label{fig:horizontal-analysis} 
\end{figure*}

We argue that \textbf{the ultimate goal of data storage is to better serve human life, and how data is \textit{accessed} is as important as how it is \textit{stored}}.
However, there are often differences in the way that data is stored and accessed. Historically, there have been efforts to bridge such a gap to better use the information (i.e., accurately recall) that exists in the world.
For example, as shown in Fig.\ref{fig:horizontal-analysis}
\begin{itemize*}
    \item (a) regarding biological neural networks (e.g., human brain), people, at an early age, are educated by a variety of well-organized courses (i.e., knowledge) so that they can elicit specific data to deal with the always complex and changing life.\footnote{Leon: The Professional Trailer :)}
    \item (b) for external device storage, people routinely structuralize the data following a certain schema (e.g., tabular) and then adopt a specialized language (e.g., SQL~\cite{melton1993understanding}) to effectively retrieve desired information from database.
    \item (c) for artificial neural network-based storage, researchers leverage self-supervised learning~\cite{devlin-etal-2019-bert,brown2020language} to memorize the data from large corpora (i.e., pre-training), then employ the network for diverse downstream tasks (e.g., sentiment classification). 
\end{itemize*}

Recently, to make the way the data is accessed by downstream tasks resemble the way the data is stored in artificial neural networks during the pre-training stage, the technique named ``prompt learning''~\cite{schick-schutze-2021-exploiting,brown2020language,liu2021pre} was introduced by reformulating downstream tasks as a language modeling style problem~\cite{rosenfeld1994adaptive}.
Although prompting methods have narrowed the difference between data storage and access with remarkable achievements in various tasks \cite{petroni-etal-2019-language,schick-schutze-2021-shot,liu2021gpt,yuan2021bartscore,tsimpoukelli2021multimodal}, it does not fundamentally eliminate the gap, as the way models store data in the pre-training stage is not transparent to diverse downstream tasks. In other words, downstream tasks (e.g., sentiment classification) do not know what access methods (i.e., prompts) could better elicit expected information from the pre-trained models, resulting in the notorious problem of \textit{prompt engineering}~\cite{zhao2021calibrate,liu2021pre,shin-etal-2020-autoprompt,jiang-etal-2020-know,liu2021makes,wei2021pretrained}.
\begin{wrapfigure}[10]{r}{5.5cm}
    \centering
    \vspace{-10pt}
    \includegraphics[width=\linewidth]{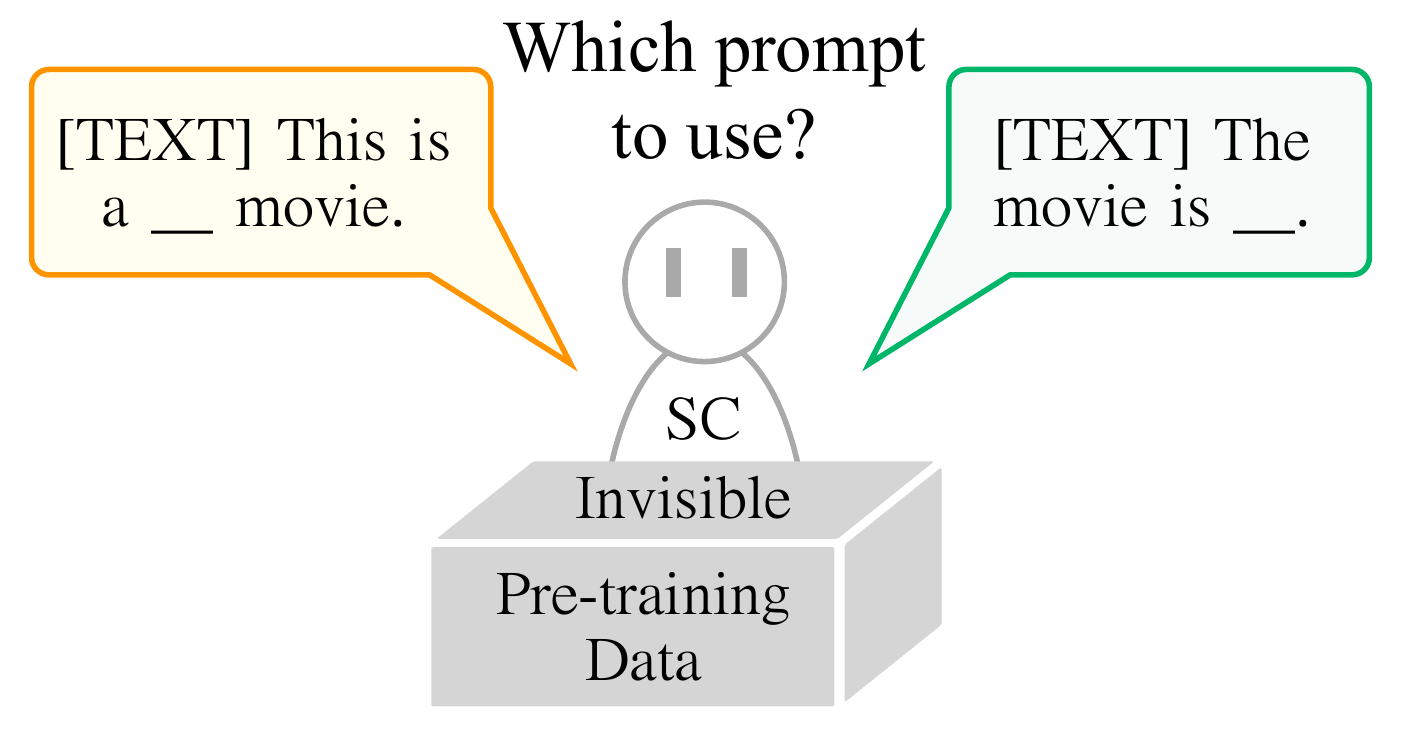}
    \caption{The sentiment classification (SC) task is guessing which prompt should be used.}
    \label{fig:prompt_engineering_issue}
\end{wrapfigure}
As shown in Fig.~\ref{fig:prompt_engineering_issue},
in order to predict the sentiment of the sentence with the help of pre-trained models, one needs to choose a query with which the pre-trained model is more familiar.\footnote{Pre-trained models have seen the pattern in the pre-training corpus and memorized it.}
However, system designers have little idea about the preferred query form since the pre-training data's distribution, or structure is not interpretable.

Methodologically, we present a new way to look at data that contains various types of \textit{information}, which could be regarded as pre-training signals that can instruct models for parameter optimization.
We structurally represent data in the unit of signals and claim that a good PLM should mark various signals during pre-training in a way that expected information could be accessed efficiently by downstream tasks.
This is similar to when using a database to store data: we first structure them into a table or JSON format so that the desired information can be accurately retrieved through specialized languages, such as SQL.

Moreover, we argue that \textbf{valuable signals are rich and exist everywhere\footnote{Valuable information can exist in curated labeled data, Wikipedia, text with a markup language, education books, knowledge base, dictionaries, etc. A good pre-trained model should cover all of them.} from the data in the world} instead of simply existing in the supervised datasets that are manually curated,\footnote{There is no absolute unlabeled data in the world. For example, at least the context information could provide the signal for self-supervised learning.} and what we need to do is to (a) identify them, (b) restructure them in a unified language, (c) integrate and store them into the pre-trained language model. We call this learning paradigm \textit{reStructured Pre-training}:
the Figure on the $3^{\text{rd}}$ page visually motivates this idea, which compares this process to \textbf{\textsc{Mining for Treasure}}.
Different data sources such as \texttt{Wikipedia} are equivalent to {mines rich in gems}. They contain rich information such as \texttt{named entities} from the hyperlink, which can provide signals for model pre-training. \textbf{A good PLM should have \textbf{a clear picture} of the composition of the various signals in the data to provide accurate information for downstream tasks according to their different needs}.
\clearpage

\section{reStructured Pre-training}

\subsection{Paradigm Shift in Modern NLP} \label{sec:paradigm_shift}

The paradigm for solving NLP tasks is changing rapidly and is still ongoing.
For example, as summarized by \cite{liu2021pretrain}, the development history of modern natural language technology can be abstracted into four paradigms. We revisit and extend it from a new perspective as summarized in Tab.~\ref{tab:five_paradigm}.

\begin{table}[!th]
\footnotesize
\setlength\tabcolsep{5pt}
\renewcommand{\arraystretch}{1.2}
\begin{tabular}{llccll}
\toprule
                  & \multirow{2}{*}{\begin{tabular}[c]{@{}l@{}}\textbf{Fully Supervised} \\ \textbf{Learning}\end{tabular}} & \multicolumn{2}{c}{\multirow{2}{*}{\textbf{Pre-train, Fine-tune}}} & \multirow{2}{*}{\begin{tabular}[c]{@{}l@{}}\textbf{Pre-train, Prompt} \\ \textbf{Predict}\end{tabular}} & \multirow{2}{*}{\begin{tabular}[c]{@{}l@{}}\textbf{reStructure, Pre-train,} \\ \textbf{Fine-tune}\end{tabular}} \\
                  &        \multicolumn{2}{l}{}                                 &                                       &                                             &                                                           \\
                  \midrule
\multirow{7}{*}{\textbf{Illustration}}      
&

\multirow{7}{*}{\parbox[c]{0em}{\includegraphics[width=1in]{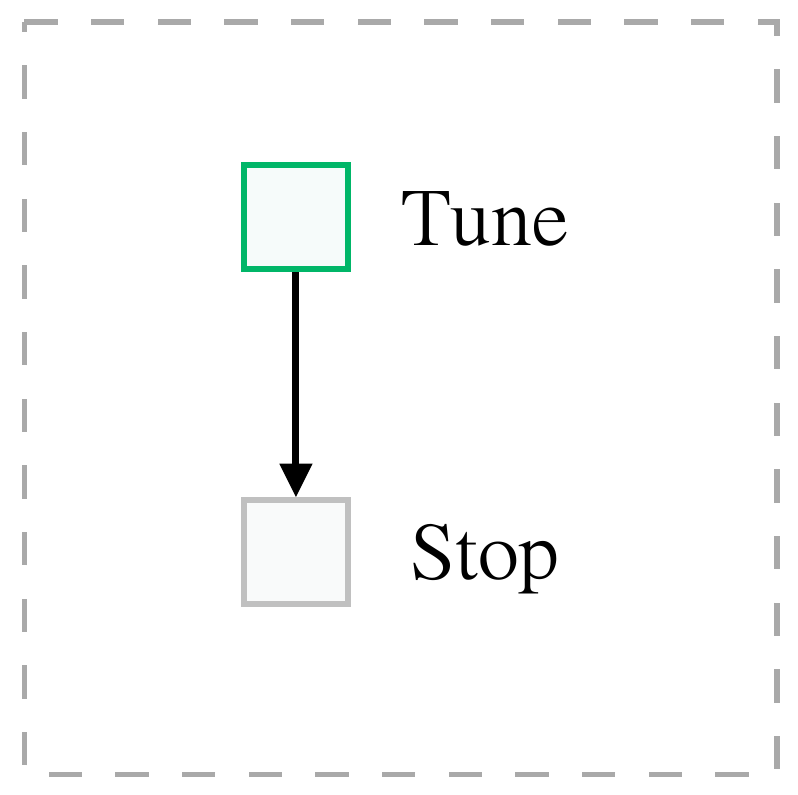}}}                                      
& \multicolumn{2}{l}{\multirow{7}{*}{\hspace{0.8cm}\parbox[c]{0em}{\includegraphics[width=1in]{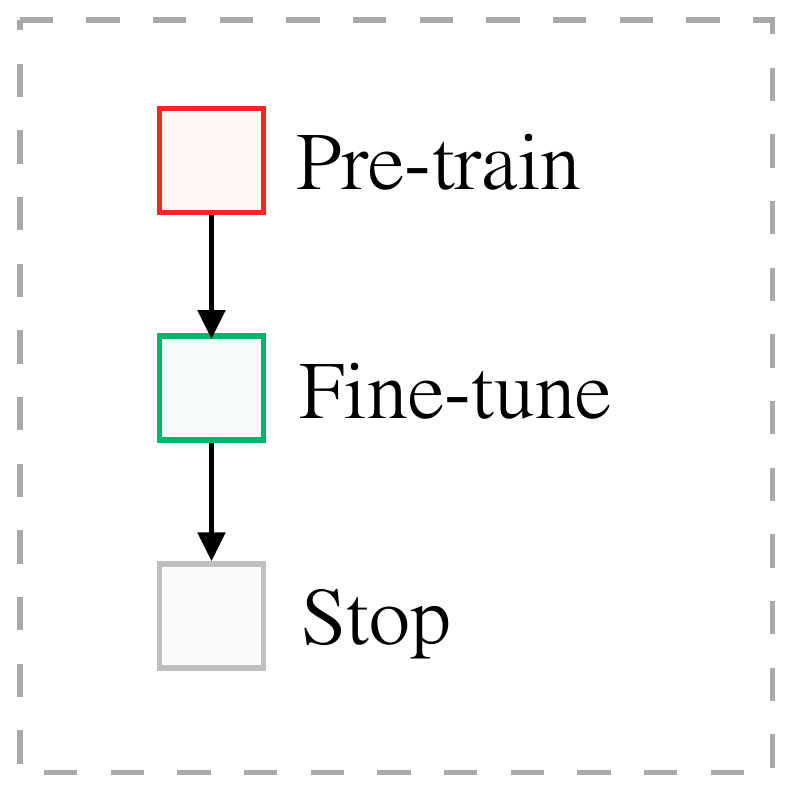}}}}                   
& \multirow{7}{*}{\parbox[c]{0em}{\includegraphics[width=1in]{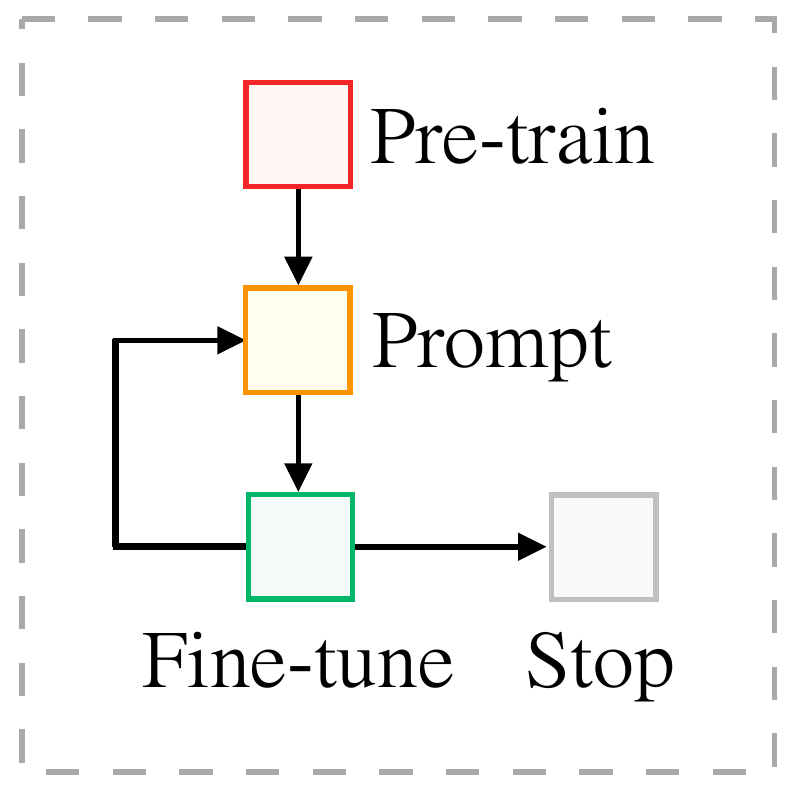}}}                                       
&   \multirow{7}{*}{\hspace{0.25cm}\parbox[c]{0em}{\includegraphics[width=1in]{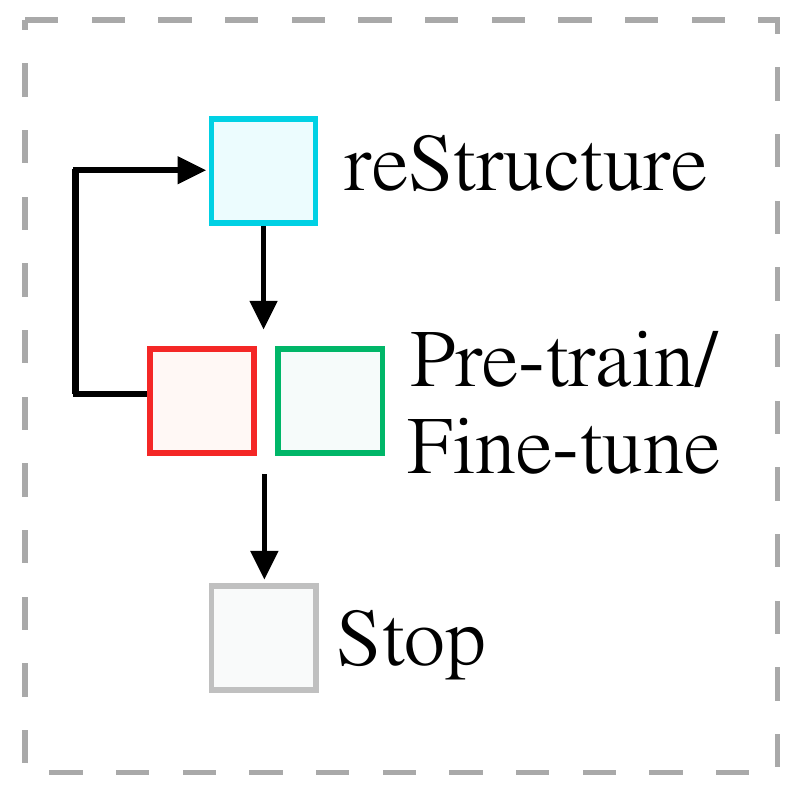}}}                                                        \\
\\
\\
\\
\\
\\
\\
\midrule
\textbf{Engineering}       
& \multicolumn{1}{c}{I: Feature}                     
& \multicolumn{1}{>{\columncolor{tablegreen}}c}{II: Architecture}                     
& \multicolumn{1}{>{\columncolor{tablegreen}}c}{III: Objective}                             
& \multicolumn{1}{c}{IV: Prompt}                                      
& \multicolumn{1}{>{\columncolor{tablegreen}}c}{V: Data reStructure}                                               \\

\textbf{Example}           
& \multicolumn{1}{c}{SVM}                         
& \multicolumn{1}{>{\columncolor{tablegreen}}c}{Word2vec}                         
& \multicolumn{1}{>{\columncolor{tablegreen}}c}{BERT}                                  
& \multicolumn{1}{c}{GPT3}                                       
& \multicolumn{1}{>{\columncolor{tablegreen}}c}{RST}                                                       \\

\textbf{Pre-training Data} 
& \multicolumn{1}{c}{-}                           
& \multicolumn{1}{>{\columncolor{tablegreen}}c}{ngram}                            
& \multicolumn{1}{>{\columncolor{tablegreen}}c}{plain text}                            
& \multicolumn{1}{c}{plain text}                                  
& \multicolumn{1}{>{\columncolor{tablegreen}}c}{reStructured text}                                         \\

\textbf{Supported Signal}  
& \multicolumn{1}{c}{- }                          
& \multicolumn{1}{>{\columncolor{tablegreen}}c}{Limited}                          
& \multicolumn{1}{>{\columncolor{tablegreen}}c}{Limited}                               
& \multicolumn{1}{c}{Limited}                                     
& \multicolumn{1}{>{\columncolor{tablegreen}}c}{Unlimited}                                                 \\

\textbf{Transparency}      
& \multicolumn{1}{c}{-}                           
& \multicolumn{1}{>{\columncolor{tablegreen}}c}{\xmark}                               
& \multicolumn{1}{>{\columncolor{tablegreen}}c}{\xmark}                                    
& \multicolumn{1}{c}{\xmark}                                          
& \multicolumn{1}{>{\columncolor{tablegreen}}c}{\cmark}    \\
\bottomrule
\end{tabular}
\caption{Five paradigms in NLP. There is no recurrent connection in paradigm I-III since models tuned for one task usually cannot be directly reused for the others due to the discrepancies between output layers.
``Transparency'' refers to, during the pre-training stage, whether the model stores data in a way that can be understood by downstream tasks so that they can access the required data efficiently.
}
\label{tab:five_paradigm}
\end{table}

(\textit{Fully Supervised Learning})
In the early time, researchers applied traditional machine learning techniques, exemplified by  SVM \cite{guyon2002gene}, PGM \cite{koller2009probabilistic} to learn from labeled data, usually without model pre-training over the large unlabeled corpus. The focus was to perform good \textbf{feature engineering} to extract salient features that are beneficial to the task at hand.
(\textit{Pre-train, Fine-tune - Non-contextual}) With the advent and popularity of neural networks, useful features of data can be automatically extracted, while, at the cost of \textbf{architecture engineering} of neural networks, e.g., which types of neural architecture will be more suitable for the sentiment classification task?
Besides, researchers at that time found that a good initialization of word representations, which is pre-trained by mining the co-occurrence information between words or n-grams \cite{mikolov2013distributed}, could further improve downstream task performance. However, a model fine-tuned on one task is usually specialized and could not be re-used by the others.\footnote{One major reason is that different types of tasks own different output layers.}
(\textit{Pre-train, Fine-tune - Contextual}) Starting from 2017, people seek to learn contextual word representations that can capture more nuanced semantics based on the context around a word \cite{DBLP:journals/corr/abs-2003-08271}. This results in large pre-trained language models (PLMs) such as BERT \cite{devlin-etal-2019-bert} and GPT \cite{radford2019language} that learn text representations from massive unlabeled text. Another benefit of these PLMs is that one can build on them with a few adjustable parameters for downstream tasks without needing to design complex model structures from scratch. This has led to a shift in focus from architecture engineering to \textbf{objective engineering} and a relaxation of the requirements for the amount of downstream training data. However, similar to the previous paradigm, a model structure could only solve one specific downstream task in general.
(\textit{Pre-train, Prompt, Predict}) This paradigm has come into fashion since the advent of GPT3 \cite{brown2020language}. Instead of adapting PLMs
to downstream tasks through objective engineering, people begin to think about reformulating the downstream tasks to pretraining-like tasks. This further relaxes the requirements for labeled data, and PLMs can even perform few-shot or zero-shot learning. However, due to the way how PLMs store data is not interpretable to downstream tasks, one needs to perform complex \textbf{prompt engineering}~\cite{liu2021pretrain}.
\subsection{reStructured Pre-training}
As of writing this, we wish to push forward the progress of this field by exploring the potential next paradigm.
Unlike existing paradigms that mainly focus on  \textit{model-centric} design, we think more from the data perspective to maximize the utility of the already available data.
Specifically, we take a data \textit{storing} $\&$ \textit{accessing} view where the pre-training stage is considered as a data storing process while downstream task (e.g.,  sentiment classification) training based on pre-trained models is regarded as data accessing process from pre-trained models, and claim that a good data storage mechanism should make the stored data more accessible.
To achieve this goal, \textbf{we look at data as an object that consists of diverse signals and argue that a good pre-trained model should (1) cover as many types of signals as possible and (2) provide precise access mechanisms for these signals when required by downstream tasks}.
i.e., a shift from \textit{pre-training over plain texts} to \textit{pre-training over structured signals}. 
In general, there are three steps within this new paradigm.

\begin{enumerate}
     \item \textbf{reStructure} There are always a variety of signals in the world's data with different formats. For example, it can be as simple as ``what the next word is'' provided by \textit{any text}; it can also be the named entity information from hyperlinks in \textit{Wikipedia} or just a word definition from a
    \textit{dictionary}. To make the PLM better educated,\footnote{Stored knowledge could be easily elicited by downstream tasks.} it is reasonable to take all available supervision in the world to train it. However, due to the diversity of the existing formats of various signals, it is necessary to restructure all of them into a unified form for model pre-training.
    \item \textbf{Pre-train} Once all pre-training data have been structuralized with a unified format, a pre-training architecture will be selected and trained over such structured data. 
    \item \textbf{Fine-tune} After pre-training, models can be further fine-tuned on restructured labeled data for better performance. Another common use case is directly applying them to downstream tasks, usually through zero-shot prompting. 
\end{enumerate}

\subsection{Evolutionary Process of Engineering Cycles} \label{sec:evolution}

 \begin{wrapfigure}[13]{r}{6cm}
    \centering
    \vspace{-10pt}
    \includegraphics[width=1\linewidth]{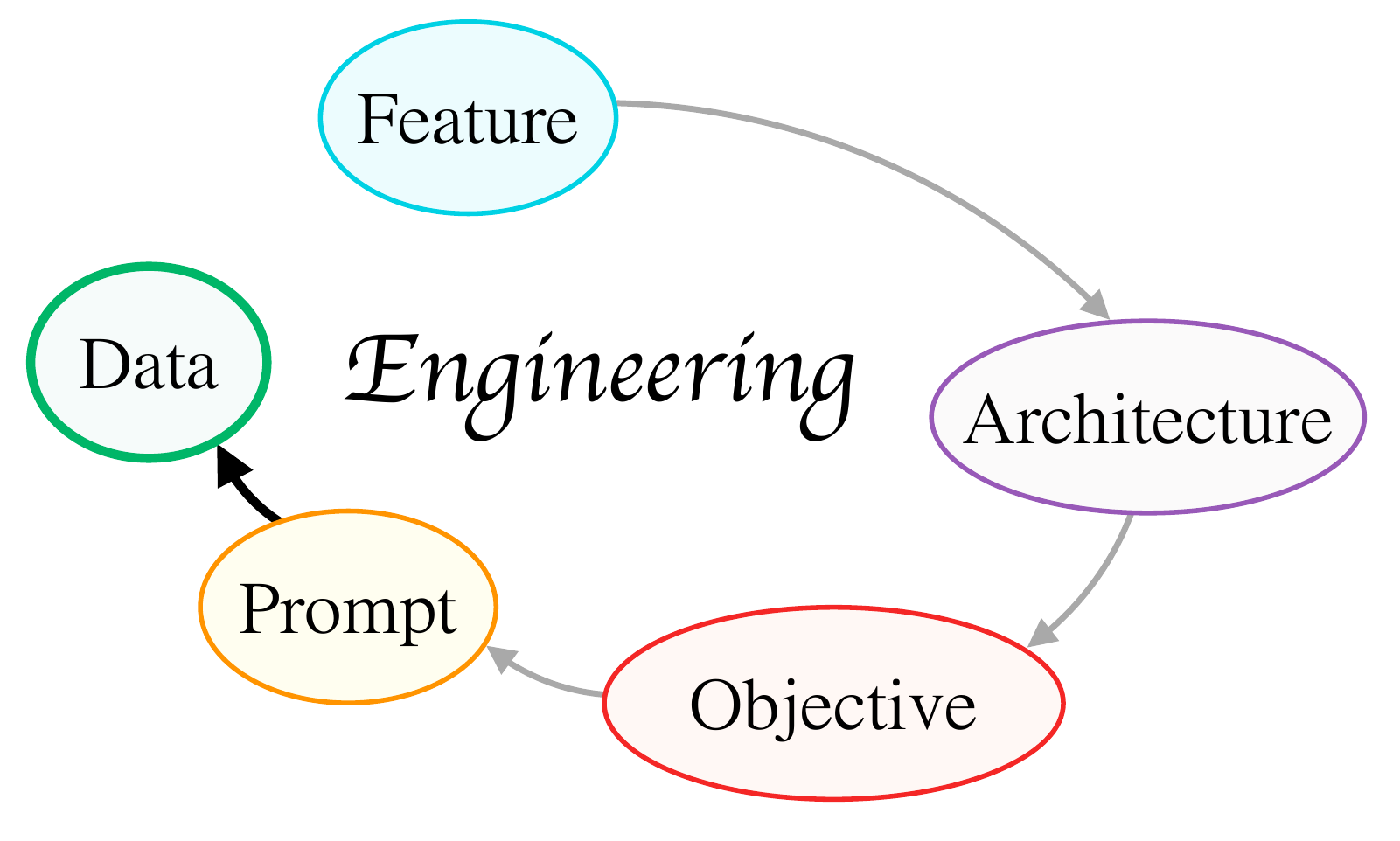}
    \caption{The evolutionary process of engineering ``cycles'' in machine learning. 
    }
    \label{fig:engineering_cycle}
\end{wrapfigure}

When we look at the evolution process of engineering in different machine learning (ML) techniques from a global view as shown in Fig.\ref{fig:engineering_cycle}, we can more easily grasp the core driving force of ML technology development, which is: \textbf{the iteration of technology always moves along the direction that system developers can design a better and more general system by doing fewer things.}
For example, although architecture engineering requires us to explore the hyper-parameters of various network structures, compared with feature engineering, which requires developers to design feature templates for data samples for different tasks manually, researchers' job has become easier.
This is not only because the hyper-parameter searching of structures can often be done automatically by machines rather than by people looking at samples but also because a well-designed structure can be effective for more datasets or tasks rather than a single one.

The new paradigm of \textit{reStructure, Pre-training, Fine-tune} highlights the importance of data, and researchers need to put more engineering effort into data processing. Note that this does not mean that more efforts than previous paradigms are needed since this paradigm only requires researchers to understand the structure of the data and process them into a unified form, rather than requiring a careful definition of each feature in the data as having been done in feature engineering.

\subsection{Design Considerations}

Any paradigm will put forward the demand of engineering efforts for machine learning researchers (or developers) (e.g., feature engineering asks researchers deeply understands what features of data matter for different tasks). The discrepancies mainly lie in the different types of engineering.
In what follows, we will detail what matters in restructured pre-training paradigm.

\begin{enumerate}
    \item \textbf{Signal Definition} \textit{Signal} is the useful information that can serve for learning knowledge for specific tasks and instructs models for learning optimization.
    As the first step for restructured learning, we first need to figure out what signals naturally exist in the world and are collected and available.\footnote{We do not need to enumerate all types of signals, which are also non-trivial and non-necessary, since ``next word'' in a text, as a special signal, is powerful enough.} This will be discussed in \S\ref{sec:signals}.
    
    \item \textbf{Data Mine Identification} In the real world, we have access to various data sources such as news websites, Wikipedia, knowledge bases, or even online videos. \textit{Data Mine} refers to a collection of data that are rich in diverse types of signals. Once signals have been defined, searching for suitable data mines is expected (\S\ref{sec:data_mines}).
    
    \item \textbf{Signal Extraction} How to effectively extract signals from data mines also matters for restructured learning. We will detail how we mine signals from the signal sources in \S\ref{sec:signal_extraction}.
    
    \item \textbf{Signal Restructuring} This process cares about how to represent all types of signals using a unified format and narrow the gap between data storage and data retrieval. We will detail it in \S\ref{signal_restructure}.
    
    \item \textbf{Pre-training and Tuning} This process cares about what would be a desirable pre-training architecture so that all restructured signals can be effectively used for model pre-training. We will detail it in \S\ref{subsection:pretrain}.

\end{enumerate}


\clearpage
\section{reStructuring Engineering} \label{sec:rst_engineering}

\subsection{Signal Definition}
\label{sec:signals}
\begin{figure}[!th]
    \centering
    \includegraphics[width=0.95\linewidth]{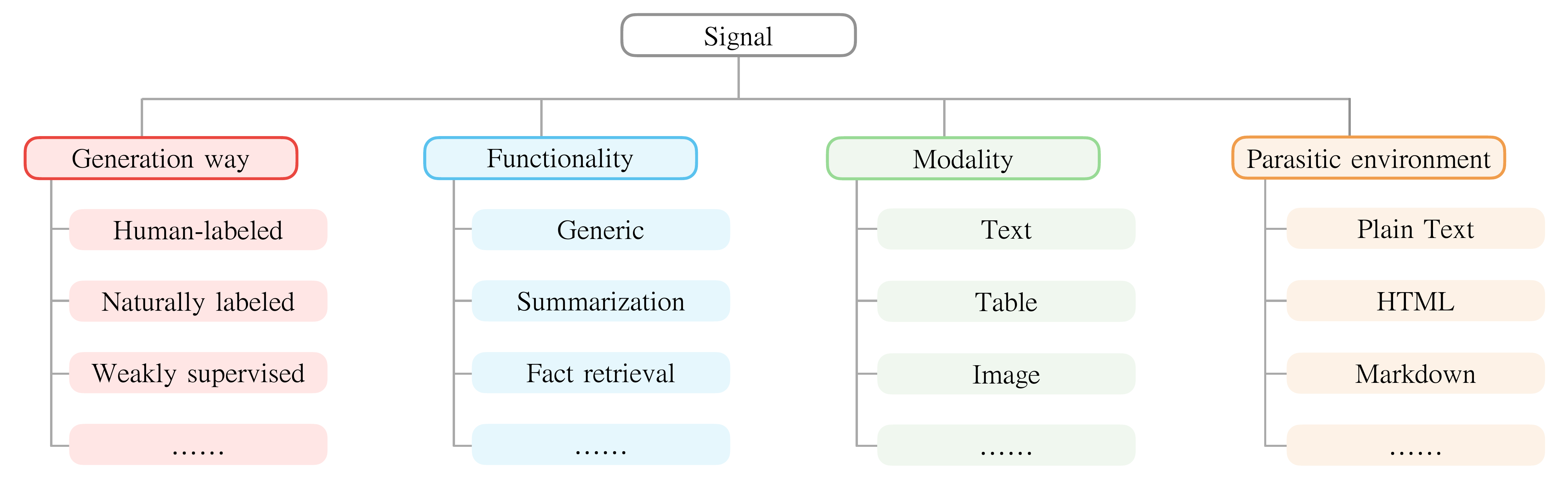}
    \caption{Four perspectives of characterizing signals.}
    \label{fig:signal_typology}
\end{figure}


\begin{wrapfigure}[11]{r}{5cm}
    \centering
    \vspace{-10pt}
    \includegraphics[width=0.9\linewidth]{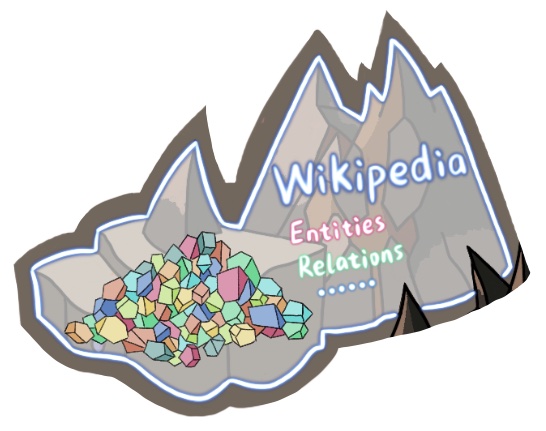}
    \caption{Signals in a data mine.}
    \label{fig:signal_in_data}
\end{wrapfigure}
Signals, like gems in a mine (Fig.~\ref{fig:signal_in_data}), are useful information present in the data that can provide supervision for machine learning models, which \textbf{can be represented as $\mathbf{n}$-tuples}. For example, $($``Mozart was born in Salzburg", ``Mozart, Salzburg"$)$ can be considered as a signal for named entity recognition. The first entry in the pair is a descriptive sentence, and the second one contains entities that are mentioned in the sentence.

Usually, signals can be clustered from different perspectives, as shown in Fig.~\ref{fig:signal_typology}. For example, signals such as ``the next word'' exist everywhere, while signals like ``the syntactic structure of a sentence'' usually require human annotations. Signals such as entity relations could also be extracted with weakly-supervised methods such as distant supervision \cite{mintz2009distant}.
Similarly, we can group signals from other dimensions, such as their functionality, modality, and parasitic environment, as shown in Fig.~\ref{fig:signal_typology}.

\begin{wrapfigure}[10]{r}{7cm}
    \centering
    \vspace{-15pt}
    \includegraphics[width=1\linewidth]{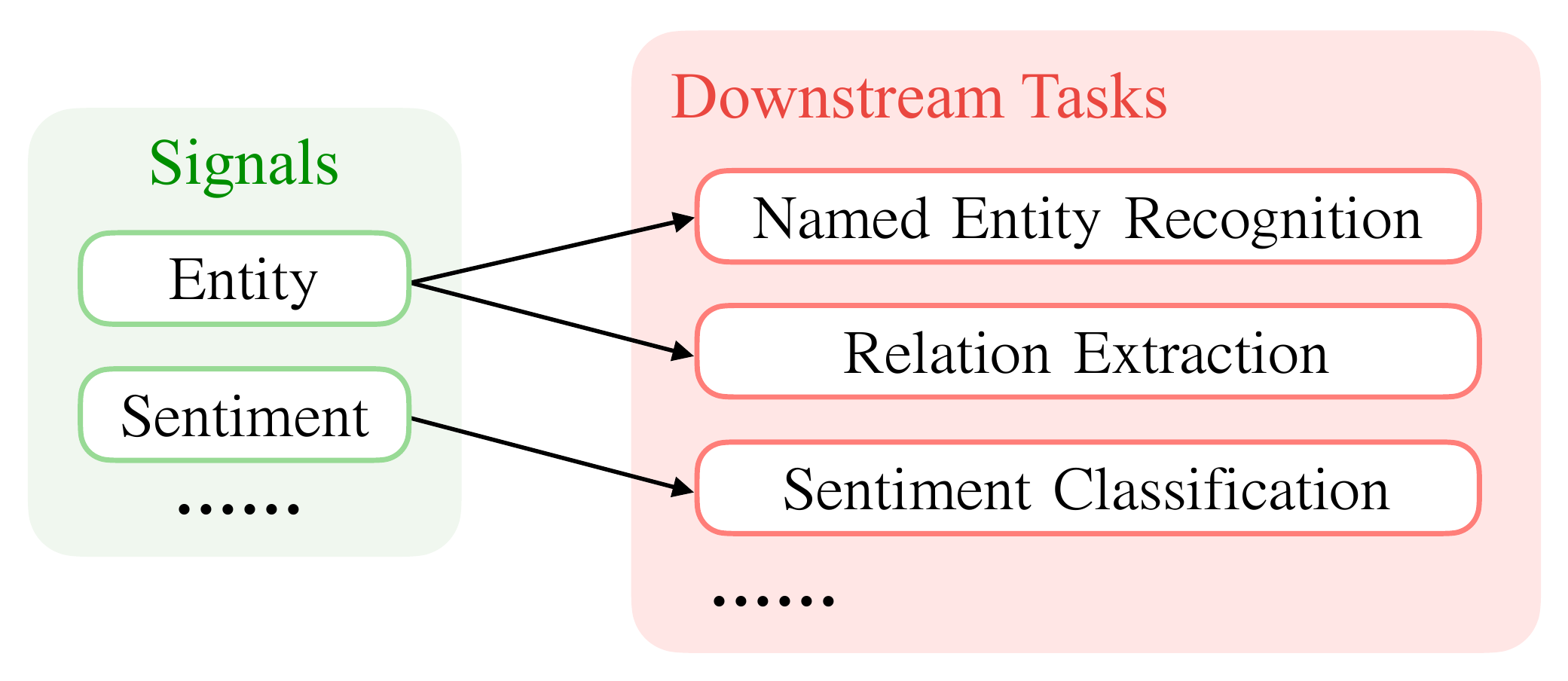}
    \caption{Relationship between signals and downstream tasks.}
    \label{fig:signal_downstream}
\end{wrapfigure}

\paragraph{Relationship between Signal and Task}
Commonly, one type of signal can offer supervised information for model training from different downstream tasks, and the model learned for one task may require multiple flavors of signals.
Fig.~\ref{fig:signal_downstream} gives an illustration where an arrow represents that one type of signal is beneficial for a downstream task. 
Below, we make a comprehensive organization of signals that would play an important role in different downstream NLP tasks.
\paragraph{}

\subsubsection{Signal \& Downstream Tasks}
\label{sec:signal-and-downstream-tasks}

\begin{itemize}

    \item \textbf{S1: Sentiment}: The sentiment of a given text. It can be positive, negative, or neutral (or more fine-grained, happy, angry, excited, etc.). This signal naturally exists in some movie reviews or product reviews on HTML pages, where the comments are accompanied by some rating. It can serve for the sentiment classification task.
    \item \textbf{S2: Category}: The category of a given text. A piece of text can contain information about a certain category or several categories (e.g., one document is labeled both politics and economies). This signal typically naturally exists on many websites where the contents are organized into different sections. It can serve for the topic classification task.
    \item \textbf{S3: Summary}: The concise summary of a given text. This signal typically exists in some news websites, where there are human-curated summaries for each news article by editors. Besides, some academic documents also have abstracts that summarize the core contents. This signal is useful for both the summarization task and sentence expansion task.
    \item \textbf{S4: Title}: The title of a given text. It is similar to S3, but with shorter text and more coverage, since paragraphs on the Internet almost always have titles.
    \item \textbf{S5: Content organization}: The high-level structural organization of the given text. The most widely used example is the table of contents for an article or book. This signal is useful for content planning when generating texts. 
    \item \textbf{S6: Word meaning}: The definitions or meanings of a word. There are human-curated resources for this signal, such as WordNet~\cite{miller1995wordnet}. This signal is useful for the task of word sense disambiguation.
    \item \textbf{S7: Synonyms}: The word or phrase that expresses a similar meaning to another one. There are human-curated resources for this signal, such as WordNet. This signal is useful for the task of paraphrasing.
    \item \textbf{S8: Antonyms}: The word or phrase that expresses a meaning opposed to the meaning of another word. There are human-curated resources for this signal, such as WordNet. This signal is useful for the task of negation.
    \item \textbf{S9: Part-of-Speech}: The part-of-speech tag for each word. This signal typically exists in human-curated datasets.
    \item \textbf{S10: Word morphologies}: The prefix, suffix of a word, or whether it is capitalized or not. It naturally exists in plain texts, and is useful for many word-related tasks such as part-of-speech tagging and named entity recognition.
    \item \textbf{S11: Reference resolution}: What the pronouns in the text (e.g., it, they) refer to. This signal typically exists in human-curated datasets and is useful for the task of co-reference resolution and relation extraction.
    \item \textbf{S12: Compositionality}: Which words are semantically-composable to form a new unit (e.g. noun phrase, clause). This signal typically exists in human-curated datasets and is useful for the task of constituency parsing.
    \item \textbf{S13: Dependency}: The dependency relationships between words within a sentence. This signal typically exists in human-curated datasets and is useful for the task of dependency parsing and relation extraction.
    \item \textbf{S14: Importance}: The salience of words, phrases, or sentences to understand the meaning of a text. For example, the high TF-IDF words are typically more important. This signal is typically automatically calculated, and is useful for the task of summarization.
    \item \textbf{S15: Previous content}: The previous content (e.g., words, phrases) of a given text. This signal naturally exists in plain text and is useful for the task of language modeling.
    \item \textbf{S16: Future content}: The future content of a given text. This signal naturally exists in plain text and is useful for the task of language modeling. 
    \item \textbf{S17: Cloze}: The content at some cloze positions. This signal can be constructed automatically. Many popular pre-training objectives (e.g., masked language modeling \cite{devlin-etal-2019-bert}, corrupted span prediction \cite{DBLP:journals/corr/abs-1910-10683}) make use of this signal.
    \item \textbf{S18: Sentence ordering}: The ordering of sentences within a given text. This signal naturally exists in plain text, and is useful for the task of language modeling.
    \item \textbf{S19: Sentence distance}: Whether two sentences are adjacent, from the same document, or from two different documents. This signal naturally exists in plain text and is useful for the task of language modeling.
    \item \textbf{S20: Discourse relation}: The semantic or rhetorical relation between two sentences. This signal typically exists in human-curated datasets and is useful for the task of discourse parsing.
    \item \textbf{S21: Grammar}: Whether a text contains grammatical errors or not. This signal typically exists in human-curated datasets and is useful for the task of grammar error correction.
    \item \textbf{S22: Fluency}: Whether a text is consistent and fluent in terms of the content it describes (naturally occurred texts are typically fluent, but we can artificially construct non-fluent texts by aggregating sentences from different places.). This signal typically exists both in human-curated datasets and any plain texts and is useful for the task of meta-evaluation for generation metrics.
    \item \textbf{S23: Paraphrase}: Whether two texts are semantically equivalent. This signal typically exists in human-curated datasets and is useful for the task of paraphrasing.
    \item \textbf{S24: NLI-Entailment}: The hypothesis is true given the premise. This signal typically exists in human-curated datasets and is useful for the task of natural language inference.
    \item \textbf{S25: NLI-Contradiction}: The hypothesis is false given the premise. This signal typically exists in human-curated datasets and is useful for the task of natural language inference.
    \item \textbf{S26: NLI-Undetermined}: The hypothesis is undetermined given the premise. This signal typically exists in human-curated datasets and is useful for the task of natural language inference.
    \item \textbf{S27: Entity mentions}: The entities within a given text. This information broadly exists in HTML web pages, where entities will be linked to their own page. This signal is useful for named entity recognition and many other entity-relevant tasks.
    \item \textbf{S28: Relations}: The relationships between different entity mentions. This information typically exists in human-curated datasets and is useful for the task of relation extraction.
    \item \textbf{S29: Text Comprehension}: This corresponds to reading comprehension tasks in which, given a text and a question, an answer can be inferred from the given text. This signal typically exists in exams and quizzes, and is useful for the task of question answering.
    \item \textbf{S30: Multilingual parallel data}: The same meanings expressed in multiple languages. This signal naturally exists in many news websites, where the same piece of news will be displayed in multiple languages. This signal is useful for the machine translation task.
    \item \textbf{S31: Multilingual matching}: Whether the text $t_a$ in language $l_a$  can translate to text $t_b$ in language $l_b$. This signal can be automatically created and is useful for the task of machine translation.
    \item \textbf{S32: Multilingual alignment}: The specific alignments for words/phrases in two languages. This signal typically exists in bilingual dictionaries and is useful for the task of machine translation. 
    \item \textbf{S33: Commonsense knowledge}: The commonsense in our daily life. For example, the sun has no eyes, and water can flow, etc. This signal typically exists in human-curated datasets and is useful for the task of commonsense reasoning.
    \item \textbf{S34: Factual knowledge}: Facts, e.g., Mozart was born in 1756. This signal typically exists in some human-curated resources, such as Wikidata. which is useful for the task of fact retrieval and information extraction.
    \item \textbf{S35: Advanced knowledge}: The laws of nature summarized by predecessors based on the objective world. For example, the laws of relativity by Albert Einstein. This signal typically exists in books and scientific articles and is useful for the task of mathematical reasoning and logical reasoning.
    \item \textbf{S36: Temporal information/Procedures}: The procedures for completing some tasks. Typical datasets include WikiHow.\footnote{\url{https://www.wikihow.com/}} This signal is useful for the task of temporal reasoning.
    \item \textbf{S37: Reasoning}: Includes commonsense reasoning or reasoning based on other knowledge. For example, given that ``today is cold", the model should reason that ``I should wear more". This signal typically exists in human-curated datasets and is useful for the task of commonsense reasoning.
    \item \textbf{S38: Table caption}: The summary of the table content. This signal naturally exists in many HTML websites and is useful for the task of table captioning.
    \item \textbf{S39: Table text}: The text that uses the table information for further descriptions. This signal naturally exists in many HTML websites and is useful for the task of table-to-text generation.
    \item \textbf{S40: Image caption}: The summary of image content. This signal naturally exists in many HTML websites and is useful for the task of image captioning.
    \item \textbf{S41: Image text}: The text that uses the image information for further descriptions. This signal naturally exists in many HTML websites and is useful for the task of image-to-text generation.
    \item \textbf{S42: Multimodal matching}: Whether a table/image is matched with the caption/text. This signal can be automatically constructed and is useful for the task of table-to-text generation and image-to-text generation.
\end{itemize}

\newpage

\subsection{Data Mines}
\label{sec:data_mines}

\begin{wrapfigure}[13]{r}{5.5cm}
    \centering
    \vspace{-10pt}
    \includegraphics[width=0.95\linewidth]{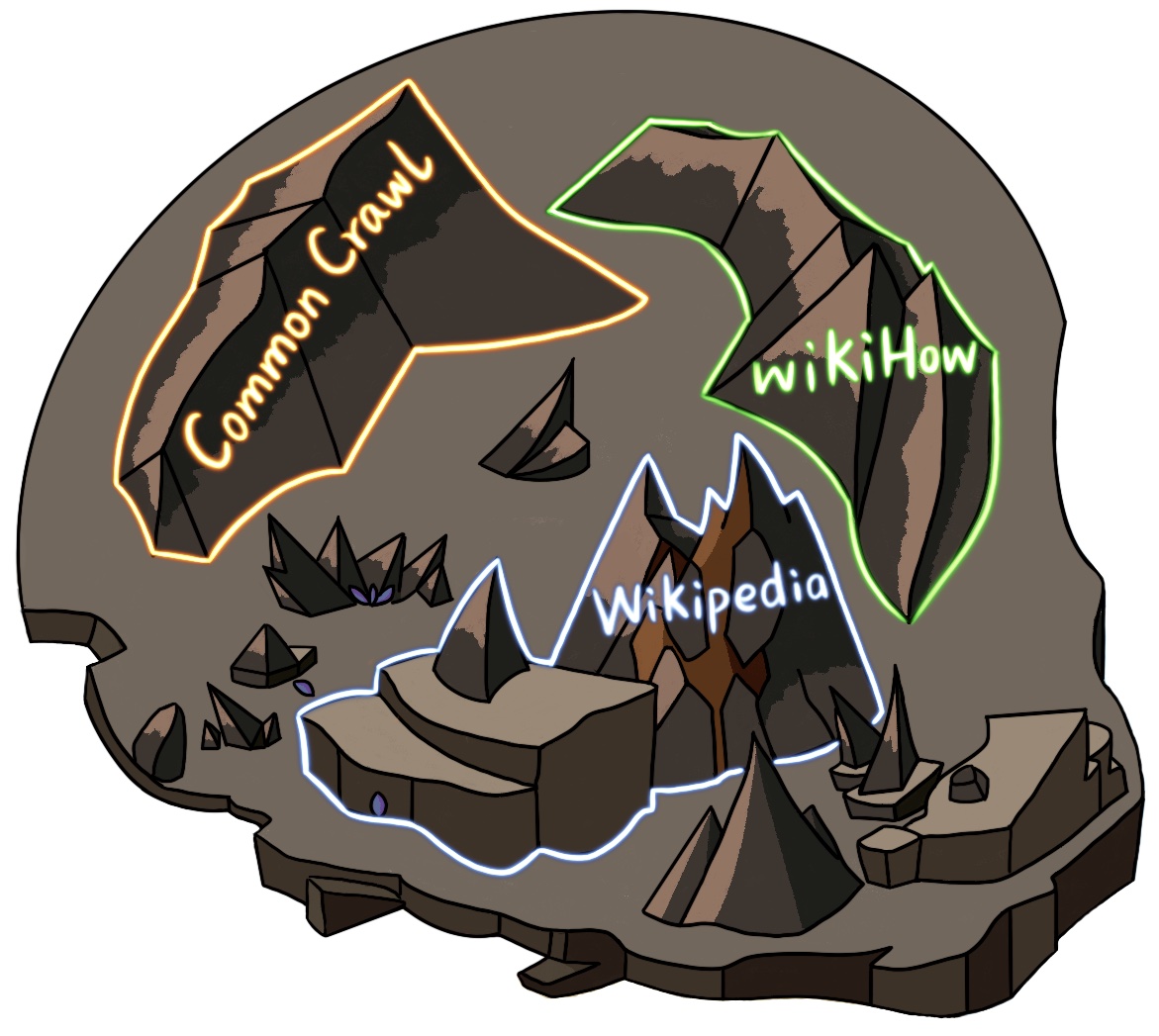}
    \caption{Data mines.}
    \label{fig:sources_gem}
\end{wrapfigure}
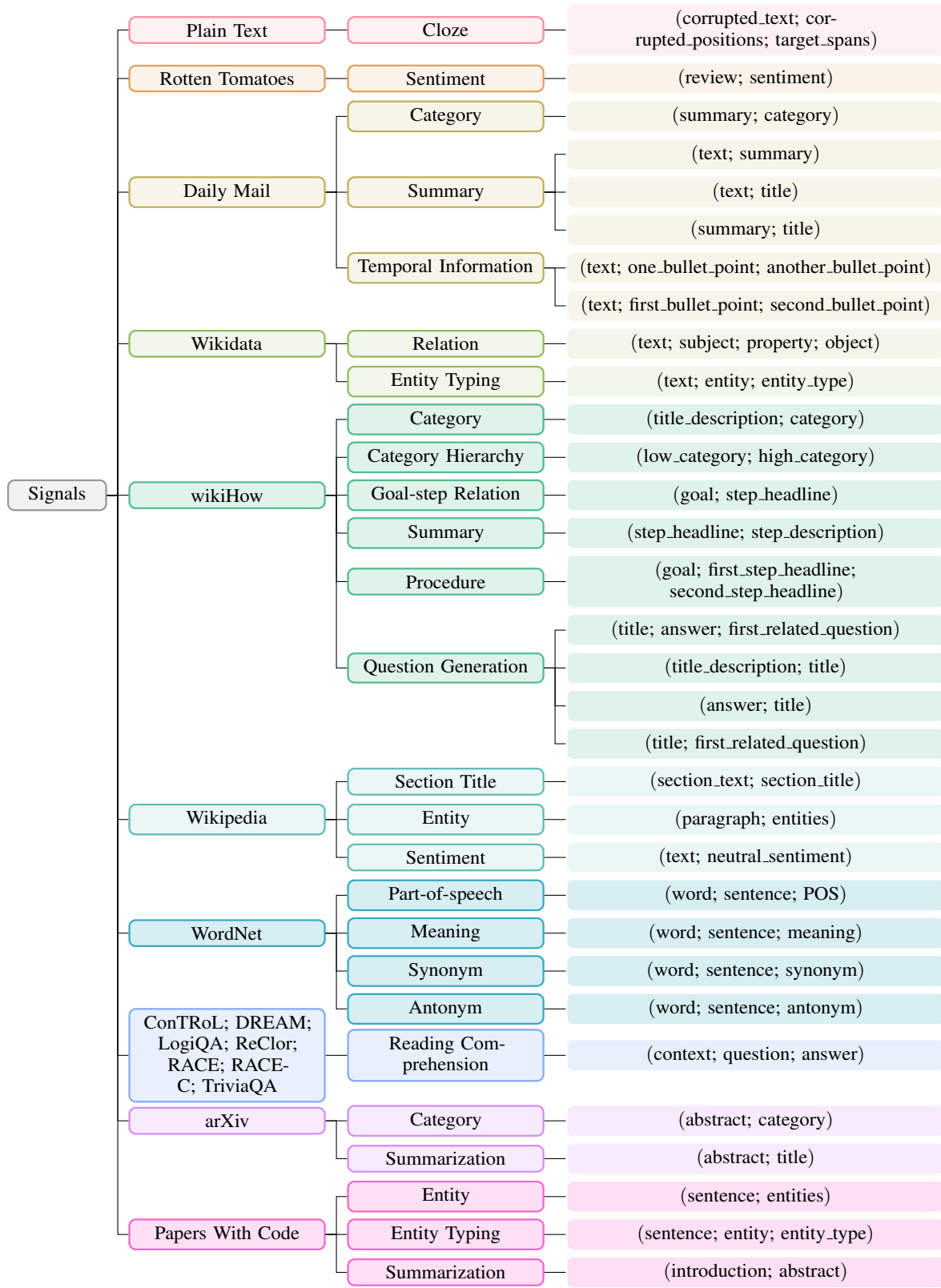
\begin{figure*}
\footnotesize
        \begin{forest}
            for tree={
                forked edges,
                grow'=0,
                draw,
                rounded corners,
                node options={align=center,},
                text width=2.7cm,
                s sep=2pt,
                calign=child edge, calign child=(n_children()+1)/2,
            },
            [Signals, fill=gray!45, parent
                [Plain Text, for tree={plaintext}
                    [Cloze,  plaintext
                        [$($corrupted\_text; corrupted\_positions; target\_spans$)$, 
                        style = plaintext_work]
                    ]
                ]
                [Rotten Tomatoes, for tree={pretrain}
                    [Sentiment,  pretrain
                        [$($review; sentiment$)$, 
                        style = pretrain_work]
                    ]
                ]
                [Daily Mail, for tree={fill=red!45,template}
                    [Category,  template
                        [$($summary; category$)$, template_work]
                    ]
                    [Summary,  template
                        [$($text; summary$)$, template_work]
                        [$($text; title$)$, template_work]
                        [$($summary; title$)$, template_work]
                    ]
                    [Temporal Information,  template
                        [$($text; one\_bullet\_point; another\_bullet\_point$)$, template_work]
                        [$($text; first\_bullet\_point; second\_bullet\_point$)$, 
                        template_work]
                    ]
                ]
                [Wikidata, for tree={fill=blue!45, answer}
                    [Relation, answer
                        [$($text; subject; property; object$)$, answer_work
                        ]
                    ]
                    [Entity Typing, answer
                        [$($text; entity; entity\_type$)$, answer_work
                        ]
                    ]
                ]
                [wikiHow, for tree={multiple}
                    [Category, multiple
                        [$($title\_description; category$)$, multiple_work]                          
                    ]
                    [Category Hierarchy, multiple
                        [$($low\_category; high\_category$)$,
                        multiple_work]
                    ]
                    [Goal-step Relation, multiple
                        [$($goal; step\_headline$)$, multiple_work]
                    ]
                    [Summary, multiple
                        [$($step\_headline; step\_description$)$, multiple_work]
                    ]
                    [Procedure, multiple
                        [$($goal; first\_step\_headline; second\_step\_headline$)$, multiple_work]
                    ]
                    [Question Generation, multiple
                        [$($title; answer; first\_related\_question$)$, multiple_work]
                        [$($title\_description; title$)$,
                        multiple_work]
                        [$($answer; title$)$, multiple_work]
                        [$($title; first\_related\_question$)$,
                        multiple_work]
                    ]
                ]   
                [Wikipedia, for tree={tuning}
                    [Section Title, tuning
                        [$($section\_text; section\_title$)$, tuning_work]               
                    ]
                    [Entity, tuning
                        [$($paragraph; entities$)$, tuning_work]               
                    ]
                    [Sentiment, tuning
                        [$($text; neutral\_sentiment$)$, tuning_work]               
                    ]
                ]
                [WordNet, for tree={wordnet}
                    [Part-of-speech, wordnet
                        [$($word; sentence; POS$)$, wordnet_example]               
                    ]
                    [Meaning, wordnet
                        [$($word; sentence; meaning$)$, wordnet_example]               
                    ]
                    [Synonym, wordnet
                        [$($word; sentence; synonym$)$, wordnet_example]               
                    ]
                    [Antonym, wordnet
                        [$($word; sentence; antonym$)$, wordnet_example]               
                    ]
                ]  
                  Question Answering
                [ConTRoL; DREAM; LogiQA; ReClor; RACE; RACE-C; TriviaQA, for tree={qa}
                    [Reading Comprehension, qa
                        [$($context; question; answer$)$, qa_example]    
                    ]
                ]  
                [arXiv, for tree={arxiv}
                    [Category, arxiv
                        [$($abstract; category$)$, arxiv_example]               
                    ]
                    [Summarization, arxiv
                        [$($abstract; title$)$, arxiv_example]               
                    ]
                ]  
                [Papers With Code, for tree={pwc}
                    [Entity, pwc
                        [$($sentence; entities$)$, pwc_example]               
                    ]
                    [Entity Typing, pwc
                        [$($sentence; entity; entity\_type$)$, pwc_example]               
                    ]
                    [Summarization, pwc
                        [$($introduction; abstract$)$, pwc_example]               
                    ]
                ]  
            ]
        \end{forest}
            \caption{Data Mines with Signals.}
            \label{fig:signal_tree}
\end{figure*}

In the real world, there are a number of data sources that are rich in different types of signal, as illustrated in Fig.~\ref{fig:sources_gem}. We refer to them as \textbf{data mines}. reStructured pre-training allows us to make full use of them. 
In what follows, we will introduce several popular data mines and corresponding signals that can be easily extracted. We organize the collected signals ($n$-tuples) in a tree diagram, as shown in Fig.~\ref{fig:signal_tree}, and elaborate on each signal below. 

\subsubsection{Plain Text}
There are some signals that are abundantly present in plain text, such as S15 (previous content), S16 (future content), and S17 (cloze) mentioned in \S\ref{sec:signal-and-downstream-tasks}. Due to their large presence and easy availability, they can usually be used as the earliest pre-training signals for the model to learn some basic language knowledge.

\subsubsection{Rotten Tomatoes}
Rotten Tomatoes\footnote{\url{https://www.rottentomatoes.com/}} is an American review-aggregation website for various TV shows and movies. Every review is accompanied by a rating score, indicating reviewers' sentiment polarities. Therefore, we use the movie and television reviews on Rotten Tomatoes to extract signals of sentiment polarity.

\subsubsection{Daily Mail}
Daily Mail\footnote{\url{https://www.dailymail.co.uk/home/index.html}} is a British news website that publishes news articles in different domains, such as sport, health, etc. For each news article, there are also bullet points created manually that summarize key information. We use the archived news articles\footnote{\url{https://www.dailymail.co.uk/home/sitemaparchive/index.html}} provided by Daily Mail and consider the news articles from 2008-7-1 to 2021-10-7. We collect the following signals from Daily Mail.

\noindent \textbf{Category} Each news article is categorized into one of the following categories: ``News", ``U.S.", ``Sport", ``TV\&Showbiz", ``Australia", ``Femail", ``Health", ``Science", ``Money", ``Video", ``Travel", ``Best Buys", ``Discounts". This category information can be easily acquired using the URL of each news article.

\noindent \textbf{Summary} The news headline can be seen as an extreme summary of the whole article, while the bullet points can be seen as a summary with details to some degree. We consider three ways to exploit this signal: (i) generate the summary constructed using all bullet points given the news article. (ii) Generate the title using the given news article. (iii) Generate the bullet points given the title (this is the task of sentence expansion, the reverse task of summarization).

\noindent \textbf{Temporal information} A news article typically contains multiple events that occur sequentially. Each bullet point can be seen as an event, and we consider two ways to exploit this signal: (i) given two events, which one happened first. (ii) Given an event, what could be the subsequent event.

\subsubsection{Wikidata}
Wikidata\footnote{\url{https://www.wikidata.org/wiki/Wikidata:Main_Page}} is an open knowledge base that documents many data items. For each data item, Wikidata records its relationship to other data items in detail, which can be useful for entity/relation extraction tasks. We use the data dump of 2021-11-18 and consider the following signals from Wikidata.

\noindent \textbf{Relation} Wikidata records over nine thousand properties (i.e., relations). We traverse all the data items and extract the signals (relational triples in which each data item is involved). We consider three ways to exploit this signal: (i) Given the subject and the object, predict the relation. (ii) Given the subject and the relation, predict the object. (iii) Given the object and the relation, predict the subject.

\noindent \textbf{Entity typing} For some data items, there is a ``P31" property, which stands for ``instance of". If $A$ is an instance of $B$, we can consider $B$ a superclass of $A$, playing the role of entity type for $A$.

\subsubsection{wikiHow}
wikiHow\footnote{\url{https://www.wikihow.com/Main-Page}} is a website that collects ``How-to" articles on a variety of topics. Each ``How-to" article in wikiHow contains important information, including problem description, steps, related questions, category, etc. We use the dataset collected by \cite{zhang-etal-2020-reasoning} and consider the following signals.

\noindent \textbf{Category}
Each article in wikiHow is categorized into a category hierarchy. For example, the category hierarchy for the article ``How to Take Action to Acquire Enough Wealth for Goals including Serving Others" is \textit{Making Money} $\Longrightarrow$ \textit{Managing Your Money} $\Longrightarrow$ \textit{Finance and Business}. There are 22 top-level categories in total, which are: ``Finance and Business", ``Health", ``Computers and Electronics", ``Cars \& Other Vehicles", ``Family Life", ``Youth", ``Hobbies and Crafts", ``Arts and Entertainment", ``Relationships", ``Food and Entertaining", ``Pets and Animals", ``Education and Communications", ``Sports and Fitness", ``Home and Garden", ``Personal Care and Style", ``Travel", ``Holidays and Traditions", ``Philosophy and Religion", ``Work World", ``Screenplays", ``Outdoor Shelters", ``Cleaning Heater Appliances". Therefore, We consider two ways to exploit this category hierarchy information: (i) given the problem description, predict the top-level category of this piece of text. (ii) The subclass-superclass relationship between categories.
        
\noindent \textbf{Goal-step relation}
Each article on wikiHow contains instructions about accomplishing a particular goal, and these can naturally serve as the goal-step signal. In particular, we consider two ways to exploit this signal: (i) given a goal, predict the steps to take. (ii) Given steps, predict the goal those steps serve for.

\noindent \textbf{Summary} Each step of the instructions in the article consists of a summary sentence at the beginning and a detailed description later. Thus, we can easily construct text-summary pairs.

\noindent \textbf{Procedure} Due to the temporal nature of the steps mentioned in the text, we consider two ways to exploit this temporal information: (i) given a goal, and two steps to be done in the process of accomplishing this goal, determine the order of these two steps. (ii) To accomplish a goal, determine what should be done in the next step after performing one step.

\noindent \textbf{Question generation} A wikiHow article contains many elements, the most important of which are: \textit{a ``How-to" question}, \textit{A description of the problem}, \textit{steps to reach the goal}, \textit{related questions}. With these elements, we can mine problem-generating signals. We consider four ways to exploit this signal: (i) given a ``How-to" question and the instructions that answer the question, predict the next question one may ask. (ii) Given the description text, propose a relevant question. (iii) Given the instructions for completing a goal, predict the ``How-to" question. (iv) Given a question, find relevant or similar questions.

\subsubsection{Wikipedia} 
Wikipedia\footnote{\url{https://www.wikipedia.org/}} is a free online encyclopedia, created and maintained through crowdsourcing. It contains various knowledge of the world, and the HTML format of Wikipedia naturally offers various signals we can mine from. We use static HTML dumps.\footnote{\url{https://dumps.wikimedia.org/other/static_html_dumps/current/en/}} In particular, we consider the following signals.

\noindent \textbf{Section title} Each page of Wikipedia contains many subsections in which the matching relationship between text and subheadings can constitute a weakly supervised signal for text classification.

\noindent \textbf{Entity} Each HTML page in Wikipedia contains many entities, which are generally displayed as hyperlinks within the text when they first appear. Thus we can get information about weakly supervised entities based on hyperlinks.

\noindent \textbf{Sentiment} Wikipedia articles present knowledge in a narrative tone and are therefore not emotionally polarized, so we can treat them as emotionally neutral text.

\subsubsection{WordNet}
WordNet\footnote{\url{https://wordnet.princeton.edu/}} is a treasure trove of rich information about words. It records the different lexical forms and meanings of the same word, as well as the corresponding example sentences. In addition, WordNet also records the relationships between words, such as synonyms, hyponyms, and meronyms. To retrieve information about words, we first collect the most frequent 333,333 words in English provided by Kaggle\footnote{\url{https://www.kaggle.com/rtatman/english-word-frequency}}. We use WordNet $3.0$ and collect the following signals for those frequent words.

\noindent \textbf{Part-of-speech} WordNet will cover part-of-speech information when organizing the meaning of a word, so the part-of-speech of a word in an example sentence is clear.

\noindent \textbf{Meaning} Some words can have multiple meanings, which are taken into account very comprehensively by WordNet so that we can match different meanings of a word with different example sentences containing the word.

\noindent \textbf{Synonyms} WordNet also provides synonyms for a word in a certain meaning, which is also important information.

\noindent \textbf{Antonyms} Similarly, WordNet also provides antonyms for a word in a certain meaning.

\subsubsection{Question Answering Datasets}
Here we mainly consider the text comprehension signal. We focus on two sources: (1) english exam questions and their corresponding answers, (2) quiz pages containing questions and answers that already exist on the Web. In particular, we utilize the following question answering datasets in the literature.

\paragraph{ConTRoL} is derived from competitive selection and recruitment test (verbal reasoning test)
for police recruitment, with expert level quality \citep{liu2020natural}. 

\noindent \textbf{DREAM} is the first dialogue-based multiple-choice reading comprehension dataset. It is collected from English-as-a-foreign-language examinations designed by human experts to evaluate the comprehension level of non-native English learners \citep{DBLP:journals/tacl/SunYCYCC19}.

\noindent \textbf{LogiQA} is sourced from publically available logical examination papers for reading comprehension, which are designed by domain experts for evaluating the logical reasoning ability of test participants \citep{DBLP:conf/ijcai/LiuCLHWZ20}.

\noindent \textbf{RACE \& RACE-C} are collected from English examinations in China. RACE collected English exams for middle and high school Chinese students between 12 and 18 \citep{DBLP:conf/emnlp/LaiXLYH17}, while RACE-C collected English exams for college students \citep{DBLP:conf/acml/LiangLY19}.

\noindent \textbf{ReClor} collected various verbal reasoning questions that are extracted from standardized graduate admission examinations \citep{DBLP:conf/iclr/YuJDF20}.

\noindent \textbf{TriviaQA} gathered question-answer pairs from 14 trivia and quiz-league websites \citep{DBLP:conf/acl/JoshiCWZ17}. The evidence documents are automatically gathered from either Wikipedia or more general Web search results.

\subsubsection{arXiv}
Besides the general textual domain above, we also consider the scientific domain. arXiv \footnote{\url{https://arxiv.org/}} is a free distribution service and an open-access archive for academic articles across different fields. We use the metadata\footnote{\url{https://www.kaggle.com/Cornell-University/arxiv}} of 1.7M+ scholarly papers across STEM provided by arXiv and consider the following signals.

\noindent \textbf{Category} The dataset contains scientific papers from the following eight domains: ``Computer Science", ``Electrical Engineering and Systems Science", ``Quantitative Biology", ``Statistics", ``Economics", ``Mathematics", ``Physics", ``Quantitative Finance". The abstract text of an article can be categorized into one or more of those domains.

\noindent \textbf{Summary} The title of a paper can be seen as a summary of its abstract text.

\subsubsection{Papers With Code}
Papers With Code\footnote{\url{https://paperswithcode.com/}} highlights trending machine learning research and the code to implement it. It provides useful information for the machine learning society. In particular, we use their provided data,\footnote{\url{https://github.com/paperswithcode/paperswithcode-data}} including papers, methods, datasets, and evaluation tables. For each paper, we used the URL information to download the PDF version and used GROBID \citep{GROBID} to parse the content into XML format, which we can easily process later. We consider the following signals.

\noindent \textbf{Entity} In the literature on machine learning, with reference to \citet{jain-etal-2020-scirex}, we are interested in the following types of entities: dataset, metric, task, method, and generic scientific terms. The former four can be acquired through the data provided by Papers With Code, while for the last one, we use a manually curated database for AI-related terminology\footnote{\url{https://github.com/jiqizhixin/Artificial-Intelligence-Terminology-Database}}. We only consider the content from the title, abstract and introduction part of a paper and use exact match to get entities in each scientific article.

\noindent \textbf{Entity typing} We consider four entity types: ``dataset", ``metric", ``task", ``method". We collect each entity that appears in the article and assign its category to one of the above if it belongs to one of the four types. Otherwise, it is categorized as ``other". 

\noindent \textbf{Summary} We use the introduction part as the text to be summarized and the abstract part as the targeted summary.


\newpage
\subsection{Signal Extraction}
\label{sec:signal_extraction}

\begin{wrapfigure}[10]{r}{4cm}
    \centering
    \vspace{-10pt}
    \includegraphics[width=0.9\linewidth]{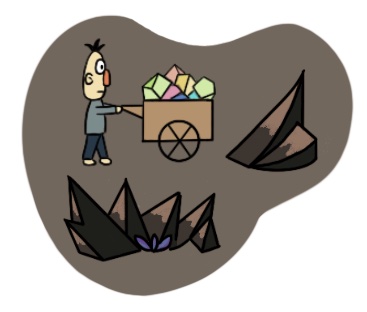}
    \caption{Signal extraction.}
    \label{fig:extract_signal}
\end{wrapfigure}

After knowing the data mines and the signals they contain, we are ready to perform signal \textit{extraction} (as illustrated in Fig.~\ref{fig:extract_signal}) together with \textit{purification}. Although the data mines mentioned above contain the signals we want, extracting these signals is still not an easy task, and the process involves getting the raw data from data mines of different modalities, data cleaning, and data normalization. Methods of existing works on this can be roughly divided into two branches: (1) ruled-based, (2) machine learning-based. In this work, we mainly focus on rule-based strategies for signal extraction and leave more high-coverage methods for future work.

\subsubsection{Plain Text}
The cloze signal is present in large quantities in plain text and does not require much pre-processing so that we can extract it directly. The target outputs are $($corrupted\_text, corrupted\_positions, target\_spans$)$ triples.

\subsubsection{Rotten Tomatoes}
\paragraph{Sentiment} To increase the diversity of the extracted reviews, we extract reviews from all DVD \& streaming movies.\footnote{\url{https://www.rottentomatoes.com/browse/dvd-streaming-all/}} For each movie, we extract up to 10,000 recent comments. Using the rating information, we are able to automatically obtain the sentiment polarity of each review. We consider comments with a score of less than 2.5 as negative comments, and comments with a score of more than 3.5 as positive comments. We remove reviews that have fewer than five words and only keep reviews that contain obvious sentiment polarity (those with a rating higher than 3.5 or lower than 2.5). The target outputs are $($review, sentiment$)$ pairs.

\subsubsection{Daily Mail}
For each article, we use the following notations: ``text" stands for the original article, ``summary" stands for the concatenation of all bullet points. ``title" stands for the heading of the article. We ignore the articles whose summary contain ``update", ``here" or ``find" since those summaries are usually redirection information, or other content not related to the articles.

\noindent\textbf{Category} We use the summary of each article as the text to be classified. Although there are thirteen categories in Daily Mail, we only consider six of them that are general enough. The categories we consider are: ``Money", ``News", ``Sport", ``Health", ``Science", ``Travel". We ignore samples with a summary of fewer than five words and truncate each summary to 400 words. The target outputs are $($summary, category$)$ pairs.

\noindent\textbf{Summary} We consider (i) text $\rightarrow$ summary (ii) text $\rightarrow$ title. For each article, we ignore it if any of the following satisfies: (i) its text has fewer than 20 words, (ii) its summary has fewer than 10 words, (iii) its text is shorter than its summary, (iv) its text is shorter than its title. We truncate the summary and text to be shorter than 400 words. The target outputs are $($text, summary$)$, $($text, title$)$ pairs.

\noindent\textbf{Summary (sentence expansion)} We use the same filtering rules as the summary signal above. The target outputs are $($text, summary$)$, $($summary, title$)$, $($text, title$)$ pairs.

\noindent\textbf{Temporal information} 
Besides the filtering rules adopted by the above two kinds of signals, we also filter out an article if its summary consists of fewer than two bullet points. The text is truncated to 400 words. The target outputs we collect are $($text, one\_bullet\_point, another\_bullet\_point$)$ triples and $($text, first\_bullet\_point, second\_bullet\_point$)$ triples. The former is used to examine the temporal order of two events, while the latter is used to examine the prediction of the immediately following event.

\subsubsection{Wikidata}
\paragraph{Relation} For relation, there are, in total, over nine thousand pre-defined properties off the shelf. However, there are many properties that are not very informative, such as ``GitHub username". Therefore, we manually filter those properties to get a cleaner subset. More specifically, the following properties are discarded by us: (i) identifier (e.g., ID) in a specific database. (ii) Properties that are specific to one culture (e.g., place of origin (Switzerland)). (iii) Properties that are related to images or URL links. When traversing data items, we first use the above filtering rules to collect $($subject, property, object$)$ triples. Then for each triple, we query the Wikipedia page of both the subject and the object and sample one of the paragraphs that contain both the subject and object, if such paragraphs exist. The target outputs are $($text, subject, property, object$)$ quads.

\paragraph{Entity typing} For entity typing, we use the ``instance of" property for entities to decide its type. When traversing data items, we keep the items that have the `instance of`" property. Besides, for each entity, we also query its Wikipedia page and sample one of the paragraphs that contain the entity if such paragraphs exist. The target outputs are $($text, entity, entity\_type$)$ triples.

\subsubsection{wikiHow}
For each wikiHow article, we are interested in the following elements: (1) the ``How to" title, (2) title description, (3) goal (title after the removal of the ``How to"), (4) steps (each step consists of a step headline and a step description), (5) related questions.

\noindent\textbf{Category} Although there are 22 top-level categories in total, we filter out ``Screenplays", ``Outdoor Shelters" and ``Cleaning Heater Appliances" since there are fewer than twenty articles in each of these classes. The target outputs are $($title\_description, category$)$ pairs. 

\noindent\textbf{Category Hierarchy} The target outputs are $($low\_category, high\_category$)$ pairs.

\noindent\textbf{Goal-step relation} For simplicity, we use the step headline to represent a step and ignore a step headline if it is longer than twenty words. For each goal, we sample a step headline for it. The target outputs are $($goal, step\_headline$)$ pairs.

\noindent\textbf{Summary} We consider using step description to generate step headline and collect all $($step\_headline, step\_description$)$ pairs as target outputs. We filter out a pair if its headline is longer than its description.

\noindent\textbf{Summary (sentence expansion)} We consider using the step headline to generate the step description. The target outputs are the same as the above summarization signal.

\noindent\textbf{Procedure} As we do for the ``goal-step relation", we use the step headline to represent a step for simplicity. We ignore a step if its step headline is longer than twenty words. When traversing all the wikiHow articles, we ignore those with fewer than two steps. The target outputs are $($goal, first\_step\_headline, second\_step\_headline$)$ triples.

\noindent\textbf{Question generation} When traversing all the wikiHow articles, we ignore those without related questions. We concatenate all step headlines as the answer to the ``how to" question. The target outputs are $($title, answer, first\_related\_question$)$ triples, $($title\_description, title$)$ pairs, $($answer, title$)$ pairs, $($title, first\_related\_question$)$ pairs.

\subsubsection{Wikipedia}
The original Wikipedia texts are very noisy and many HTML pages contain much uninformative information (e.g. welcome message). Therefore, we use some heuristics for data cleaning after inspection on the data. For each textual paragraph, we ignore it if it contains one of the following words: ``edit", ``ip address", ``redirect", ``whois  rdns  rbls", ``username", ``licence", ``welcome", ``please", ``thank", ``wiki", ``click", ``file", ``license", ``copyright".
                 
\noindent\textbf{Section title} For each Wikipedia page, we collect section texts and section titles. We ignore a section if its section text is shorter than ten words, and we also ignore those HTML pages that have fewer than two sections. We truncate each section's text to 256 words. The target outputs are $($section\_text, section\_title$)$ pairs.

\noindent\textbf{Entity} For each Wikipedia page, we collect all paragraph texts and the entities in each paragraph. This is accomplished using the hyperlink information on the HTML page. However, the following problems need to be addressed (i) many hyperlinks are superscript numbers linked to references. (ii) An entity is only hyperlinked on its first appearance. Subsequent appearances no longer carry hyperlinks. To address (i), we simply ignore hyperlinked text starting with "[" and ending with "]". To address (ii), we first record all the hyperlinked text in a Wikipedia HTML page as the full set of entities, then mark the text span in the text that matches one of those entities. The target outputs are $($paragraph, entities$)$ pairs, and we ignore one paragraph if it has fewer than ten words.

\noindent\textbf{Sentiment} For each Wikipedia page, we collect paragraphs that are no shorter than 10 words. Then we truncate each of those paragraphs to 256 words and sample one of them as the text to be judged for sentiment polarity. The target outputs are $($text, neutral\_sentiment$)$ pairs.

\subsubsection{WordNet} Although we have collected the most frequently used words, some of them might be redundant. For example, ``love" and ``loves" have the same meaning as a verb. Therefore, for each word in each part of speech, we conduct lemmatization and keep the lemma only. This will result in a shorter word list. Furthermore, we remove stop words and words that have fewer than 3 characters. This results in a word list consisting of 51,941 words. Given a word, we search for all meanings of the word and retrieve one example sentence in each meaning. 

\noindent\textbf{Part-of-speech}
WordNet defines ``part-of-speech'' (POS) as either noun, verb, adjective, or adverb. Since the POS of a word in a certain meaning is known, we can easily get $($word, sentence, POS$)$ triples as our target outputs.

\noindent\textbf{Meaning} The target outputs are $($word, sentence, meaning$)$ triples.

\noindent\textbf{Synonym \& Antonym} For each word in each meaning, WordNet provides synonyms and antonyms for that word. We can thus get $($word, sentence, synonym$)$ triples and $($word, example sentence, antonym$)$ triples as target outputs.

\subsubsection{Question Answering}
We restrict each reading passage (i.e., context) to be no longer than 400 words. For the TriviaQA dataset, we only use evidence obtained using web search. We only keep questions that end with a ``?" mark or ``:" mark, or questions that are cloze questions (e.g., some questions in RACE). The target outputs for question answering signals are $($context, question, answer$)$ triples.

\subsubsection{arXiv}
For arXiv papers, we are interested in the following elements: (i) top-level category, (ii) title, (iii) abstract. 

\noindent\textbf{Category} We only keep the papers that belong to only one of the eight top-level categories. The target outputs are $($abstract, category$)$ pairs.

\noindent\textbf{Summary} For summarization, we consider generating a title from the abstract. For each paper, we ignore it if (i) its title has fewer than three words or (ii) its abstract has fewer than 20 words, or (iii) its abstract is shorter than its title. We truncate the abstract to 400 words. The target outputs are $($abstract, title$)$ pairs.

\subsubsection{Papers With Code}
Papers With Code offers the following four categories of entities: datasets, tasks, methods, and evaluation metrics. We filter out items with fewer than three characters or longer than six words. For items extracted from the AI terminology database,\footnote{\url{https://github.com/jiqizhixin/Artificial-Intelligence-Terminology-Database}} we keep the items that are longer than two words and categorize them into the ``other" entity type. For each item, we keep both the full name and short abbreviation. Now, we have a list of entities and the corresponding categories for each entity. For each paper, we consider sentences from its title, abstract and introduction.

\noindent\textbf{Entity} We use exact match to mine all the entities within a sentence. We consider all text spans within a sentence that have a length shorter or equal to four words. When comparing an entity with a text span, we normalize both using the following rules. (i) If the item has only one word, remove all characters that are not letters or numbers. (ii) If the item contains more than one word, after removal of all characters that are not letters or numbers, we lowercase the whole item. (iii) If different lengths of text span starting from a certain position can match items in our entity list, we choose the longest text span as an entity. The target outputs are $($sentence, entities$)$ pairs.

\noindent\textbf{Entity typing}
For each entity within a sentence, the type of the entity is known according to our entity list, so we can collect $($sentence, entity, entity\_type$)$ triples as our target outputs.

\noindent\textbf{Summary} 
We consider using the introduction of a paper to generate its abstract. We truncate the abstract and introduction to 400 words and filter samples with an introduction shorter than the abstract. The target outputs are $($introduction, abstract$)$ pairs.

\subsubsection{Other Data Processing}

We normalize multiple tab characters, newline characters, and space characters into a single space character. We ignore characters that are not ASCII characters. We use a specified language detector \citep{nakatani2010langdetect} to detect the language of a piece of text and keep the text if it is English with a probability greater than 0.9999.

\subsection{Tooling}
We provide our collected signals through DataLab \citep{xiao-etal-2022-datalab}, a convenient tool for data loading and processing. Signal tuples can be easily accessed through the code snippet \texttt{load\_dataset("rst", \{signal\_type\})}. The statistics are shown in Tab.\ref{tab:signal_stat}, and we hope this would be a useful resource for researchers.

\begin{table}[!t]
\footnotesize
\renewcommand{\arraystretch}{1.1}
\caption{\label{tab:signal_stat}Statistics of our collected signals.}
\begin{tabular}{p{0.11\textwidth}p{0.17\textwidth}>{\raggedleft\arraybackslash}p{0.1\textwidth}p{0.32\textwidth}p{0.17\textwidth}}
\toprule
Mine            & Signal Tuple                                        & \multicolumn{1}{l}{\#Sample} & Use in DataLab                                       & Some Applications                           \\
\midrule
Rotten Tomatoes & (review,rating)                                    & 5,311,109                    & \texttt{load\_dataset("rst", "rotten\_tomatoes\_sentiment")}  & Sentiment classification                 \\
Daily Mail      & (text,category)                                    & 899,904                      & \texttt{load\_dataset("rst", "daily\_mail\_category")}        & Topic classification                      \\
Daily Mail      & (title,text,summary)                              & 1,026,616                    & \texttt{load\_dataset("rst", "daily\_mail\_summary")}         & Summarization, Sentence expansion        \\
Daily Mail      & (text,events)                                      & 1,006,412                    & \texttt{load\_dataset("rst", "daily\_mail\_temporal")}       & Temporal reasoning                         \\
Wikidata        & (text,entity,entity\_type)                        & 2,214,274                    & \texttt{load\_dataset("rst", "wikidata\_entity")}             & Entity typing                             \\
Wikidata        & (subject,object,relation, text)                   & 1,526,674                    & \texttt{load\_dataset("rst", "wikidata\_relation")}           & Relation extraction, Fact retrieval        \\
wikiHow         & (text,category)                                    & 112,109                      & \texttt{load\_dataset("rst", "wikihow\_text\_category")}      & Topic classification                \\
wikiHow         & (low\_category, high\_category)                     & 4,868                        & \texttt{load\_dataset("rst", "wikihow\_category\_hierarchy")} & Relation extraction   \\
wikiHow         & (goal,steps)                                       & 47,956                       & \texttt{load\_dataset("rst", "wikihow\_goal\_step")}          & Intent detection                            \\
wikiHow         & (text,summary)                                     & 703,278                      & \texttt{load\_dataset("rst", "wikihow\_summary")}             & Summarization, Sentence expansion          \\
wikiHow         & (first\_step,second\_step, goal)                   & 47,787                       & \texttt{load\_dataset("rst", "wikihow\_procedure")}           & Temporal reasoning                         \\
wikiHow         & (question,description, related\_questions, answer) & 47,705                       & \texttt{load\_dataset("rst", "wikihow\_question")}            & Question generation                        \\
Wikipedia       & (text,entities)                                    & 22,231,011                   & \texttt{load\_dataset("rst", "wikipedia\_entities")}          & Entity recognition                         \\
Wikipedia       & (texts,titles)                                     & 3,296,225                    & \texttt{load\_dataset("rst", "wikipedia\_sections")}          & Summarization, Topic classification        \\
WordNet         & (word,sentence,pos)                               & 27,123                       & \texttt{load\_dataset("rst", "wordnet\_pos")}                 & Part-of-speech tagging                     \\
WordNet         & (word,sentence,meaning, possible\_meanings)       & 27123                        & \texttt{load\_dataset("rst", "wordnet\_meaning")}             & Word sense disambiguation                  \\
WordNet         & (word,sentence,synonyms)                          & 17,804                       & \texttt{load\_dataset("rst", "wordnet\_synonym")}             & Paraphrasing                               \\
WordNet         & (word,sentence,antonyms)                          & 6,408                        & \texttt{load\_dataset("rst", "wordnet\_antonym")}             & Negation                                   \\
ConTRoL         & (premise,hypothesis,label)                        & 8,323                        & \texttt{load\_dataset("rst", "qa\_control")}                  & Natural language inference                 \\
DREAM           & (context,question,options, answer)                & 9,164                        & \texttt{load\_dataset("rst", "qa\_dream")}                    & Reading comprehension                      \\
LogiQA          & (context,question,options, answer)                & 7,974                        & \texttt{load\_dataset("rst", "qa\_logiqa")}                   & Reading comprehension                      \\
ReClor          & (context,question,options, answer)                & 5,138                        & \texttt{load\_dataset("rst", "qa\_reclor")}                   & Reading comprehension                      \\
RACE            & (context,question,options, answer)                & 44,880                       & \texttt{load\_dataset("rst", "qa\_race")}                     & Reading comprehension                      \\
RACE-C          & (context,question,options, answer)                & 5,093                        & \texttt{load\_dataset("rst", "qa\_race\_c")}                  & Reading comprehension                       \\
TriviaQA        & (context,question,answer)                         & 46,636                       & \texttt{load\_dataset("rst", "qa\_triviaqa")}                 & Reading comprehension                      \\
Arxiv           & (text,category)                                    & 1,696,348                    & \texttt{load\_dataset("rst", "arxiv\_category")}              & Topic classification                       \\
Arxiv           & (text,summary)                                     & 1,696,348                    & \texttt{load\_dataset("rst", "arxiv\_summary")}               & Summarization, Sentence expansion          \\
Paperswithcode  & (text,entities,datasets, methods,tasks,metrics) & 4,731,233                    & \texttt{load\_dataset("rst", "paperswithcode\_entity")}       & Entity recognition                         \\
Paperswithcode  & (text,summary)                                     & 120,924                      & \texttt{load\_dataset("rst", "paperswithcode\_summary")}      & Summarization, Sentence expansion          \\ 
\bottomrule
\end{tabular}
\end{table}

\subsection{Signal reStructuring}
\label{signal_restructure}
After we extract the different signals from the various data mines, the next important step is to unify them into a fixed form so that rich information can be stored together in the model during pre-training.
This non-trivial goal became achievable since the advent of prompt technology \citep{brown2020language,liu2021pretrain}, which,
in principle, could unify almost all types of signals into a language model-style form with the help of suitable prompt designing. 

We divide the signals into two main categories: generic signals (cloze signal in plain text in \S\ref{sec:signal-and-downstream-tasks}) and task-relevant signals (all other signals in \S\ref{sec:signal-and-downstream-tasks}). The former contains basic linguistic knowledge and can benefit all downstream tasks to some extent, while the latter would benefit some specific downstream tasks.

\paragraph{Generic signals} With $($corrupted\_text, corrupted\_positions, target\_spans$)$ triples, we construct the following prompt:\\
\begin{itemize}
    \item \textcolor{red}{source}: \{corrupted\_text\}\\
\textcolor{blue}{target}: \{corrupted\_position1\}\{target\_span1\}\{corrupted\_position2\}\{target\_span2\}...
\end{itemize}
To give a concrete example, suppose we have $($Thank you $<$X$>$ me to your party $<$Y$>$ week., $<$X$>$ $\vert$ $<$Y$>$, for inviting $\vert$ last$)$, the prompted source would be ``Thank you $<$X$>$ me to your party $<$Y$>$ week." and the prompted target would be ``$<$X$>$ for inviting $<$Y$>$ last $<$Z$>$"

\paragraph{Task-relevant signals}
We design the following two forms of prompts for all other types of signals.
\begin{itemize}
    \item multiple-choice format
    \item generation format
\end{itemize}
 For a multiple-choice format prompt, we bind the available options to the end of it while for a generation format prompt, we do not give such hint. To give a concrete example, a multiple-choice format prompt for the sentiment classification task could be the following: \texttt{I like this movie. Is this text ``positive" or ``negative"?} while a generation format prompt could be the following: \texttt{I like this movie. What's the sentiment of the previous text?}. We use two special markers: ``\texttt{TEXT}:" and ``\texttt{QUERY}:" to separate the general context and the intended task to be completed. For each type of signal, we construct multiple prompts so that the model can learn various query forms. We design a total of 1124 prompts for the 30 signals in \S\ref{sec:signal-and-downstream-tasks} (except the cloze signal in plain text), with an average of 37 prompts per signal. The prompts for different signals are listed in \S\ref{app:prompts}.

\subsection{Pre-training \& Fine-tuning} \label{subsection:pretrain}

Given the entirety of all the available data, suppose that we can extract $K$ types of signals in total, then we have
\begin{align}
     \{\text{S}_1, \text{S}_2, \cdots, \text{S}_K\} =   \text{Extract}(\text{Data})
     \label{eq:extract_signal}
\end{align}
$\text{Extract}(\cdot)$ represents the extraction function of signals.
$\text{S}_k = s_1^k, s_2^k, \cdots, s_{N_k}^k$ represents all ($N_k$) signal tuples that could provide supervision from the $k$th type.

Suppose for the $k$th signal type, we have $M_k$ restructuring functions in total, each restructuring function would map a signal tuple to an \textit{input-output} pair as shown in the equations below
\begin{align}
    r_{M_k}^k(s_{N_k}^k) &= (x_{N_kM_k}^k, y_{N_kM_k}^k) \label{eq:input-output}
\end{align}
where $r_{M_k}^k$ represents the $M_k$th restructuring function for the $k$th signal type.

After the restructuring process, each signal will be reformatted as follows: Eq.~\ref{eq:restructure_signal}
\begin{align}
\label{eq:restructure_signal}
    \tilde{S}_k = (x_{11}^k, y_{11}^k), \cdots, (x_{1M_k}^k, y_{1M_k}^k), \cdots, (x_{N1}^k, y_{N1}^k), \cdots,  (x_{N_kM_k}^k, y_{N_kM_k}^k).
\end{align}

To give a concrete example, suppose we have a triple $($Mozart, was born
in, Salzburg$)$ from the relation signal. We can restructure it into different textual input-output pairs such as $($Where was Mozart born?, Salzburg$)$ and $($Who was born in Salzburg?, Mozart$)$ by applying different restructuring functions.

Then the pre-training or fine-tuning process is typically associated with solving the following optimization problem:
\begin{align}
\label{eq:optimization}
    \min_{\theta}\frac{1}{K}\sum_{k=1}^K\frac{1}{M_k}\frac{1}{N_k}\sum_{j=1}^{M_k}\sum_{i=1}^{N_k}\lambda_{kj} \cdot l(f_{\theta}(x_{ij}^k), y_{ij}^k)
\end{align}

\noindent where $r_j^k(s_i^k) = {(x_{ij}^k, y_{ij}^k)}$ represents the input-output pair obtained by applying the $j$th restructuring function of signal $k$ to the $i$th tuple of signal $s$. $f(\cdot)$ could either be a pre-trained model or fine-tuned model parameterized by $\theta$. $\lambda$ denotes a signal type-dependent (or even restructuring function-dependent) weight that could be specified beforehand. $l$ is the loss function that measures the distance between predicted target and true target.

The advantage of restructuring learning can also be reflected in the above formal description: when we use a pre-trained model $f(\cdot)$ for a specific downstream task, we no longer need to bother with which prompts to use because we just need to use $r_j^k$. In other words, the knowledge stored in the pre-training model has been structured by $r_j^k$ so that we can find the information required by downstream tasks more accurately with the handle $r_j^k$.

\clearpage

\section{Experiment on 55 Popular NLP Datasets} \label{sec:exp_setup_nlp_tasks}

\subsection{Tasks and Datasets}
\label{sec:tasks_and_datasets}
Evaluation, as an important part of the entire ML system development lifecycle, is often underestimated. For example, the selection process of tasks and datasets has not been systematically designed.
In this paper, we aim to access the proposed paradigm from different levels for a thorough analysis.
To this end, we chose \textbf{9} different categories of NLP tasks, not only including the text classification and matching tasks but also covering structure prediction (e.g., relation extraction, named entity recognition (NER)), intention prediction, fact retrieval, reasoning, and generation tasks, \textbf{55}
datasets in total. 
Below, we will detail the dataset information w.r.t each task category.

\paragraph{Topic classification}
aims to assign a piece of text to one of the predefined topics. We consider the following evaluation datasets: 
subj \cite{pang-lee-2004-sentimental}, qc \cite{li-roth-2002-learning}, yahoo\_answers\_topics \cite{DBLP:conf/nips/ZhangZL15}, hate\_speech18 \cite{gibert2018hate}, tweet\_eval/emotion \cite{mohammad2018semeval}, tweet\_eval/hate \cite{basile-etal-2019-semeval}, tweet\_eval/irony \cite{van2018semeval}, and tweet\_eval/offensive \cite{mohammad2016semeval}.

\paragraph{Sentiment classification}
aims to identify the sentiment polarity (e.g., positive, negative) conveyed in a text. We consider the following datasets for evaluation: financial\_phrasebank \cite{Malo2014GoodDO}, mr \cite{pang-lee-2005-seeing}, and sst2 \cite{socher-etal-2013-recursive}.

\paragraph{Information extraction} is the task of extracting structured information automatically from unstructured data.
We consider the following datasets for evaluation: conll03 \cite{tjong-kim-sang-de-meulder-2003-introduction}, OntoNotes 5.0\footnote{\url{https://catalog.ldc.upenn.edu/LDC2013T19}} (notebc, notebn, notemz, notenw, notewb, notetc), wikiann \cite{pan-etal-2017-cross}, wnut17 \cite{derczynski-etal-2017-results}, semeval\_rel \cite{hendrickx-etal-2010-semeval}, and wiki80 \cite{han-etal-2019-opennre}. For named entity recognition tasks, we develop a novel two-step approach. We first identify all the entities contained in the text and then determine the entity type for each entity.

\paragraph{Natural language inference} 
aims to determine the relationship between two given texts (e.g, entailment or contradiction). We consider the following datasets for evaluation: anli-r1 \cite{nie2019adversarial}, anli-r2 \cite{nie2019adversarial}, anli-r3 \cite{nie2019adversarial}, cb \cite{Marneffe2019TheCI}, multi\_nli\_matched \cite{N18-1101}, multi\_nli\_mismatched \cite{N18-1101}, rte \cite{DBLP:conf/iclr/WangSMHLB19}, sick \cite{marelli-etal-2014-sick}, and snli \cite{snli:emnlp2015}.

\paragraph{Intent detection}
aims to recognize the intent (e.g. booking a flight) contained in a piece of speech or text. We consider the following datasets for evaluation: atis \cite{hemphill-etal-1990-atis}, banking77 \cite{Casanueva2020}, clinc150 \cite{larson-etal-2019-evaluation}, fb \cite{schuster-etal-2019-cross-lingual}, hint3-curekart \cite{arora-etal-2020-hint3}, hint3-powerplay11 \cite{arora-etal-2020-hint3}, hint3-sofmattress \cite{arora-etal-2020-hint3}, nlued \cite{XLiu.etal:IWSDS2019}, slurp \cite{slurp}, and snips \cite{https://doi.org/10.48550/arxiv.1805.10190}.

\paragraph{Fact retrieval}
aims to extract different types of factual knowledge (e.g., Beethoven's birth year) from the PLMs. We consider the following dataset for evaluation: LAMA-TREx \cite{DBLP:conf/emnlp/PetroniRRLBWM19}.

\paragraph{Temporal reasoning}
aims to determine the sequence of events and the duration. We consider the following datasets for evaluation: TRACIE \cite{ZRNKSR21} and UDS-T temporal ordering \cite{vashishtha-etal-2020-temporal}. 

\paragraph{Word sense disambiguation}
aims to distinguish different meanings of the same word in different contexts. We consider the following datasets for evaluation: semeval2007 \cite{pradhan-etal-2007-semeval}, semeval2013 \cite{navigli-etal-2013-semeval}, semeval2015 \cite{moro-navigli-2015-semeval}, and senseval2 \cite{edmonds-cotton-2001-senseval}.

\paragraph{Summarization}
aims to summarize a long article or articles into  a few short sentences. We consider the following datasets for evaluation: SciTLDR \cite{cachola-etal-2020-tldr}, Reddit-TIFU \cite{kim-etal-2019-abstractive}, Multi-Xscience \cite{lu-etal-2020-multi-xscience}, WikiSum \cite{cohen-etal-2021-wikisum}, GovReport \cite{huang-etal-2021-efficient}, BillSum \cite{kornilova-eidelman-2019-billsum}, and BigPatent \cite{sharma-etal-2019-bigpatent}.

\subsection{Data Mines and Signals}
We consider 10 data mines that are highly popular in the NLP community, as described in \S\ref{sec:data_mines}. 
For each data mine, we extract 1-6 types of signals, as shown in Fig.~\ref{fig:signal_tree}. We transform these signals into training data through prompting to pre-train the model, as mentioned in \S\ref{signal_restructure}.

\begin{table}[!ht] 
\footnotesize
  \setlength\tabcolsep{3.6pt}
  \renewcommand{\arraystretch}{1.2}
  \caption{Training signals for different tasks (part1). ``Num." denotes the number of signal tuples. For each individual type of signal, the signal tuples are selected randomly.
  }
  \label{tab:task_signal_part1}
\begin{tabular}{lllrl}
\toprule
\textbf{Task}                                                                             & \textbf{Signals}                              & \textbf{Prompt Type}             & \multicolumn{1}{l}{\textbf{Num.}} & \textbf{Notes}                                                                                                           \\
\midrule
\multirow{4}{*}{\begin{tabular}[c]{@{}l@{}}Topic \\ Classification\end{tabular}}          & DailyMail category                            & Multiple-choice                  & 30,000                           & Take at most 5,000 for each category                                                                                     \\
                                                                                          & arXiv category                                & Multiple-choice                  & 36,185                           & Take at most 5,000 for each category                                                                                     \\
                                                                                          & wikiHow text category                         & Multiple-choice                  & 67,728                           & Take at most 5,000 for each category                                                                                     \\
                                                                                          & Wikipedia section title                       & Multiple-choice                  & 100,000                          &                                                                                                                          \\
                                                                                          \midrule
\multirow{2}{*}{\begin{tabular}[c]{@{}l@{}}Sentiment \\ Classification\end{tabular}}      & Rotten Tomatoes sentiment                     & Multiple-choice                  & 100,000                          & 50,000 positive, 50,000 negative                                                                                         \\
                                                                                          & Wikipedia sentiment                           & Multiple-choice                  & 50,000                           &                                                                                                                          \\
                                                \midrule
\multirow{6}{*}{\begin{tabular}[c]{@{}l@{}}Information \\ Extraction\end{tabular}}        & Paperswithcode entity                         & Generation                       & 70,000                           & 50,000 with entities, 20,000 without entities                                                                            \\
                                                                                          & \multirow{2}{*}{Paperswithcode entity typing} & \multirow{2}{*}{Multiple-choice} & \multirow{2}{*}{25,000}          & \multirow{2}{*}{\begin{tabular}[c]{@{}l@{}}Take at most 5,000 for each entity type \\ (including ``other")\end{tabular}} \\
                                                                                          &                                               &                                  &                                  &                                                                                                                          \\
                                                                                          & Wikidata entity typing                        & Multiple-choice                  & 50,000                           &                                                                                                                          \\
                                                                                          & Wikidata relation                             & Multiple-choice                  & 100,000                          &                                                                                                                          \\
                                                                                          & Wikipedia entity                              & Generation                       & 150,000                          & 100,000 with entities, 50,000 without entities                                                                           \\
                                                                                          \midrule
\multirow{6}{*}{\begin{tabular}[c]{@{}l@{}}Natural \\ Language \\ Inference\end{tabular}} & ConTRoL                                       & Multiple-choice                  & 8,323                            &                                                                                                                          \\
                                                                                          & DREAM                                         & Multiple-choice                  & 9,164                            &                                                                                                                          \\
                                                                                          & LogiQA                                        & Multiple-choice                  & 7,974                            &                                                                                                                          \\
                                                                                          & RACE \& RACE-C                                & Multiple-choice                  & 106,602                          &                                                                                                                          \\
                                                                                          & ReClor                                        & Multiple-choice                  & 5,138                            &                                                                                                                          \\
                                                                                          & DailyMail temporal information                & Multiple-choice                  & 40,000                           &                                                                                       \\
                                                                                          \midrule
\begin{tabular}[c]{@{}l@{}}Intent \\ Detection\end{tabular}                               & wikiHow goal-step relation                    & Multiple-choice                  & 95,912                           & Use both goal-to-step and step-to-goal                                                                                   \\
\midrule
\multirow{9}{*}{Fact Retrieval}                                                           & WordNet meaning                               & Generation                       & 17,690                           &                                                                                                                          \\
                                                                                          & WordNet part-of-speech                        & Generation                       & 27,123                           &                                                                                                                          \\
                                                                                          & WordNet synonym                               & Generation                       & 33,496                           &                                                                                                                          \\
                                                                                          & WordNet antonym                               & Generation                       & 3,872                            &                                                                                                                          \\
                                                                                          & wikiHow category hierarchy                    & Generation                       & 4,868                            &                                                                                                                          \\
                                                                                          & Wikidata relation                             & Generation                       & 300,000                          &                                                                                                                          \\
                                                                                          & Wikidata entity typing                        & Generation                       & 50,000                           &                                                                                                                          \\
                                                                                          & \multirow{2}{*}{Paperswithcode entity typing} & \multirow{2}{*}{Generation}      & \multirow{2}{*}{20,000}          & \multirow{2}{*}{\begin{tabular}[c]{@{}l@{}}Take at most 5,000 for each entity type \\ (excluding ``other")\end{tabular}} \\
                                                                                          &                                               &                                  &                                  &                                                                                                                          \\
                                                                                          \midrule
\multirow{2}{*}{\begin{tabular}[c]{@{}l@{}}Temporal \\ Reasoning\end{tabular}}                             & DailyMail temporal information                & Multiple-choice                  & 100,000                          &                                                                                                                          \\
                                                                                          & wikiHow procedure                             & Multiple-choice                  & 100,000                          &                                                                                                                          \\
                                                                                          \midrule
\multirow{4}{*}{\begin{tabular}[c]{@{}l@{}}Word Sense \\ Disambiguation\end{tabular}}     & WordNet meaning                               & Generation                       & 17,690                           &                                                                                                                          \\
                                                                                          & WordNet part-of-speech                        & Generation                       & 27,123                           &                                                                                                                          \\
                                                                                          & WordNet synonym                               & Generation                       & 33,496                           &                                                                                                                          \\
                                                                                          & WordNet antonym                               & Generation                       & 3,872                            &                                                                                                                          \\
                                                                                          \midrule
\multirow{4}{*}{Summarization}                                                            & DailyMail summary                       & Generation                       & 200,000                          &                                                                                                                          \\
                                                                                          & Paperswithcode summary                  & Generation                       & 200,000                          &                                                                                                                          \\
                                                                                          & arXiv summary                           & Generation                       & 200,000                          &                                                                                                                          \\
                                                                                          & wikiHow summary                         & Generation                       & 200,000                          &                                                                                                                          \\
                                                                                          \midrule
                                                                                         
\multirow{9}{*}{All Tasks}                                                                        & Rotten Tomatoes                               & Multiple-choice                  & 100,000                          & 50,000 positive, 50,000 negative                                                                                         \\
                                                                                          & DailyMail category                            & Multiple-choice                  & 30,000                           & Take at most 5,000 for each category                                                                                     \\
                                                                                          & DailyMail summary                       & Multiple-choice                  & 50,000                           &                                                                                                                          \\
                                                                                          & DailyMail temporal information                & Multiple-choice                  & 100,000                          &                                                                                                                          \\
                                                                                          & Paperswithcode entity                         & Multiple-choice                  & 70,000                           & 50,000 with entities, 20,000 without entities                                                                            \\
                                                                                          & \multirow{2}{*}{Paperswithcode entity typing} & \multirow{2}{*}{Multiple-choice} & \multirow{2}{*}{25,000}          & \multirow{2}{*}{\begin{tabular}[c]{@{}l@{}}Take at most 5,000 for each entity type \\ (including ``other")\end{tabular}} \\
                                                                                          &                                               &                                  &                                  &                                                                                                                          \\
                                                                                          & arXiv category                                & Multiple-choice                  & 36,185                           & Take at most 5,000 for each category                                                                                     \\
                                                                                          & arXiv summary                           & Multiple-choice                  & 50,000                           &                                                                                                                          \\
                                                                               
                                                                            \bottomrule
\end{tabular}
\end{table}

\begin{table}[!ht] 
\footnotesize
  \setlength\tabcolsep{1.6pt}
  \renewcommand{\arraystretch}{1.2}
  \caption{Signals for different tasks (part2). ``Num." denotes the number of signal tuples. For each individual type of signal, the signal tuples are selected randomly.}
  \label{tab:task_signal_part2}
\begin{tabular}{lllrl}
\toprule
\textbf{Task}                                                                             & \textbf{Signals}                              & \textbf{Prompt Type}             & \multicolumn{1}{l}{\textbf{Num.}} & \textbf{Notes}                                                                                                           \\
\midrule             

\multirow{50}{*}{All Tasks}               & Wikidata entity                               & Multiple-choice                  & 100,000                          &                                                                                                                          \\
                                                                                          & Wikidata relation                             & Multiple-choice                  & 200,000                          &                                                                                                                          \\
                                                                                          & wikiHow text category                         & Multiple-choice                  & 67,728                           & Take at most 5,000 for each category                                                                                     \\
                                                                                          & wikiHow category hierarchy                    & Multiple-choice                  & 4,868                            &                                                                                                                          \\
                                                                                          & wikiHow goal-step relation                    & Multiple-choice                  & 95,912                           &                                                                                                                          \\
                                                                                          & wikiHow summary                         & Multiple-choice                  & 50,000                           &                                                                                                                          \\
                                                                                          & wikiHow procedure                             & Multiple-choice                  & 100,000                          &                                                                                                                          \\
                                                                                          & wikiHow question generation                   & Multiple-choice                  & 150,000                          &                                                                                                                          \\
                                                                                          & Wikipedia section title                       & Multiple-choice                  & 100,000                           &                                                                                                                          \\
                                                                                          & Wikipedia entity                              & Multiple-choice                  & 70,000                           & 50,000 with entities, 20,000 without entities                                                                            \\
                                                                                          & Wikipedia sentiment                           & Multiple-choice                  & 50,000                           &                                                                                                                          \\
                                                                                          & WordNet meaning                               & Multiple-choice                  & 17,690                           &                                                                                                                          \\
                                                                                          & WordNet part-of-speech                        & Multiple-choice                  & 27,123                           &                                                                                                                          \\
                                                                                          & WordNet synonym                               & Multiple-choice                  & 33,496                           &                                                                                                                          \\
                                                                                          & WordNet antonym                               & Multiple-choice                  & 3,872                            &                                                                                                                          \\
                                                                                          & ConTRoL                                       & Multiple-choice                  & 8,323                            &                                                                                                                          \\
                                                                                          & DREAM                                         & Multiple-choice                  & 9,164                            &                                                                                                                          \\
                                                                                          & LogiQA                                        & Multiple-choice                  & 7,974                            &                                                                                                                          \\
                                                                                          & RACE \& RACE-C                                & Multiple-choice                  & 106,602                          &                                                                                                                          \\
                                                                                          & ReClor                                        & Multiple-choice                  & 5,138                            &                                                                                                                          \\
                                                                                          & Rotten Tomatoes                               & Generation                       & 100,000                          & 50,000 positive, 50,000 negative                                                                                         \\
                                                                                          & DailyMail category                            & Generation                       & 30,000                           & Take at most 5,000 for each category                                                                                     \\
                                                                                          & DailyMail summary (sentence expansion)                  & Generation                       & 90,000                           &                                                                                                                          \\
                                                                                          & DailyMail summary                       & Generation                       & 400,000                          &                                                                                                                          \\
                                                                                          & DailyMail temporal information                & Generation                       & 50,000                           &                                                                                                                          \\
                                                                                          & Paperswithcode entity                         & Generation                       & 70,000                           & 50,000 with entities, 20,000 without entities                                                                            \\
                                                                                          & \multirow{2}{*}{Paperswithcode entity typing} & \multirow{2}{*}{Generation}      & \multirow{2}{*}{20,000}          & \multirow{2}{*}{\begin{tabular}[c]{@{}l@{}}Take at most 5,000 for each entity type \\ (excluding ``other")\end{tabular}} \\
                                                                                          &                                               &                                  &                                  &                                                                                                                          \\
                                                                                          & Paperswithcode summary                  & Generation                       & 100,000                          &                                                                                                                          \\
                                                                                          & arXiv category                                & Generation                       & 36,185                           & Take at most 5,000 for each category                                                                                     \\
                                                                                          & arXiv summary                           & Generation                       & 100,000                          &                                                                                                                          \\
                                                                                          & Wikidata entity typing                        & Generation                       & 100,000                          &                                                                                                                          \\
                                                                                          & Wikidata relation                             & Generation                       & 200,000                          &                                                                                                                          \\
                                                                                          & wikiHow text category                         & Generation                       & 67,728                           & Take at most 5,000 for each category                                                                                     \\
                                                                                          & wikiHow category hierarchy                    & Generation                       & 4,868                            &                                                                                                                          \\
                                                                                          & wikiHow goal-step relation                    & Generation                       & 95,912                           &                                                                                                                          \\
                                                                                          & wikiHow summary                         & Generation                       & 50,000                           &                                                                                                                          \\
                                                                                          & wikiHow summary (sentence expansion)                    & Generation                       & 50,000                           &                                                                                                                          \\
                                                                                          & wikiHow procedure                             & Generation                       & 50,000                           &                                                                                                                          \\
                                                                                          & wikiHow question generation                   & Generation                       & 150,000                          &                                                                                                                          \\
                                                                                          & Wikipedia section title                       & Generation                       & 50,000                           &                                                                                                                          \\
                                                                                          & Wikipedia entity                              & Generation                       & 200,000                          & 150,000 with entities, 50,000 without entities                                                                           \\
                                                                                          & Wikipedia sentiment                           & Generation                       & 50,000                           &                                                                                                                          \\
                                                                                          & WordNet meaning                               & Generation                       & 27,123                            &                                                                                                                          \\
                                                                                          & WordNet part-of-speech                        & Generation                       & 27,123                            &                                                                                                                          \\
                                                                                          & WordNet synonym                               & Generation                       & 33,496                            &                                                                                                                          \\
                                                                                          & WordNet antonym                               & Generation                       & 3,872                             &                                                                                                                          \\
                                                                                          & ConTRoL                                       & Generation                       & 8,323                             &                                                                                                                          \\
                                                                                          & DREAM                                         & Generation                       & 9,164                             &                                                                                                                          \\
                                                                                          & TriviaQA                                      & Generation                       & 37,811                            &                                            \\                                                                 
\bottomrule
\end{tabular}
\end{table}




\subsection{Model Setups} 
\label{subsec:model_setups}
\paragraph{reStructured Models}
We explore two kinds of restructuring ways based on different choices of $\lambda$ in Eq.~\ref{eq:optimization}.
\begin{enumerate}
    \item \textbf{Generalist} 
    (RST-All) that uses all signals we collected in \S\ref{sec:data_mines}. For this model, we set $\lambda$ to be a constant for all signal tuples. Details are shown in Tab.~\ref{tab:task_signal_part1}-\ref{tab:task_signal_part2}. 
    \item \textbf{Specialist} (RST-Task) that instantiates $\lambda$ in a heuristic way. For each task, we set the weights of the signal tuples that we consider beneficial to it to be constant and the weights of the other signal tuples to be 0. This is equivalent to considering only the specific signal tuples when optimizing Eq.~\ref{eq:optimization}. The signals we consider for each task are listed in Tab.~\ref{tab:task_signal_part1}.
\end{enumerate}


\paragraph{T0pp} 
T0 \cite{sanh2021multitask} is a series of models that use multitask supervised datasets for fine-tuning T5. T0pp is the model with the largest number of supervised datasets (55 in total) used for fine-tuning and is the most performant model in the T0 series.


\paragraph{GPT3}\citep{brown2020language} is an autoregressive PLM with 175 billion parameters and has shown strong few-shot generalization ability across different tasks with in-context examples.


\subsection{Training}
\label{sec:mTain}
We adopt an encoder-decoder framework as the pre-training backbone since it can naturally support prompted input-output pairs as described in Eq.\ref{eq:input-output}.
Additionally, to reduce the cost of pre-training and instruct models to memorize knowledge in a curriculum way, we schedule the pre-training order of different signals and make the S17 (cloze) signal trained first. By doing this, we can utilize the checkpoints of existing PLMs and make restructured learning more widely applicable.
Based on the computational resources we own and open available PLMs, we finally choose T5-XXL \cite{DBLP:journals/corr/abs-1910-10683} to instantiate our restructuring framework.\footnote{When we perform experiments in this work, this was one of the largest public models. Although more than a dozen times smaller than GPT3, the results of this work are much better, raising the prospect of using larger models in the future.}

Since we design multiple prompts for each signal type, we randomly select one of these prompts for each signal tuple when performing signal restructuring.
During training, we use a batch size of 4 and a learning rate of 0.001. Following \citet{DBLP:journals/corr/abs-1910-10683}, we use the Adafactor \cite{DBLP:conf/icml/ShazeerS18} optimizer and save a checkpoint model every 10,000 steps. The input sequence length and output sequence length are set to 1024 and 256 tokens, respectively. All the models are trained using Google Cloud TPU v3-8. 

For the information extraction model, there are training data using multiple-choice format prompts as well as training data that use generation format prompts. According to our empirical observation, we found that first training on multiple-choice format prompts then training on both multiple-choice format prompts and generation format prompts can result in reasonable performance in both multiple-choice format tasks (e.g., assign a category to an entity) and generation format tasks (e.g., generate the entities within a sentence). However, directly training on both multiple-choice and generation format prompts will significantly sacrifice the performance of classification format tasks. Our solution can be seen as a kind of curriculum learning \cite{DBLP:conf/icml/BengioLCW09}, where we first train the model to perform easier tasks (i.e., classification tasks) and then expose more difficult tasks to the model (i.e., generation tasks). We also use this strategy when training the model with all the available signals. The best checkpoints are chosen based on the performance on validation sets. Details are included in Appendix \S\ref{app:training-details}.

\subsection{Evaluation}
For evaluation, we use the standard performance measurements (e.g., accuracy, f1) for each dataset in \S\ref{sec:tasks_and_datasets}. Please note that we do not evaluate models on datasets that T0pp has fine-tuned with downstream task samples for a fair comparison. As a reference, T0pp uses a total of 55 supervised datasets, while we use only 7 supervised datasets.

Due to our computation budgets, we only sample 5,000 samples for each classification test set if it contains more than 5,000 samples. For generation test sets, we sample 2,000 samples for each one if it contains more than 2,000 samples. For a dataset without a test split, we use the validation (or training if no validation split exists) split instead. For multiple-choice format tasks, we use a scoring method as follows. For each option, we compute the average log-likelihood of option tokens given the source input and take the one with the highest likelihood as the predicted option. For generation format tasks, we use beam search decoding \cite{DBLP:journals/corr/abs-1211-3711}. Note that T0pp's results are run by us using their released model checkpoint. For each dataset, we design five prompts to measure the model performance. Please refer to Appendix \S\ref{app:eval-prompts} for details.

\subsection{Results}
We evaluate RSTs on a total of 55 datasets and then compare them to GPT3 and T0pp, respectively.
Regarding {GPT3} (\S\ref{subsec:gpt3}): Due to the \textit{highly expensive costs of API usage}, we do not evaluate it on all 55 datasets, but instead, following \citet{sanh2021multitask}, use their reported results on the datasets overlapped with ours;
Regarding {T0pp} (\S\ref{subsec:t0pp}), we are able to evaluate it on all 55 datasets because they have released model checkpoints.

\begin{table}[!ht]
\centering
  \footnotesize
  \setlength\tabcolsep{4.5pt}
  \renewcommand{\arraystretch}{1.02}
  \caption{Results on typical NLP tasks. In the metric column, ``acc" is the abbreviation for ``accuracy" and ``r1" is the abbreviation for ROUGE-1. ``f1" means ``F1" score. The scores of RST models are underlined if they are better than those of the T0pp model. The best average score and maximum score for each dataset are bolded.}
  \label{tab:results}
\begin{tabular}{llcrrrrrrrrr}
\toprule
\multirow{2}{*}{}                                                                                  & \multicolumn{1}{c}{\multirow{2}{*}{\textbf{Dataset}}} &  \multicolumn{1}{c}{\multirow{2}{*}{\textbf{Metric}}} & \multicolumn{3}{c}{\textbf{T0pp}}                                                                  & \multicolumn{3}{c}{\textbf{RST-Task}}                                                                      & \multicolumn{3}{c}{\textbf{RST-All}}                                                                   \\
\cmidrule(lr){4-6} \cmidrule(lr){7-9} \cmidrule(lr){10-12}
                                                                                                   & \multicolumn{1}{c}{}                                  & &  \multicolumn{1}{c}{\textbf{avg}} & \multicolumn{1}{c}{\textbf{max}} & \multicolumn{1}{c}{\textbf{std}} & \multicolumn{1}{c}{\textbf{avg}} & \multicolumn{1}{c}{\textbf{max}} & \multicolumn{1}{c}{\textbf{std}} & \multicolumn{1}{c}{\textbf{avg}} & \multicolumn{1}{c}{\textbf{max}} & \multicolumn{1}{c}{\textbf{std}} \\
                                                                                                   \midrule
\multirow{8}{*}{\textbf{\begin{tabular}[c]{@{}l@{}}Topic \\ Classification\end{tabular}}}          & subj                 & acc          &     48.70                            & 58.00                            & 8.53                  & \underline{\textbf{60.24}}                            & \underline{\textbf{83.90}}                            & 14.16                                                   & \underline{54.52}                            & \underline{62.00}                            & 5.47                             \\
                                                                                 & qc      & acc              & 57.43                            & 66.06                            & 7.89                                   & \underline{\textbf{69.12}}                            & \underline{\textbf{71.69}}                            & 2.70                      & 37.95                            & 40.96                            & 2.57                             \\
                                                                                                   & yahoo\_answers\_topics  & acc   & 35.13                            & 41.10                            & 4.00                                 & \underline{\textbf{58.69}}                            & \underline{\textbf{59.40}}                            & 0.45                                                     & \underline{53.86}                            & \underline{54.56}                            & 0.58                             \\
                                                                                                   & hate\_speech18  & acc    & 79.64                            & 86.46                            & 9.20                                          & 74.78                            & 81.56                            & 4.43                                                  & \underline{\textbf{86.54}}                            & \underline{\textbf{86.80}}                            & 0.22                             \\
                                                                                                   & tweet\_eval/emotion & acc  & 70.54                            & 72.34                            & 1.93                                      & \underline{\textbf{80.99}}                            & \underline{\textbf{81.77}}                            & 0.69                                                    & 67.38                            & 69.74                            & 2.33                             \\
                                                                                                   & tweet\_eval/hate  & acc      & 57.95                            & 58.15                            & 0.16                                  & \underline{\textbf{62.45}}                            & \underline{\textbf{64.28}}                            & 2.12                                                      & \underline{62.44}                            & \underline{62.90}                            & 0.63                             \\
                                                                                                   & tweet\_eval/irony   & acc    & 61.63                            & 63.14                            & 1.35                                    & \underline{64.29}                            & \underline{\textbf{78.57}}                            & 8.56                                                    & \underline{\textbf{70.97}}                            & \underline{78.06}                            & 5.91                             \\
                                                                                                   & tweet\_eval/offensive & acc     & 72.07                            & 72.21                            & 0.13                                 & \underline{72.23}                            & \underline{\textbf{77.21}}                            & 4.23                                                     & \underline{\textbf{72.88}}                            & \underline{73.95}                            & 0.88                             \\
                                                                                                   \midrule
\multirow{3}{*}{\textbf{\begin{tabular}[c]{@{}l@{}}Sentiment \\ Classification\end{tabular}}}      & financial\_phrasebank       & acc  & 36.17                            & 36.44                            & 0.31     & \underline{60.75}                            & \underline{61.17}                            & 0.35                                                     & \underline{\textbf{61.33}}                            & \underline{\textbf{62.50}}                            & 0.67                             \\
                                                                                                   & mr     & acc           & 89.04                            & 92.50                            & 3.22                                            & \underline{\textbf{92.10}}                            & \underline{\textbf{92.59}}                            & 0.44                                                  & \underline{89.08}                            & 90.43                            & 1.24                             \\
                                                                                                   & sst2     & acc     & 92.32                            & \textbf{96.44}                            & 3.52                                             & \underline{\textbf{93.26}}                            & 93.58                            & 0.36                                                     & 89.20                            & 89.91                            & 0.88                             \\
                                                                                                   \midrule
\multirow{11}{*}{\textbf{\begin{tabular}[c]{@{}l@{}}Information \\ Extraction\end{tabular}}}       & conll03 & f1      & 17.87                            & 23.37                            & 4.35                                            & \underline{\textbf{54.72}}                            & \underline{\textbf{55.33}}                            & 0.64                                             & \underline{50.58}                            & \underline{52.55}                            & 1.40                             \\
                                                                                                   & notebc & f1        & 10.89                            & 13.83                            & 2.75                                                            & \underline{\textbf{28.51}}                            & \underline{\textbf{31.38}}                            & 1.72                                 & \underline{24.07}                            & \underline{25.73}                            & 1.41                             \\
                                                                                                   & notebn & f1         & 16.07                            & 19.94                            & 3.49               & \underline{\textbf{33.07}}                            & \underline{\textbf{34.72}}                            & 1.27                                & \underline{25.17}                            & \underline{25.66}                            & 0.49                             \\
                                                                                                   & notemz & f1      & 12.72                            & 17.78                            & 3.65                                                             & \underline{\textbf{30.36}}                     & \underline{\textbf{32.28}}                            & 2.87                                 & \underline{20.67}                            & \underline{21.85}                            & 1.04                             \\
                                                                                                   & notenw & f1       & 16.04                            & 18.54                            & 1.92                                                           & \underline{\textbf{28.53}}                            & \underline{\textbf{29.69}}                            & 1.47                                  & \underline{20.63}                            & \underline{20.99}                            & 0.25                             \\
                                                                                                   & notewb & f1        & 9.41                             & 11.75                            & 2.18                                                          & \underline{16.45}                            & \underline{\textbf{18.30}}                            & 1.05                                  & \underline{\textbf{17.24}}                            & \underline{18.12}                             & 0.54                             \\
                                                                                                   & notetc & f1      & \textbf{9.59}                             & \textbf{12.29}                           & 2.52                                                         & 8.06                             & 10.20                        & 1.20                                 & 3.58                             & 3.86                          & 0.42                             \\
                                                                                                   & wikiann      & f1    & 24.69                            & 31.66                            & 6.57                                           & \underline{\textbf{55.81}}                            & \underline{\textbf{57.20}}                            & 0.88                                                     & \underline{47.92}                            & \underline{48.60}                            & 0.48                             \\
                                                                                                   & wnut17     & f1      & 11.00                            & 14.09                            & 1.81                                           & \underline{22.28}                            & \underline{\textbf{23.62}}                            & 0.85                                                    & \underline{\textbf{22.33}}                            & \underline{22.83}                            & 0.72                             \\
                                                                                                   & semeval\_rel & f1       & 15.62                            & 19.16                            & 2.88                          & \underline{20.48}                            & \underline{\textbf{25.28}}                            & 4.77                                                        & \underline{\textbf{21.06}}                            & \underline{23.12}                            & 1.38                             \\
                                                                                                   & wiki80   & acc       & 36.04                            & 41.30                            & 3.90                                              & \underline{53.57}                            & \underline{54.58}                            & 1.01                                                  & \underline{\textbf{62.12}}                            & \underline{\textbf{62.82}}                            & 0.51                             \\
                                                                                                   \midrule
\multirow{9}{*}{\textbf{\begin{tabular}[c]{@{}l@{}}Natural \\ Language \\ Inference\end{tabular}}} & anli-r1                     & acc          & 45.96                            & 52.00                            & 6.33                     & \underline{\textbf{71.28}}                            & \underline{\textbf{74.30}}                            & 3.57                                                     & \underline{61.10}                            & \underline{66.10}                            & 6.97                             \\
                                                                                                   & anli-r2   & acc      & 41.34                            & 46.10                            & 4.44                                              & \underline{\textbf{60.08}}                            & \underline{\textbf{63.20}}                            & 3.94                                                  & \underline{48.66}                            & \underline{51.50}                            & 3.30                             \\
                                                                                                   & anli-r3     & acc      & 40.17                            & 44.42                            & 3.86                                             & \underline{\textbf{57.37}}                            & \underline{\textbf{61.33}}                            & 4.63                                                 & \underline{50.20}                            & \underline{53.92}                            & 4.21                             \\
                                                                                                   & cb       & acc       & 76.07                            & 83.93                            & 11.89                                                  & \underline{\textbf{82.50}}                            & \underline{\textbf{91.07}}                            & 8.78                                             & \underline{79.64}                            & \underline{87.50}                            & 9.50                             \\
                                                                                                   & multi\_nli\_matched & acc   & 55.14                            & 62.22                            & 9.67                                        & \underline{\textbf{75.47}}                            & \underline{\textbf{77.96}}                            & 4.09                                                  & \underline{69.62}                            & \underline{73.74}                            & 7.50                             \\
                                                                                                   & multi\_nli\_mismatched    & acc                 & 56.56                            & 63.50                            & 10.12                  & \underline{\textbf{76.31}}                            & \underline{\textbf{78.68}}                            & 3.53                                                  & \underline{70.78}                            & \underline{75.34}                            & 8.08                             \\
                                                                                                   & rte     & acc         & 79.06                            & \textbf{87.00}                            & 10.38                                             & \underline{\textbf{83.90}}                            & 85.92                            & 1.87                                                 & \underline{79.42}                            & 81.23                            & 2.17                             \\
                                                                                                   & sick     & acc       & 37.31                            & 55.81                            & 14.27                                               & \underline{\textbf{54.25}}                            & \underline{\textbf{69.75}}                            & 11.54                                               & \underline{47.65}                            & \underline{57.28}                            & 9.49                             \\
                                                                                                   & snli     & acc      & 58.67                            & 65.38                            & 9.09                                            & \underline{\textbf{78.44}}                            & \underline{\textbf{82.44}}                            & 3.78                                                     & \underline{76.37}                            & \underline{80.46}                            & 6.34                             \\
                                                                                                   \midrule
\multirow{10}{*}{\textbf{\begin{tabular}[c]{@{}l@{}}Intent \\ Detection\end{tabular}}}             & atis              & acc    & 5.82                             & 7.17                             & 1.38                                             & \underline{8.78}                             & \underline{11.31}                            & 1.63                                             & \underline{\textbf{14.22}}                            & \underline{\textbf{15.34}}                            & 1.19                             \\
                                                                                                   & banking77  & acc      & 33.63                            & 47.92                            & 10.97                                           & \underline{\textbf{65.45}}                            & \underline{65.91}                            & 0.38                                                   & \underline{64.77}                            & \underline{\textbf{67.66}}                            & 2.88                             \\
                                                                                                   & clinc150    & acc    & 40.70                            & 54.98                            & 10.45                                           & \underline{57.30}                            & \underline{\textbf{62.76}}                            & 4.41                                                    & \underline{\textbf{58.26}}                            & \underline{60.16}                            & 1.36                             \\
                                                                                                   & fb        & acc        & 66.58                            & 74.27                            & 8.74                                           & \underline{\textbf{92.47}}                            & \underline{\textbf{93.54}}                            & 0.96                                                   & \underline{90.14}                            & \underline{92.44}                            & 2.62                             \\
                                                                                                   & hint3-curekart    & acc    & 30.59                            & 37.47                            & 6.38                                       & \underline{\textbf{64.79}}                            & \underline{\textbf{67.97}}                            & 2.39                                                   & \underline{42.56}                            & \underline{49.55}                            & 5.83                             \\
                                                                                                   & hint3-powerplay11   & acc    & 32.30                            & 42.72                            & 7.23                                 & \underline{\textbf{47.57}}                            & \underline{\textbf{51.78}}                            & 3.35                                                       & 15.61                            & 18.11                            & 2.05                             \\
                                                                                                   & hint3-sofmattress  & acc     & 27.35                            & 39.53                            & 9.02                                    & \underline{\textbf{52.41}}                            & \underline{\textbf{54.94}}                            & 2.25                                                    & \underline{48.61}                            & \underline{50.88}                            & 3.61                             \\
                                                                                                   & nlued           & acc      & 36.91                            & 52.76                            & 11.60                                       & \underline{54.25}                            & \underline{\textbf{59.56}}                            & 5.38                                                  & \underline{\textbf{54.28}}                            & \underline{55.44}                            & 1.08                             \\
                                                                                                   & slurp       & acc         & 38.29                            & 51.31                            & 11.25                                          & \underline{55.06}                            & \underline{\textbf{61.03}}                            & 4.98                                                & \underline{\textbf{55.85}}                            & \underline{56.76}                            & 0.62                             \\
                                                                                                   & snips       & acc      & 82.57                            & 88.00                            & 4.30                                           & \underline{84.63}                            & 86.71                            & 2.38                                                   & \underline{\textbf{89.91}}                            & \underline{\textbf{90.71}}                            & 1.10                             \\
                                                                                                   \midrule
\textbf{Fact Retrieval}                                                                            & LAMA-TREx                 & acc   & 15.38                            & 22.19                            & 7.49                                   & \underline{\textbf{36.89}}                            & \underline{\textbf{37.48}}                            & 0.50                                                     & \underline{30.80}                            & \underline{31.78}                            & 0.64                             \\
\midrule
\multirow{2}{*}{\textbf{\begin{tabular}[c]{@{}l@{}}Temporal \\ Reasoning\end{tabular}}}            & TRACIE      & acc        & \textbf{56.81}                            & 57.92                            & 1.02                                                & 56.74                            & \underline{\textbf{59.48}}                            & 2.33                                            & 56.28                            & 57.85                            & 1.97                             \\
                                                                                                   & UDS-T    & acc      & 54.80                            & 55.76                            & 0.75                                           & \underline{57.76}                            & \underline{\textbf{63.60}}                            & 5.69                                                      & \underline{\textbf{57.88}}                            & \underline{62.10}                            & 3.56                             \\
                                                                                                   \midrule
\multirow{4}{*}{\textbf{\begin{tabular}[c]{@{}l@{}}Word Sense \\ Disambiguation\end{tabular}}}     & semeval2007              & f1   & 23.82                            & 26.20                            & 1.40                                     & \underline{\textbf{31.60}}                            & \underline{\textbf{32.30}}                            & 0.48                                                & \underline{28.96}                            & \underline{29.20}                            & 0.22                             \\
                                                                                                   & semeval2013    & f1       & 38.10                            & 40.50                            & 1.75                                       & \underline{\textbf{43.78}}                            & \underline{\textbf{44.00}}                            & 0.18                                                   & \underline{43.00}                            & \underline{43.50}                            & 0.39                             \\
                                                                                                   & semeval2015     & f1     & 36.30                            & 36.90                            & 0.56                                     & \underline{44.26}                            & \underline{44.60}                            & 0.21                                                      & \underline{\textbf{44.46}}                            & \underline{\textbf{45.30}}                            & 0.86                             \\
                                                                                                   & senseval2     & f1      & 37.10                            & 39.40                            & 1.51                                       & \underline{\textbf{43.88}}                            & \underline{\textbf{44.10}}                            & 0.18                                                     & \underline{41.76}                            & \underline{41.80}                            & 0.09                             \\
                                                                                                   \midrule
\multirow{7}{*}{\textbf{Summarization}}                                                            & SciTLDR                   & r1               & 25.18                            & 25.27                            & 0.11                       & \underline{\textbf{34.92}}                            & \underline{\textbf{36.73}}                            & 2.56                                                & \underline{34.00}                            & \underline{36.59}                            & 3.14                             \\
                                                                                                   & Reddit-TIFU     & r1    & 19.47                            & 20.29                            & 1.34                                        & \underline{\textbf{22.67}}                            & \underline{\textbf{23.68}}                            & 1.43                                                    & \underline{20.48}                            & \underline{21.45}                            & 1.23                             \\
                                                                                                   & Multi-Xscience     & r1      & 26.74                            & 26.98                            & 0.41                                  & \underline{28.75}                            & \underline{29.03}                            & 0.26                                                     & \underline{\textbf{29.64}}                            & \underline{\textbf{29.69}}                            & 0.06                             \\
                                                                                                   & WikiSum         & r1     & 36.92                            & 36.99                            & 0.06                                       & \underline{\textbf{41.47}}                            & \underline{\textbf{41.68}}                            & 0.24                                                    & \underline{40.00}                            & \underline{40.29}                            & 0.26                             \\
                                                                                                   & GovReport      & r1        & 34.60                            & 34.61                            & 0.02                                    & \underline{42.34}                            & \underline{42.51}                            & 0.11                                                     & \underline{\textbf{44.05}}                            & \underline{\textbf{44.17}}                            & 0.14                             \\
                                                                                                   & BillSum        & r1      & 28.84                            & 28.84                            & 0.00                                    & \underline{40.20}                            & \underline{40.45}                            & 0.17                                                       & \underline{\textbf{41.47}}                            & \underline{\textbf{41.61}}                            & 0.19                             \\
                                                                                                   & BigPatent       & r1     & 30.06                            & 30.06                            & 0.00                                      & \underline{\textbf{39.06}}                            & \underline{\textbf{39.21}}                            & 0.09                                                     & \underline{36.70}                            & \underline{36.85}                            & 0.13           \\
                                                                                                   \bottomrule
\end{tabular}
\end{table}

\begin{table}[!ht]
\centering
\footnotesize
 \setlength\tabcolsep{1pt}
 \caption{The model performance corresponding to each prompt on all datasets (Part 1). ``RST-T" represents the RST-Task model and ``RST-A" represents the RST-All model.}
 \label{tab:prompt_result1}
\begin{tabular}{l|llll}
\toprule
\multirow{14}{*}{\textbf{\begin{tabular}[c]{@{}l@{}}Topic \\ Classification\end{tabular}}}          & \multirow{7}[1]{*}{\includegraphics[scale=0.22]{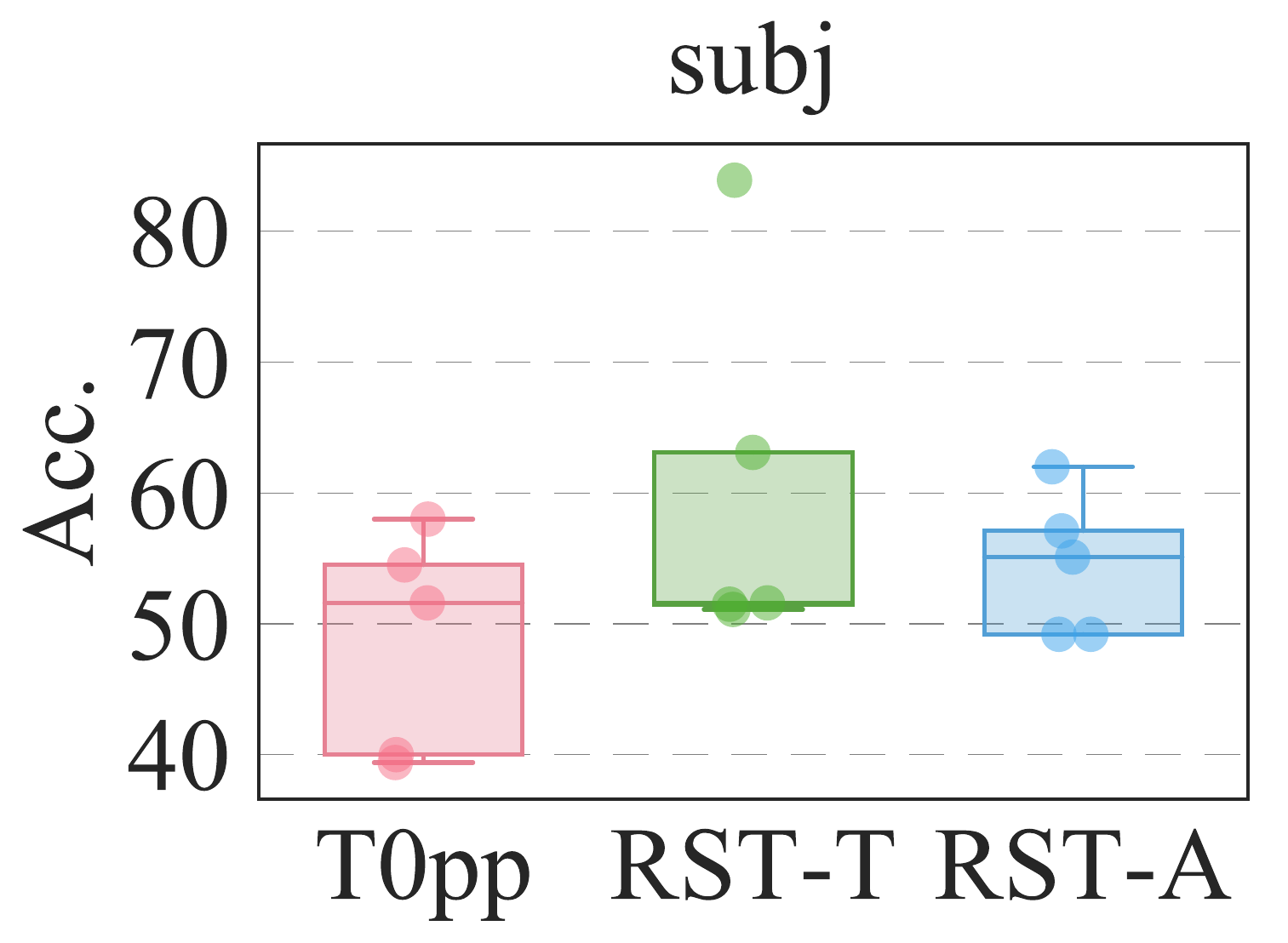}}  & \multirow{7}[1]{*}{\includegraphics[scale=0.22]{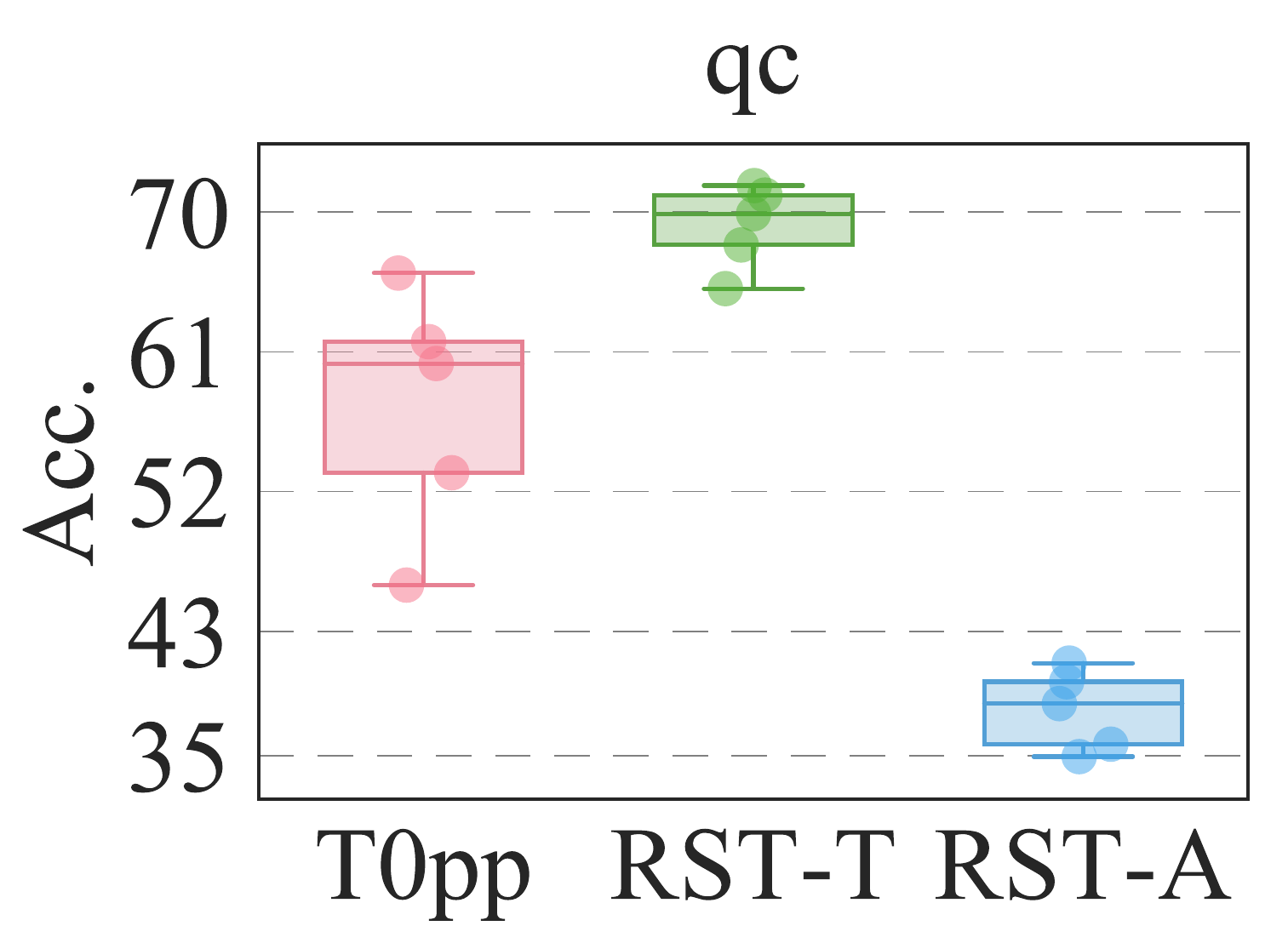}} & \multirow{7}[1]{*}{\includegraphics[scale=0.22]{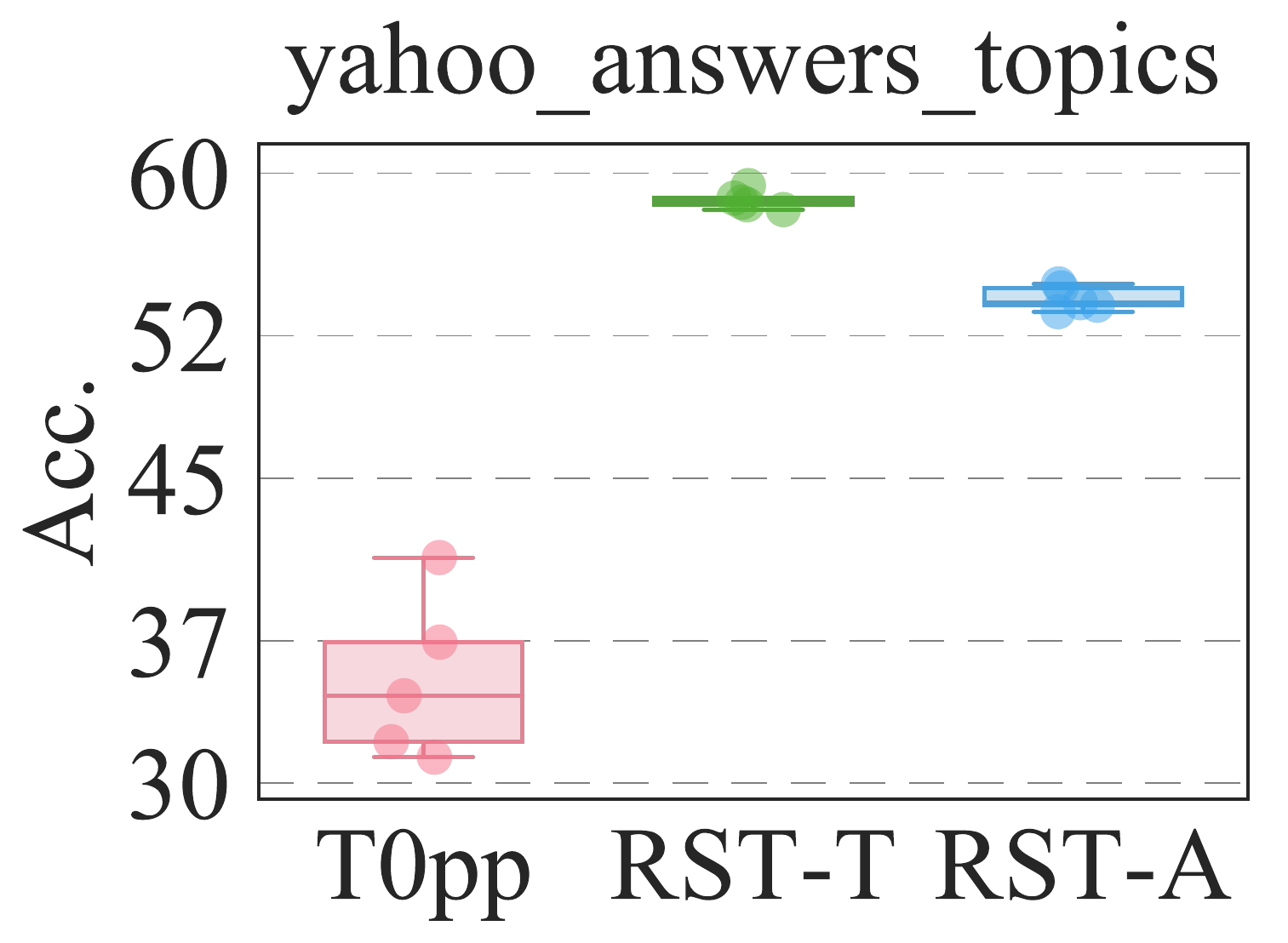}} & \multirow{7}[1]{*}{\includegraphics[scale=0.22]{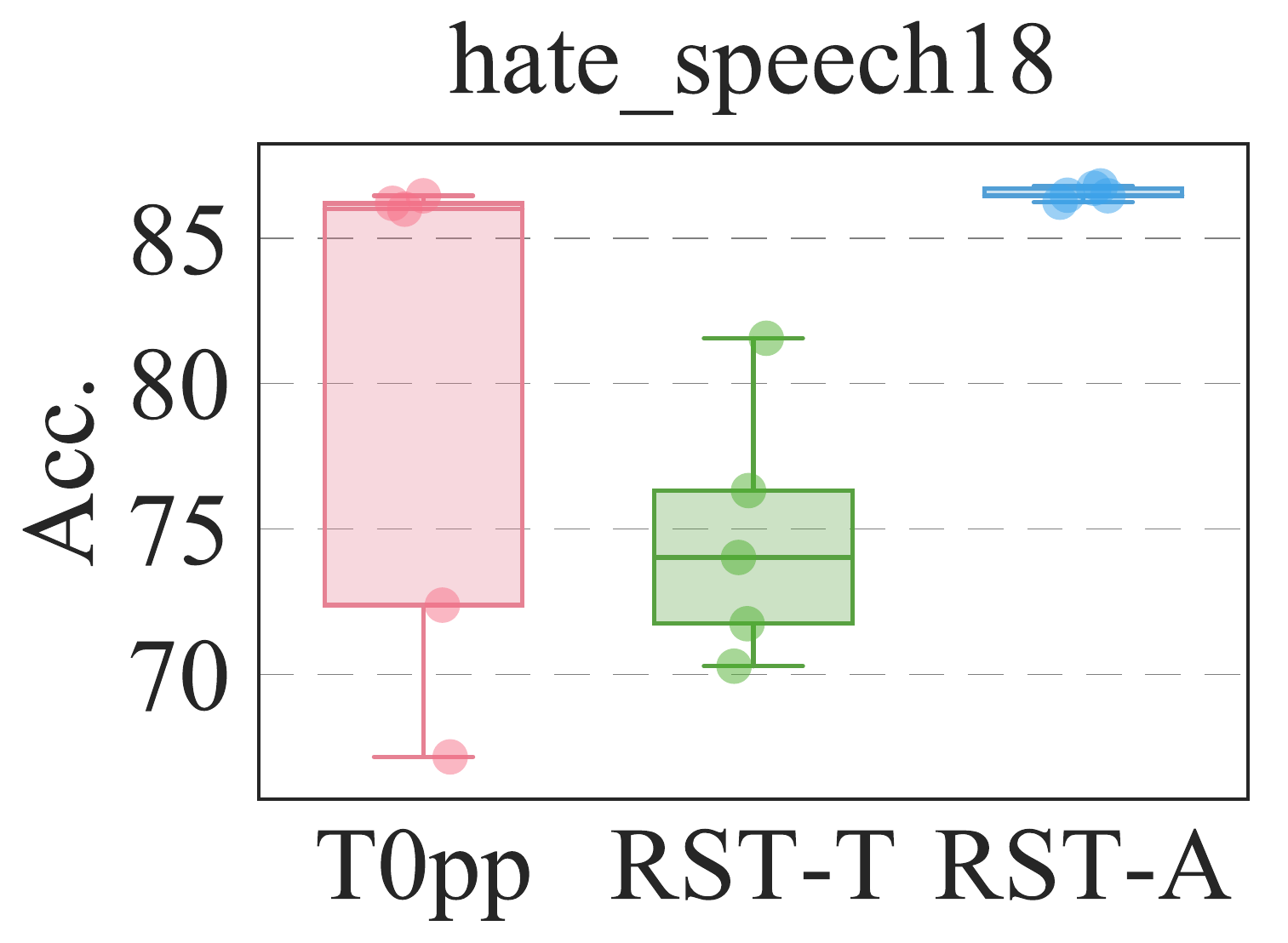}}\\
                                                                                                                  \\

                                                                                                                   \\
                                                                                                                  \\
                                                                                                                  \\
                                                                                                                  \\
                                                                                                                  \\
                                                                                                         
& \multirow{7}[1]{*}{\includegraphics[scale=0.22]{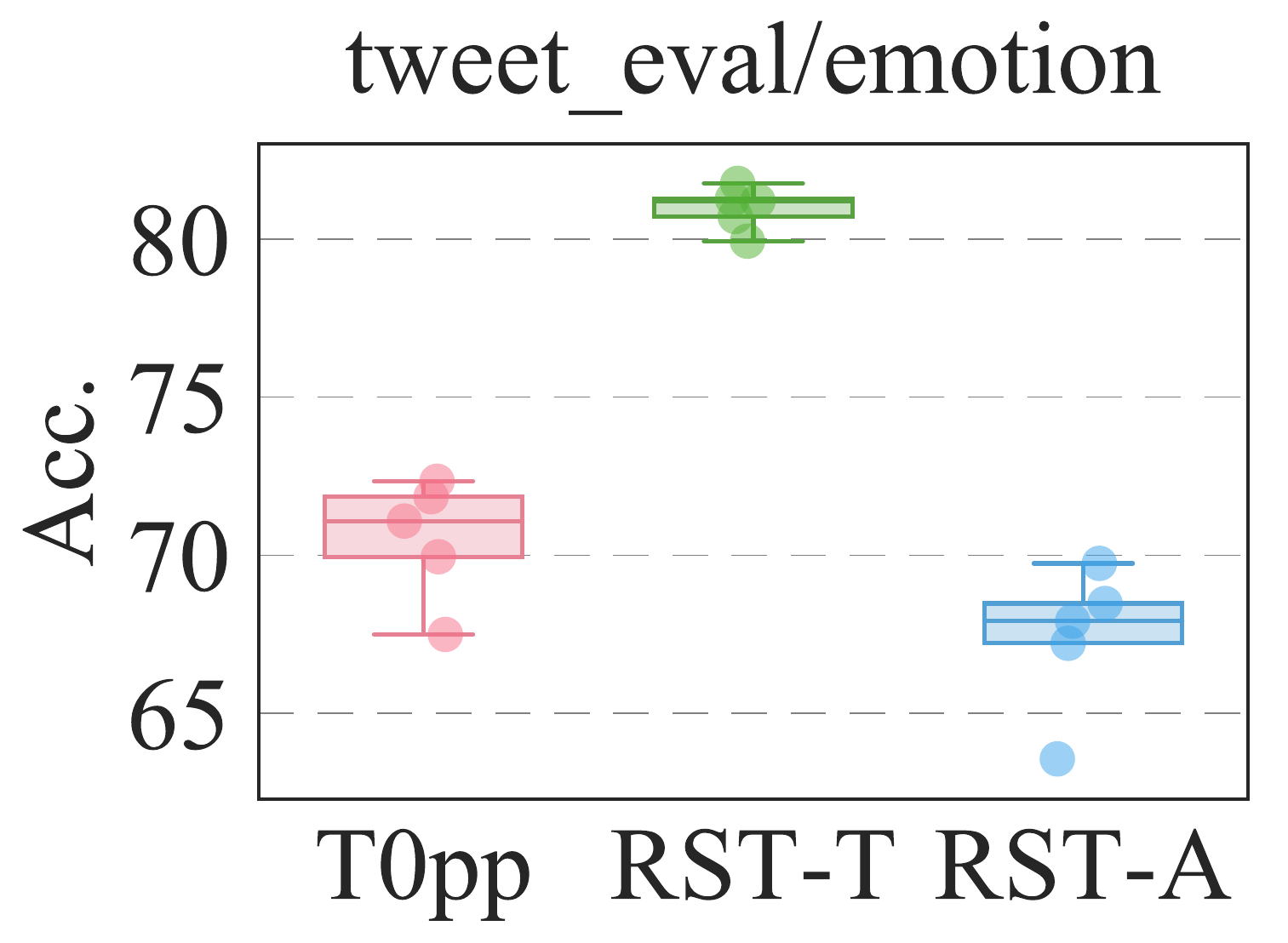}}  & \multirow{7}[1]{*}{\includegraphics[scale=0.22]{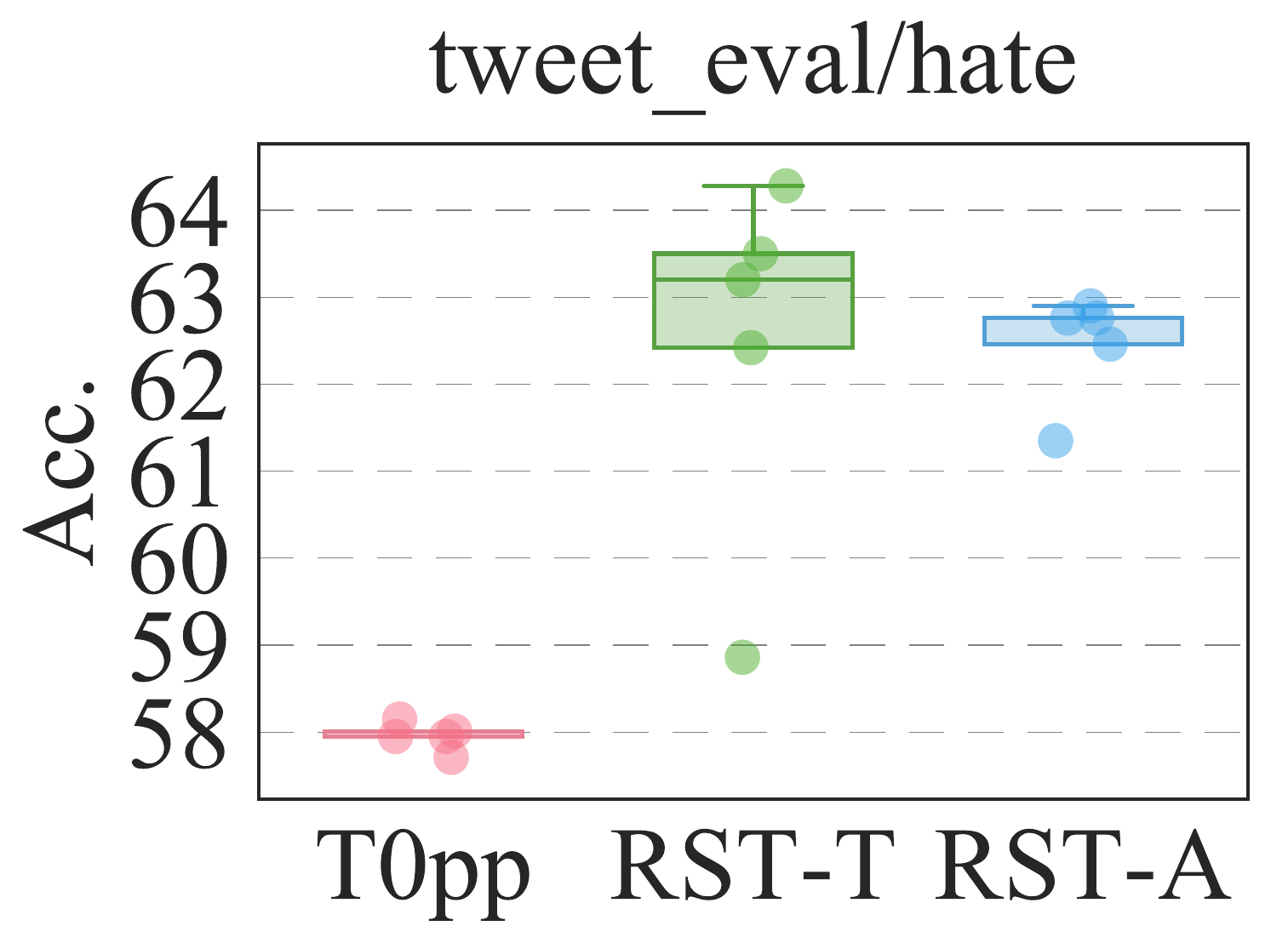}} & \multirow{7}[1]{*}{\includegraphics[scale=0.22]{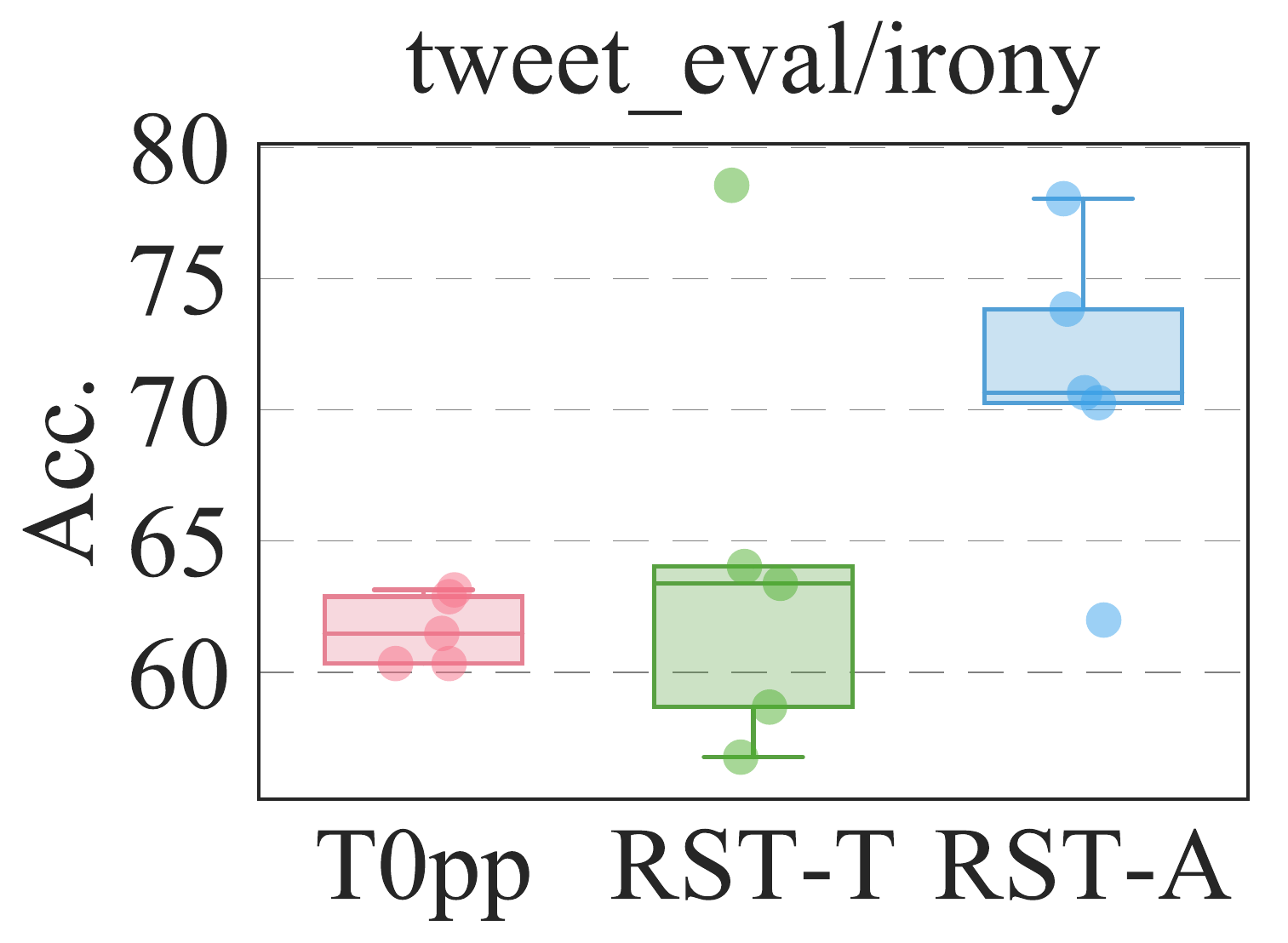}} & \multirow{7}[1]{*}{\includegraphics[scale=0.22]{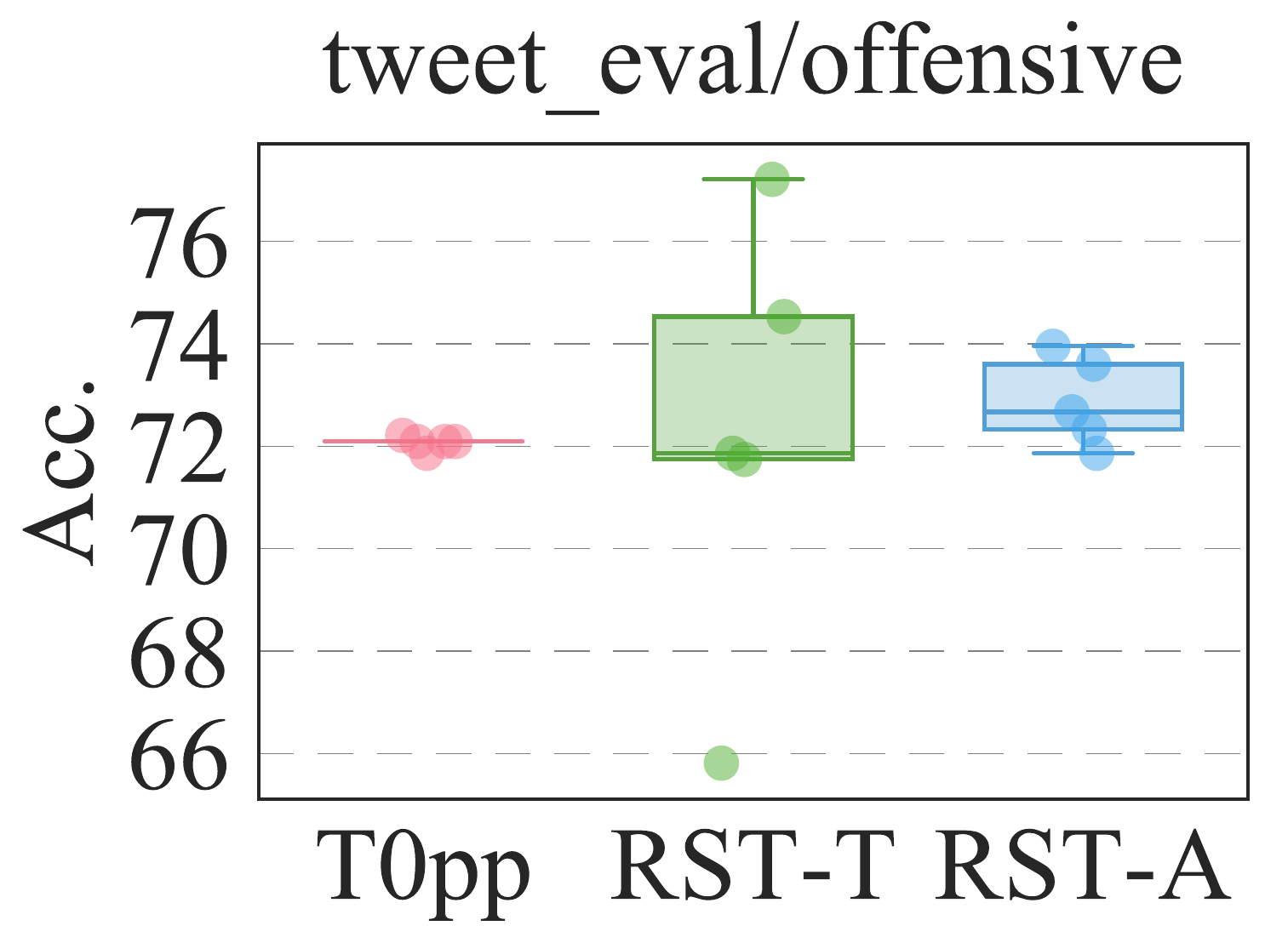}}\\
                             &                    \\

                                                                                                  &                    \\
                                                                                                  &                    \\
                                                                                                  &                    \\
                                                                                                  &                    \\
                                                                                                  \\
                                            \midrule
\multirow{7}{*}{\textbf{\begin{tabular}[c]{@{}l@{}}Sentiment \\ Classification\end{tabular}}}      &  \multirow{7}[1]{*}{\includegraphics[scale=0.22]{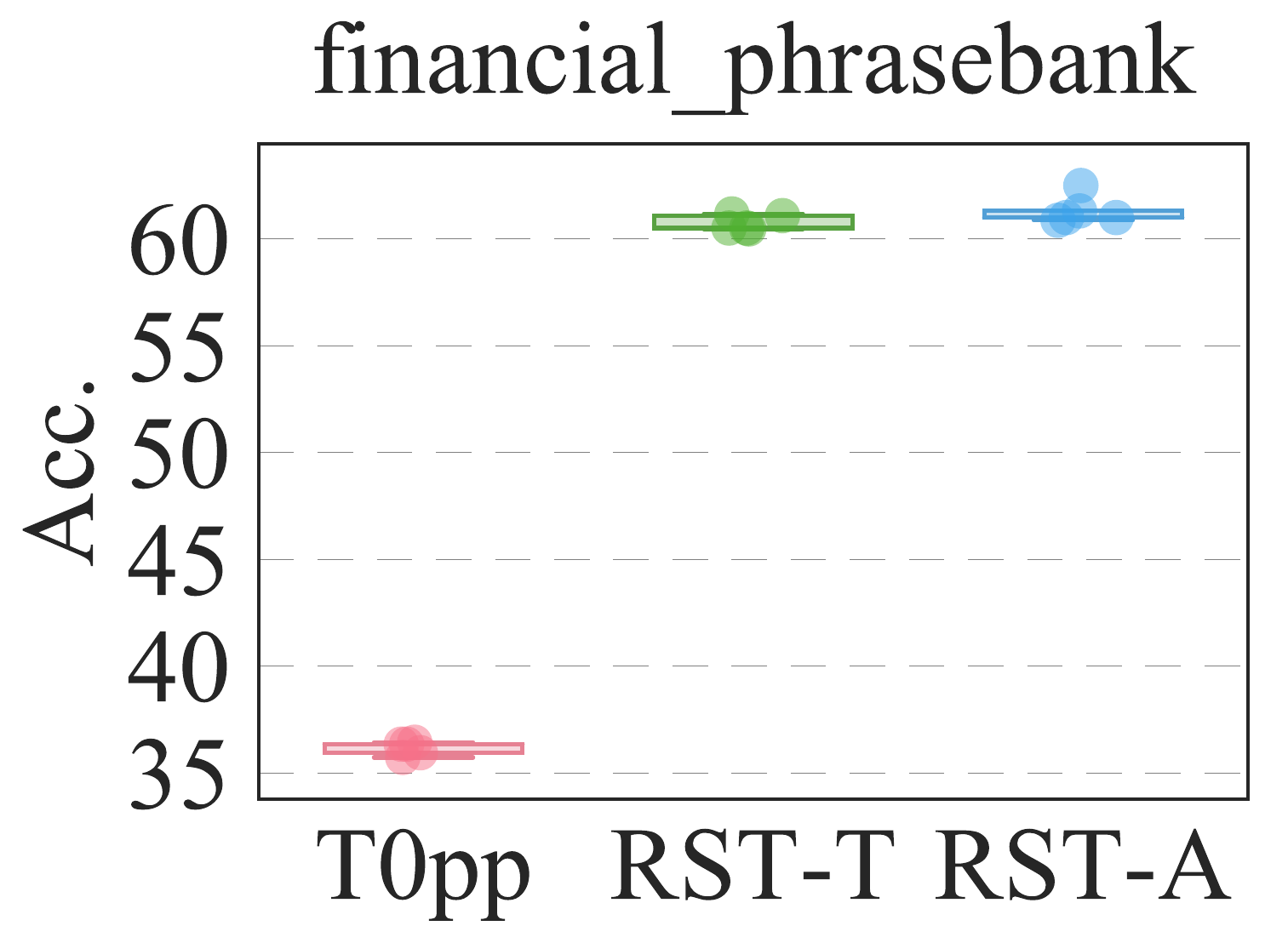}}  & \multirow{7}[1]{*}{\includegraphics[scale=0.22]{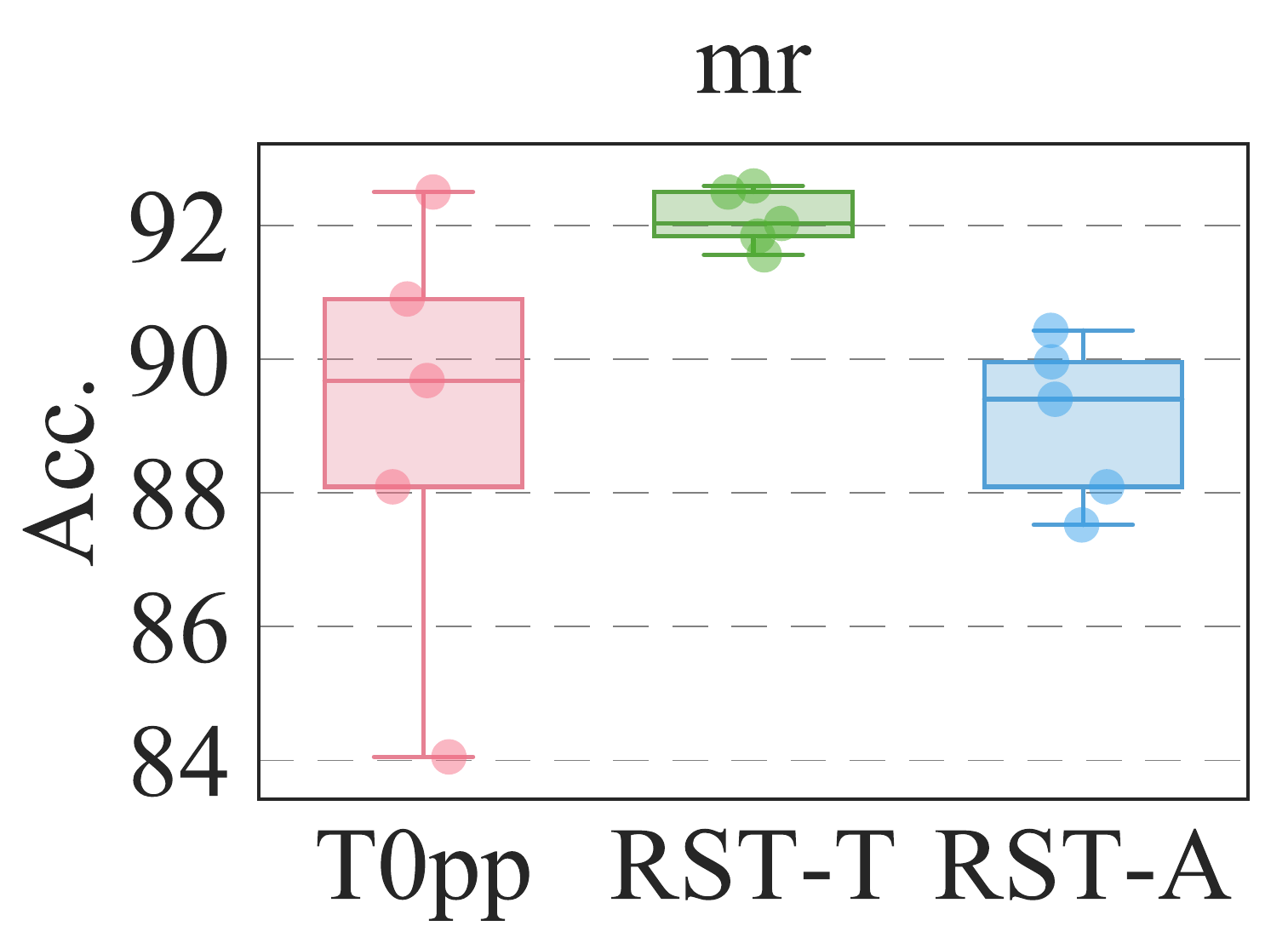}} & \multirow{7}[1]{*}{\includegraphics[scale=0.22]{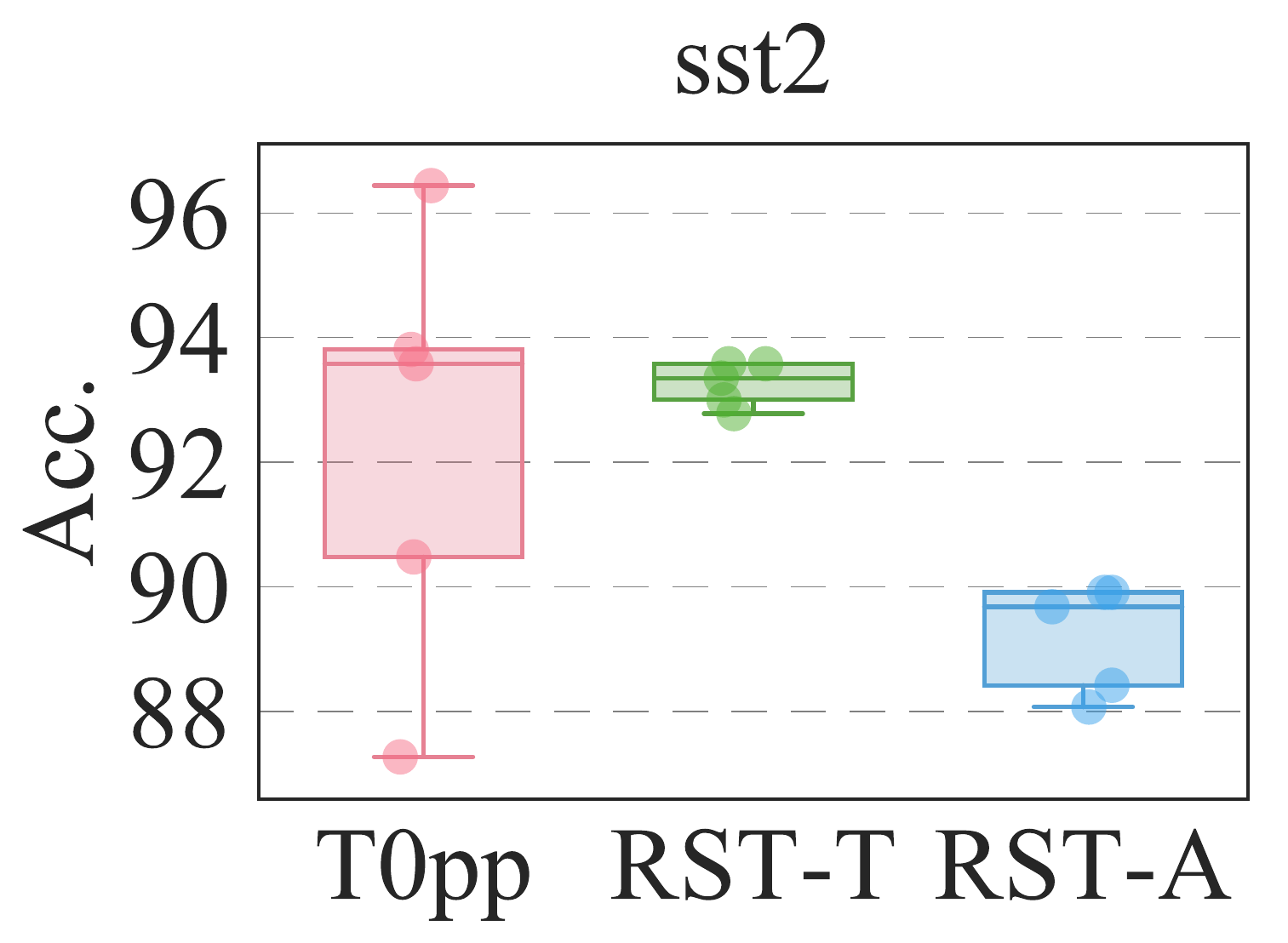}} &  \\
                                                                                                   &                    \\
                                                                                                   &                    \\
                                                                                                   \\
                                                                                                   \\
                                                                                                   \\
                                                                                
                                                                                                   \\
                                                                                                   \midrule
\multirow{21}{*}{\textbf{\begin{tabular}[c]{@{}l@{}}Information \\ Extraction\end{tabular}}}     &   \multirow{7}[1]{*}{\includegraphics[scale=0.22]{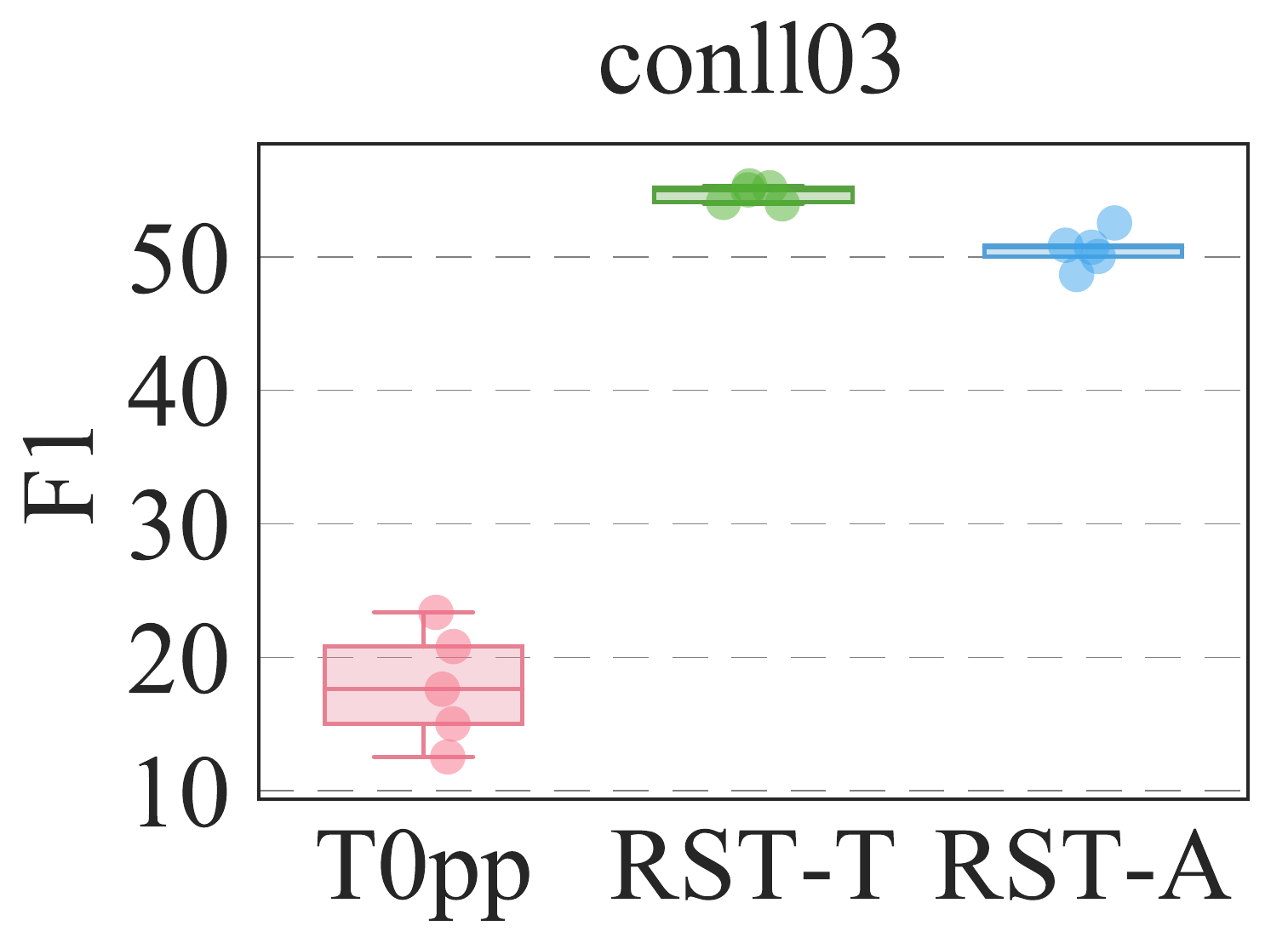}}  & \multirow{7}[1]{*}{\includegraphics[scale=0.22]{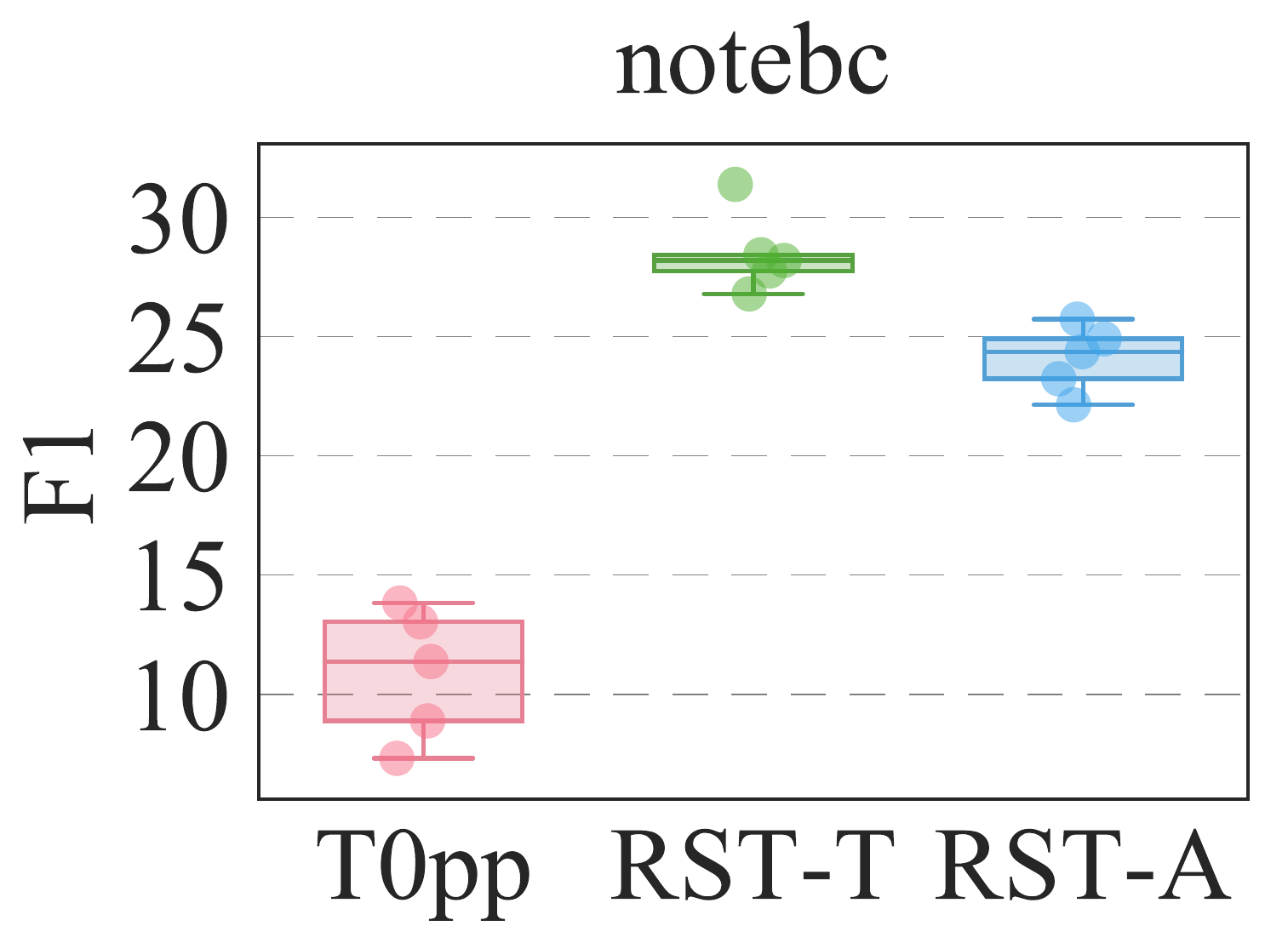}} & \multirow{7}[1]{*}{\includegraphics[scale=0.22]{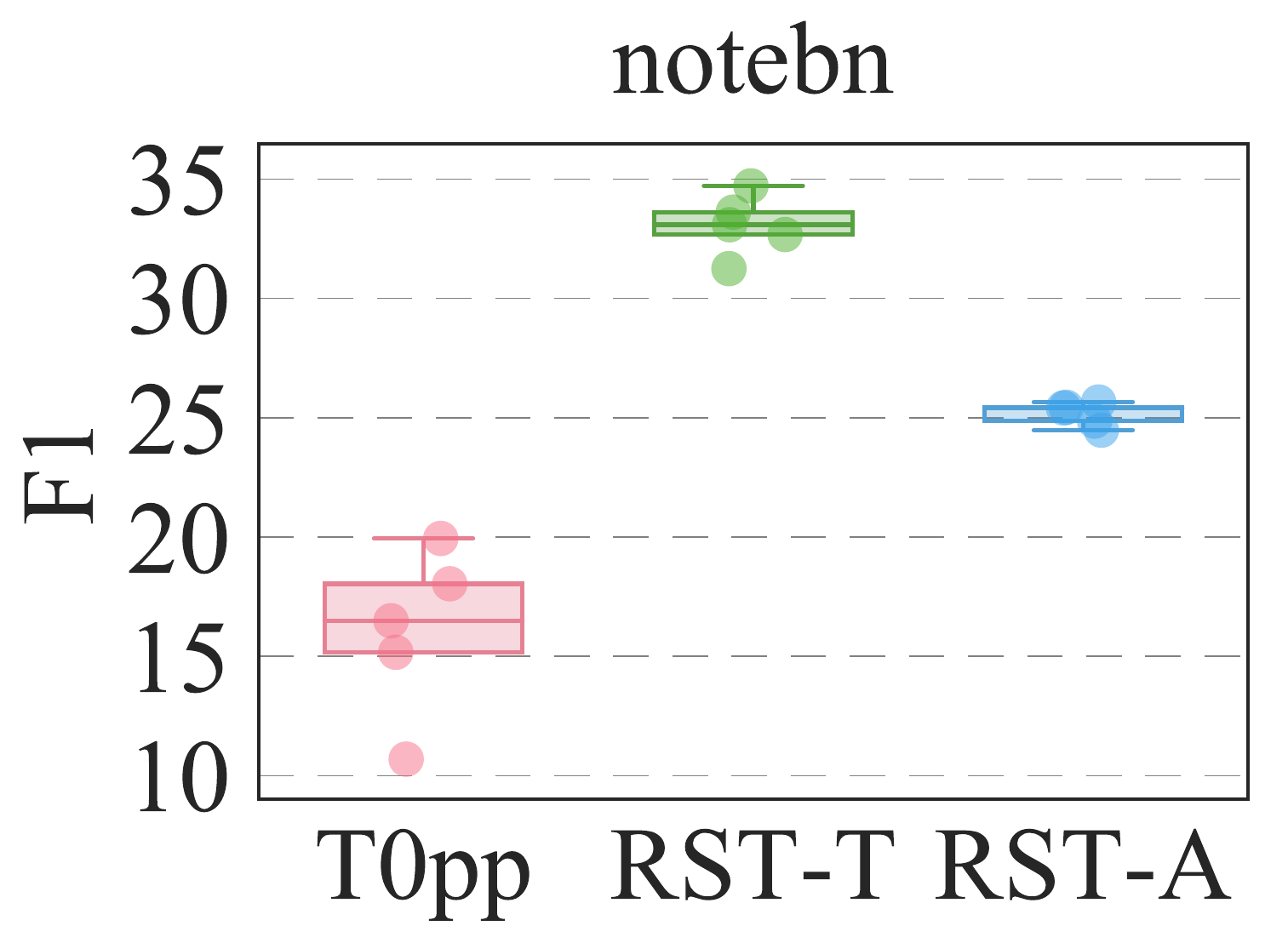}} & \multirow{7}[1]{*}{\includegraphics[scale=0.22]{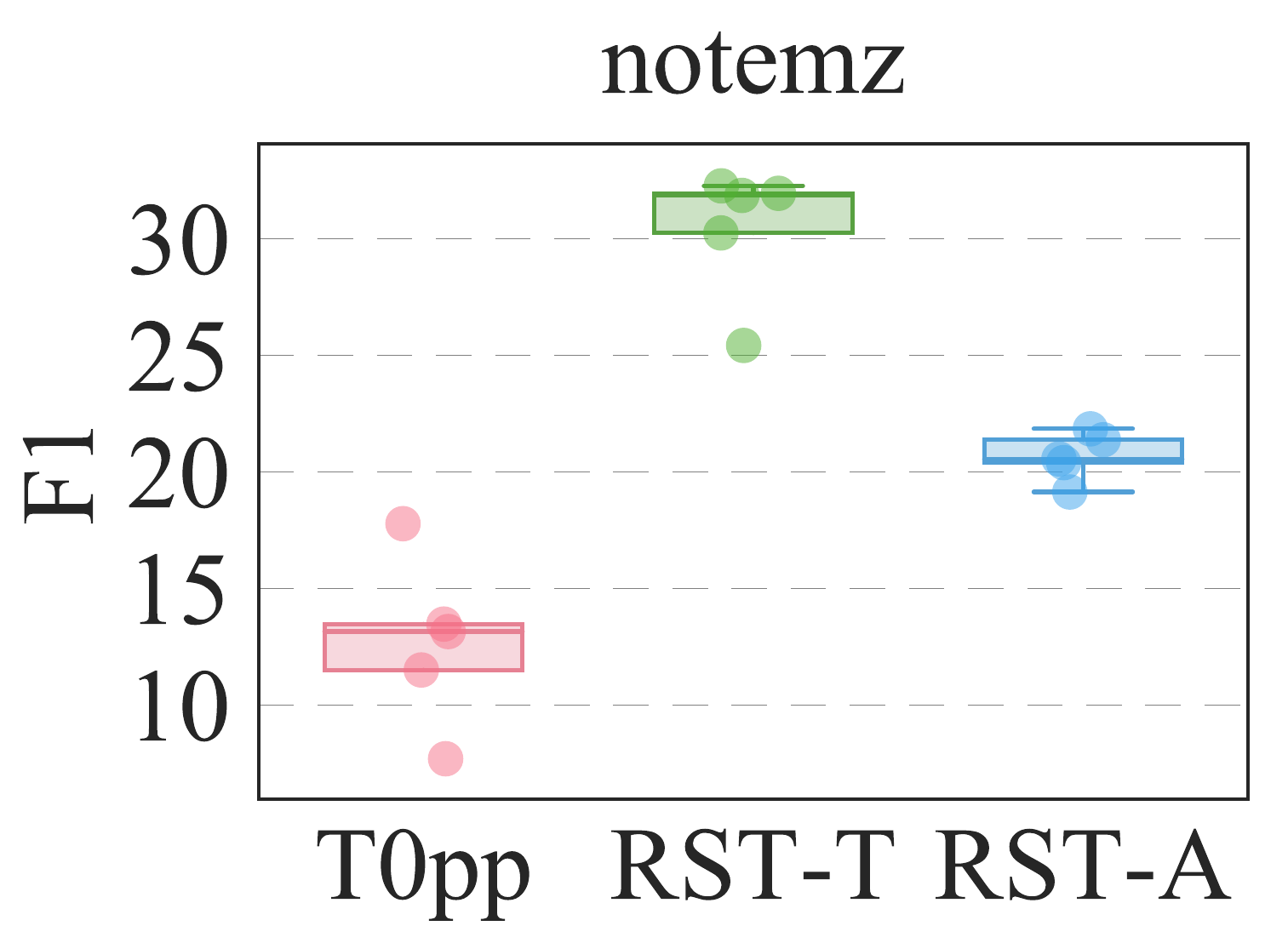}}                                                  \\
                                                                                                   &                    \\
                                                                                                   &                    \\
                                                                                                   &                    \\
                                                                                                   &                    \\
                                                                                                   &                    \\
                                                                                
                                                                    \\        
                                                                                 &
 \multirow{7}[1]{*}{\includegraphics[scale=0.22]{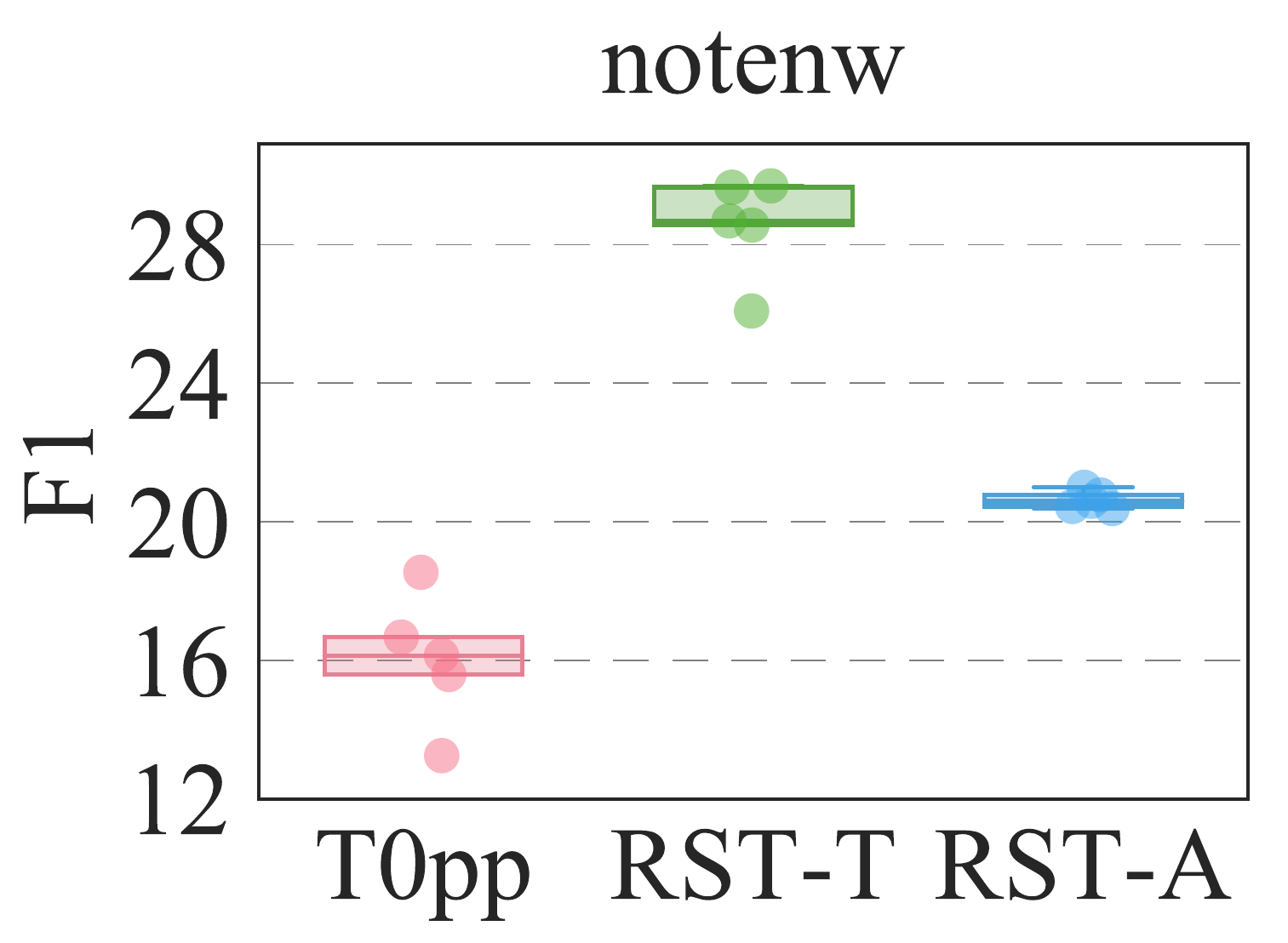}}  & \multirow{7}[1]{*}{\includegraphics[scale=0.22]{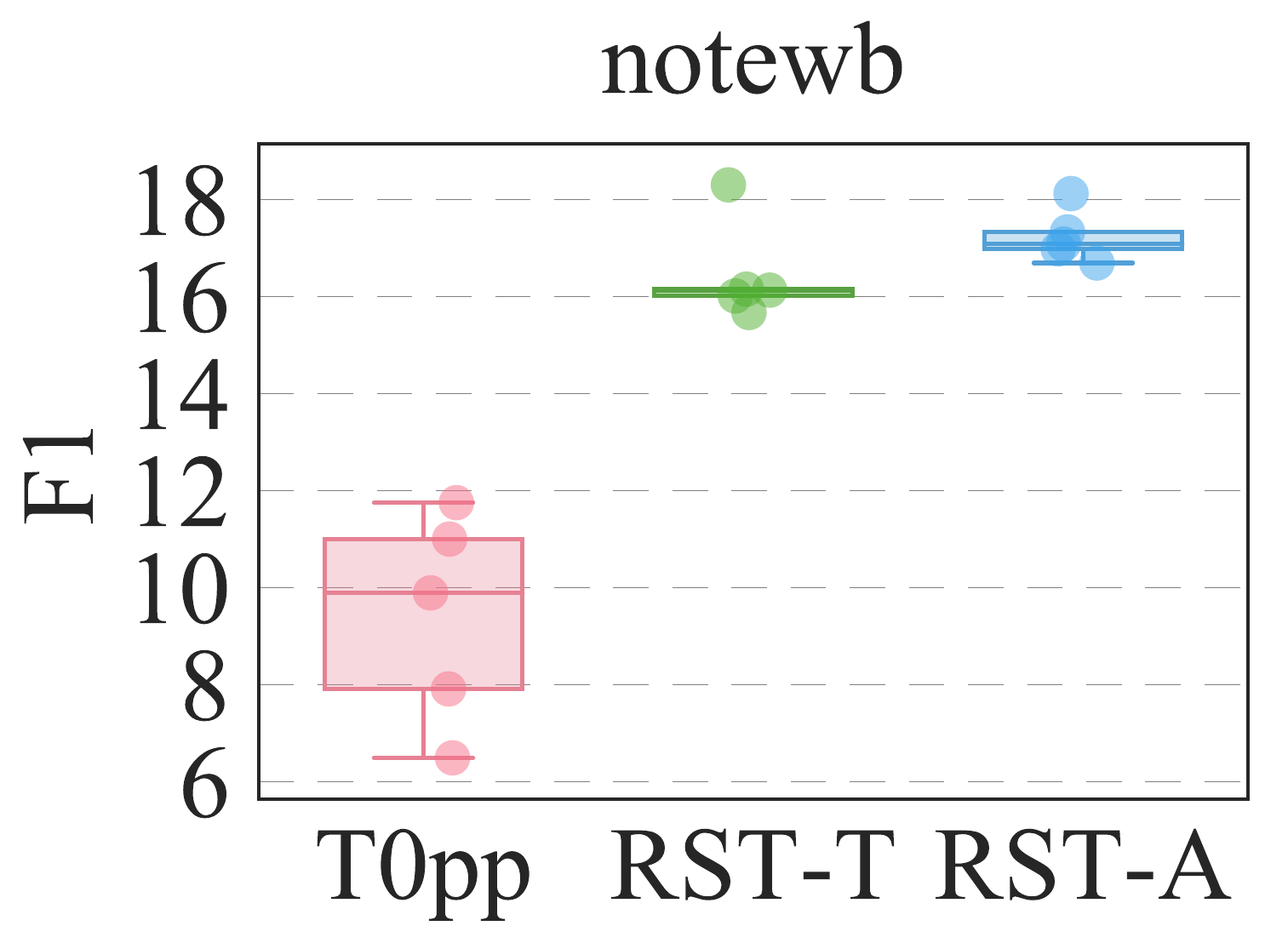}} & \multirow{7}[1]{*}{\includegraphics[scale=0.22]{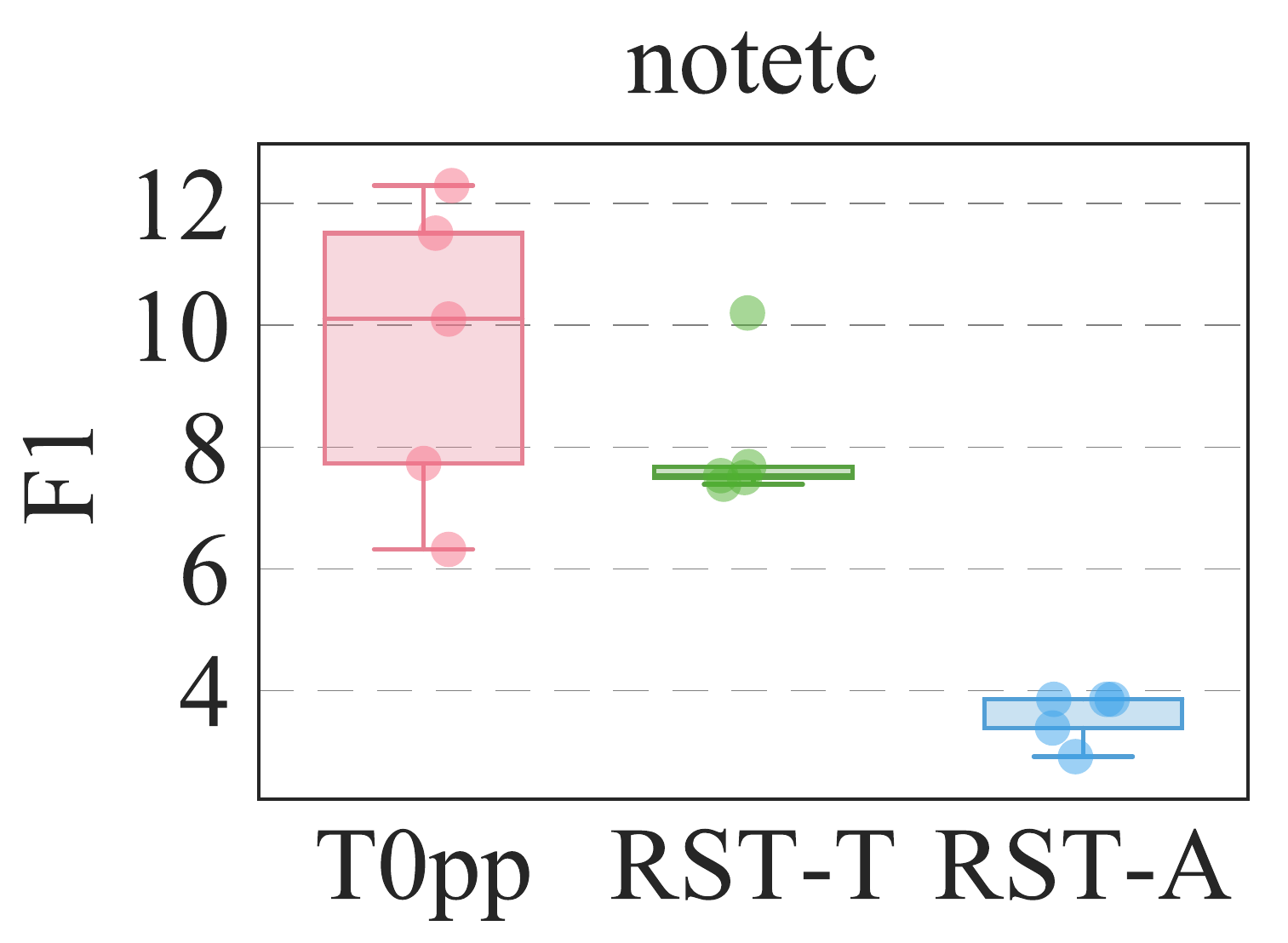}} & \multirow{7}[1]{*}{\includegraphics[scale=0.22]{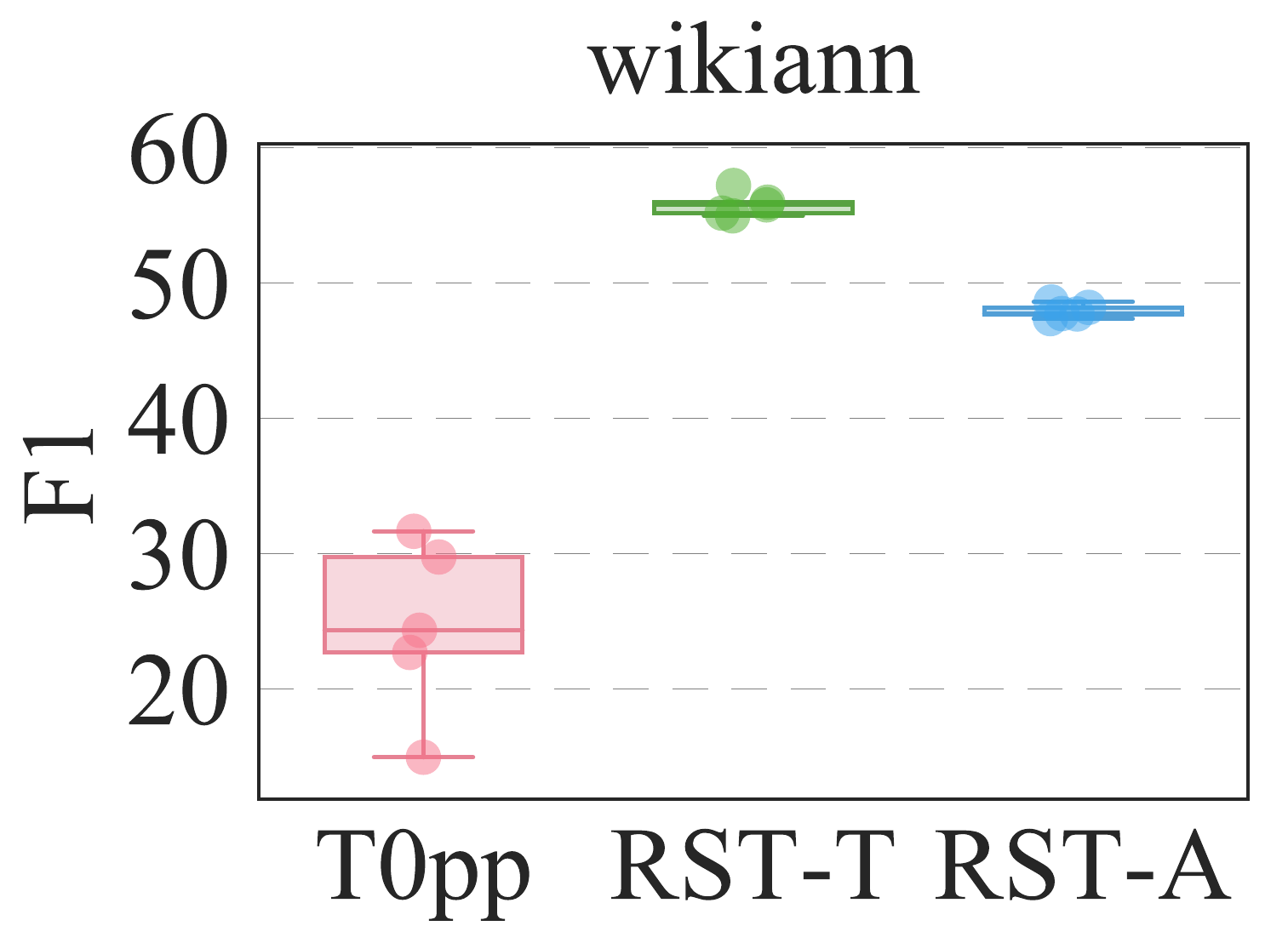}}   \\                                   \\
 \\
 \\
 \\
 \\
 \\
        &
 \multirow{7}[1]{*}{\includegraphics[scale=0.22]{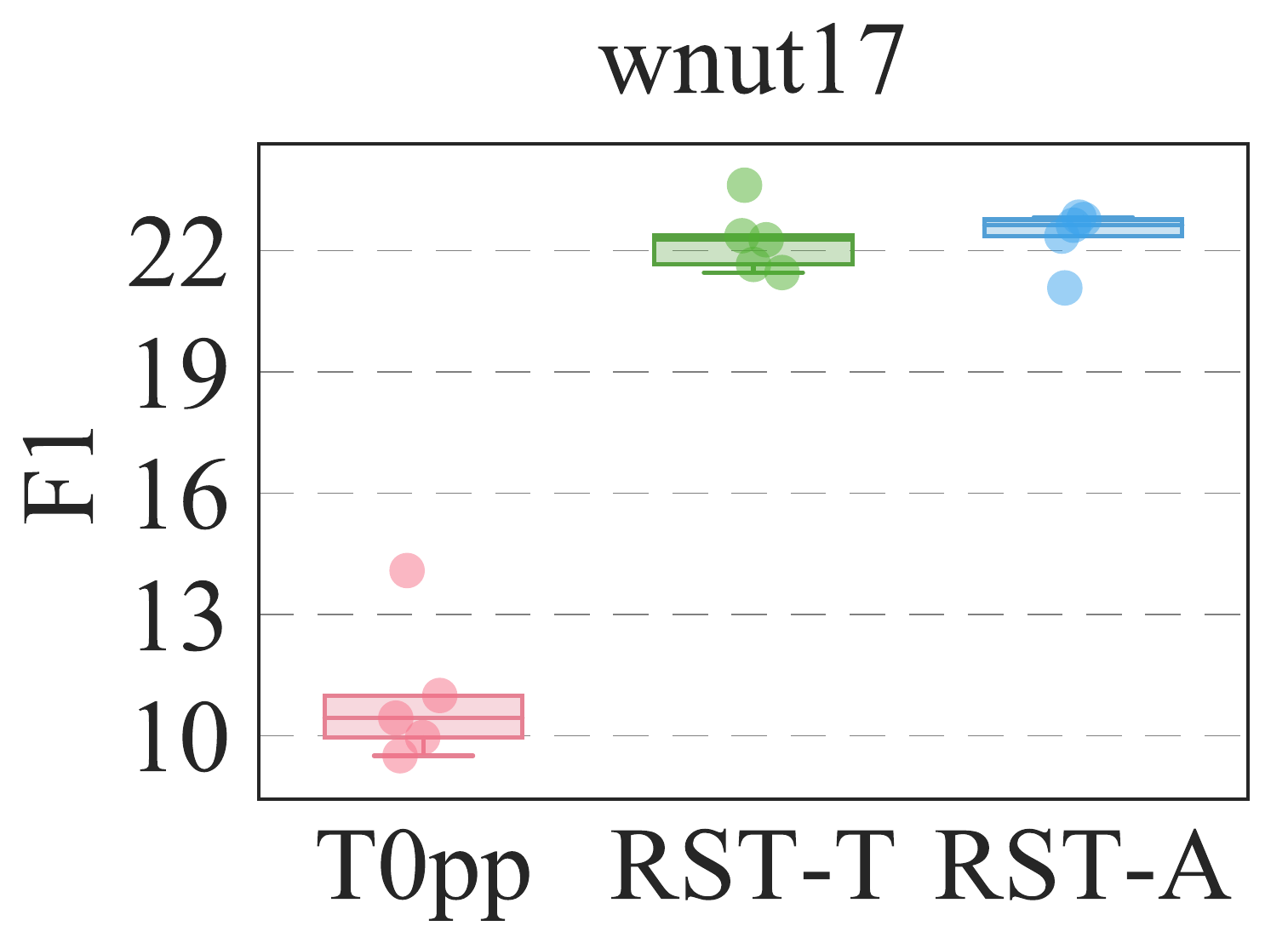}}  & \multirow{7}[1]{*}{\includegraphics[scale=0.22]{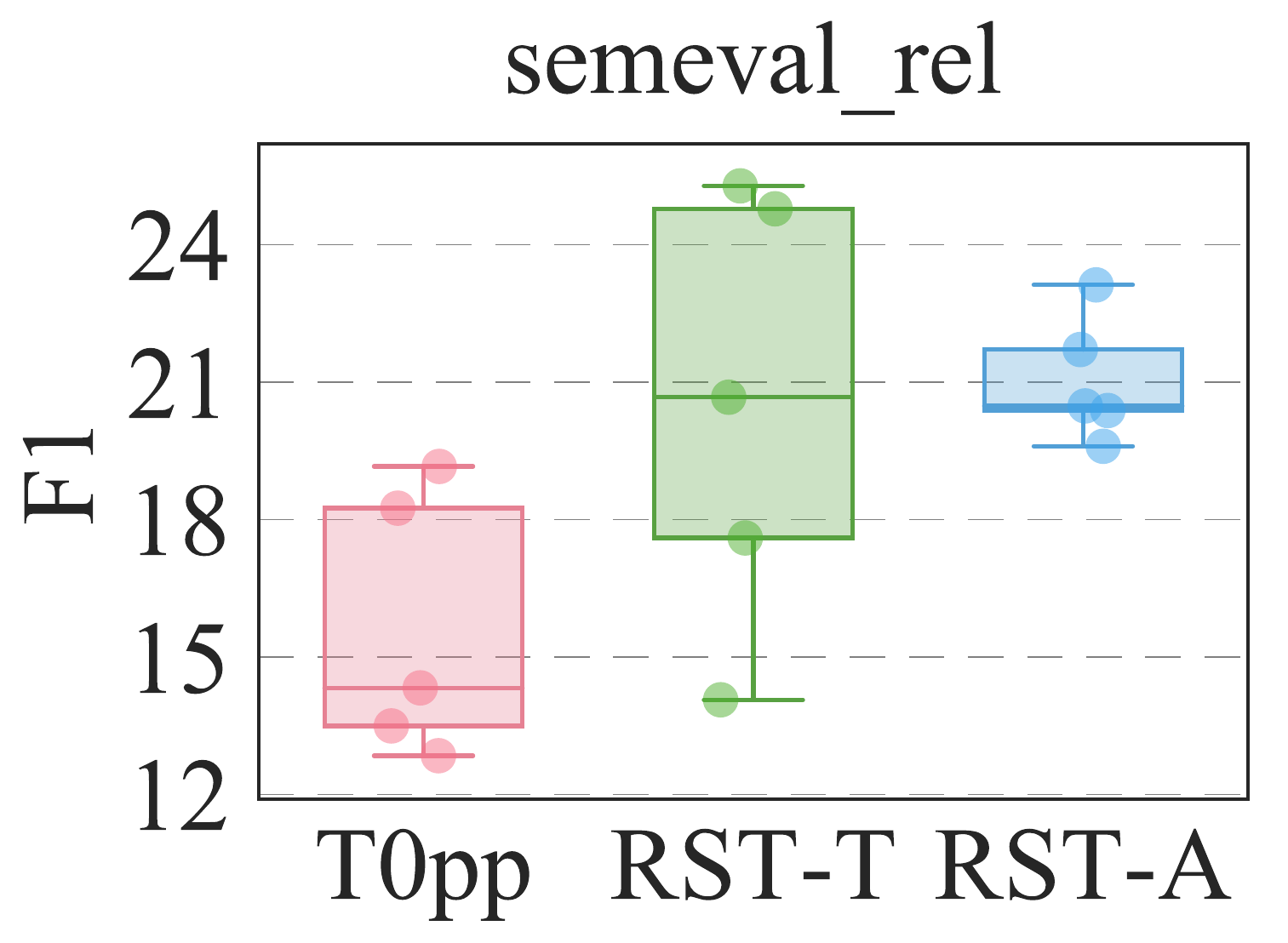}} & \multirow{7}[1]{*}{\includegraphics[scale=0.22]{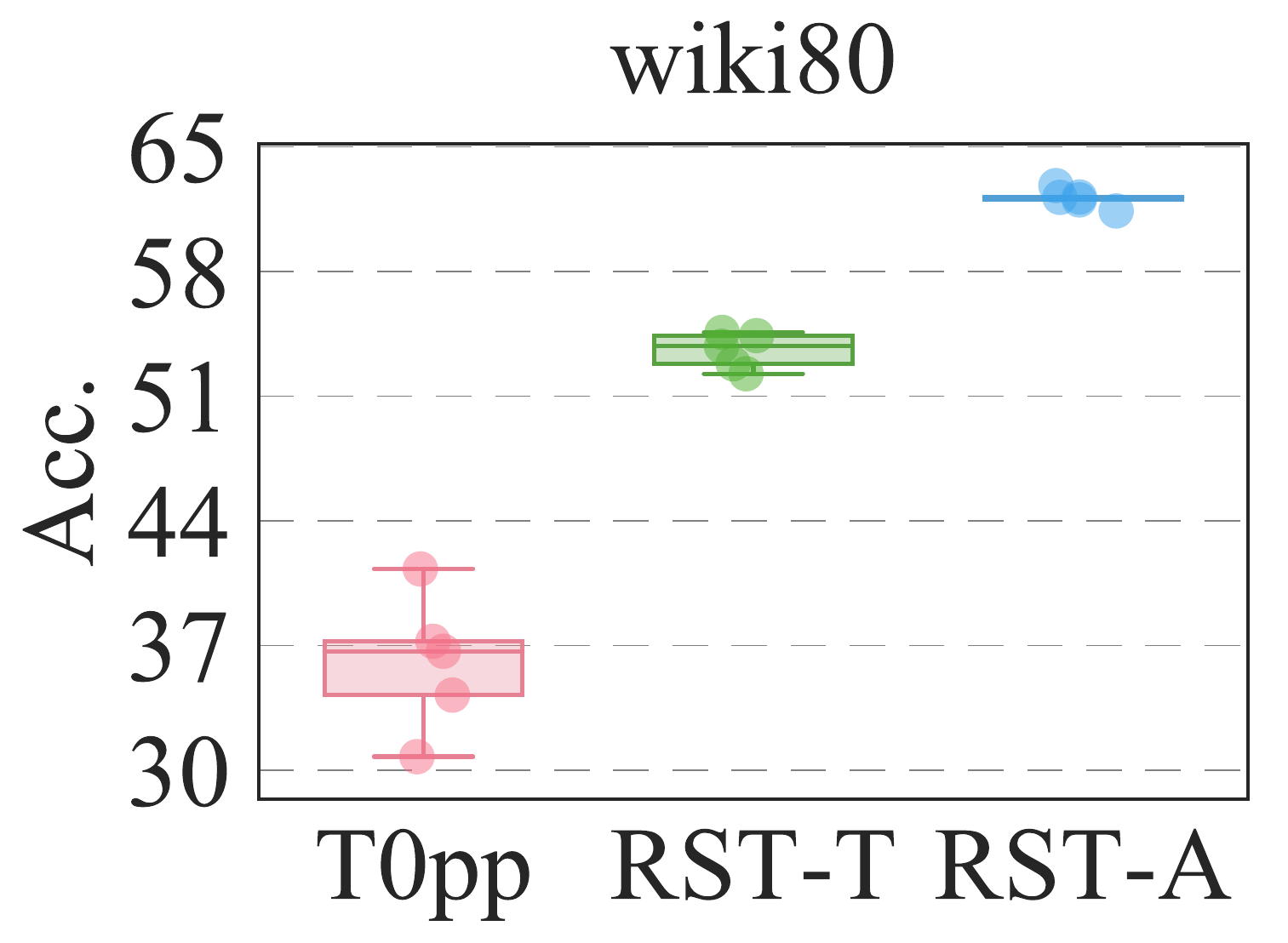}} & 
 \\
 \\
 \\
 \\
 \\
 \\
 \\
                                                                                                   \midrule
\multirow{21}{*}{\textbf{\begin{tabular}[c]{@{}l@{}}Natural \\ Language \\ Inference\end{tabular}}} &
 \multirow{7}[1]{*}{\includegraphics[scale=0.22]{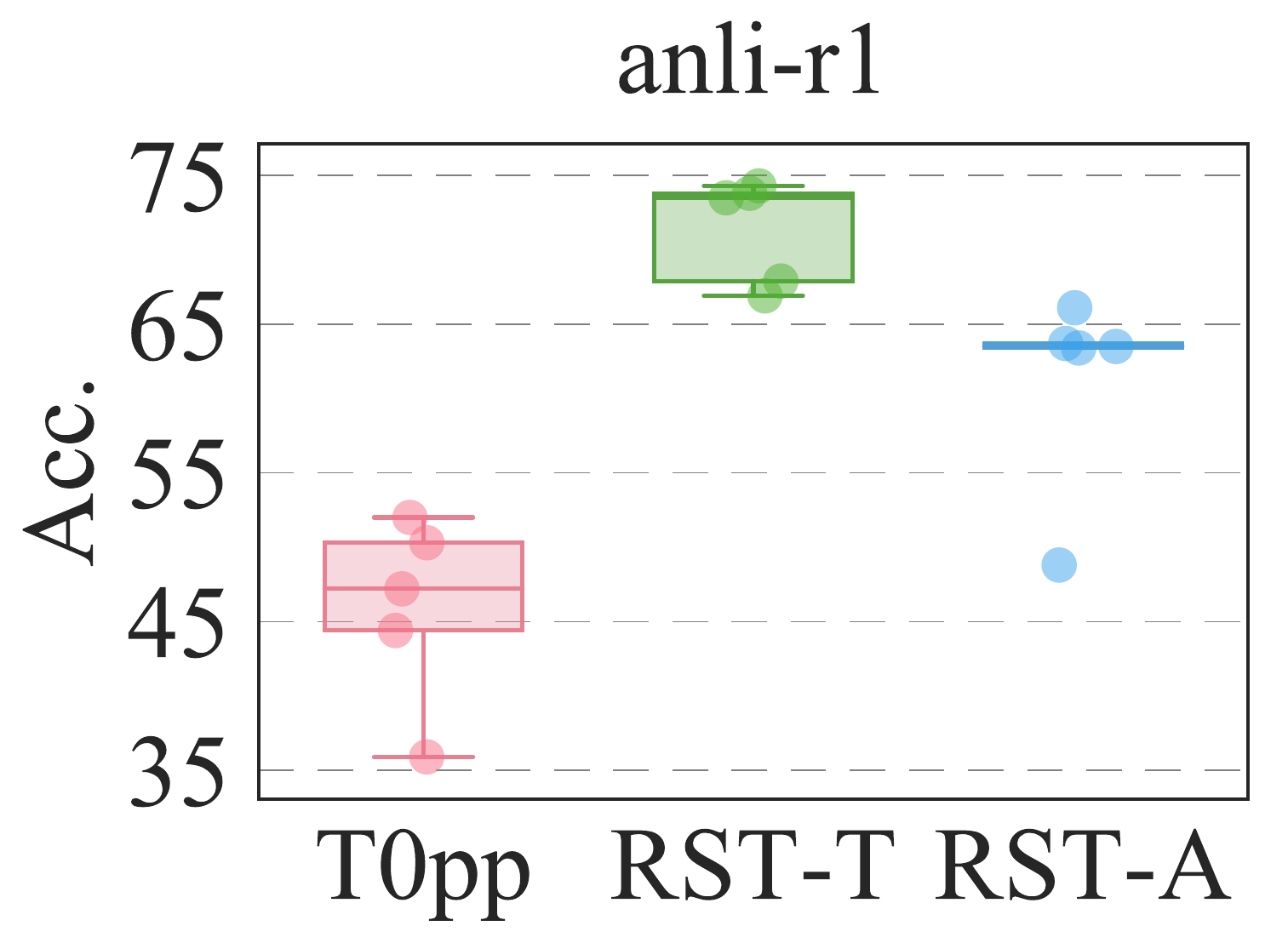}}  & \multirow{7}[1]{*}{\includegraphics[scale=0.22]{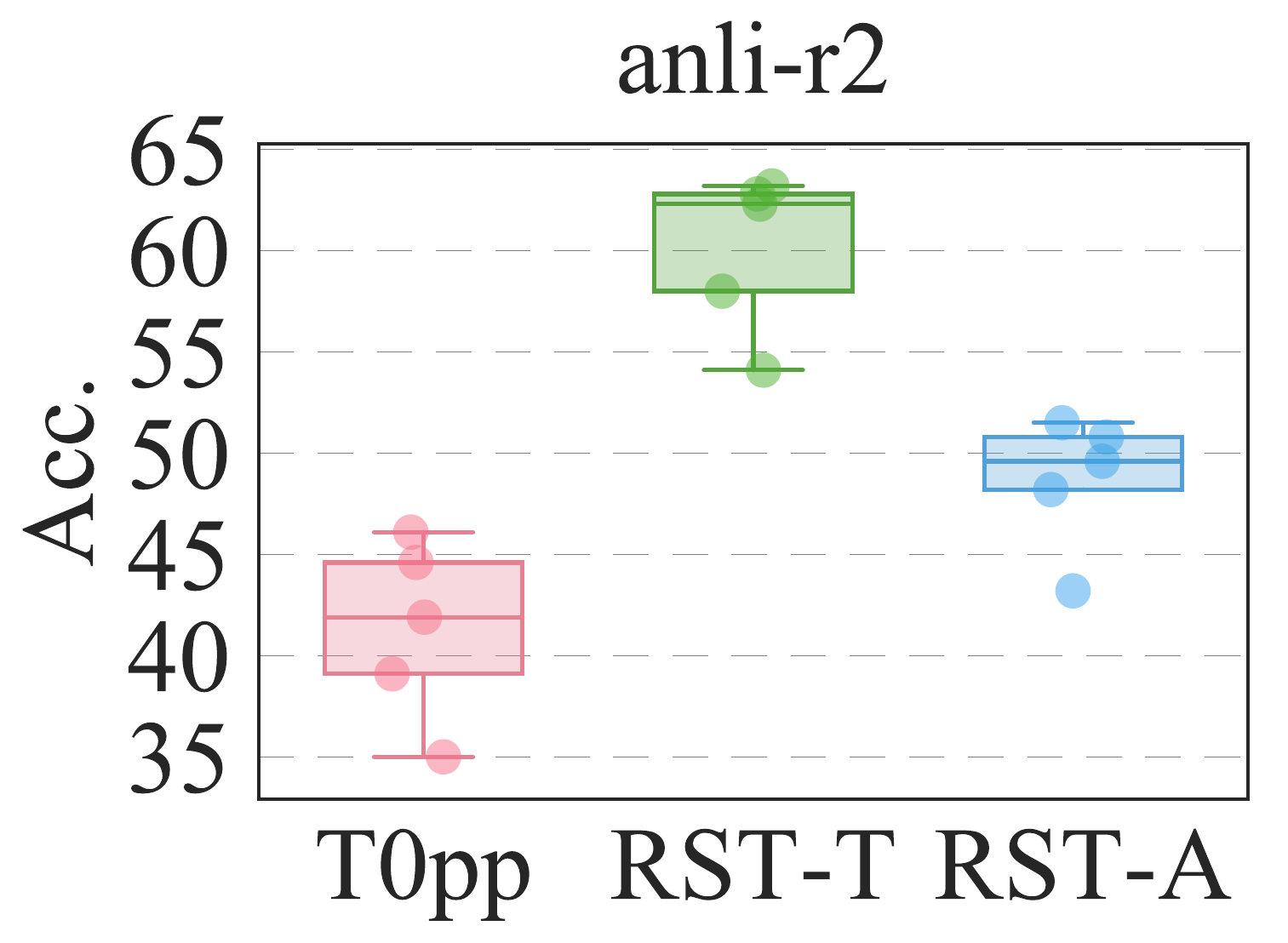}} & \multirow{7}[1]{*}{\includegraphics[scale=0.22]{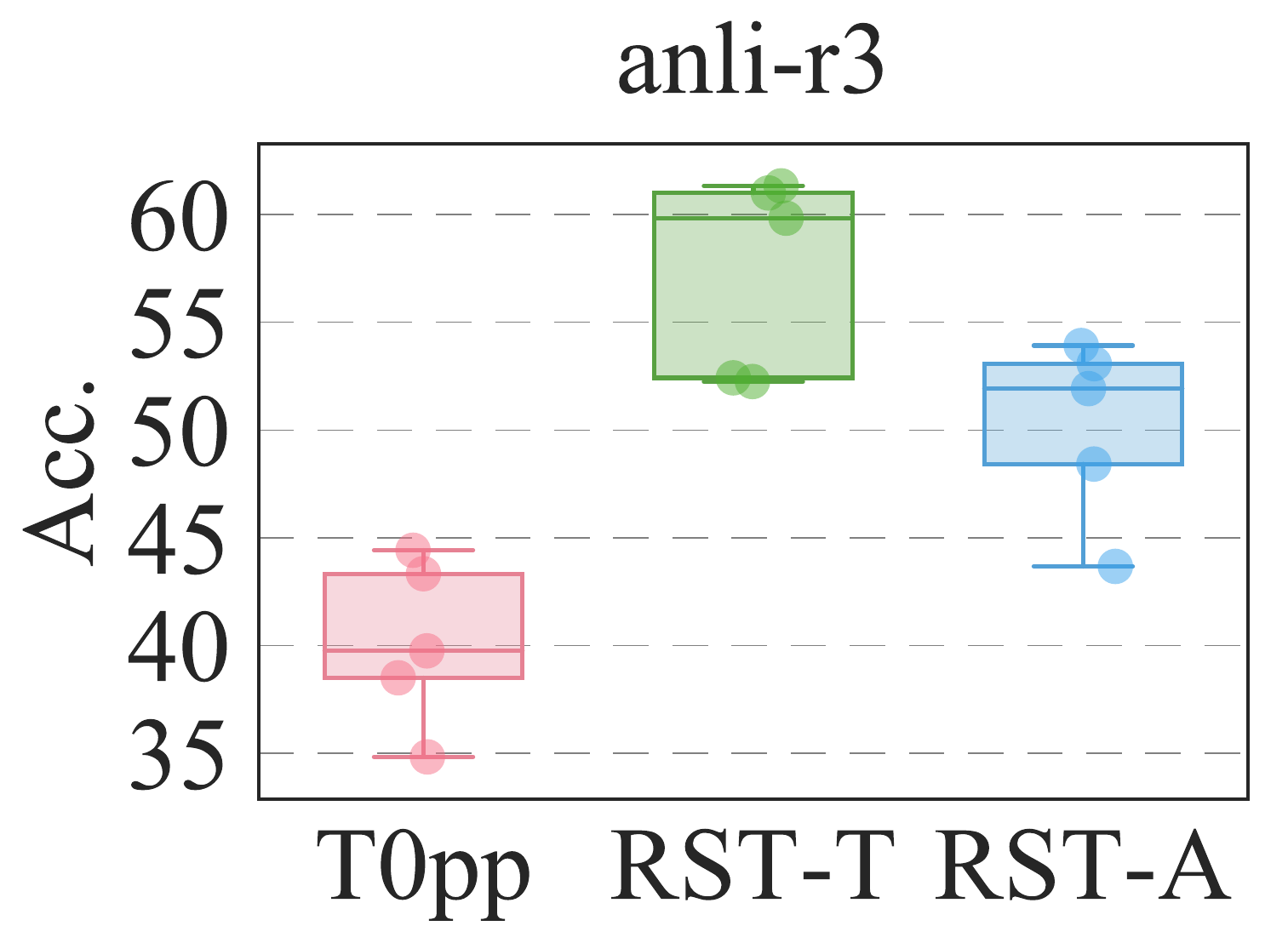}} & \multirow{7}[1]{*}{\includegraphics[scale=0.22]{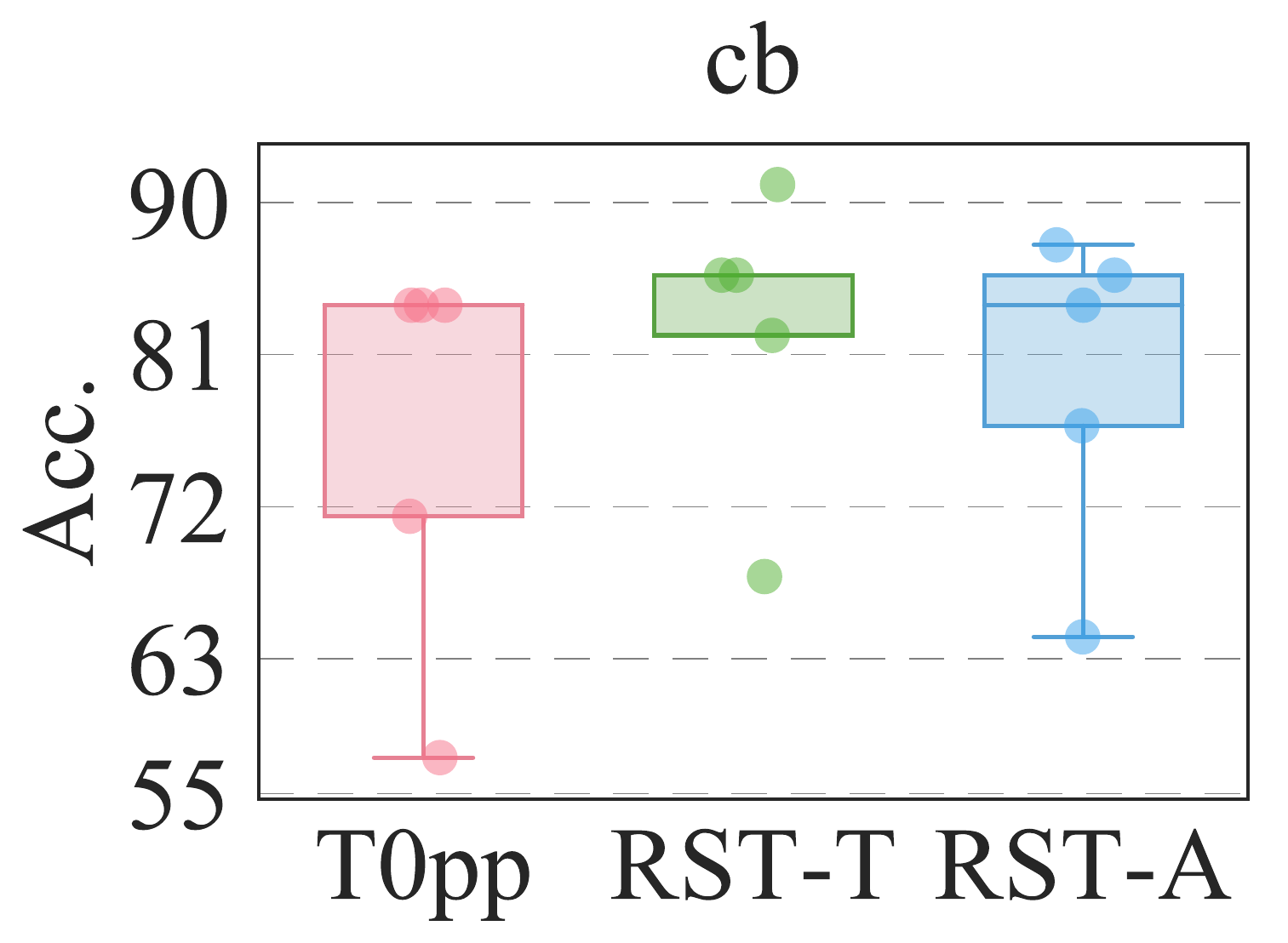}}   \\
                                                                                                   &                    \\
                                                                                                   &                    \\
                                                                                                   &                    \\
                                                                                                   &                    \\
                                                                                                   &                    \\
                                                                                                   &                    \\                                             &
 \multirow{7}[1]{*}{\includegraphics[scale=0.22]{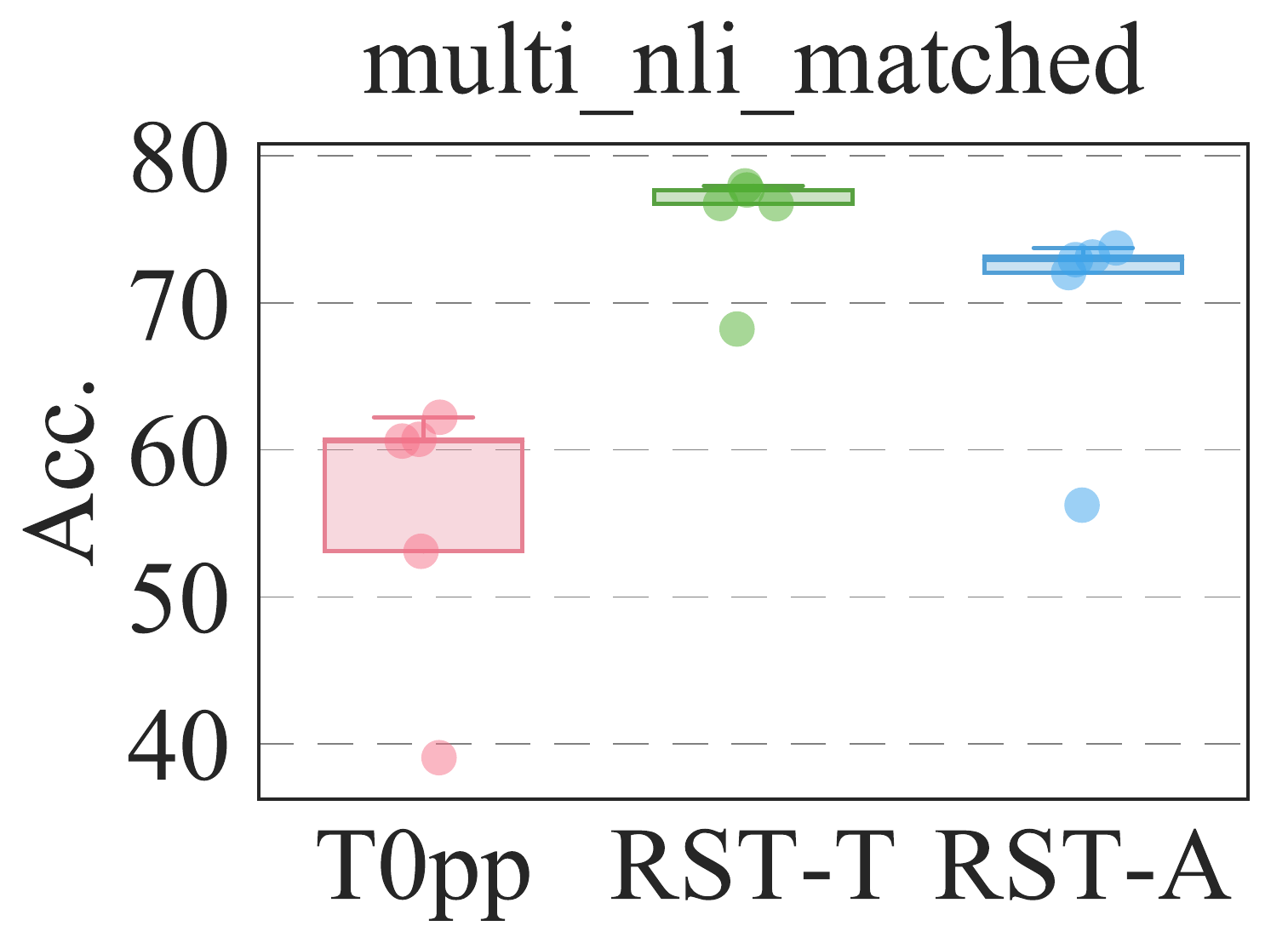}}  & \multirow{7}[1]{*}{\includegraphics[scale=0.22]{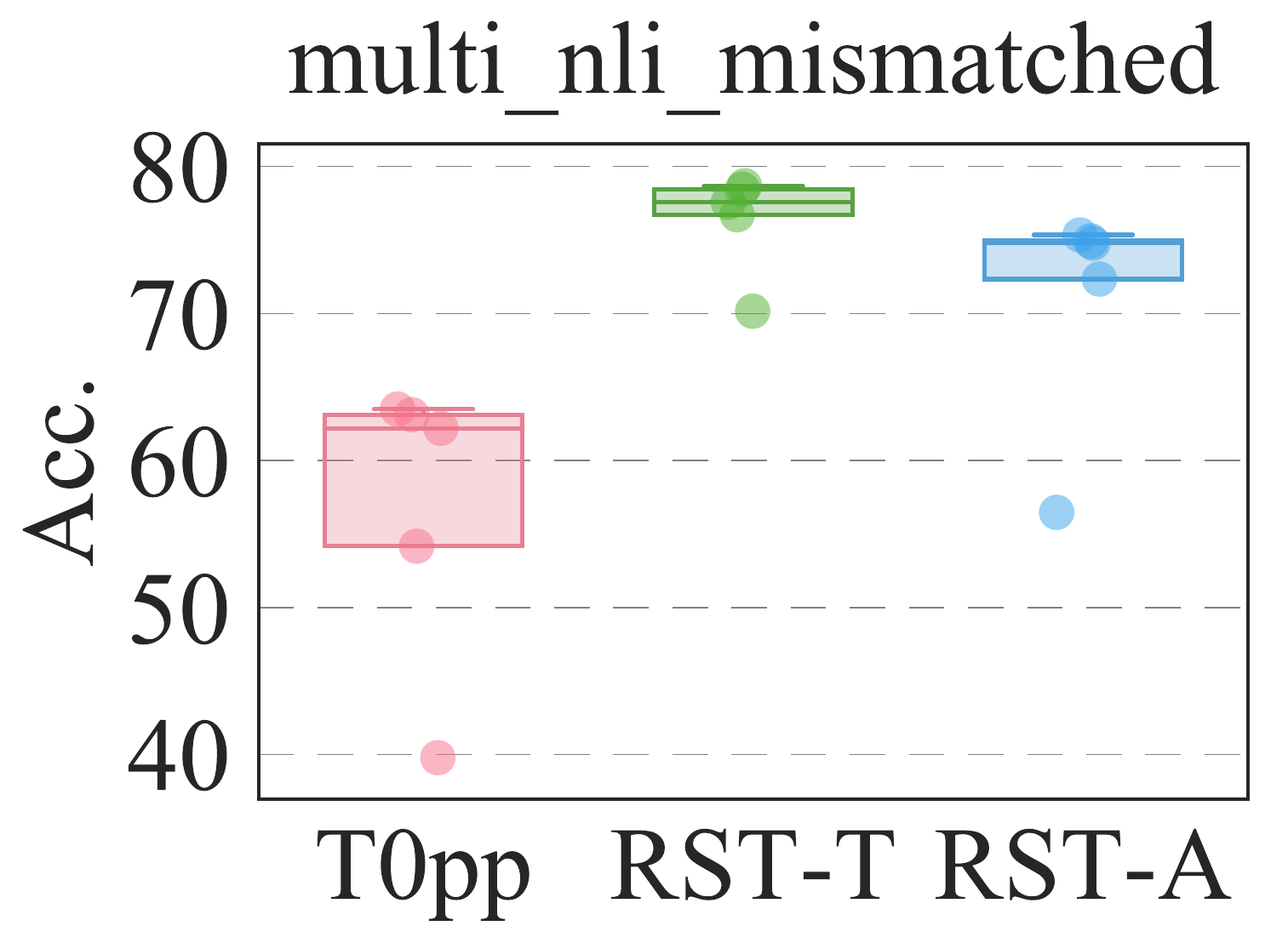}} & \multirow{7}[1]{*}{\includegraphics[scale=0.22]{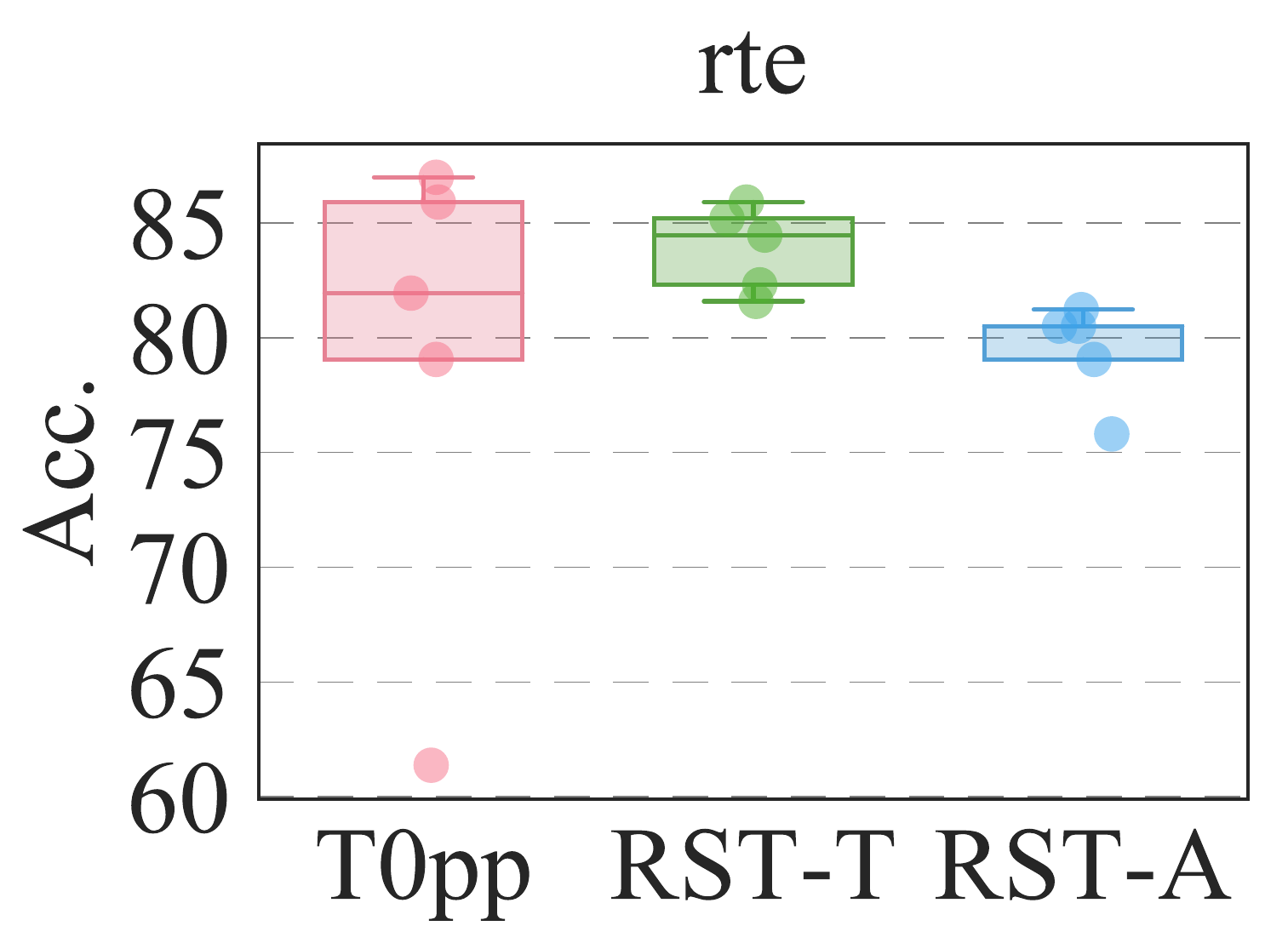}} & \multirow{7}[1]{*}{\includegraphics[scale=0.22]{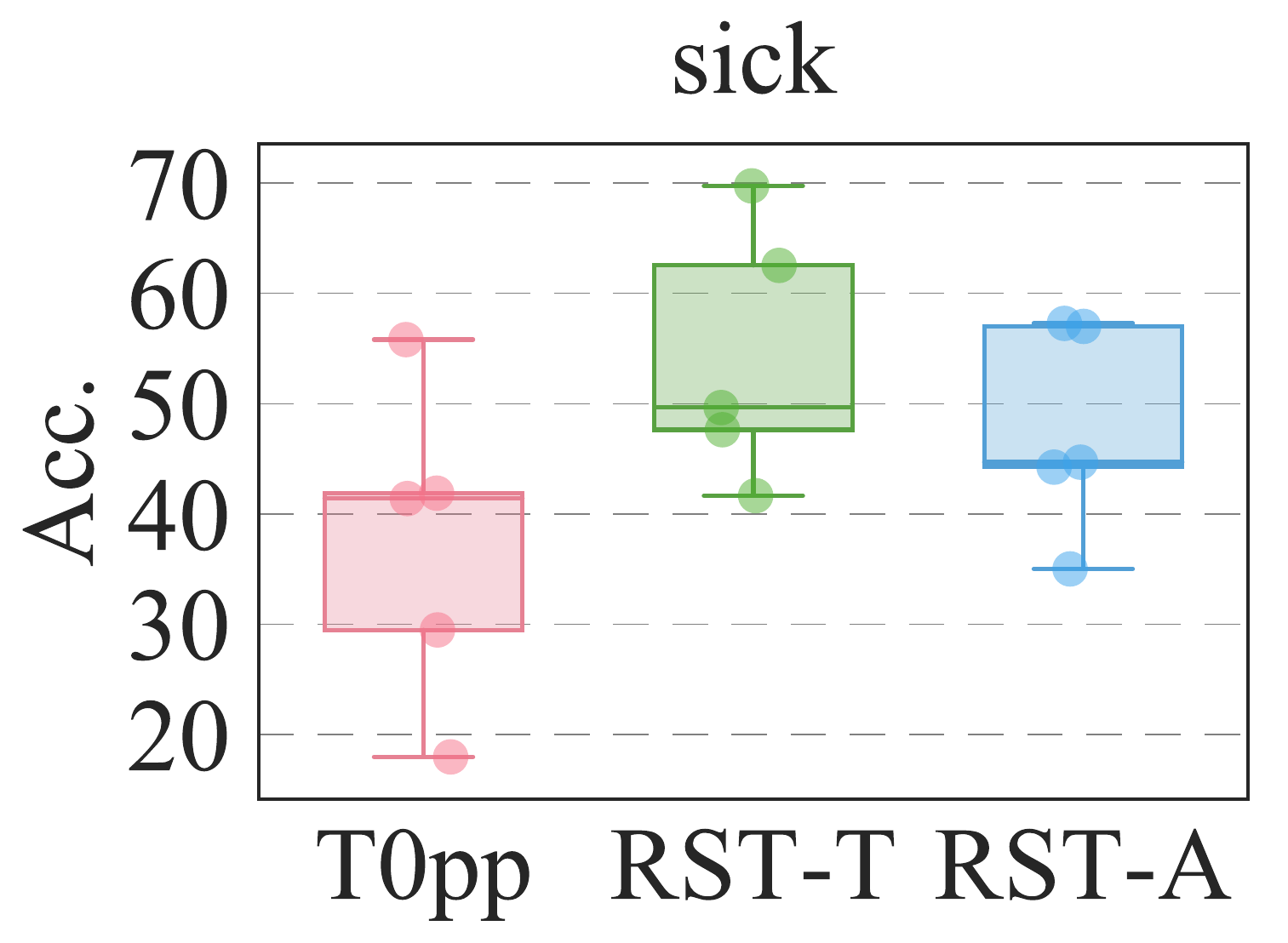}}\\
 \\
 \\
 \\
 \\
 \\
 \\
 & \multirow{7}[1]{*}{\includegraphics[scale=0.22]{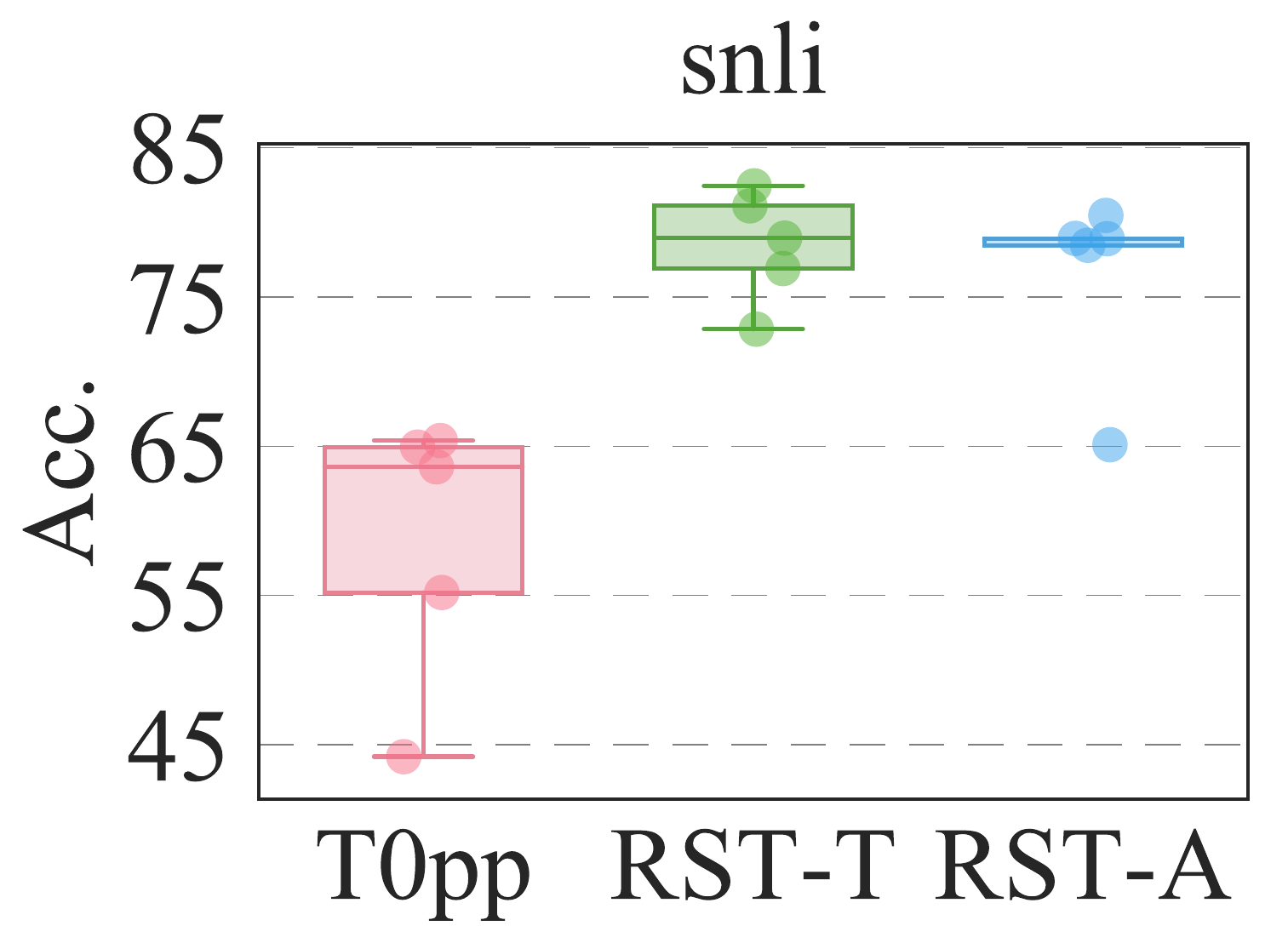}}  &  &  & \\
 \\
 \\
 \\
 \\
 \\
 \\
                                                                                                   \bottomrule
\end{tabular}
\end{table}

\begin{table}[!ht]
\centering
\footnotesize
 \setlength\tabcolsep{1pt}
  \caption{The model performance corresponding to each prompt on all datasets (Part 2). ``RST-T" represents the RST-Task model and ``RST-A" represents the RST-All model.}
 \label{tab:prompt_result2}
\begin{tabular}{l|llll}
\toprule
\multirow{22}{*}{\textbf{\begin{tabular}[c]{@{}l@{}}Intent \\ Detection\end{tabular}}}               & \multirow{7.5}[1]{*}{\includegraphics[scale=0.22]{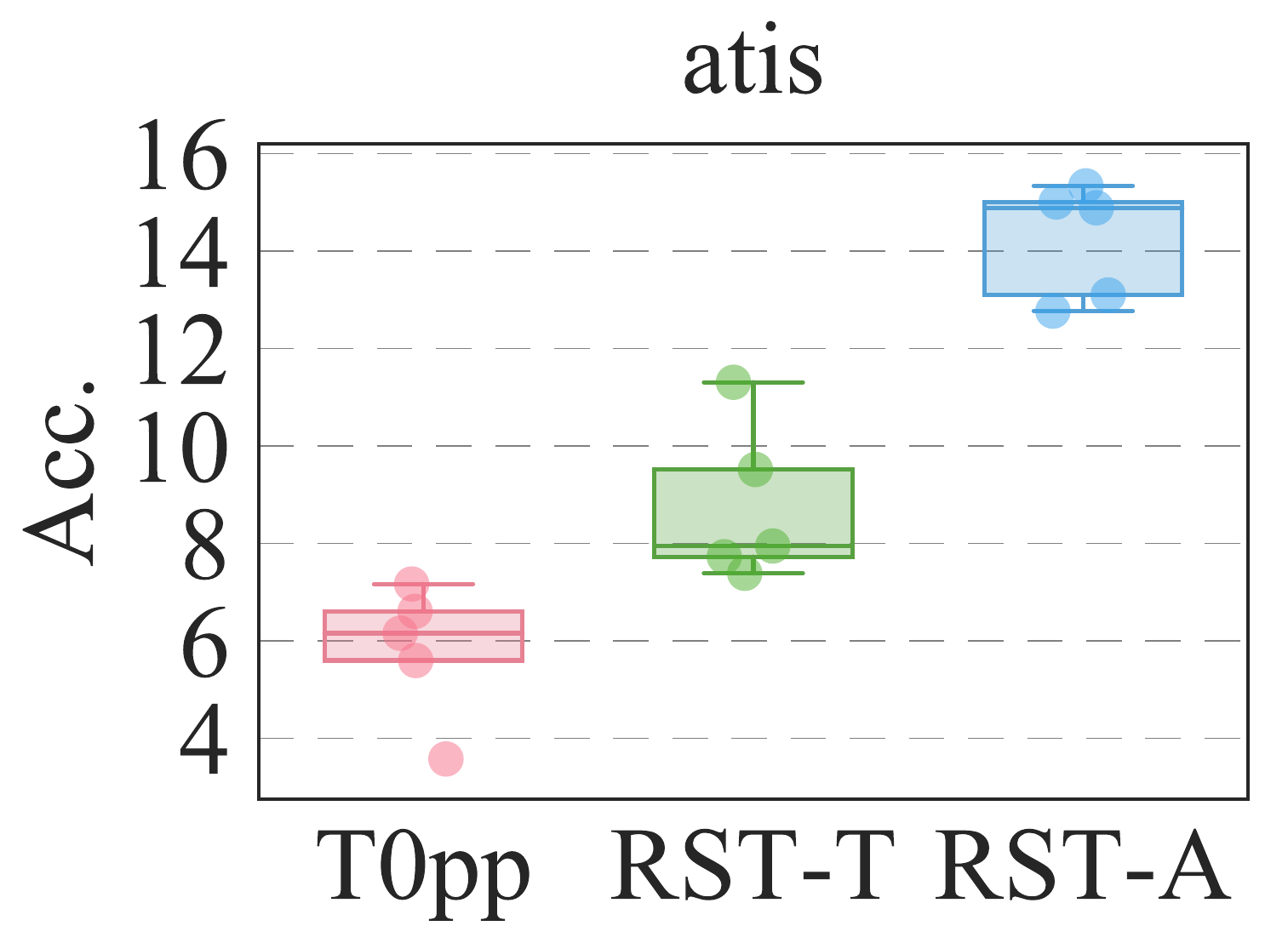}}  & \multirow{7.5}[1]{*}{\includegraphics[scale=0.22]{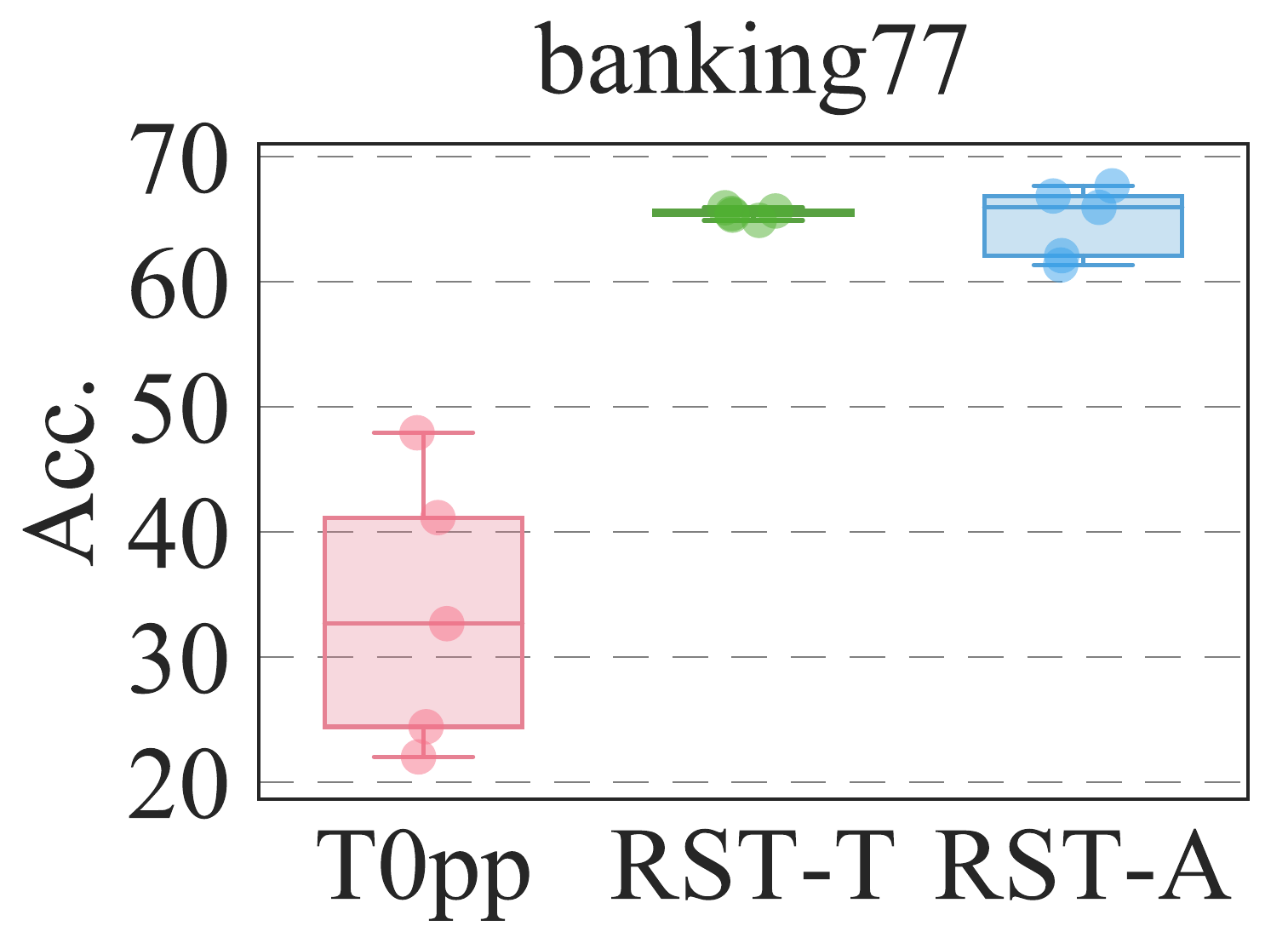}} & \multirow{7.5}[1]{*}{\includegraphics[scale=0.22]{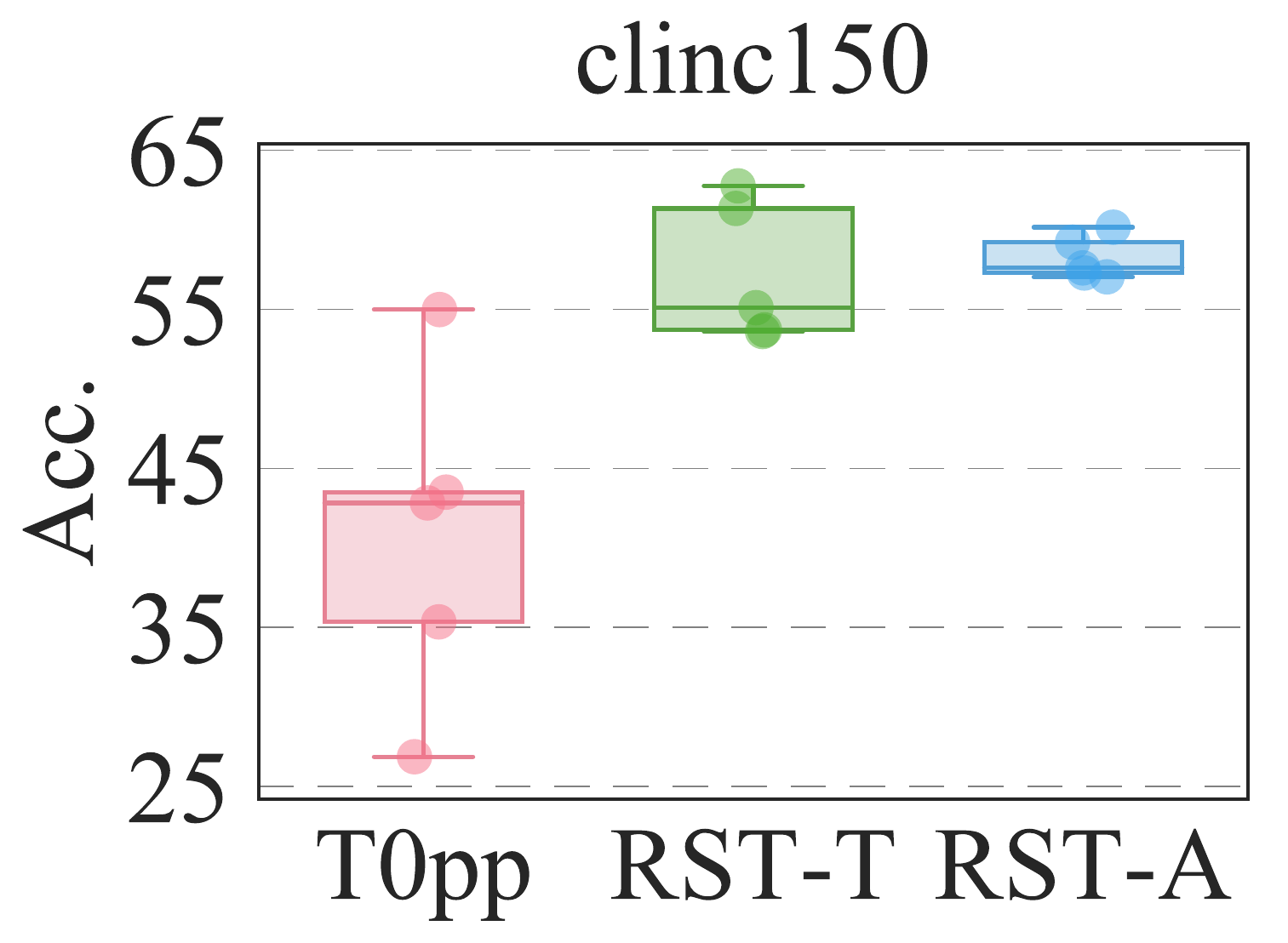}} & \multirow{7.5}[1]{*}{\includegraphics[scale=0.22]{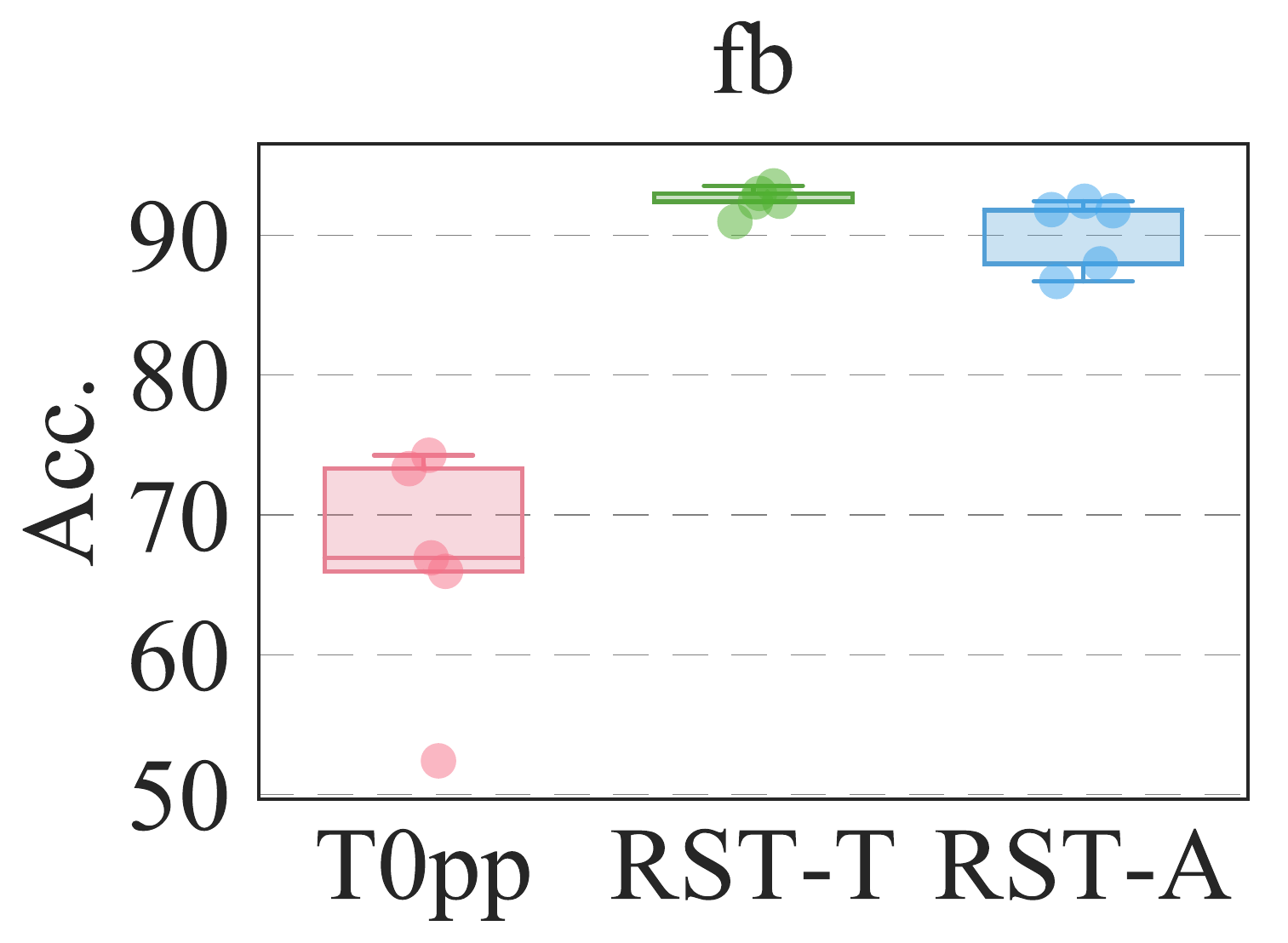}}\\
                                                                                                   &                    \\
                                                                                                   &                    \\
                                                                                                   &                    \\
                                                                                                   &                    \\
                                                                                                   &                    \\
                                                                                                   &                    \\
                                                                                                   &                                              \multirow{8}[1]{*}{\includegraphics[scale=0.22]{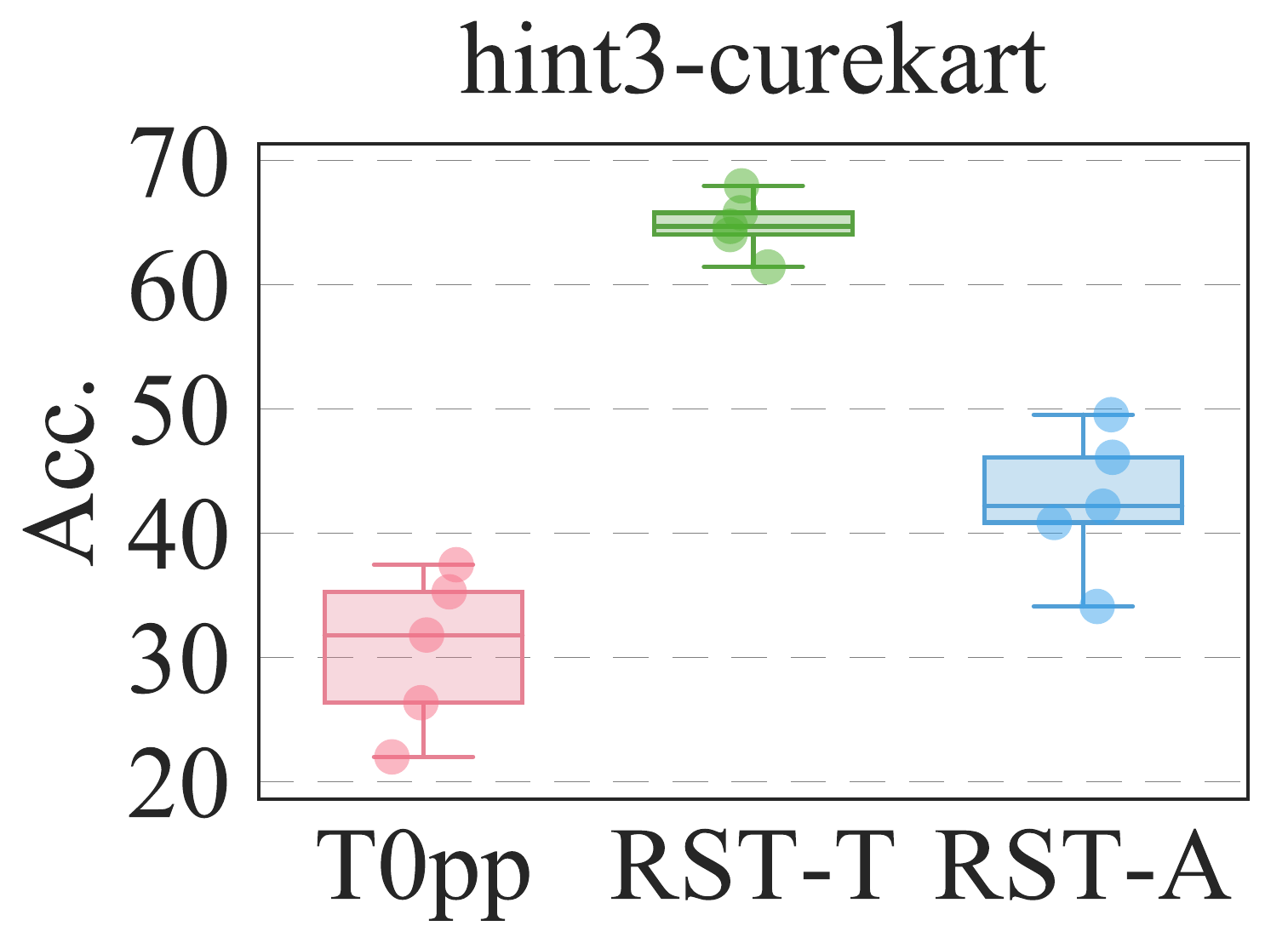}}  & \multirow{8}[1]{*}{\includegraphics[scale=0.22]{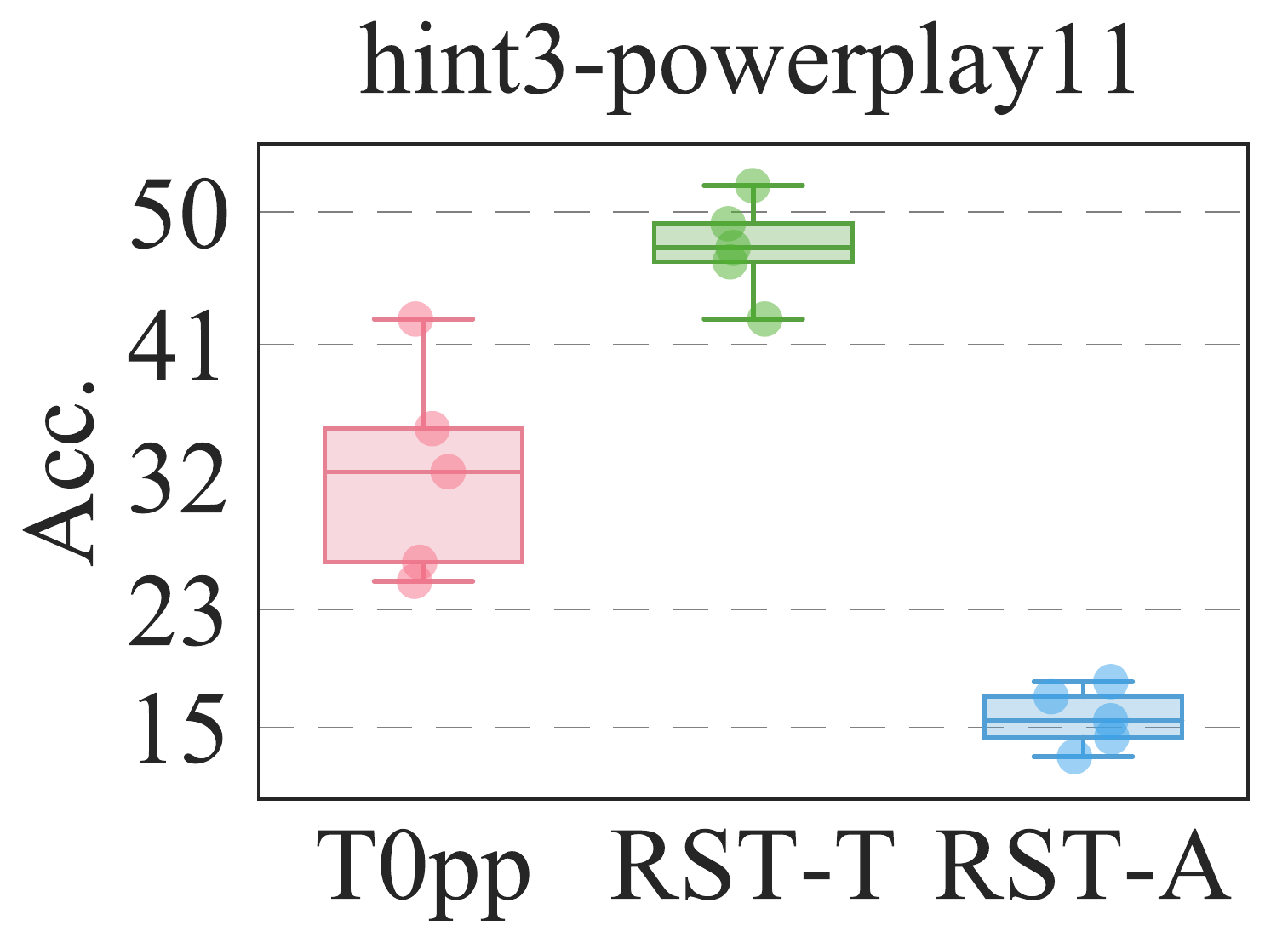}} & \multirow{8}[1]{*}{\includegraphics[scale=0.22]{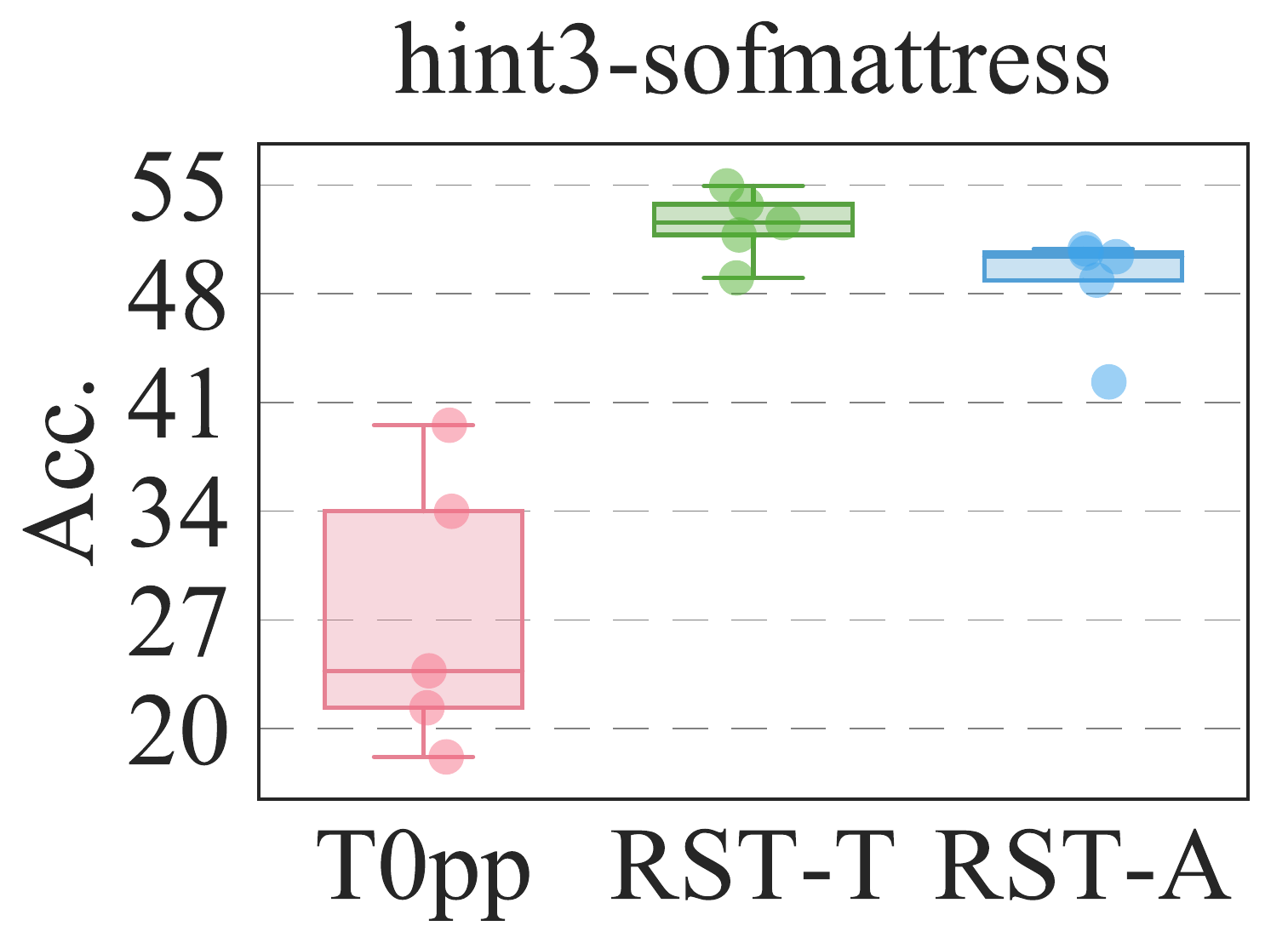}} & \multirow{8}[1]{*}{\includegraphics[scale=0.22]{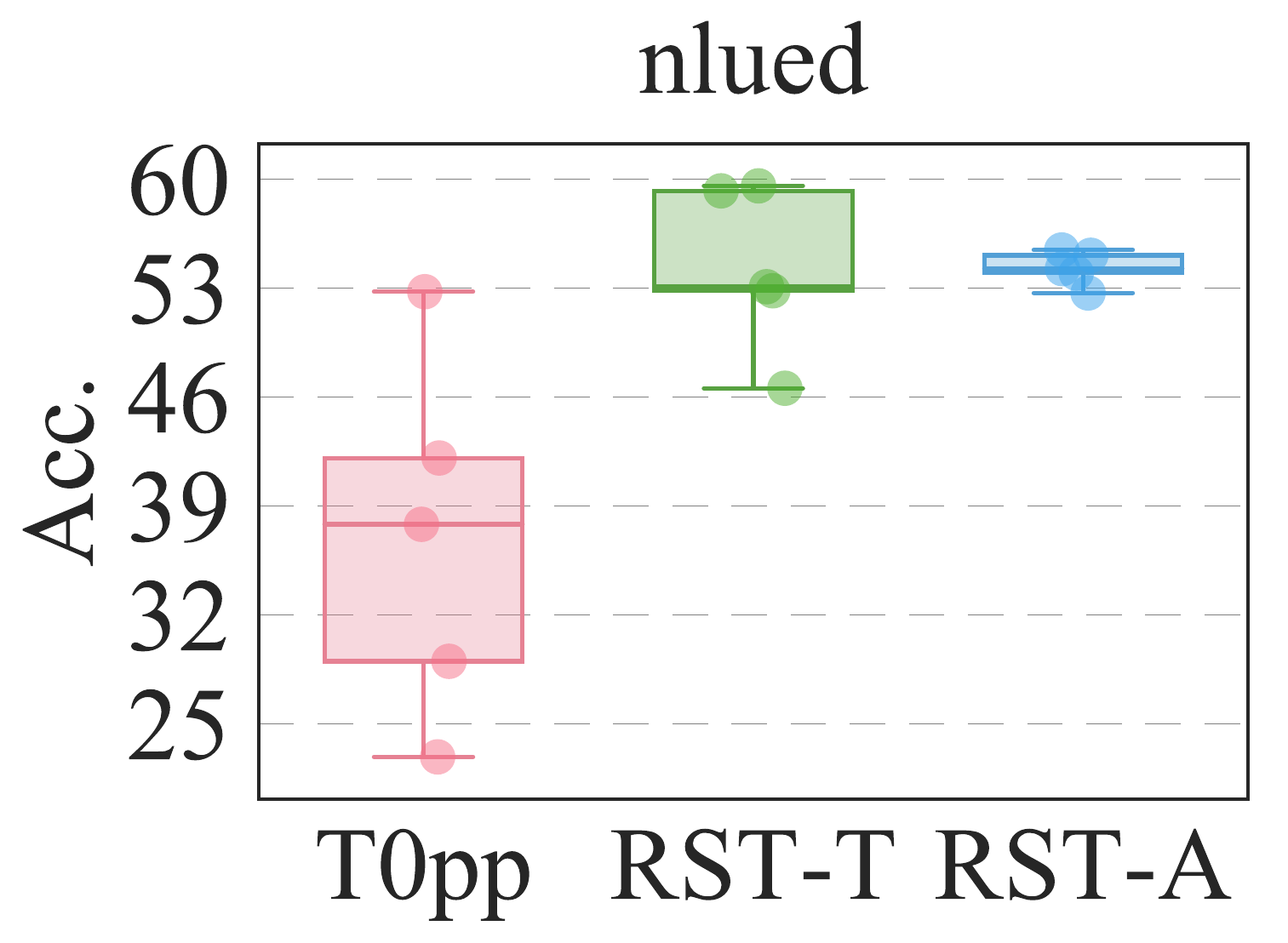}}\\
                                                                                                        \\
\\ 
\\
\\
\\
\\
   &                                              \multirow{8}[1]{*}{\includegraphics[scale=0.22]{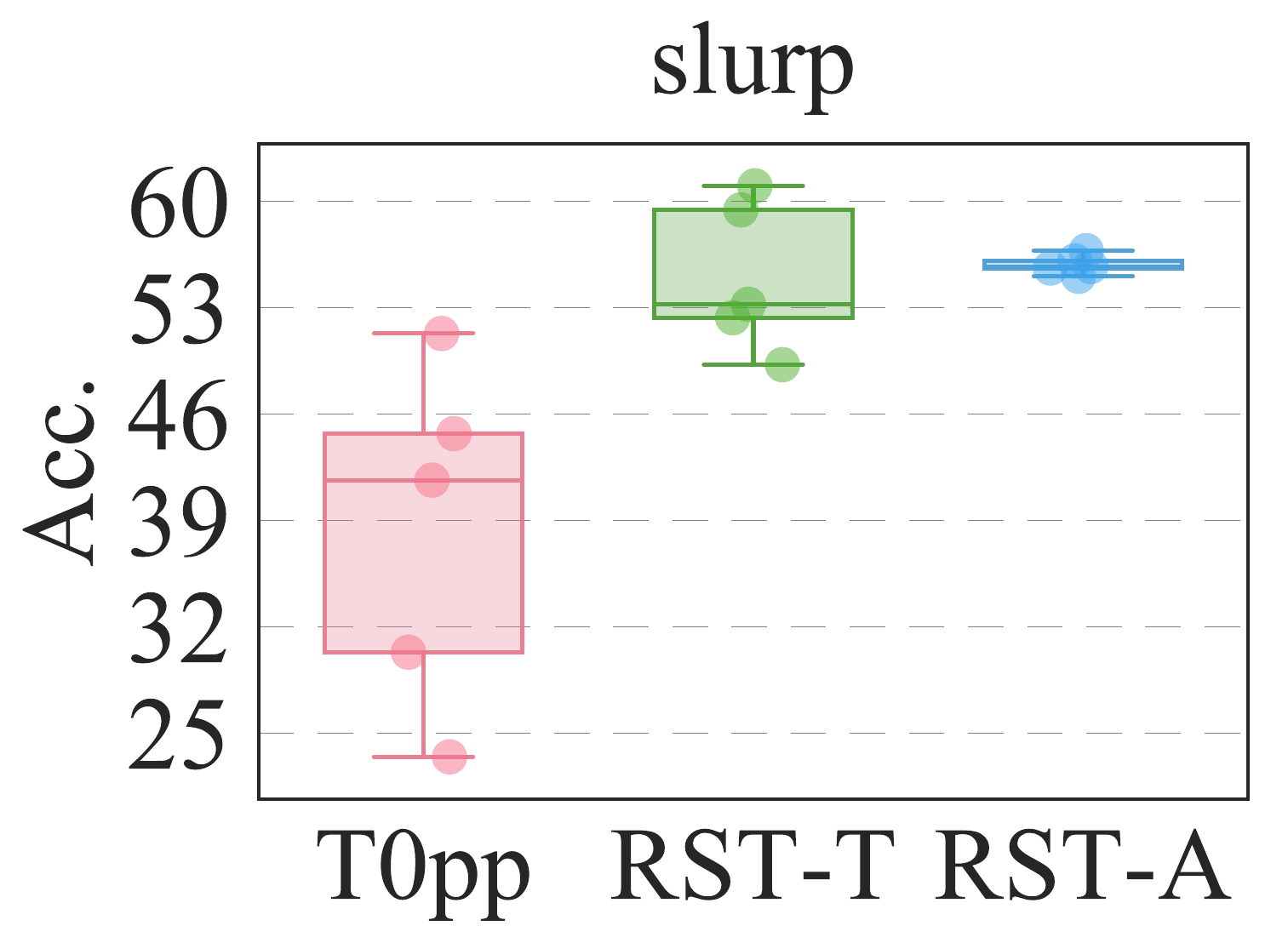}}  & \multirow{8}[1]{*}{\includegraphics[scale=0.22]{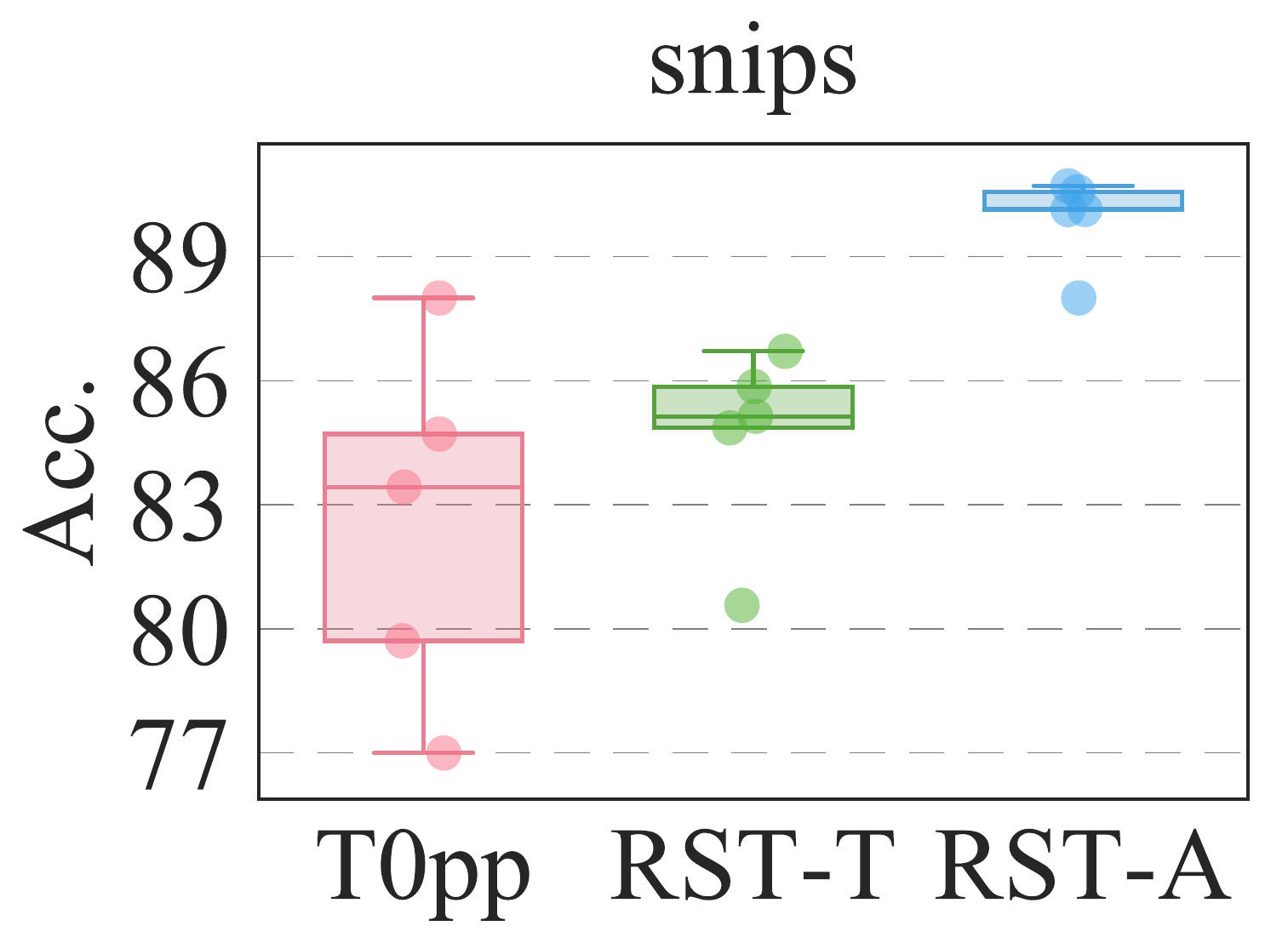}} &  & \\
   \\
   \\
   \\
   \\
   \\
   \\
   \\
\midrule
\multirow{8}{*}{\textbf{\begin{tabular}[c]{@{}l@{}}Fact Retrieval\end{tabular}}}                                                                                      &                                              \multirow{7.5}[1]{*}{\includegraphics[scale=0.22]{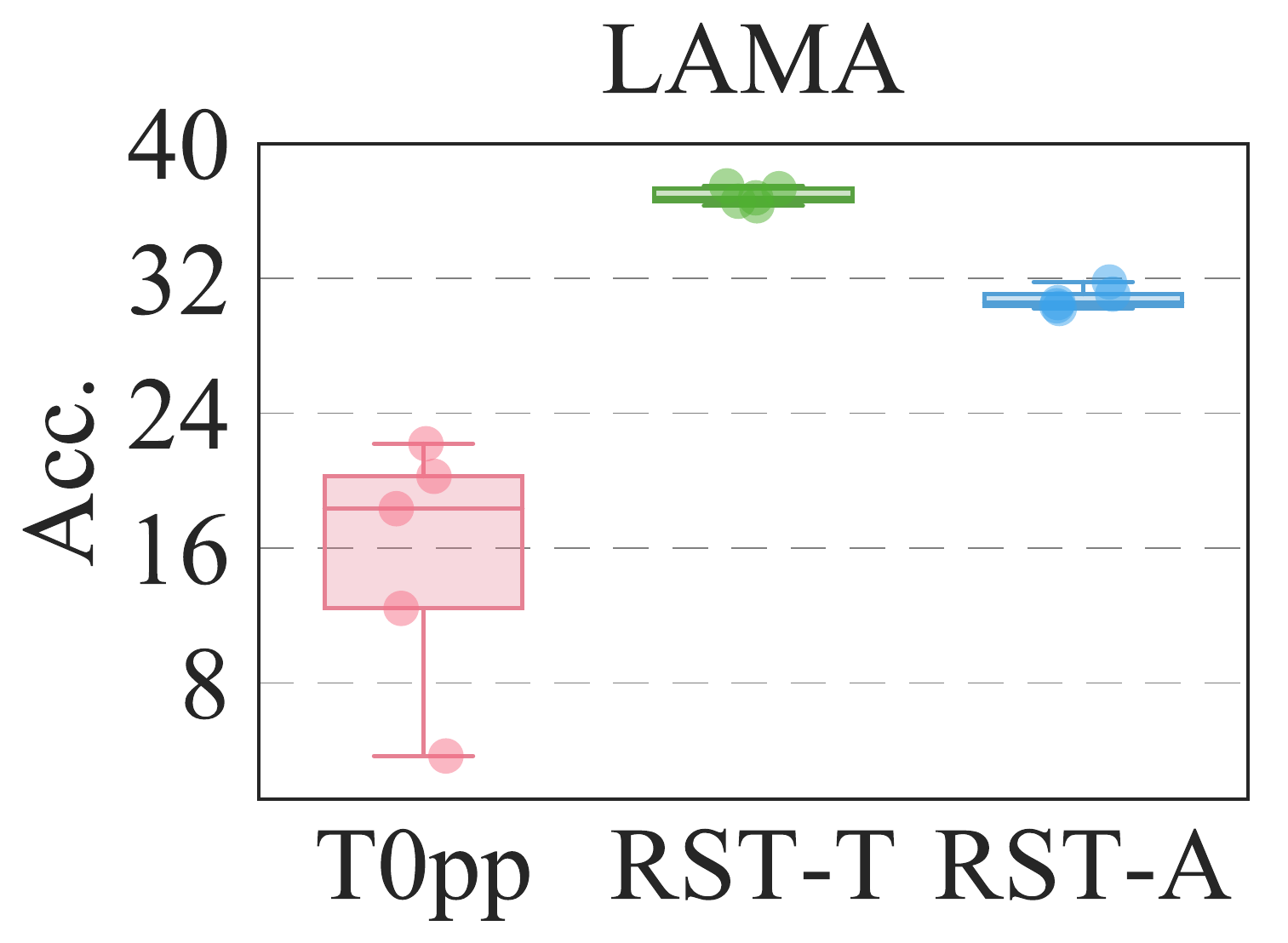}}  &  & & \\
\\
\\
\\
\\
\\
\\
\\
\midrule
\multirow{8}{*}{\textbf{\begin{tabular}[c]{@{}l@{}}Temporal \\ Reasoning\end{tabular}}}              &                                              \multirow{8}[1]{*}{\includegraphics[scale=0.22]{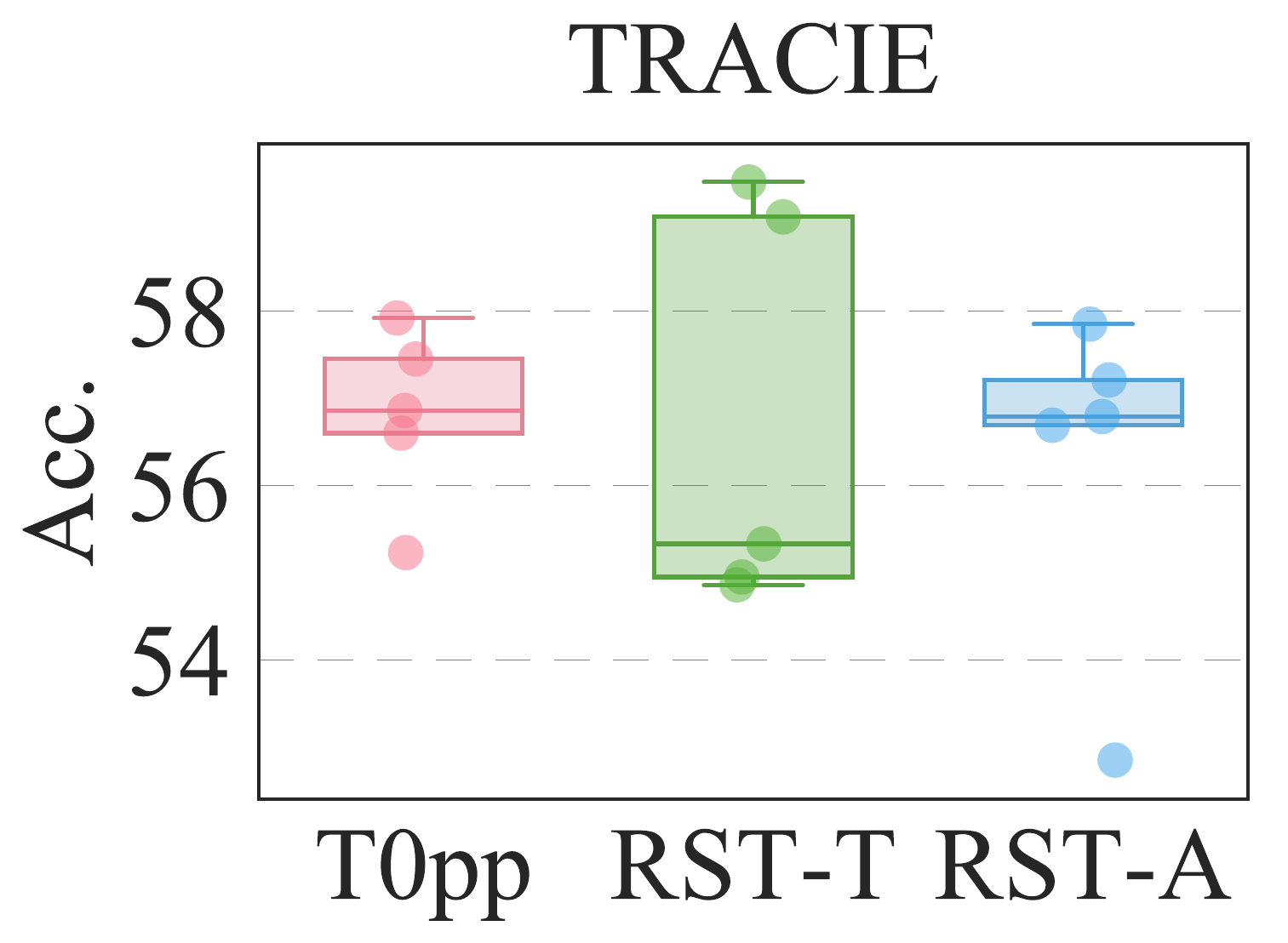}}  & \multirow{8}[1]{*}{\includegraphics[scale=0.22]{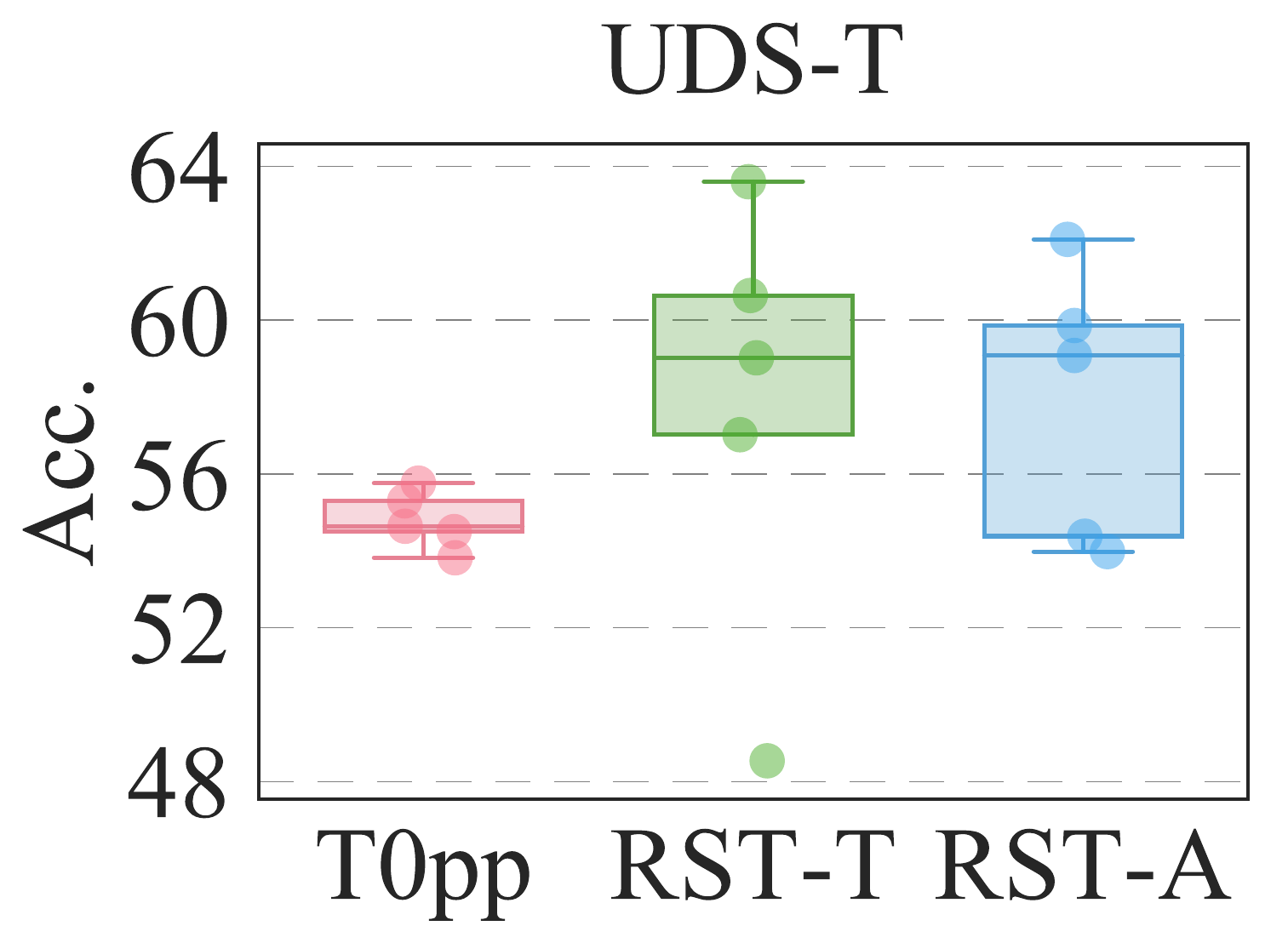}} & & \\
\\
\\
\\
\\
\\
\\
\\
                                                                             
                                                                                                   \midrule
\multirow{8}{*}{\textbf{\begin{tabular}[c]{@{}l@{}}Word Sense \\ Disambiguation\end{tabular}}}     &                                                \multirow{8}[1]{*}{\includegraphics[scale=0.22]{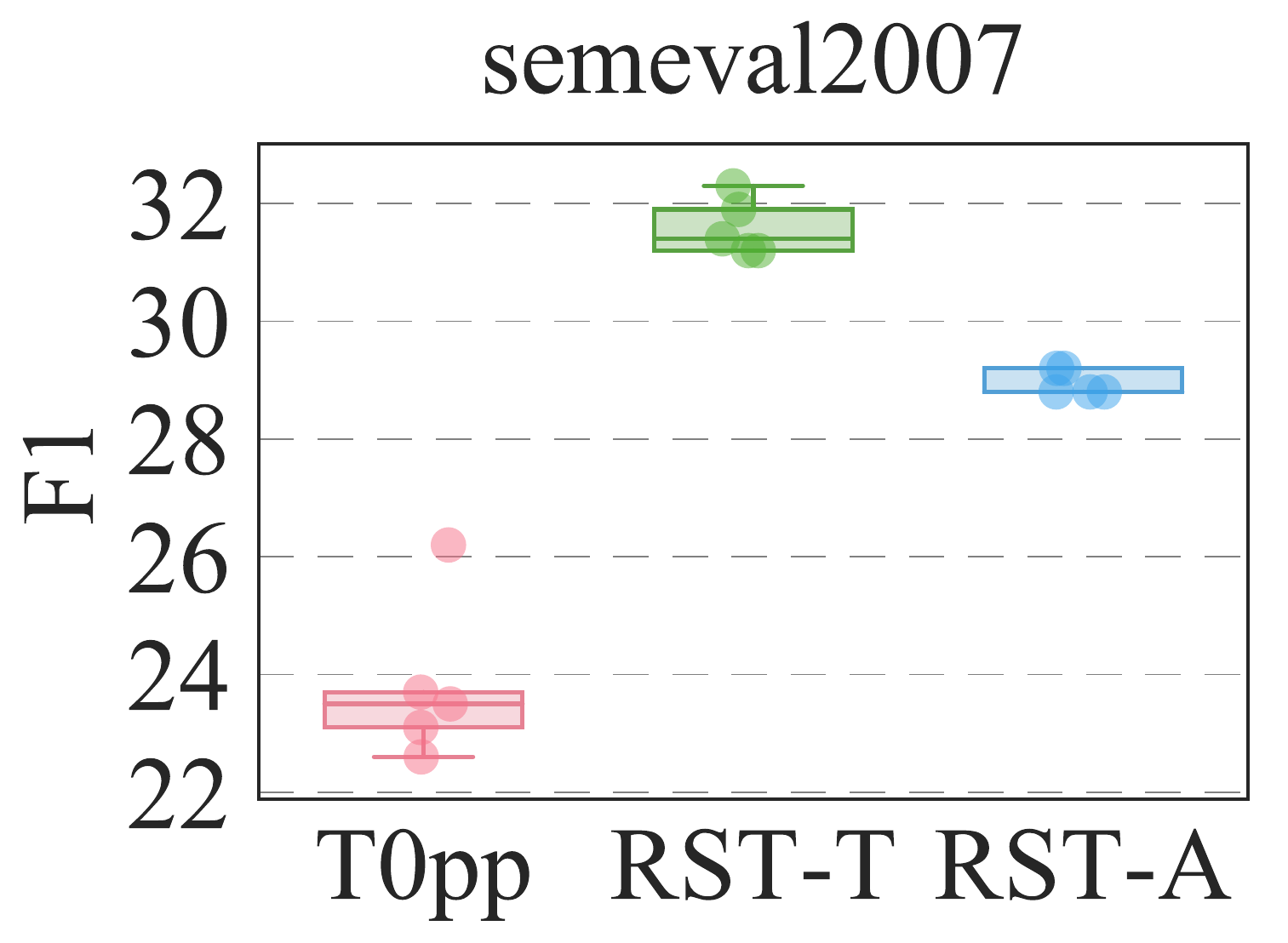}}  & \multirow{8}[1]{*}{\includegraphics[scale=0.22]{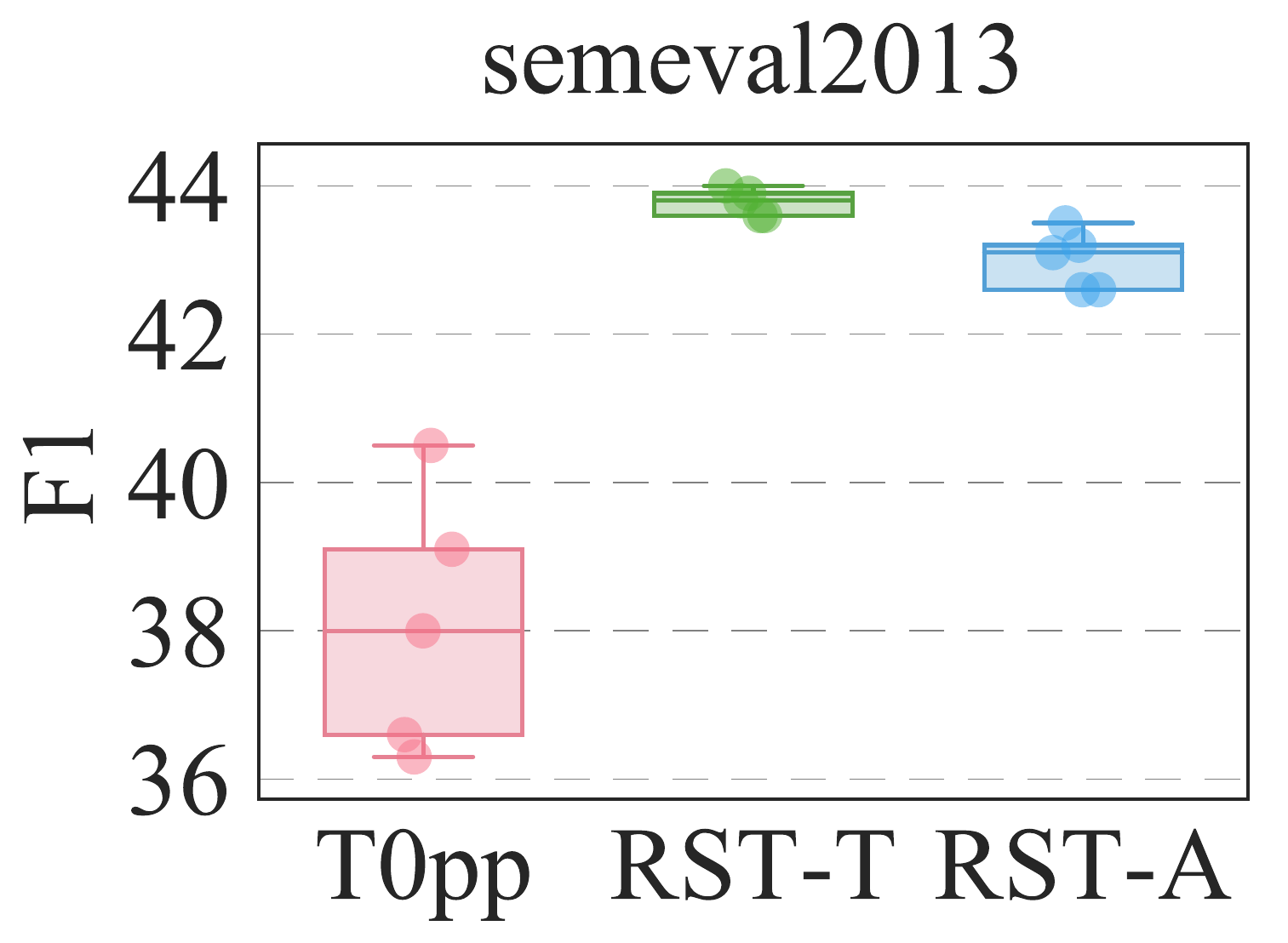}} & \multirow{8}[1]{*}{\includegraphics[scale=0.22]{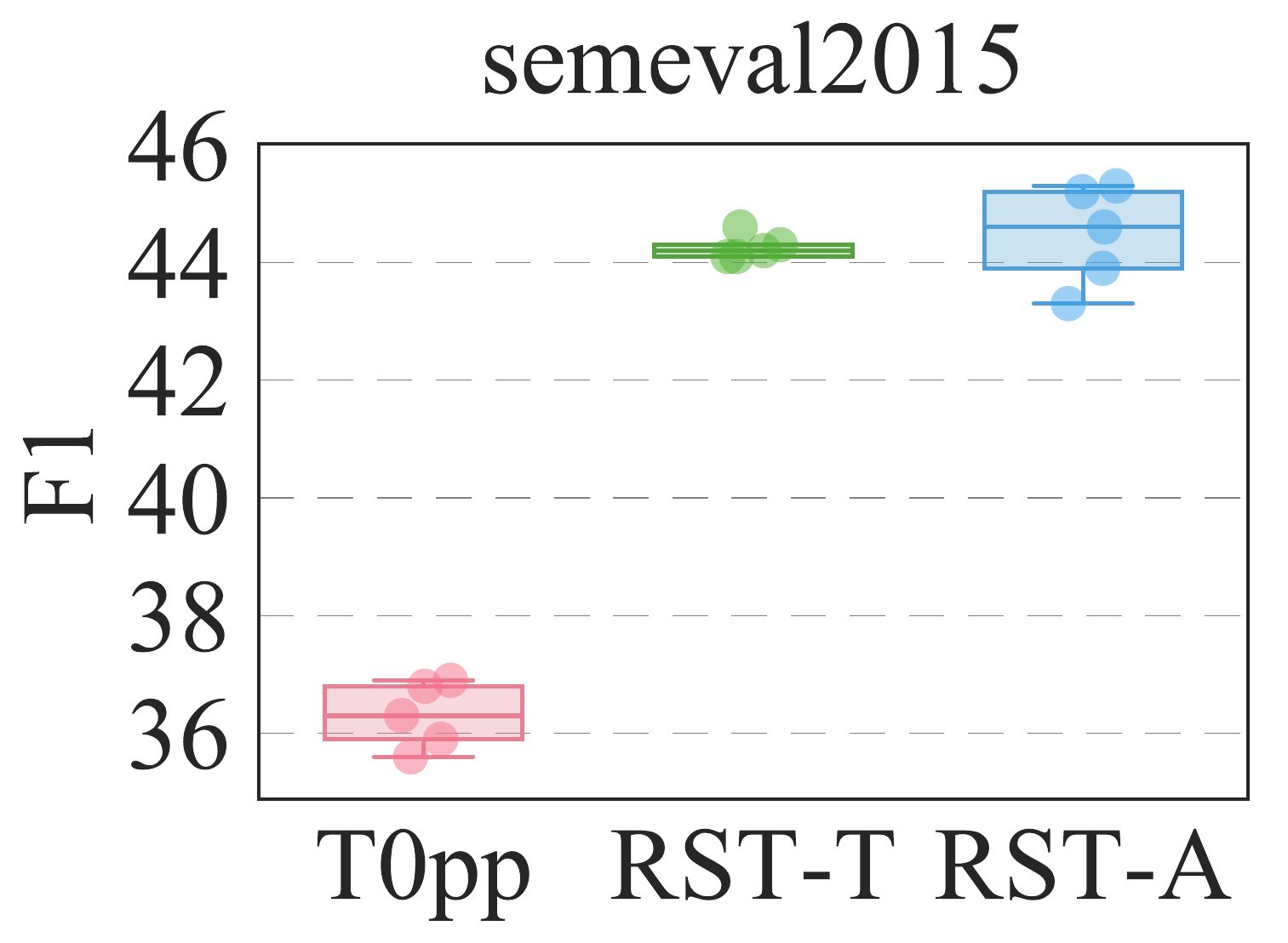}} & \multirow{8}[1]{*}{\includegraphics[scale=0.22]{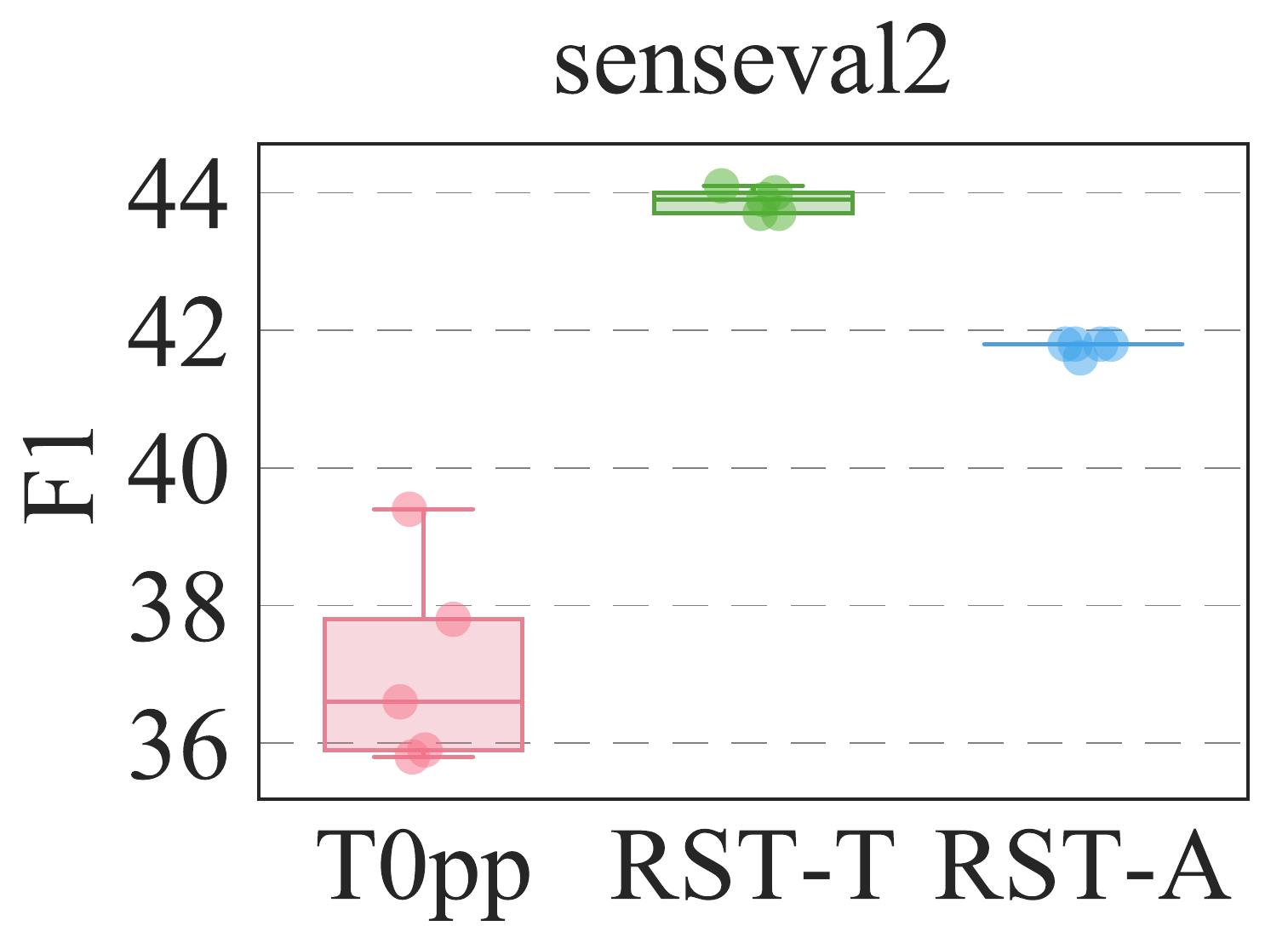}}\\  \\
                                                                                                   &                    \\
                                                                                                   &                    \\
                                                                                                   &                    \\
                                                                                                   &                    \\
                                                  \\    
                                                  \\
                                                  \midrule
\multirow{15}{*}{\textbf{Summarization}}                                                              &                                              \multirow{8}[1]{*}{\includegraphics[scale=0.22]{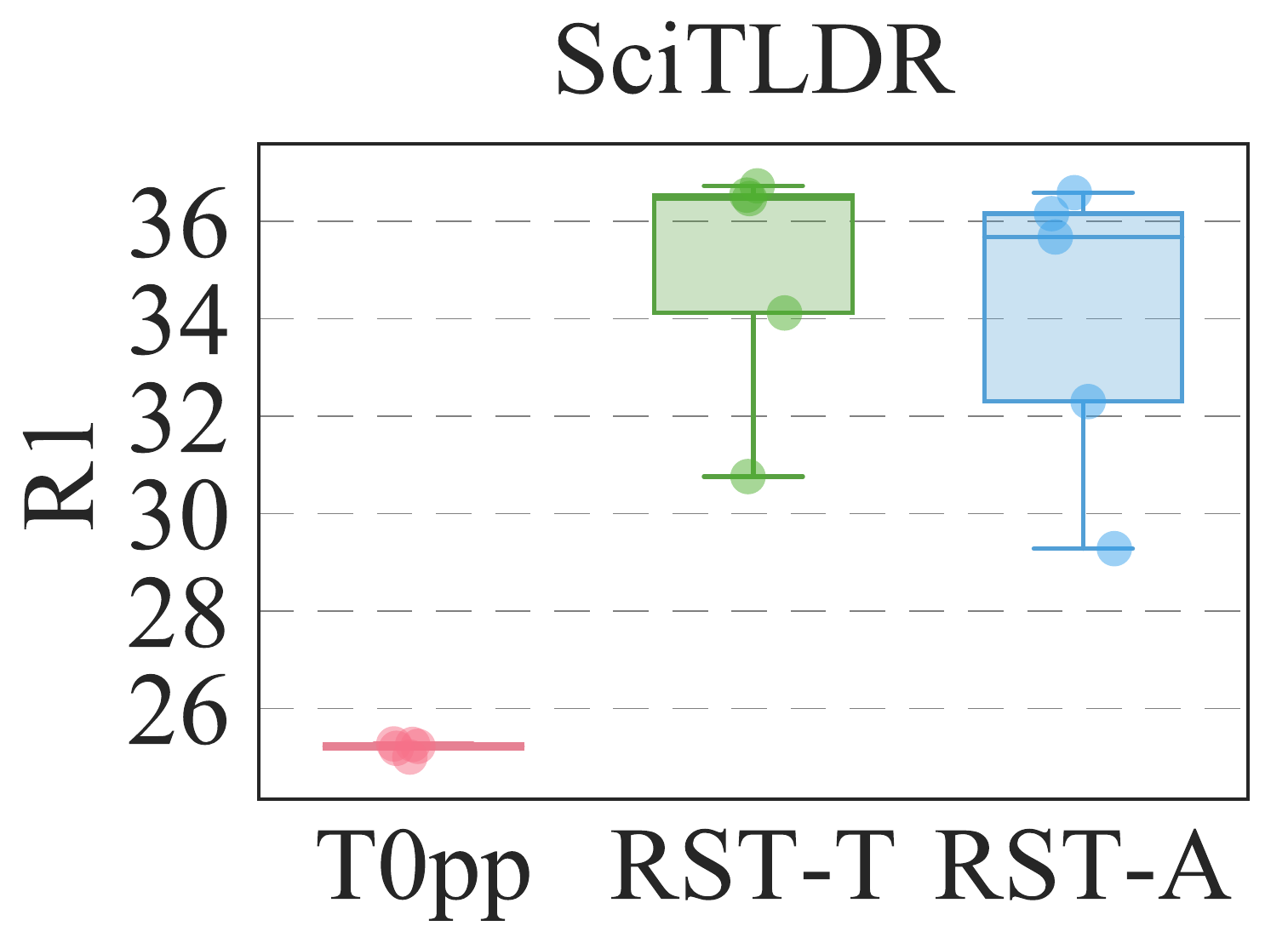}}  & \multirow{8}[1]{*}{\includegraphics[scale=0.22]{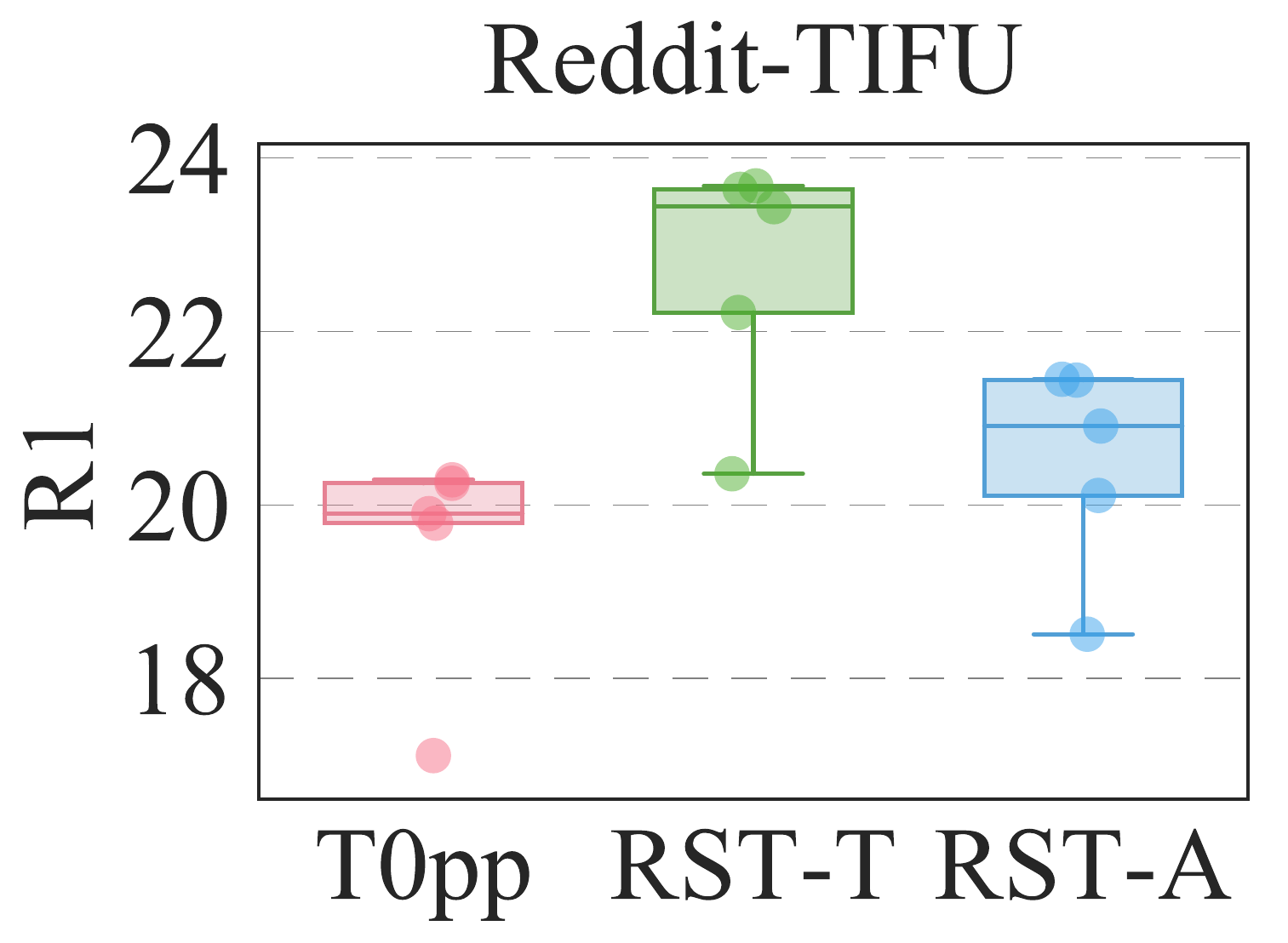}} & \multirow{8}[1]{*}{\includegraphics[scale=0.22]{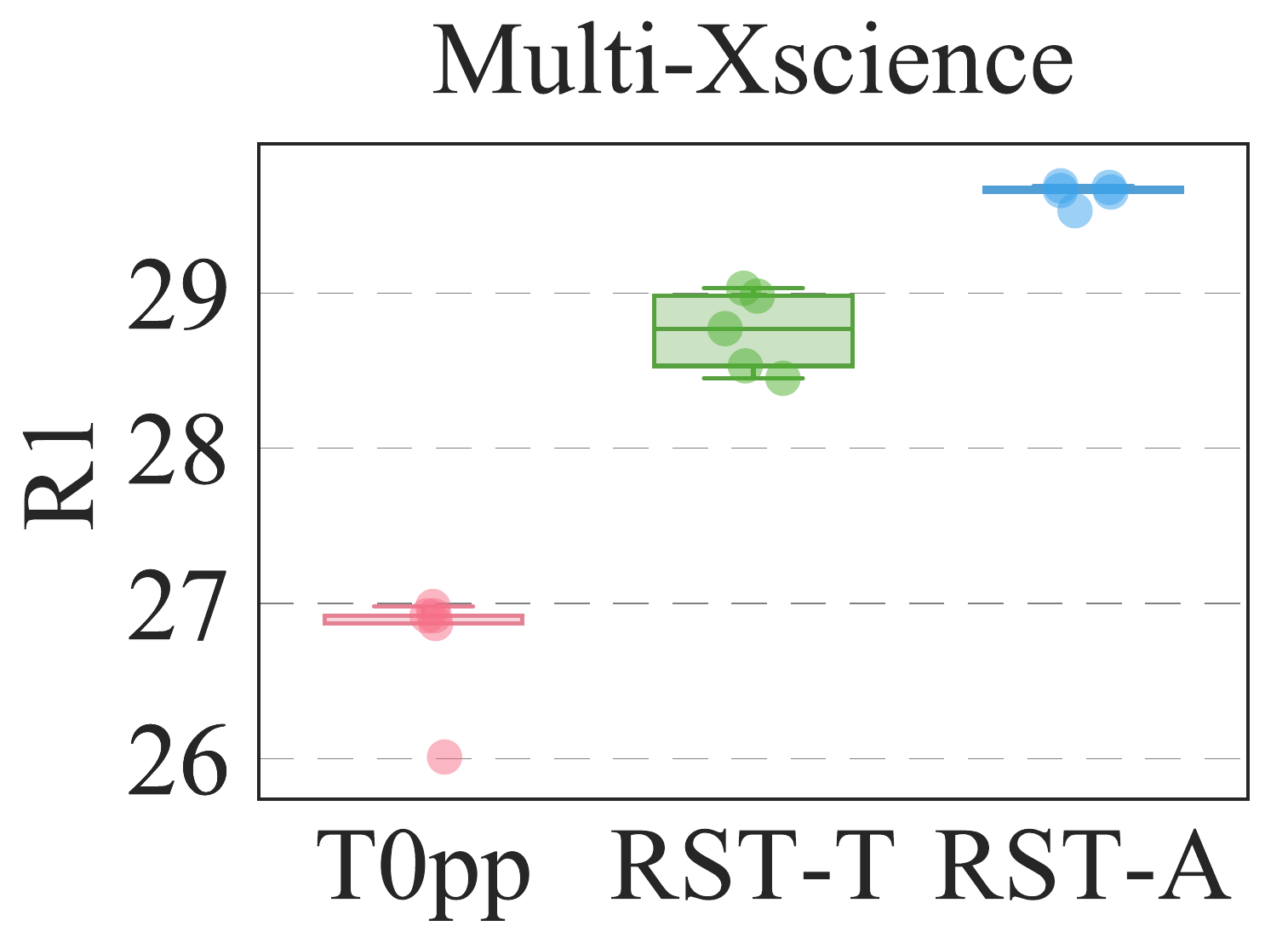}} & \multirow{8}[1]{*}{\includegraphics[scale=0.22]{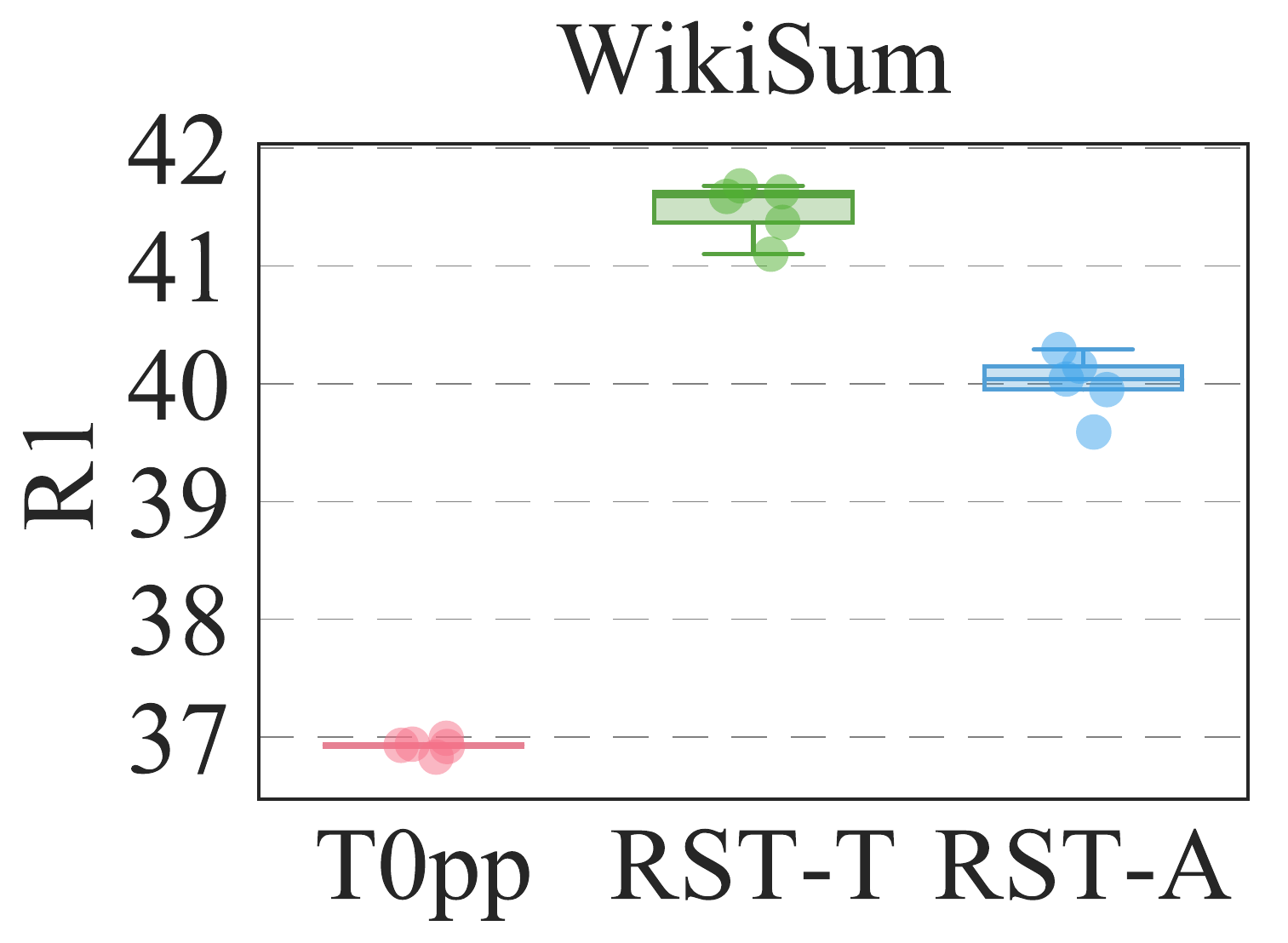}}\\  \\
                                                                                                   &                    \\
                                                                                                   &                    \\
                                                                                                   &                    \\
                                                                                                   &                    \\
                                                                                                   &                    \\
                                                                                                   &           \multirow{8}[1]{*}{\includegraphics[scale=0.22]{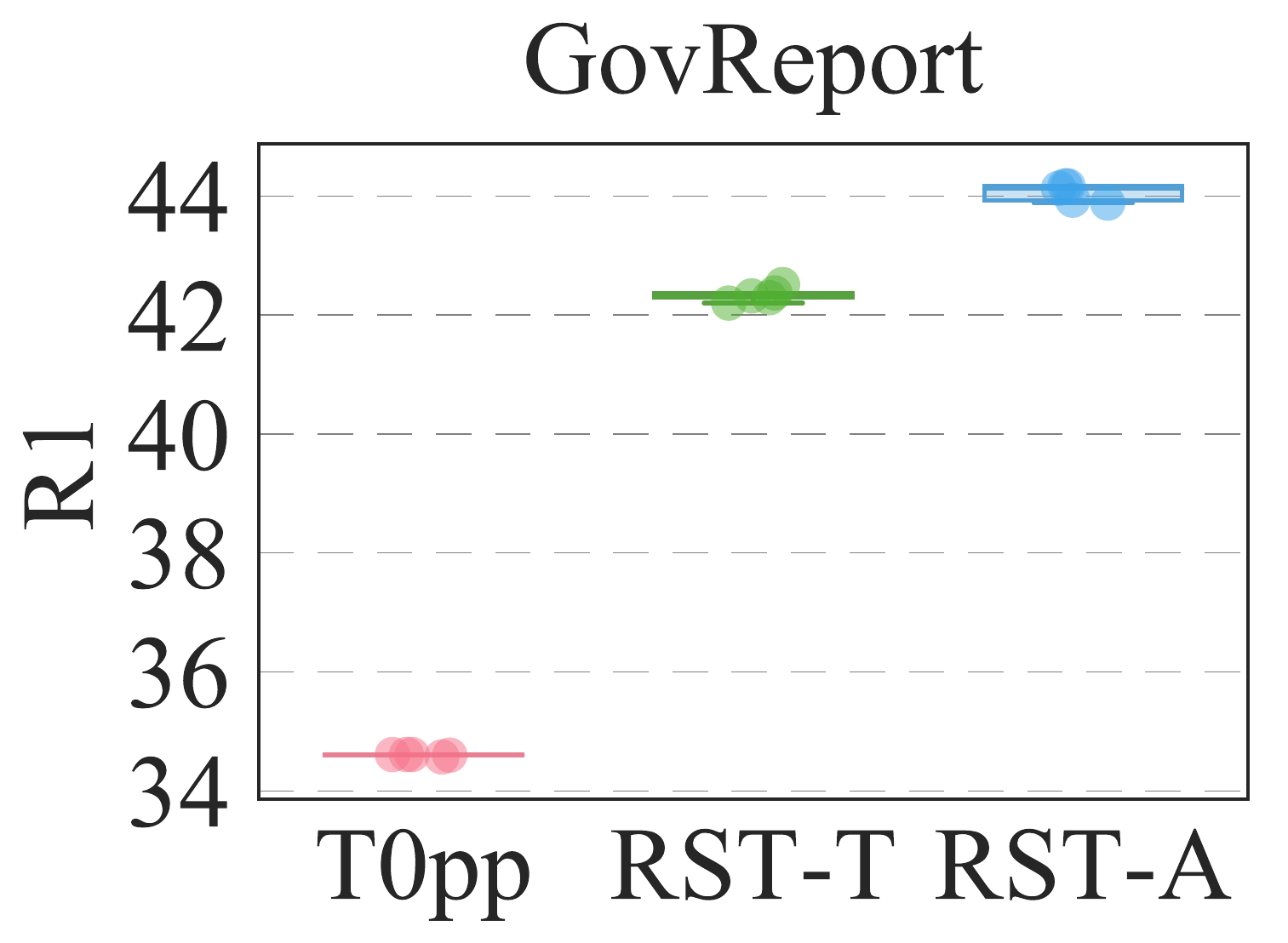}}  & \multirow{8}[1]{*}{\includegraphics[scale=0.22]{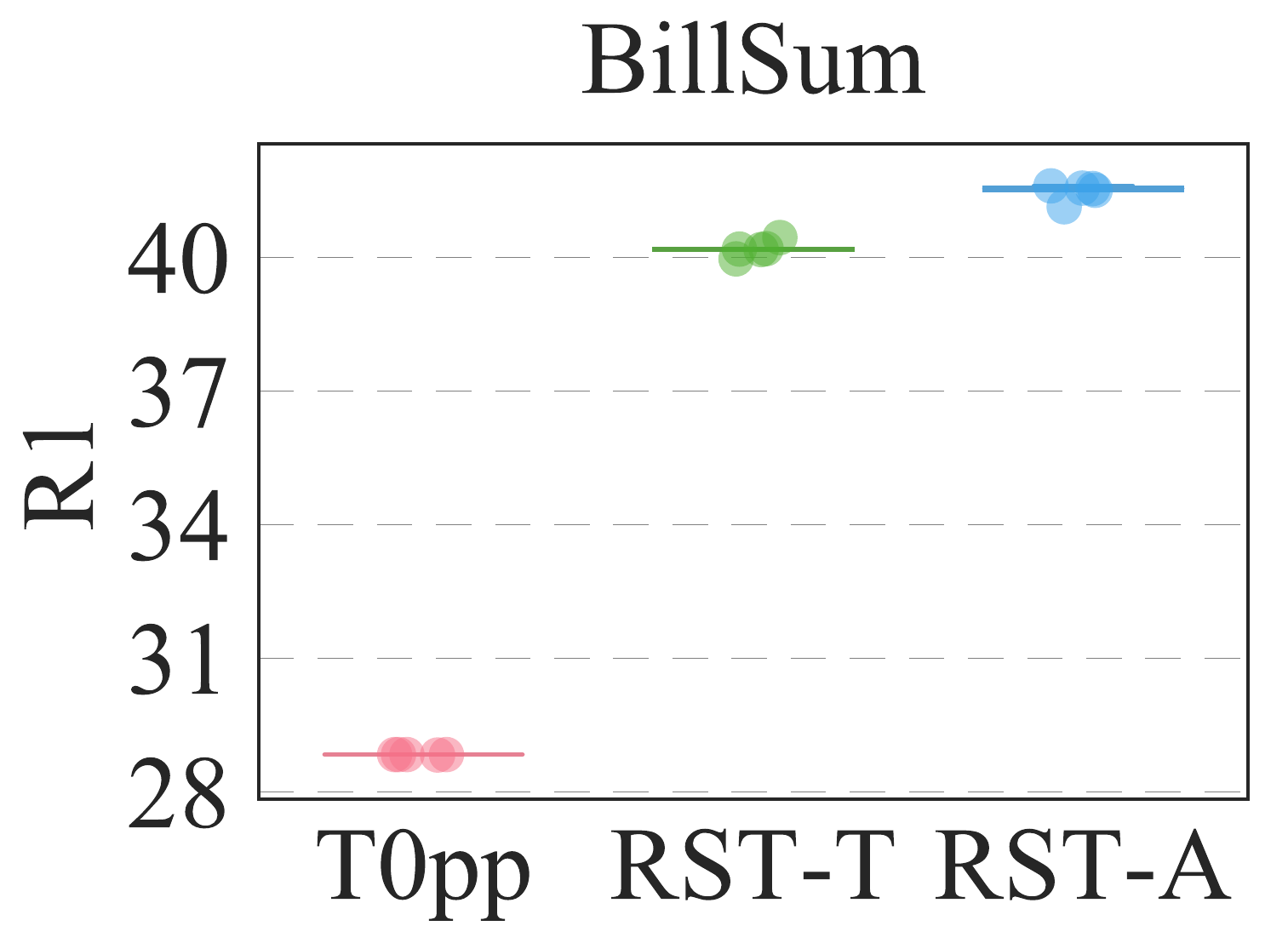}} & \multirow{8}[1]{*}{\includegraphics[scale=0.22]{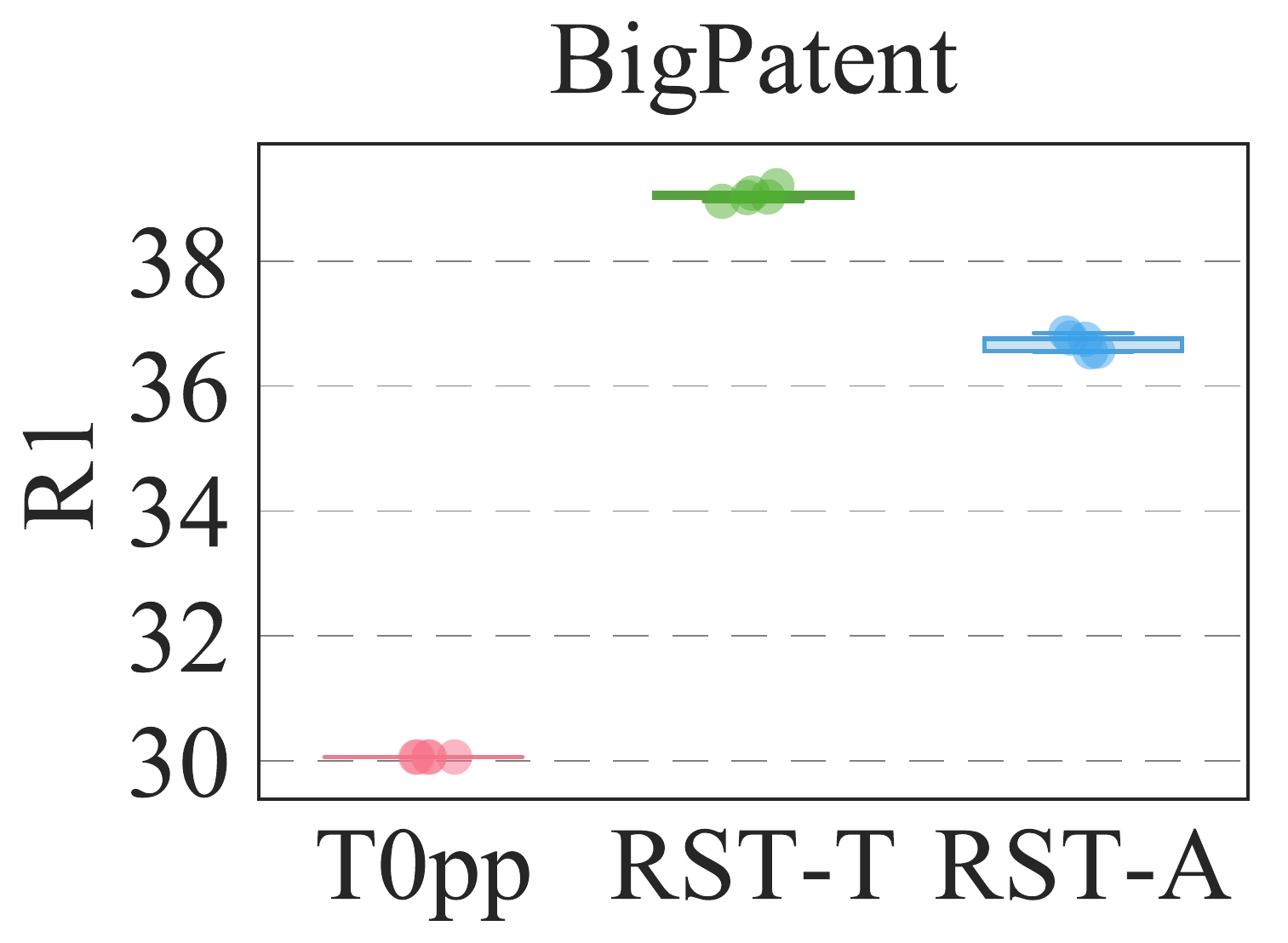}} & \\
                                                                                                      \\
                                  \\                                   \\                                 \\
                                    \\                                \\  
                                    \\
                                    \bottomrule
\end{tabular}
\end{table}

\subsubsection{Comparisons to GPT3} \label{subsec:gpt3}
\begin{figure}[!t]
\centering
\includegraphics[height=0.145\linewidth]{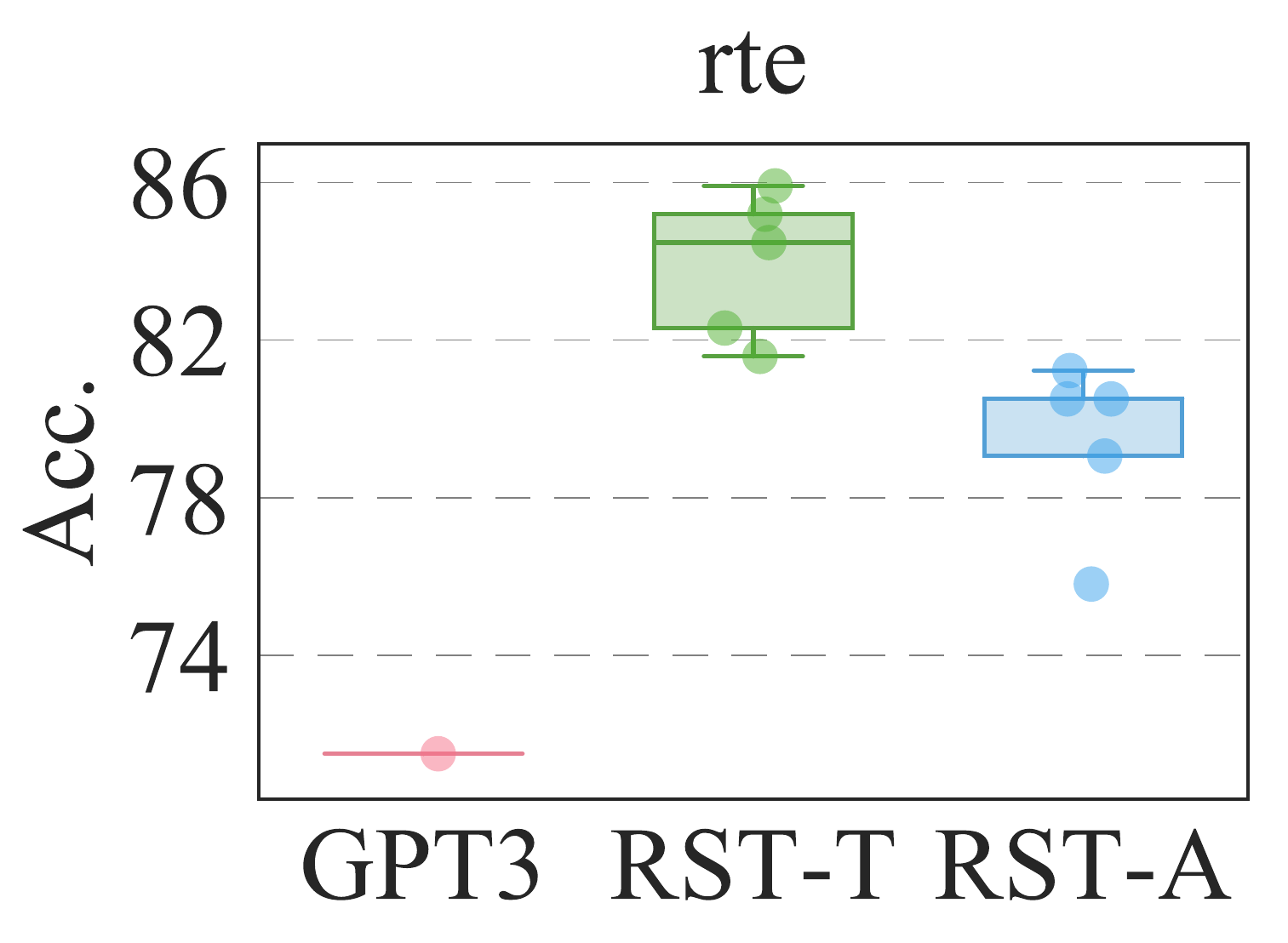}%
\hspace{-1px}
\includegraphics[height=0.145\linewidth]{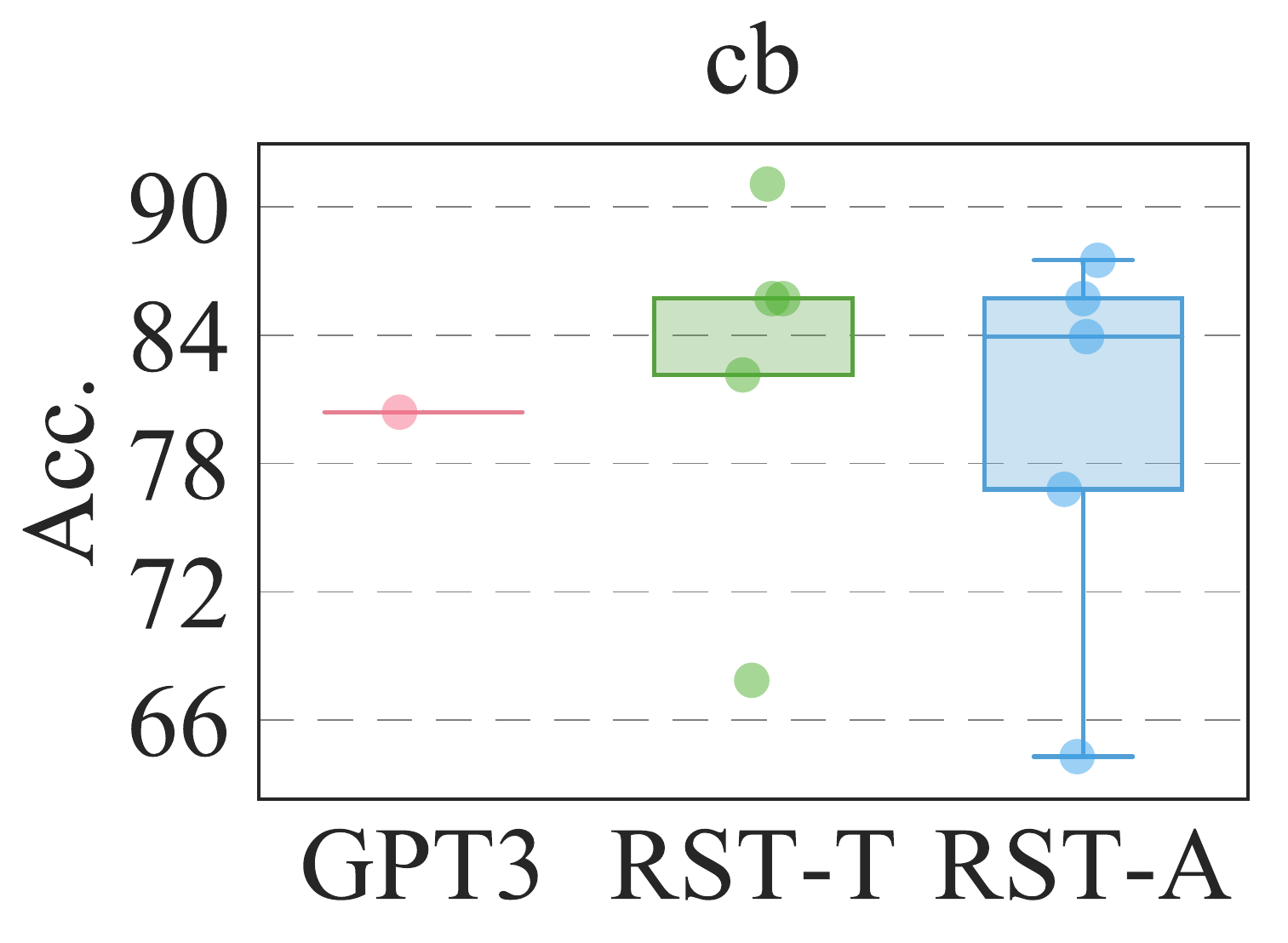}
\hspace{-4px}
\includegraphics[height=0.145\linewidth]{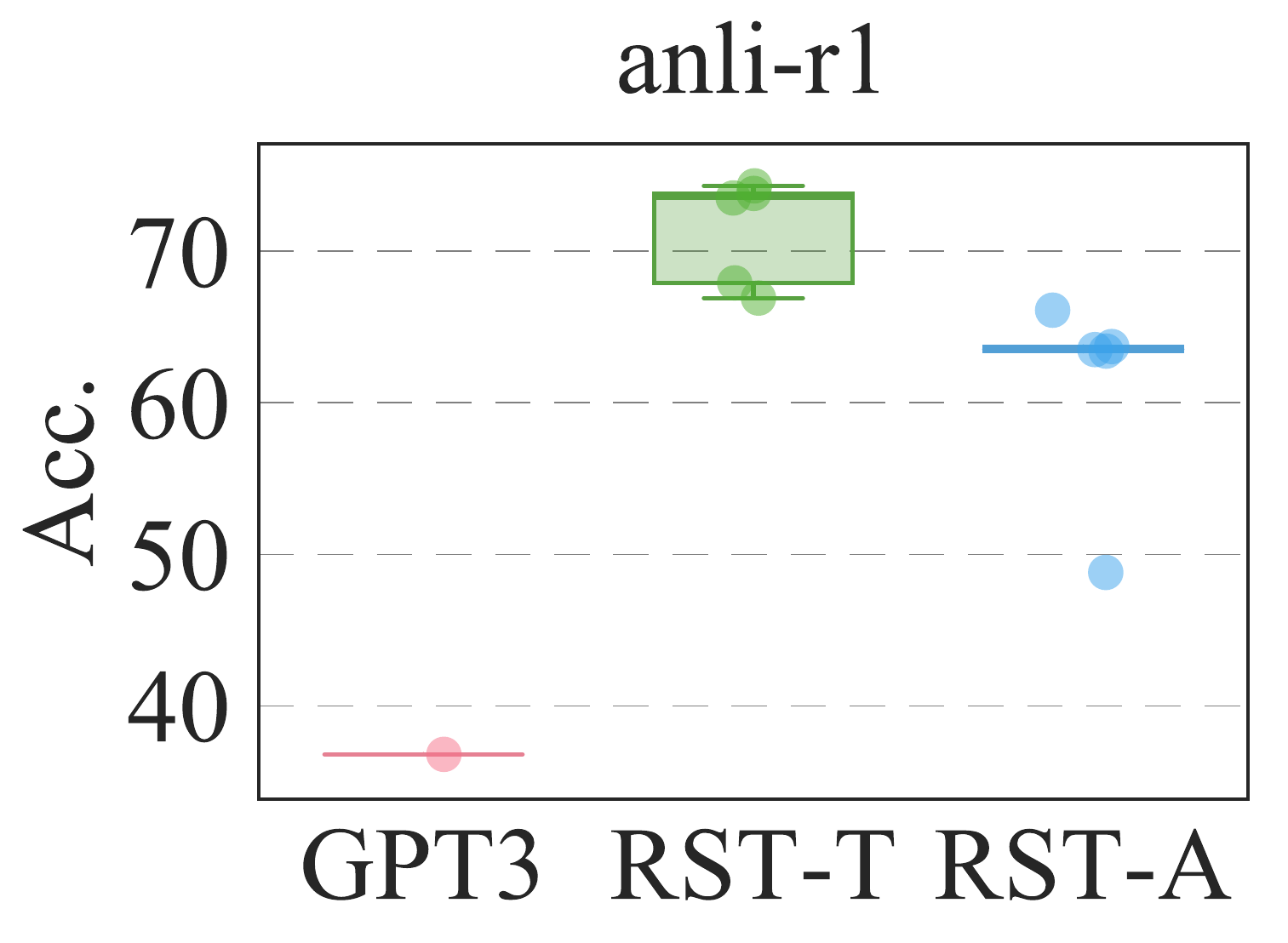}
\hspace{-4px}
\includegraphics[height=0.145\linewidth]{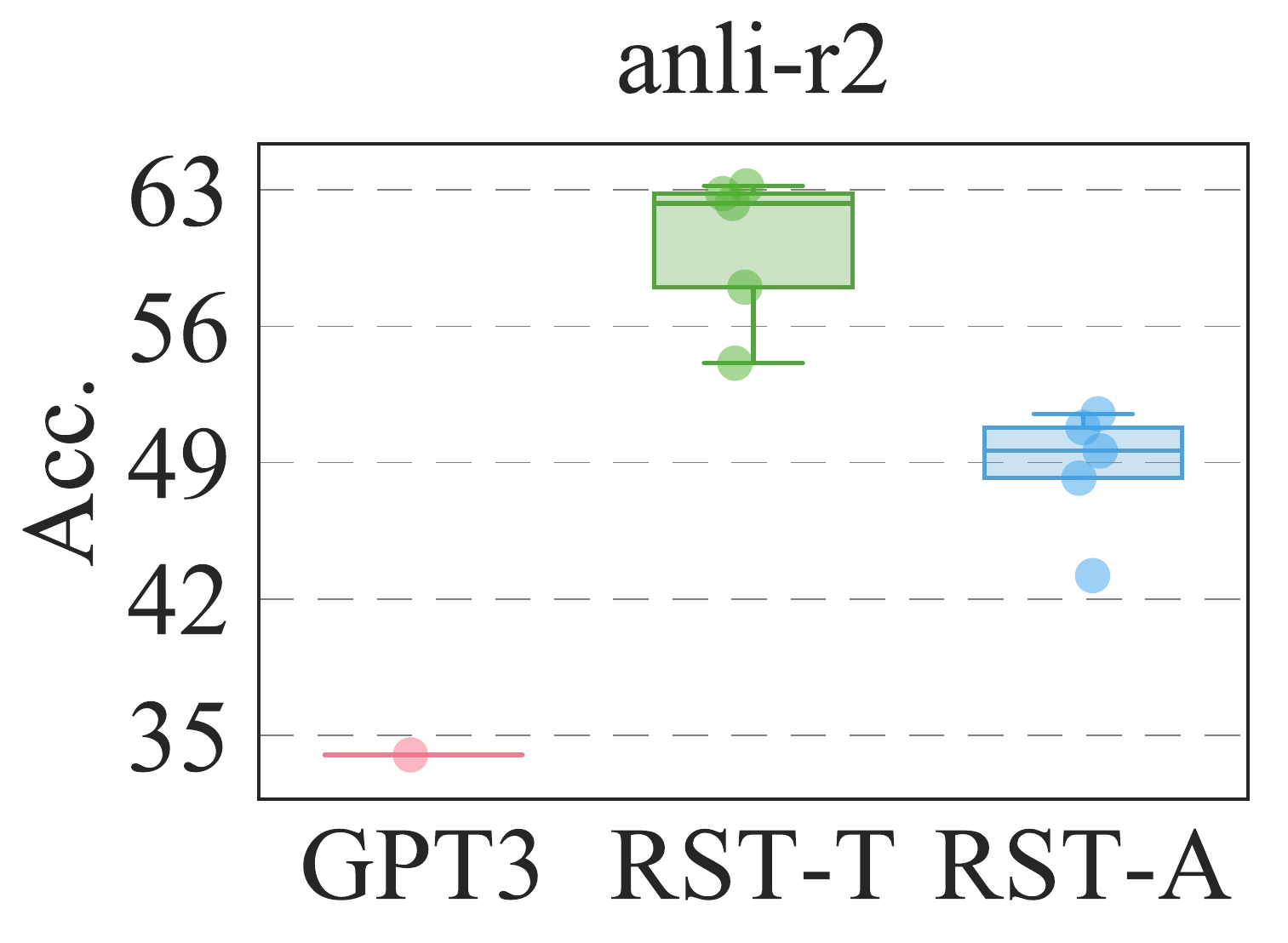}
\hspace{-4px}
\includegraphics[height=0.145\linewidth]{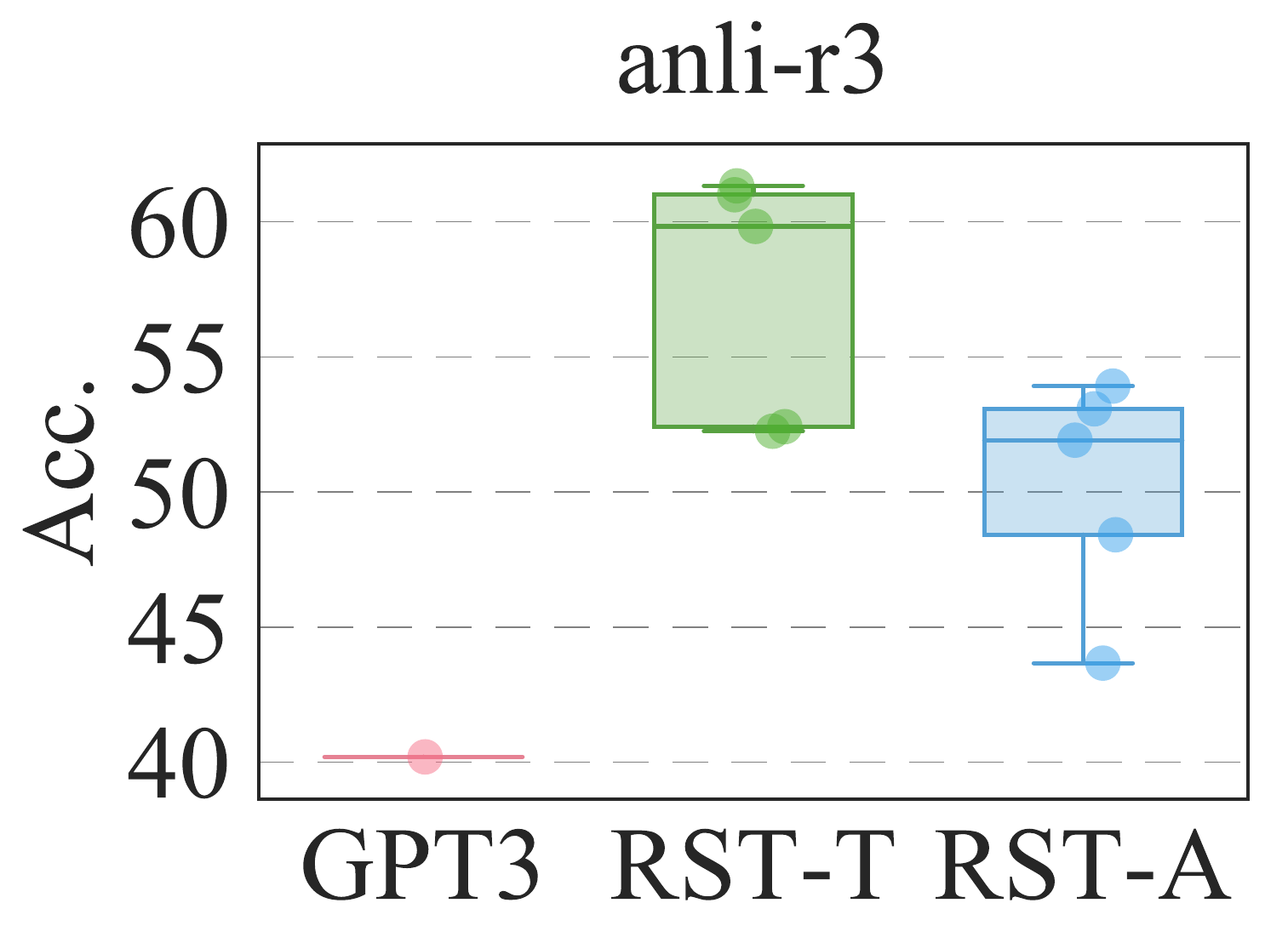}

\caption{The results on \texttt{rte}, \texttt{cb}, \texttt{anli-r1}, \texttt{anli-r2}, \texttt{anli-r3} datasets. Note that GPT3 only reports the result got using a single prompt for each dataset which contains few-shot examples, while we evaluate RST models in a zero-shot setting.}%
\label{fig:gpt3_boxplots} 
\end{figure}
We use the overlapping datasets of our evaluation datasets and the evaluation datasets of GPT3 to compare the difference between the performance of RST models and GPT3. It is worth noting that our evaluation dataset is chosen to be the complete dataset (i.e., we do not sample a subset), and we exclude the test split of a dataset if we have trained on its training split (e.g., TriviaQA) to ensure a fair comparison. This results in five considered datasets from the natural language inference task. The results are shown in Fig.~\ref{fig:gpt3_boxplots}, where the performance of GPT3 is taken from \citet{brown2020language}. 

On four datasets other than the \texttt{cb} dataset, both \textit{RST-All} and \textit{RST-Task} have better zero-shot performance (using any of our designed five prompts) than GPT3's few-shot learning. Besides, the \texttt{cb} dataset is the smallest of these considered datasets, with only 56 samples in the validation set, so different prompts will have larger fluctuations in performance on this dataset.

\subsubsection{Comparisons to T0pp} \label{subsec:t0pp}
The results are shown in Tab.~\ref{tab:results}-\ref{tab:prompt_result2}. We use the average and best performance of five prompts on each dataset to inspect the models' ability to perform different tasks. Overall, we make the following observations.

\begin{itemize}
    \item \textit{RST-All} beats \textit{T0pp} on 49 datasets out of 55 measured w.r.t average performance and wins in 47/55 cases w.r.t maximum performance. Additionally, \textit{RST-Task} outperforms \textit{T0pp} on 52 datasets out of 55 measured by the average performance and surpasses \textit{T0pp} in 50/55 cases w.r.t maximum performance. This indicates the superiority of restructured learning.
    \item The performance of the \textit{Generalist} (RST-All) is relatively worse than the \textit{Specialist} (RST-Task). This is understandable because putting all the signals into one monolithic model will inevitably lead to overwriting of the parameters in the learning process. However, the \textit{Generalist} model is still valuable because (1) its results are not much worse than those of the \textit{Specialist} model, and (2) the resource cost of maintaining one monolithic model is greatly reduced.
    \item Compared with \textit{T0pp}, both \textit{RST-Task} and \textit{RST-All} have a smaller performance standard deviation for different prompts in 39/55 and 40/55 cases, respectively. This indicates that when we put more engineering efforts into the restructuring of pre-training data, prompt engineering efforts will be reduced,
    which can alleviate the instability of the performance of different prompts in prompt learning. In addition, the performance standard deviation of \textit{RST-All} for different prompts is smaller in 31/55 cases compared to \textit{RST-Task}, suggesting that pre-training over more signals may further enhance the model robustness when using different prompts.
    \item For some sentiment classification datasets in the movie review domain, such as \texttt{mr} and \texttt{sst2}, the \textit{RST-Task} model can achieve an average accuracy of over 90\%. This indicates that for sentiment classification tasks with only two labels (positive and negative), solely pre-training can already allow the model to achieve sufficiently good downstream performance as long as the pre-training signal and downstream tasks do not differ significantly in terms of the domain.
    \item For the NER task, the \textit{RST-Task} model can obtain more than a 100\% increase in average performance on \texttt{conll03}, \texttt{notebc}, \texttt{notebn}, \texttt{notemz}, \texttt{wikiann}, and \texttt{wnut17} compared with \textit{T0pp}. Similarly, the \textit{RST-All} model can obtain more than 100\% increase in average performance on \texttt{conll03}, \texttt{notebc}, \texttt{wnut17} compared with \textit{T0pp}. On the \texttt{conll} dataset, both the \textit{RST-Task} model and \textit{RST-All} model can achieve an f1 of larger than 50 without any training sample (i.e., zero-shot). This shows that using naturally occurring signals from data mines has great potential for this task. Moreover, our proposed two-step approach to identifying named entities and their types also provides a new idea to do NER by prompting.
\end{itemize}

\subsection{Analysis}

\begin{figure}[!t]
    \centering
    \includegraphics[width=1\linewidth]{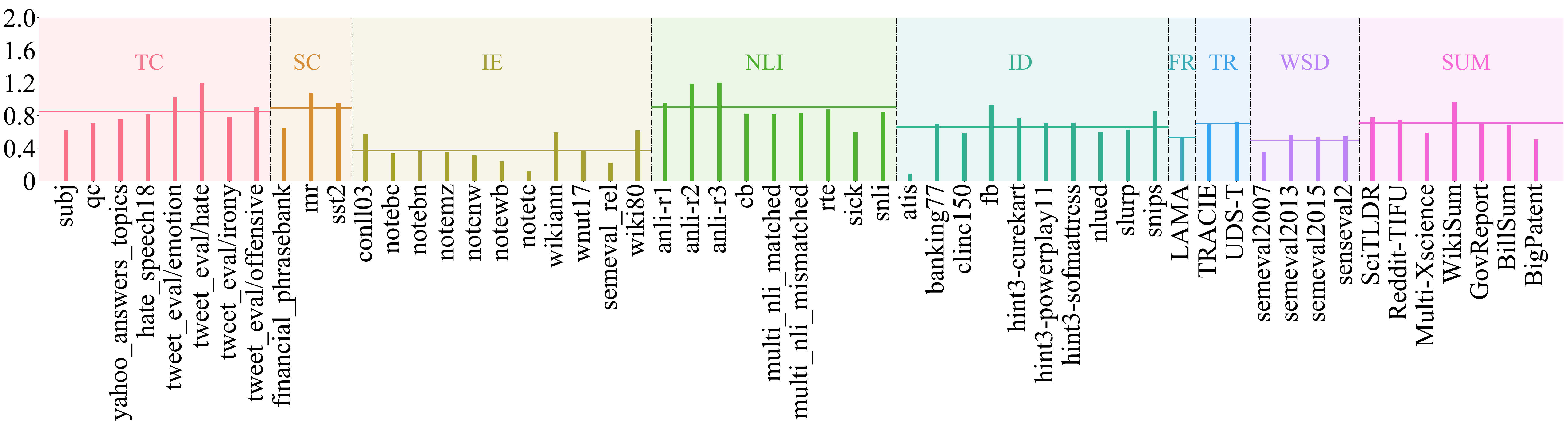}
    \caption{The percentage of performance achieved through RST-Task zero-shot prompting, compared to the SOTA fine-tuning. The horizontal line within each task area represents the average of the individual datasets within that task. We use the following abbreviations. TC: topic classification, SC: sentiment classification, IE: information extraction, NLI: natural language inference, ID: intent detection, FR: fact retrieval, TR: temporal reasoning, WSD: word sense disambiguation, SUM: summarization.}
    \label{fig:lollipop}
\end{figure}
\subsubsection{Specialist vs. Generalist}
The results in Tab.~\ref{tab:results} have demonstrated the superiority of the restructuring learning paradigm. A natural question would be: What is the difference in performance between the \textit{RST-Task} and \textit{RST-All} models compared to each other? From Tab.~\ref{tab:results}, we can see that \textit{RST-All} performs consistently lower than \textit{RST-Task} on the natural language inference task. 
This indicates that using additional signals from other tasks weakens the model's ability to make natural language inferences compared to using only the signals relevant to natural language inferences. For the other tasks, there is no such significant phenomenon. However, \textit{RST-Task} would have greater potential for each task overall.

\subsubsection{What tasks do \textit{RST} models excel at?}
In this section, we ask: what tasks does the best-performing model \textit{RST-Task} excel at?
To answer this question, we compare the performance of the \textit{RST-Task} model in the zero-shot setting with the performance that current state-of-the-art (SOTA) models\footnote{we use paperswithcode to find the SOTA models that can use training data of each task if applicable. Otherwise, we will refer to recent papers that deal with those datasets.} can achieve on each evaluation dataset. The results are shown in Fig.~\ref{fig:lollipop}.

\paragraph{\textit{RST-Task} is best at}
topic classification, sentiment classification, and natural language inference tasks. The commonality of these three tasks is that they are in the form of multiple-choice categories, and there are not many categories in total, usually fewer than ten. 

\paragraph{\textit{RST-Task} is worst at}
the information extraction task. As shown in Fig.~\ref{fig:lollipop} the model performs poorly on both NER and relation extraction datasets. We speculate the following reasons: (i) for the NER task, we choose a two-step approach to determine the class of entities. This avoids enumerating all possible text spans and reduces the time complexity but makes the model accumulate errors in two steps. (ii) For the relation extraction task, the number of categories is very large, and each relation has its inverse relation, which is certainly more challenging for the model. 

This suggests that (i) if we want better performance on the NER task, we need better sources of signals (e.g., from supervised datasets) or find a more efficient way of balancing time complexity and performance. (ii) If we want a better performance on the relation extraction task, we should construct multiple-choice prompts with more options and include some options that are difficult to distinguish from the correct option.


\clearpage

\newpage

\section{Experiment on \textsc{Gaokao}: Towards Benchmarking Human-level AI} \label{sec:gaokao}

Artificial Intelligence (AI) \cite{mccarthy2006proposal} comes from the ultimate dream of building a system that can achieve human-level intelligence (HLAI) or artificial general intelligence (AGI) \cite{turing1950computing,feigenbaum1963computers}. 
However, after several failed attempts at such visions \cite{treleaven1982japan,roland2002strategic}, researchers have turned to domain-specific problems and specialized solutions.
Over the past decades, no matter how many iterations of machine learning techniques have gone through, e.g, from SVM \cite{guyon2002gene}, PGM \cite{koller2009probabilistic} to deep neural networks \cite{lecun2015deep}, researchers have never stopped working towards human-level AI.\footnote{\url{https://cs.stanford.edu/groups/nips05-AI-Workshop/}}

\paragraph{}
Recently, with the successful application of the large pre-trained language model~\cite{brown2020language}, human-level AI is being discussed again \cite{reed2022generalist}.
In this work, we argue that before we want to answer the question of whether we achieve human-level AI, establishing a benchmark that can track how well we make progress towards it plays a more important role. Although there are some existing application-oriented multi-task benchmarks \cite{wang-etal-2018-glue,hu2020xtreme}, the tasks in them are commonly not originally designed for human intelligence. Therefore, even though AI systems have achieved good results on these benchmarks, it is still unclear how far we are from the human-level AI. 
It is important that AI system evaluation aligns with what we ultimately care about.

\begin{table}[!th]
\footnotesize
\setlength\tabcolsep{3pt}
\renewcommand{\arraystretch}{1.15}
\caption{Seven subcategories of questions in Gaokao-English and example questions for each subcategory.}
\label{tab:gaokao_example}
\begin{tabular}{lll}
\toprule
\textbf{Subcategory}           &      \textbf{Task Formulation}             & \textbf{Example Question}    \\
\midrule
\multirow{4}{*}{\textbf{Listening}}    &  Speech Recognition & \textbf{Requirement}: Based on the listening materials, choose the right answer 
\\
& Code Switching & from the given options.
\\
&  Dialogue Understanding  & \textbf{Question}: Where does this conversation take place?
\\
&  Multiple-choice QA   & \textbf{Options}: (A) In a classroom (B) In a hospital (C) In a museum 
\\
\midrule
\multirow{4}{*}{\textbf{\begin{tabular}[c]{@{}l@{}}Cloze 
\\
(multiple-choice)\end{tabular}}} & \multirow{4}{*}{Multiple-choice QA}  & \textbf{Requirement}: Based on the context, choose the right answer to fill in 
\\
& &the blank from the given options.
\\
&     & \textbf{Text}: ...... The \_\_\_ might damage the beauty of the place......                \\
&    & \textbf{Options}: (A) stories (B) buildings (C) crowds (D) reporters                        \\
\cmidrule(lr){2-3}
\multirow{5}{*}{\textbf{Cloze (hint)}}       &  \multirow{5}{*}{Open-domain QA}     & \textbf{Requirement}: Based on the context and hint, write down the correct 
\\ 
& & answer to fill in the blank.
\\
&      & \textbf{Text}: A 90-year-old has been awarded "Woman Of The Year" for \_\_\_ 
\\
& & Britain's oldest employee......    
\\
&      & \textbf{Hint}: be  
\\
\midrule
\multirow{7}{*}{\textbf{\begin{tabular}[c]{@{}l@{}}Reading 
\\ (multiple-choice)\end{tabular}}} & \multirow{7}{*}{Multiple-choice QA} & \textbf{Requirement}: Based on the text, choose the correct option from the  
\\
& &  given choices to answer the question.\\
&   & \textbf{Text}: Need a Job This Summer? The provincial government and its  \\ 
& & partners offer many programs to help students find summer jobs......                                                    \\
&      & \textbf{Question}: What is the age range required by Stewardship Youth \\ & &  Ranger Program?                                                          \\
&     & \textbf{Options}: (A) 15--18 (B) 15--24 (C) 15--29 (D) 16-17                                                         \\
\cmidrule(lr){2-3}
\multirow{9}{*}{\textbf{Reading (cloze)}}       &  \multirow{9}{*}{Multiple-choice QA}  & \textbf{Requirement}: Based on the context, choose the best option from the  \\ 
& & given choices to fill in the blank.                       \\
&   & \textbf{Text}: ...... \_\_\_ Like the child on the diving board, you will stay undecided\\ & & ......            \\
&        & \textbf{Options}: \\ 
&& (A) Without motivation, you can neither set a goal nor reach it.                                                     \\
&        & (B) So how should you motivate yourself?           \\
&       & (C) This can affect your work.                      \\
&       & (D) They can change according to circumstances.     \\
\midrule
\multirow{11}{*}{\textbf{\begin{tabular}[c]{@{}l@{}}Writing \\ (grammar)\end{tabular}}}    &  \multirow{11}{*}{\begin{tabular}[c]{@{}l@{}}Text Editing \\ Grammar Error Correction\end{tabular}}      & \textbf{Requirement}: There are ten gramatical errores in the given text in \\ 
& & total, each involving the addition, modification or deletion of a word. \\ 
& & Please correct them.                                      \\
&      & \textbf{Text}: I became interesting in playing football thanks to a small accident.\\ && One afternoon where I was in primary school, I was walking by the\\ && school playground. Suddenly football fell just in front of me but almost \\ &&  hit me. I stopped the ball and kicked it hardly back to the playground. \\ &&To everyone's surprising,  the ball went into the net. All the football \\ && player on the playground cheered loudly, say that I had a talent for \\ && football. From now on, I started to play my football with classmates \\ && after school. I am a good player now. \\
                                           \cmidrule(lr){2-3}
\multirow{8}{*}{\textbf{\begin{tabular}[c]{@{}l@{}}Writing \\ (essay writing)\end{tabular}}}   &     \multirow{8}{*}{\begin{tabular}[c]{@{}l@{}}Controllable Text Generation \\ Machine Translation\end{tabular}}         & \textbf{Requirement}: Write an article based on the question and requirements. \\ && \textbf{Question}: Suppose you are Li Hua, studying in London during the \\ && summer vacation, and learning that the local art museum will hold an \\ && exhibition of Chinese paintings. Please write a letter to apply for \\ && volunteering, including: 1. Purpose of writing the letter; 2. Personal \\ && strengths: 3. What you can do. \\&&
\textbf{Requirement}: 1. The number of words is about 100; 2. Details can be \\ && added appropriately to make the text coherent; 3. The conclusion has \\ && been written for you. Please use the end: "Yours, Li Hua".                 \\
\bottomrule
\end{tabular}
\end{table}

\paragraph{Desiderata for Benchmarking Human-level AI}
To track the progress of achieving human-level AI, a benchmark should fulfill the following desiderata:

\begin{itemize}
    \item \textbf{Comprehensiveness}: Benchmarks should involve a variety of modalities (e.g., speech, text, and image), tasks (e.g., classification, question answering, text generation), and domains (news, economic and novel), which also lays the basis for the ``practicality''.
    \item \textbf{Practicality}: Tasks should be practically useful in real-world scenarios and align with human demands. For example, people do not always need to verify whether a sentence can be entailed by another sentence in real life but will need to write an email asking for help. Therefore, the task of generating text based on specific requirements is more aligned with real-world demands than natural language inferences.
    \item \textbf{Discriminability}: Benchmarks are expected to make reasonable arrangements in terms of task difficulty and are capable of discriminating models with different levels of abilities.
    \item \textbf{Accessibility}. Both model and human performance on tasks of the benchmark should be easily collected so that we can make a fair comparison between AI models and human and understand their comparative advantages.
    \item \textbf{Extendibility}: Benchmarks should be flexibly extended and updated, which not only evaluate whether the model is overfitting but also assess the incremental learning ability of AI models.
\end{itemize}
With the above desiderata in mind, we are searching for a good fit and argue that Gaokao will be a good example to meet these characteristics.

\subsection{\textsc{GaoKao}  Benchmark}

Gaokao, also known as the national college entrance examination, is a nationally unified and selective examination for qualified high school graduates, which is the \textbf{most} authoritative entrance examination in China.
Every year many experts with different subject backgrounds will gather together to design a set of Gaokao papers involving different subjects (e.g., Chinese, English, maths, physics, chemistry, and geography) to select different types of talents.
In this paper, we focus on the Gaokao-English test and show the potential of building a benchmark based on Gaokao-English papers.
\paragraph{Comprehensiveness}
As illustrated in Tab.\ref{tab:gaokao_example}, Gaokao-English provides a {comprehensive} evaluation suite that consists of 
four categories of questions:
(1) listening comprehension, (2) reading comprehension, (3) cloze filling, and (4) writing, which involve a variety of NLP tasks including dialogue comprehension, question answering, reading comprehension, text generation, grammar error correction, etc. as well as tasks from speech recognition and image recognition.
Moreover, these questions are often designed to come from different topics, including natural science, art, culture, etc.


\paragraph{Practicality}
Questions in Gaokao-English examine students' abilities such as comprehension, communication, and writing, which are {practically} needed in their daily lives. For example, the writing part aims to assess students' comprehensive language ability to see whether students can use the English knowledge and skills learned to communicate and fulfill the needs of different situations.

\paragraph{Discriminability}
Questions in Gaokao are carefully designed by multiple experts in combination with practical applications and skills to be assessed.  The distribution of question types and difficulties are arranged reasonably.

\paragraph{Accessibility}
Gaokao papers are public and available online (including audio materials). Since Gaokao-English questions are essentially tasks in NLP, their evaluation methods for AI models are well-defined.
In addition, Gaokao scores (such as average scores) are also publicly available and easily accessible.

\paragraph{Extendibility}

Gaokao papers are updated annually, and the question types are also changing, allowing us to measure models' generalization ability and their incremental learning capacity.
Additionally, considering that the overall actual performance of students is transparent from year to year, this makes it possible to compare models to humans over time.

\subsection{Benchmark Datasets}

\paragraph{Description}
We have collected ten English exam papers for the college entrance exams, including 2018 National Paper I/III, 2019 National Paper I/II/III, 2020 National Paper I/II/III, and 2021 National Paper A/B\footnote{A: Jia. B: Yi.}. These papers follow the same question types, and we divide all exam question types into the following seven subcategories, as shown in Tab.~\ref{tab:gaokao_example}.

\paragraph{Statistics}
The full mark for each Gaokao-English paper is \textbf{150}. As shown in Tab.~\ref{tab:gaokao_info}, \textit{listening}, \textit{cloze}, \textit{reading}, and \textit{writing} account for \textbf{30}, \textbf{45}, \textbf{40}, \textbf{35}, respectively.
Commonly, the writing part is subjective that requires human evaluation, while the others are objective that could be graded automatically.

\begin{table}[!th]
\centering
\footnotesize
\setlength\tabcolsep{6pt}
\renewcommand{\arraystretch}{1.1}
\caption{Configuration of Gaokao-English paper.}
\label{tab:gaokao_info}
\begin{tabular}{lccccccc}
\toprule
                 &                             & \multicolumn{2}{c}{\textbf{Cloze}}                          & \multicolumn{2}{c}{\textbf{Reading}}                              & \multicolumn{2}{c}{\textbf{Writing}}         \\
                 \cmidrule(lr){3-4} \cmidrule(lr){5-6} \cmidrule(lr){7-8}
                 & \multirow{-2}{*}{\textbf{Listening}} & \textbf{multiple-choice} & \textbf{hint} & \textbf{multiple-choice} & \textbf{cloze} & \textbf{grammar}                       & \textbf{essay writing}       \\
                 \midrule
\textbf{Num.}             & 20                          & 20                                        & 10                             & 15                                        & 5                               & 1                               & 1                    \\
\textbf{Point / Question} & 1.5                         & 1.5                                       & 1.5                            & 2                                         & 2                               & 10                              & 25                   \\
\textbf{Total Points}     & 30                          & 30                                        & 15                             & 30                                        & 10                              & 10                              & 25                   \\
\textbf{Auto. Grading}    & \cmark        & \cmark                      & \cmark           & \cmark                      & \cmark            & \cmark            & \xmark \\
\bottomrule
\end{tabular}
\end{table}


\subsection{\includegraphics[scale=0.1]{imgs/qin-half.png} \textsc{Gaokao} AI: \textsc{Qin}}\label{subsec:qin}

\begin{figure}[!th]
    \centering
    \includegraphics[width=0.85\linewidth]{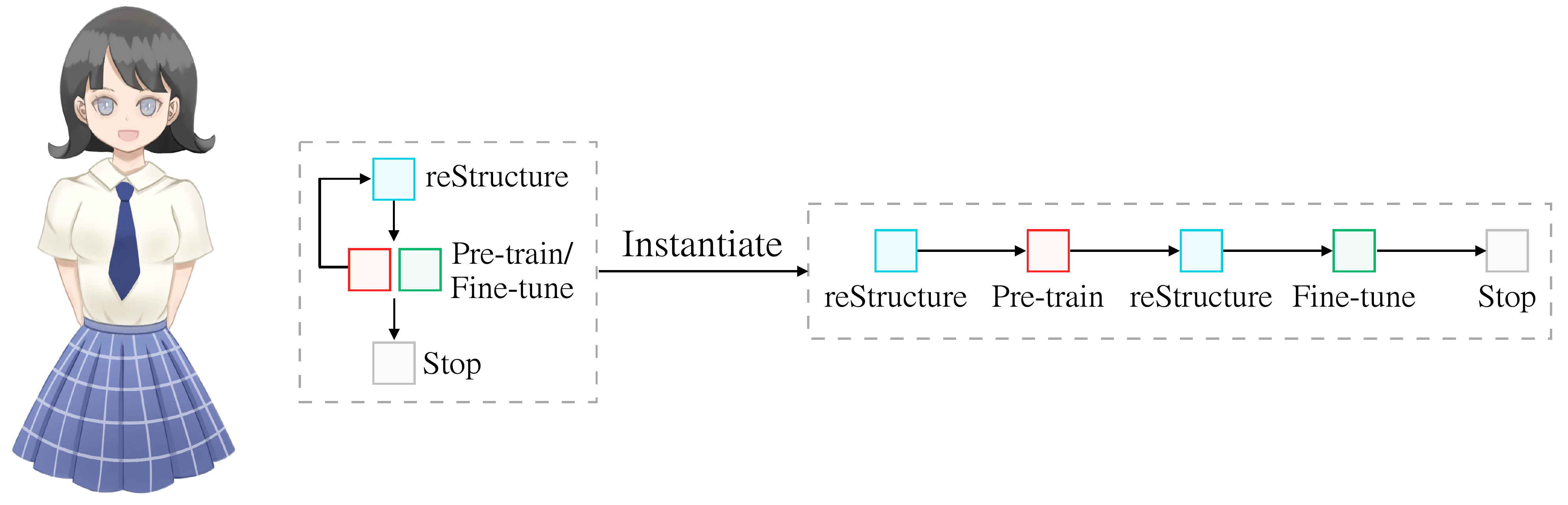}
    \caption{Gaokao AI: Qin and An example use case of RST.}
    \label{fig:gaokaoai}
\end{figure}

We illustrate how to use the restructure engineering cycle illustrated in Tab.\ref{tab:five_paradigm} to build an AI system for Gaokao-English, namely \textsc{Qin}. The whole procedure is shown in Fig.~\ref{fig:gaokaoai}:
(1) We first take the \textit{RST-All} model we have pre-trained using all restructured signals as done in \S\ref{subsec:model_setups},
(2) and then fine-tune the model using some newly restructured signals specifically designed for Gaokao-English. 

As shown in Tab.~\ref{tab:gaokao_example}, seven subcategories of questions are examined in the Gaokao-English. Based on the knowledge they require, we further group each question into one of the following three classes: \textit{Comprehension} (listening, reading (multiple-choice)), \textit{Language Usage} (cloze (multiple-choice), cloze (hint), reading (cloze)), \textit{Writing} (writing (grammar), writing (essay writing)). We fine-tune a specialized model for each class separately.

\subsubsection{Signal Collection}
We collect a variety of signals for model restructuring learning. Please refer to Appendix \ref{app:gaokao-english-signal} for further details.

\paragraph{Comprehension} The datasets we use for the comprehension model are multiple-choice QA datasets that are used to train the T0pp model. The signals we collect are $($context, question, answer$)$ triples. 

\paragraph{Language Usage} We prepare signals for each question subcategory. (i) For cloze (multiple-choice) questions, we use the CLOTH dataset \cite{xie-etal-2018-large} for training and validation. The signals we collect are $($context, cloze\_position, taget$)$ triples. (ii) For cloze (hint) questions, we automatically construct training and validation data by randomly masking out some words in context and provide certain hints by searching their derivationally related forms using WordNet. The signals we collect are $($context, cloze\_position, hint, target$)$ quads. (iii) For reading (cloze) questions, we automatically construct training and validation data by randomly masking some sentences in context. The signals we collect are $($context, cloze\_position, target$)$ triples.

\paragraph{Writing} We prepare signals for each question subcategory. (i) For writing (grammar) questions, we use the data from BEA 2019 Shared Task \cite{bryant-etal-2019-bea}. The signals we collect are $($original\_text, corrected\_text$)$ pairs. (ii) For writing (essay writing) questions, we collect questions from real high school exams and Chinese graduate school entrance exams. The signals we collect are $($question, requirement, article$)$ triples.

\subsubsection{reStructure Engineering}
We use the following prompts to transform the original signal tuples into prompted samples, as shown in Tab.~\ref{tab:gaokao-prompt}.

\begin{table}[!ht]
\footnotesize
\setlength\tabcolsep{5pt}
\renewcommand{\arraystretch}{1.1}
\caption{Prompts for Gaokao questions. ``MC" stands for ``multiple-choice". \{options\_with\_or\} (e.g. ``option1", ``option2", or ``option3") stands for all available options provided by the question.}
\label{tab:gaokao-prompt}
\begin{tabular}{lll}
\toprule
\textbf{Subcategory}                             & \textbf{Signal}                                                    & \textbf{Prompt}                                                                                                                                                   \\
\midrule
Reading (MC)                            & \multirow{2}{*}{(context, question, answer)}              & \textcolor{red}{source}: \texttt{TEXT}: \{context\} \texttt{QUERY}: \{question\} \{options\_with\_or\}?                                                                                     \\
Listening                               &                                                           & \textcolor{blue}{target}: \{answer\}                                                                                                                                       \\
\midrule
\multirow{3}{*}{\begin{tabular}[c]{@{}l@{}}Cloze (MC) \\ Reading (cloze)\end{tabular}}
           & \multirow{3}{*}{(context, cloze\_position, taget)}        & \textcolor{red}{source}: \texttt{TEXT}: \{context\} \texttt{QUERY}: What should be filled in at the \\ & &  \{cloze\_position\} position? \{options\_with\_or\}?                                     \\

                                        &                                                           & \textcolor{blue}{target}: \{target\}                                                                                                                                       \\
                                        \midrule
\multirow{3}{*}{Cloze (hint)}           & \multirow{3}{*}{(context, cloze\_position, hint, target)} & \textcolor{red}{source}: \texttt{TEXT}: \{context\} \texttt{QUERY}: What should be filled in at the \\ & & \{cloze\_position\} position given the hint ``\{hint\}"? \\
                                        &                                                           & \textcolor{blue}{target}: \{target\}                                                                                                                                       \\
\midrule                                
\multirow{3}{*}{\begin{tabular}[c]{@{}l@{}}Writing \\ (grammar)\end{tabular}}      & \multirow{3}{*}{(original\_text, corrected\_text)}        & \textcolor{red}{source}: \texttt{TEXT}: \{original\_text\} \texttt{QUERY}: Please fix the grammatical \\ & & errors in the above paragraph.                                                        \\
                                        &                                                           & \textcolor{blue}{target}: \{corrected\_text\}                                                                                                                              \\
\midrule                                        
\multirow{2}{*}{\begin{tabular}[c]{@{}l@{}}Writing \\ (essay writing)\end{tabular}} & \multirow{2}{*}{(question, requirement, article)}         & \textcolor{red}{source}: \texttt{QUERY}: \{question\} \{requirement\}                                                                                                              \\
                                        &                                                           & \textcolor{blue}{target}: \{article\} \\ 
                                        \bottomrule
\end{tabular}
\end{table}

\begin{table}[!t]
\centering
\footnotesize
\setlength\tabcolsep{1.7pt}
\renewcommand{\arraystretch}{1.2}
\caption{Results on Gaokao (part1). We use the following abbreviations. MC: multiple-choice, NP: national paper, ASR: automatic speech recognition. The model sizes are as follows. T0pp: 11B, GPT3: 175B, RST: 11B. The highest total score for each year is bolded.}
\label{tab:gaokao_results_part1}
\begin{tabular}{lrrrrrrrrrrrrrrr}
\toprule
                          & \multicolumn{3}{c}{2018 NP-I}                                                  & \multicolumn{3}{c}{2018 NP-III}                                                  & \multicolumn{3}{c}{2019 NP-I}                                                  & \multicolumn{3}{c}{2019 NP-II}                                                  & \multicolumn{3}{c}{2019 NP-III}                                                  \\
                          \cmidrule(lr){2-4}\cmidrule(lr){5-7}\cmidrule(lr){8-10}\cmidrule(lr){11-13}\cmidrule(lr){14-16}
                          & \multicolumn{1}{c}{T0pp} & \multicolumn{1}{c}{GPT3} & \multicolumn{1}{c}{RST} & \multicolumn{1}{c}{T0pp} & \multicolumn{1}{c}{GPT3} & \multicolumn{1}{c}{RST} & \multicolumn{1}{c}{T0pp} & \multicolumn{1}{c}{GPT3} & \multicolumn{1}{c}{RST} & \multicolumn{1}{c}{T0pp} & \multicolumn{1}{c}{GPT3} & \multicolumn{1}{c}{RST} & \multicolumn{1}{c}{T0pp} & \multicolumn{1}{c}{GPT3} & \multicolumn{1}{c}{RST} \\
                          \midrule
Listening 
& \multicolumn{1}{>{\columncolor{tablegreen}}r}{30.0}
& \multicolumn{1}{>{\columncolor{tablegreen}}r}{28.5}
& \multicolumn{1}{>{\columncolor{tablegreen}}r}{30.0}
& 27.0
& 22.5
& 30.0
& \multicolumn{1}{>{\columncolor{tablegreen}}r}{25.5}
& \multicolumn{1}{>{\columncolor{tablegreen}}r}{28.5}
& \multicolumn{1}{>{\columncolor{tablegreen}}r}{25.5}
& 24.0
& 28.5
& 28.5
& \multicolumn{1}{>{\columncolor{tablegreen}}r}{27.0}
& \multicolumn{1}{>{\columncolor{tablegreen}}r}{28.5}
& \multicolumn{1}{>{\columncolor{tablegreen}}r}{28.5}
\\
Listening (ASR)
& \multicolumn{1}{>{\columncolor{tablegreen}}r}{25.5}
& \multicolumn{1}{>{\columncolor{tablegreen}}r}{28.5}
& \multicolumn{1}{>{\columncolor{tablegreen}}r}{30.0}
& 28.5
& 21.0
& 27.0
& \multicolumn{1}{>{\columncolor{tablegreen}}r}{24.0}
& \multicolumn{1}{>{\columncolor{tablegreen}}r}{27.0}
& \multicolumn{1}{>{\columncolor{tablegreen}}r}{27.0}
& 16.5
& 28.5
& 28.5    
& \multicolumn{1}{>{\columncolor{tablegreen}}r}{16.5}
& \multicolumn{1}{>{\columncolor{tablegreen}}r}{28.5}
& \multicolumn{1}{>{\columncolor{tablegreen}}r}{28.5}
\\
Cloze (MC)
& \multicolumn{1}{>{\columncolor{tablegreen}}r}{21.0}
& \multicolumn{1}{>{\columncolor{tablegreen}}r}{30.0}
& \multicolumn{1}{>{\columncolor{tablegreen}}r}{28.5}
& 24.0
& 28.5
& 30.0
& \multicolumn{1}{>{\columncolor{tablegreen}}r}{15.0}
& \multicolumn{1}{>{\columncolor{tablegreen}}r}{22.5}
& \multicolumn{1}{>{\columncolor{tablegreen}}r}{28.5}
& 19.5
& 27.0
& 30.0
& \multicolumn{1}{>{\columncolor{tablegreen}}r}{18.0}
& \multicolumn{1}{>{\columncolor{tablegreen}}r}{25.5}
& \multicolumn{1}{>{\columncolor{tablegreen}}r}{25.5}
\\
Cloze (hint)
& \multicolumn{1}{>{\columncolor{tablegreen}}r}{1.5}
& \multicolumn{1}{>{\columncolor{tablegreen}}r}{6.0}
& \multicolumn{1}{>{\columncolor{tablegreen}}r}{12.0}
& 0.0
& 4.5
& 13.5
& \multicolumn{1}{>{\columncolor{tablegreen}}r}{0.0}
& \multicolumn{1}{>{\columncolor{tablegreen}}r}{4.5}
& \multicolumn{1}{>{\columncolor{tablegreen}}r}{10.5}
& 0.0
& 6.0
& 12.0
& \multicolumn{1}{>{\columncolor{tablegreen}}r}{0.0}
& \multicolumn{1}{>{\columncolor{tablegreen}}r}{6.0}
& \multicolumn{1}{>{\columncolor{tablegreen}}r}{12.0}
\\
Reading (MC) 
& \multicolumn{1}{>{\columncolor{tablegreen}}r}{24.0}
& \multicolumn{1}{>{\columncolor{tablegreen}}r}{22.0}
& \multicolumn{1}{>{\columncolor{tablegreen}}r}{18.0}
& 24.0
& 22.0
& 30.0
& \multicolumn{1}{>{\columncolor{tablegreen}}r}{28.0}
& \multicolumn{1}{>{\columncolor{tablegreen}}r}{24.0}
& \multicolumn{1}{>{\columncolor{tablegreen}}r}{30.0}
& 26.0
& 26.0
& 28.0
& \multicolumn{1}{>{\columncolor{tablegreen}}r}{24.0}
& \multicolumn{1}{>{\columncolor{tablegreen}}r}{22.0}
& \multicolumn{1}{>{\columncolor{tablegreen}}r}{30.0}
\\
Reading (cloze)
& \multicolumn{1}{>{\columncolor{tablegreen}}r}{0.0}
& \multicolumn{1}{>{\columncolor{tablegreen}}r}{2.0}
& \multicolumn{1}{>{\columncolor{tablegreen}}r}{10.0}
& 6.0
& 4.0
& 6.0
& \multicolumn{1}{>{\columncolor{tablegreen}}r}{6.0}
& \multicolumn{1}{>{\columncolor{tablegreen}}r}{4.0}
& \multicolumn{1}{>{\columncolor{tablegreen}}r}{10.0}
& 2.0
& 6.0 
& 10.0
& \multicolumn{1}{>{\columncolor{tablegreen}}r}{4.0}
& \multicolumn{1}{>{\columncolor{tablegreen}}r}{8.0}
& \multicolumn{1}{>{\columncolor{tablegreen}}r}{6.0}
\\
Writing (grammar)
& \multicolumn{1}{>{\columncolor{tablegreen}}r}{0.0}
& \multicolumn{1}{>{\columncolor{tablegreen}}r}{6.0}
& \multicolumn{1}{>{\columncolor{tablegreen}}r}{6.0}
& 0.0
& 9.0
& 7.0
& \multicolumn{1}{>{\columncolor{tablegreen}}r}{0.0}
& \multicolumn{1}{>{\columncolor{tablegreen}}r}{8.0}
& \multicolumn{1}{>{\columncolor{tablegreen}}r}{7.0}
& 1.0
& 8.0
& 8.0
& \multicolumn{1}{>{\columncolor{tablegreen}}r}{0.0}
& \multicolumn{1}{>{\columncolor{tablegreen}}r}{10.0}
& \multicolumn{1}{>{\columncolor{tablegreen}}r}{7.0}
\\
Writing (essay writing)
& \multicolumn{1}{>{\columncolor{tablegreen}}r}{10.0}
& \multicolumn{1}{>{\columncolor{tablegreen}}r}{22.0}
& \multicolumn{1}{>{\columncolor{tablegreen}}r}{21.0}
& 12.0
& 21.0
& 21.0
& \multicolumn{1}{>{\columncolor{tablegreen}}r}{0.0}
& \multicolumn{1}{>{\columncolor{tablegreen}}r}{20.0}
& \multicolumn{1}{>{\columncolor{tablegreen}}r}{22.0}
& 8.0
& 21.0
& 22.0
& \multicolumn{1}{>{\columncolor{tablegreen}}r}{10.0}
& \multicolumn{1}{>{\columncolor{tablegreen}}r}{21.0}
& \multicolumn{1}{>{\columncolor{tablegreen}}r}{22.0}
\\
\midrule
Total score
& \multicolumn{1}{>{\columncolor{tablegreen}}r}{86.5}
& \multicolumn{1}{>{\columncolor{tablegreen}}r}{116.5}
& \multicolumn{1}{>{\columncolor{tablegreen}}r}{\textbf{125.5}}
& 93.0
& 111.5
& \textbf{137.5}
& \multicolumn{1}{>{\columncolor{tablegreen}}r}{74.5}
& \multicolumn{1}{>{\columncolor{tablegreen}}r}{111.5}
& \multicolumn{1}{>{\columncolor{tablegreen}}r}{\textbf{133.5}}
& 80.5
& 122.5
& \textbf{138.5}
& \multicolumn{1}{>{\columncolor{tablegreen}}r}{83.0}
& \multicolumn{1}{>{\columncolor{tablegreen}}r}{121.0}
& \multicolumn{1}{>{\columncolor{tablegreen}}r}{\textbf{131.0}}
\\
Total score (ASR)
& \multicolumn{1}{>{\columncolor{tablegreen}}r}{82.0}
& \multicolumn{1}{>{\columncolor{tablegreen}}r}{116.5}
& \multicolumn{1}{>{\columncolor{tablegreen}}r}{\textbf{125.5}}
& 94.5
& 110.0
& \textbf{134.5}
& \multicolumn{1}{>{\columncolor{tablegreen}}r}{73.0}
& \multicolumn{1}{>{\columncolor{tablegreen}}r}{110.0}
& \multicolumn{1}{>{\columncolor{tablegreen}}r}{\textbf{135.0}}
& 73.0
& 122.5
& \textbf{138.5}
& \multicolumn{1}{>{\columncolor{tablegreen}}r}{72.5}
& \multicolumn{1}{>{\columncolor{tablegreen}}r}{121.0}
& \multicolumn{1}{>{\columncolor{tablegreen}}r}{\textbf{131.0}}
\\
\bottomrule
\end{tabular}
\end{table}

\begin{table}[!ht]
\centering
\footnotesize
\setlength\tabcolsep{1.7pt}
\renewcommand{\arraystretch}{1.2}
\caption{Results on Gaokao (part2). We use the following abbreviations. MC: multiple-choice, NP: national paper, ASR: automatic speech recognition. The model sizes are as follows. T0pp: 11B, GPT3: 175B, RST: 11B. The highest total score for each year is bolded.}
\label{tab:gaokao_results_part2}
\begin{tabular}{lrrrrrrrrrrrrrrr}
\toprule
                          & \multicolumn{3}{c}{2020 NP-I}                                                  & \multicolumn{3}{c}{2020 NP-II}                                                  & \multicolumn{3}{c}{2020 NP-III}                                                  & \multicolumn{3}{c}{2021 NP-A}                                                & \multicolumn{3}{c}{2021 NP-B}                                                 \\
                          \cmidrule(lr){2-4}\cmidrule(lr){5-7}\cmidrule(lr){8-10}\cmidrule(lr){11-13}\cmidrule(lr){14-16}
                          & \multicolumn{1}{c}{T0pp} & \multicolumn{1}{c}{GPT3} & \multicolumn{1}{c}{RST} & \multicolumn{1}{c}{T0pp} & \multicolumn{1}{c}{GPT3} & \multicolumn{1}{c}{RST} & \multicolumn{1}{c}{T0pp} & \multicolumn{1}{c}{GPT3} & \multicolumn{1}{c}{RST} & \multicolumn{1}{c}{T0pp} & \multicolumn{1}{c}{GPT3} & \multicolumn{1}{c}{RST} & \multicolumn{1}{c}{T0pp} & \multicolumn{1}{c}{GPT3} & \multicolumn{1}{c}{RST} \\
                          \midrule
Listening                 
& \multicolumn{1}{>{\columncolor{tablegreen}}r}{27.0}                       
& \multicolumn{1}{>{\columncolor{tablegreen}}r}{28.5}
& \multicolumn{1}{>{\columncolor{tablegreen}}r}{27.0}
& 27.0                       
& 28.5
& 27.0 
& \multicolumn{1}{>{\columncolor{tablegreen}}r}{27.0}                       
& \multicolumn{1}{>{\columncolor{tablegreen}}r}{28.5}
& \multicolumn{1}{>{\columncolor{tablegreen}}r}{27.0}  
& 27.0                       
& 25.5  
& 28.5 
& \multicolumn{1}{>{\columncolor{tablegreen}}r}{27.0}                       
& \multicolumn{1}{>{\columncolor{tablegreen}}r}{25.5}
& \multicolumn{1}{>{\columncolor{tablegreen}}r}{28.5} 
\\
Listening (ASR)           
& \multicolumn{1}{>{\columncolor{tablegreen}}r}{22.5}                     
& \multicolumn{1}{>{\columncolor{tablegreen}}r}{28.5}
& \multicolumn{1}{>{\columncolor{tablegreen}}r}{27.0}  
& 22.5                     
& 28.5   
& 27.0
& \multicolumn{1}{>{\columncolor{tablegreen}}r}{22.5}                     
& \multicolumn{1}{>{\columncolor{tablegreen}}r}{28.5} 
& \multicolumn{1}{>{\columncolor{tablegreen}}r}{27.0}  
& 25.5                     
& 24.0   
& 28.5       
& \multicolumn{1}{>{\columncolor{tablegreen}}r}{25.5}                     
& \multicolumn{1}{>{\columncolor{tablegreen}}r}{24.0} 
& \multicolumn{1}{>{\columncolor{tablegreen}}r}{28.5}   
\\
Cloze (MC)
& \multicolumn{1}{>{\columncolor{tablegreen}}r}{16.5}                     
& \multicolumn{1}{>{\columncolor{tablegreen}}r}{22.5} 
& \multicolumn{1}{>{\columncolor{tablegreen}}r}{28.5}  
& 22.5                     
& 22.5   
& 28.5 
& \multicolumn{1}{>{\columncolor{tablegreen}}r}{13.5}                     
& \multicolumn{1}{>{\columncolor{tablegreen}}r}{25.5}   
& \multicolumn{1}{>{\columncolor{tablegreen}}r}{30.0} 
& 13.5                     
& 24.0    
& 27.0  
& \multicolumn{1}{>{\columncolor{tablegreen}}r}{15.0}                       
& \multicolumn{1}{>{\columncolor{tablegreen}}r}{21.0}    
& \multicolumn{1}{>{\columncolor{tablegreen}}r}{27.0}  
\\
Cloze (hint)              
& \multicolumn{1}{>{\columncolor{tablegreen}}r}{0.0}                        
& \multicolumn{1}{>{\columncolor{tablegreen}}r}{1.5}    
& \multicolumn{1}{>{\columncolor{tablegreen}}r}{12.0}   
& 0.0                        
& 4.5    
& 10.5 
& \multicolumn{1}{>{\columncolor{tablegreen}}r}{1.5}                        
& \multicolumn{1}{>{\columncolor{tablegreen}}r}{7.5}    
& \multicolumn{1}{>{\columncolor{tablegreen}}r}{10.5}    
& 0.0                        
& 7.5  
& 7.5      
& \multicolumn{1}{>{\columncolor{tablegreen}}r}{0.0}                        
& \multicolumn{1}{>{\columncolor{tablegreen}}r}{4.5}  
& \multicolumn{1}{>{\columncolor{tablegreen}}r}{12.0} 
\\
Reading (MC) 
& \multicolumn{1}{>{\columncolor{tablegreen}}r}{18.0}                       
& \multicolumn{1}{>{\columncolor{tablegreen}}r}{26.0} 
& \multicolumn{1}{>{\columncolor{tablegreen}}r}{20.0} 
& 28.0                       
& 28.0      
& 28.0    
& \multicolumn{1}{>{\columncolor{tablegreen}}r}{24.0}                       
& \multicolumn{1}{>{\columncolor{tablegreen}}r}{26.0} 
& \multicolumn{1}{>{\columncolor{tablegreen}}r}{28.0} 
& 24.0                       
& 22.0 
& 28.0 
& \multicolumn{1}{>{\columncolor{tablegreen}}r}{26.0}                       
& \multicolumn{1}{>{\columncolor{tablegreen}}r}{28.0}  
& \multicolumn{1}{>{\columncolor{tablegreen}}r}{28.0}   
\\
Reading (cloze)           
& \multicolumn{1}{>{\columncolor{tablegreen}}r}{2.0}                        
& \multicolumn{1}{>{\columncolor{tablegreen}}r}{8.0}   
& \multicolumn{1}{>{\columncolor{tablegreen}}r}{10.0}   
& 2.0                        
& 10.0 
& 10.0  
& \multicolumn{1}{>{\columncolor{tablegreen}}r}{0.0}                        
& \multicolumn{1}{>{\columncolor{tablegreen}}r}{2.0}  
& \multicolumn{1}{>{\columncolor{tablegreen}}r}{8.0}    
& 2.0                        
& 4.0    
& 10.0   
& \multicolumn{1}{>{\columncolor{tablegreen}}r}{0.0}                        
& \multicolumn{1}{>{\columncolor{tablegreen}}r}{6.0}  
& \multicolumn{1}{>{\columncolor{tablegreen}}r}{8.0} 
\\
Writing (grammar)         
& \multicolumn{1}{>{\columncolor{tablegreen}}r}{1.0}                        
& \multicolumn{1}{>{\columncolor{tablegreen}}r}{7.0}  
& \multicolumn{1}{>{\columncolor{tablegreen}}r}{7.0}  
& 0.0                        
& 10.0   
& 5.0    
& \multicolumn{1}{>{\columncolor{tablegreen}}r}{0.0}                        
& \multicolumn{1}{>{\columncolor{tablegreen}}r}{10.0}     
& \multicolumn{1}{>{\columncolor{tablegreen}}r}{6.0}  
& 0.0                        
& 8.0    
& 8.0    
& \multicolumn{1}{>{\columncolor{tablegreen}}r}{0.0}                        
& \multicolumn{1}{>{\columncolor{tablegreen}}r}{8.0}    
& \multicolumn{1}{>{\columncolor{tablegreen}}r}{4.0}    
\\
Writing (essay writing)    
& \multicolumn{1}{>{\columncolor{tablegreen}}r}{5.0}                        
& \multicolumn{1}{>{\columncolor{tablegreen}}r}{22.0}   
& \multicolumn{1}{>{\columncolor{tablegreen}}r}{21.0}   
& 1.0                        
& 20.0        
& 20.0    
& \multicolumn{1}{>{\columncolor{tablegreen}}r}{0.0}                        
& \multicolumn{1}{>{\columncolor{tablegreen}}r}{20.0}    
& \multicolumn{1}{>{\columncolor{tablegreen}}r}{21.0}   
& 5.0                        
& 21.0      
& 19.0      
& \multicolumn{1}{>{\columncolor{tablegreen}}r}{7.0}                        
& \multicolumn{1}{>{\columncolor{tablegreen}}r}{22.0}    
& \multicolumn{1}{>{\columncolor{tablegreen}}r}{20.0} 
\\
\midrule
Total score               
& \multicolumn{1}{>{\columncolor{tablegreen}}r}{69.5}                     
& \multicolumn{1}{>{\columncolor{tablegreen}}r}{115.5}    
& \multicolumn{1}{>{\columncolor{tablegreen}}r}{\textbf{125.5}}  
& 80.5                     
& 123.5     
& \textbf{129.0}    
& \multicolumn{1}{>{\columncolor{tablegreen}}r}{66}                     
& \multicolumn{1}{>{\columncolor{tablegreen}}r}{119.5}   
& \multicolumn{1}{>{\columncolor{tablegreen}}r}{\textbf{130.5}}    
& 71.5                    
& 112.0         
& \textbf{128.0}   
& \multicolumn{1}{>{\columncolor{tablegreen}}r}{75.0}                       
& \multicolumn{1}{>{\columncolor{tablegreen}}r}{115.0}    
& \multicolumn{1}{>{\columncolor{tablegreen}}r}{\textbf{127.5}}  
\\
Total score (ASR)         
& \multicolumn{1}{>{\columncolor{tablegreen}}r}{65.0}                       
& \multicolumn{1}{>{\columncolor{tablegreen}}r}{115.5}   
& \multicolumn{1}{>{\columncolor{tablegreen}}r}{\textbf{125.5}}    
& 76.0                       
& 123.5     
& \textbf{129.0}   
& \multicolumn{1}{>{\columncolor{tablegreen}}r}{61.5}                       
& \multicolumn{1}{>{\columncolor{tablegreen}}r}{119.5}    
& \multicolumn{1}{>{\columncolor{tablegreen}}r}{\textbf{130.5}}    
& 70.0                       
& 110.5     
& \textbf{128.0}  
& \multicolumn{1}{>{\columncolor{tablegreen}}r}{73.5}                     
& \multicolumn{1}{>{\columncolor{tablegreen}}r}{113.5}
& \multicolumn{1}{>{\columncolor{tablegreen}}r}{\textbf{127.5}}  
\\
\bottomrule
\end{tabular}
\end{table}

\subsubsection{Model Tuning}
\paragraph{Comprehension \& Language Usage} We follow the basic training setting as in \S\ref{sec:mTain} and fine-tune the model for 200,000 steps. We save a checkpoint every 10,000 steps, and the best model is chosen based on the accuracy on the validation set.


\paragraph{Writing} We follow the basic training setting as in \S\ref{sec:mTain} and fine-tune the model for 2,000 steps. We save a checkpoint every 100 steps, and the best model is chosen based on its average rank on the grammar error correction and the essay writing validation sets.

\subsection{Baseline}
The baselines we compared with are T0pp and GPT3. We use the ``text-davinci-002" GPT3 model, which contains 175B parameters, more than 15 times the number of parameters of our model. We consider zero-shot prompting for T0pp and 2-shot in-context learning for GPT3. More specifically, for GPT3, for each question type, we use two questions of the same type from 2017 National Paper I as in-context examples. We also introduce the average scores of students who participate in the Gaokao.

\subsection{Evaluation} 
\label{subsec:gaokao_evaluation}
The evaluation metric for all question types (except essay writing) is accuracy. For essay writing, we ask two high school English teachers in China to score each essay. We take the minimum score as the final score. For the listening part, we investigate two settings: (i) directly use the audio transcript. (ii) use the Google speech-to-text service to transform the original MP3 audio file into text and use some heuristics to break the whole transcript into contexts of different questions. For the essay writing part, we use Google Translation API\footnote{\url{https://cloud.google.com/translate/docs/reference/rest}} to translate the original Chinese question into its English version.

\subsection{Results}
\label{subsec:gaokao_results}
The results are shown in Tab.~\ref{tab:gaokao_results_part1}-\ref{tab:gaokao_results_part2}. Overall, we make the following observations.
\begin{enumerate}
    \item For each English national paper we considered, RST achieves the highest total score for both listening sets, with an average score of 130.6.
    \item Compared with T0pp, RST performs far better than it with the same model size. The total score achieved by RST is, on average, 54.5 points higher than T0pp in all considered settings, with a maximum of 69 points (46\% of the total score).
    \item Compared with GPT3, RST can achieve significantly better results with a model size 16 times smaller. The total score achieved by RST is, on average, 14.0 points higher than T0pp in all considered settings, with a maximum of 26 points (17\% of the total score). 
    \item For T0pp, the listening scores obtained using gold and speech-to-text transcripts differ considerably, with an average of 4.2 points. This compares to 0.6 and 0.45 for GPT3 and RST, respectively, indicating that T0pp's performance is sensitive to the quality of the text. Usually, the transcripts obtained from speech-to-text are often wrong for the segmentation boundary between different speakers' speech and also have some textual errors. These lead to differences in the performance of models under the two settings. However, the text quality does not affect GPT3 and RST much, indicating their robustness.
\end{enumerate}

\begin{figure*}[!t]
\centering
\includegraphics[width=1\linewidth]{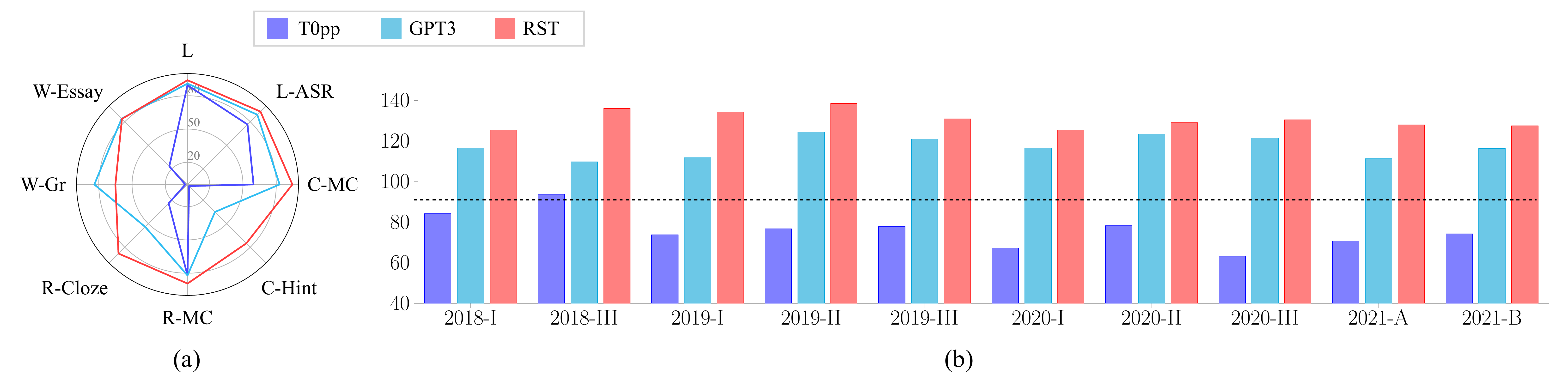}
\caption{Gokao performance of T0pp, GPT3, RST. (a) We plot the percentage of scores for different models on different question subcategories. We use the following abbreviations. L: listening, C: cloze, R: reading, W: writing, MC: multiple-choice, Gr: grammar. 
(b) We plot the average total score (considering ASR and original transcripts) of the different models on each paper. The black horizontal line represents the average score of all students.
The average score of students varies slightly from year to year, and here we take the maximum value.
}
\label{fig:gaokao_analysis}
\end{figure*}

\subsection{Analysis}
\subsubsection{Fine-grained analysis}
We conduct a fine-grained analysis to understand how different models perform on different question subcategories. In Fig.~\ref{fig:gaokao_analysis}-(a), it is obvious that RST and GPT3 can outperform T0pp on every question subcategory. \\
\noindent\textbf{T0pp} does well on the listening and reading (multiple-choice) questions but poorly on the cloze (hint) and grammar questions. This is because the model has seen many of the former question types but not the latter during fine-tuning, which raises doubts about the model's zero-shot generalization after multitask fine-tuning. \\
\noindent\textbf{GPT3} performs better on cloze (hint) questions than T0pp because the prompt form of cloze (hint), although difficult to occur in natural texts, is not completely absent. In addition, GPT3 performs very well in grammar and essay questions and outperforms T0pp and RST in general, demonstrating that the super-large scale language model is sufficient to gain strong generative power after pre-training on a huge corpus and can perform well on various generative downstream tasks.\\
\noindent\textbf{RST} outperforms T0pp and GPT3 on all question subcategories except writing (grammar), and for some of those, even though the signals we construct are weak supervisions (e.g., cloze(hint)), we are able to achieve much better results than T0pp and GPT3. This further demonstrates the effectiveness of pre-training over signals and the importance of constructing and collecting signals.

\subsubsection{Comparisons to humans}
Fig.~\ref{fig:gaokao_analysis}-(b) shows the models' performance and the students' average performance on the national papers in recent years. It is obvious that  T0pp's total score on 9/10 papers is below the students' average, while RST and GPT3 exceed the students' average performance. In particular, in five of these ten papers, the total score achieved by RST exceeds 130  (usually considered the target score for students to strive for). 

It is worth noting that it does not mean that the model is perfect. According to the feedback from high school teachers, there is still a gap between the essays written by the model and those written by top students. The main problem with model-written essays is the lack of novelty and details. In addition, our experiment setting does not completely simulate the Gaokao exam because the possibility of achieving a high score in the exam also depends on many aspects, such as the beauty of the writing font.



\subsection{Latest 2022 Gaokao}
\textbf{The 2022 Gaokao-English exams (2022.06.08) have just ended, and we are keeping up with current events to find out how models perform on the latest year's Gaokao papers.} We consider GPT3 and RST since they are the best performing models according to \S\ref{subsec:gaokao_results}. The evaluation is the same as described in \S\ref{subsec:gaokao_evaluation} except that for listening, we only consider the setting which takes the released mp3 file as input. The results are shown in Tab.~\ref{tab:gaokao_2022}. \textbf{RST achieves a total score of 134, which is much higher than the 108 score achieved by GPT3.}

\begin{table}[ht]
\centering
\footnotesize
\setlength\tabcolsep{4pt}
\renewcommand{\arraystretch}{1.1}
\caption{Results of GPT3 and RST in Gaokao-English 2022.}
\label{tab:gaokao_2022}
\begin{tabular}{lcccccccc}
\toprule
                        & Listening                     & Cloze             & Cloze                          & Reading                                   & Reading & Writing & Writing         &                               \\
\multirow{-2}{*}{Model} & (ASR) & (multiple-choice) & (hint) & (multiple-choice) & (cloze) & (grammar)                       & (essay writing) & \multirow{-2}{*}{Total Score} \\
\midrule
GPT3                    & 28.5                          & 21.0                                        & 4.5                            & 26.0                                        & 4.0                               & 5.0                               & 19.0              &   \textbf{108}                      \\
RST                     & 30.0                            & 25.5                                      & 12.0                             & 30.0                                        & 10.0                              & 7.0                               & 19.5              &    \textbf{134}                \\
\bottomrule
\end{tabular}
\end{table}

\section{Related Work} 



When we demonstrate the effectiveness of a machine learning method, it is usually carried out within a specific scope of application, and a large scope of justification commonly comes at much more workload, e.g., more experiments and evaluated data.
The goal of this work is obviously not to improve the performance of one or several specific task(s) through data manipulation, which discriminates our work from most existing works, not to mention other contributions we have listed in the \textit{contribution} section. For the first time, we are learning from so many different signals in the world in a unified way, not trying to distinguish between supervised and unsupervised data, but caring about how much we can use the information that nature gives us. We will detail some selected aspects below.


\paragraph{Knowledge enhanced PLM}
There are mainly two lines of works that try to incorporate those signals into pre-trained language models. The first line of works introduces auxiliary objectives during \textit{pre-training}. Representative works include \cite{zhang-etal-2019-ernie, DBLP:conf/aaai/SunWLFTWW20, DBLP:journals/corr/abs-2107-02137}. However, they typically use a limited number of supervisions since they focus on mining supervision signals in plain text. The other line of works tries to incorporate various signals during \textit{fine-tuning}, as done in \cite{DBLP:journals/corr/abs-2109-01652, sanh2021multitask}. Although the types of training signals they consider are much diverse, they are restricted by the existence of curated labeled datasets.
In this work, we ask a question: \textit{``Can we bridge the gap between unlabeled data and labeled data and propose a unified framework for pre-training?"}. To make a step towards answering this question, we propose a new paradigm which we dub \textit{``reStructure, Pre-train, Fine-tune/Prompt"} (RST). 

\paragraph{Multitask Prompted Training}
The emergence of prompting methods has opened up new possibilities for multitask learning. By applying appropriate prompts, the input and output of different tasks can be transformed into texts with similar templates, thus making it easy to perform multitask learning. Representative works include FLAN \cite{DBLP:journals/corr/abs-2109-01652}, T0 \cite{sanh2021multitask}, and Polyglot Prompt \cite{fu2022polyglot}. However, these works focus on fine-tuning the language model with supervised datasets from downstream tasks, while we are concerned with pre-training the language model with diverse restructured signals (more than 26 types of signals from 10 different data mines).
Other works, such as PPT \cite{gu-etal-2022-ppt} also introduce ``prompts'' into the ``pre-training'' stage. However they focus on ``pre-trained prompt'' to get a good initialization of soft prompts for prompt tuning instead of ``prompted pre-training''.

\paragraph{Data-centric Machine Learning}
Data-centric machine learning shifts the focus from fiddling with model architectures to ensuring data quality \cite{miranda2021datacentric}. A number of studies \cite{axelrod-etal-2015-class, xu-koehn-2017-zipporah, dou-etal-2020-dynamic, https://doi.org/10.48550/arxiv.2112.07844} have shown that improving data quality (e.g., selecting more informative samples for training) can lead to significant model improvement. On the one hand, deep learning models are being democratized, and everyone is able to take advantage of performant language models, either by running them locally (BERT) or through APIs (GPT3). On the other hand, in recent years, practices represented by MLOps\footnote{https://ml-ops.org/} are gradually being adopted to support the whole ML lifecycle. We have witnessed the emergence of tools geared towards data transformation, labeling, and tracking~\cite{xiao-etal-2022-datalab}. In this context, data-centric machine learning is gradually becoming a major trend, and restructured pre-training is a typical example of exploring data-centric ML in language pre-training. By mining data mines, we extract, clean, and transform the signals so that the model can achieve far better than baseline performance on mainstream NLP tasks even without any training data from downstream tasks. 
However, RST does not exclude fine-tuning on downstream tasks, which is well-supported because, in the opinion of RST, supervised datasets on downstream tasks are just different data mines. For example, we can continue restructuring more signals from labeled data to further enhance the model.

\paragraph{Weakly supervised/Distant Supervised Learning}
In some domains where labeled data are not readily available in large quantities, researchers usually collect data for machine learning in a weakly-supervised way \cite{mintz2009distant,li2012wiki,Zhou2018ABI}.
In this paper, weakly-supervised learning is one of the ways we use to collect more types of signals. Compared with existing works that usually focus on one specific task, we mainly care about how to use the valuable information that the world has provided as much as possible, rather than a specific method used for signal extraction. Moreover, we have considered 26 different types of signals in the world.

\section*{Acknowledgements}

The authors would like to thank Google Cloud Research Credits Program and TPU Research Cloud Program (TRC), which lay the foundation for model training. We also would like to thank High Flyer for the generous GPU support, which accelerates the inference process of our models. Also, thanks to Jie Fu, who helped us search for potential computational resources when we need. Thanks to Zhijiang Lab for the temporary GPU support. All other types of expenses, including annotation, cloud storage, etc., are \textit{self-funded} by Pengfei Liu.

\noindent We are very grateful to all the annotators of Gaokao papers, which are very valuable to the establishment of the Gaokao benchmark: Jiao Zhang, Wenjuan Shao, Yuru Deng, Xuelei Liang.
Also, thanks to Weizhe Yuan for careful post-edits of annotated papers.
Thank Zhao Yang for searching for qualified high school teachers for Gaokao essay scoring.
Additionally, many thanks to Zhendang Wan, who is a famous English teacher from No.1 High school of Huaibei and has been patiently helping with the essay scoring work and giving constructive suggestions.
Thank Hiroaki Hayashi, Jie Fu, Jinlan Fu for proofreading the paper.
Thanks to Graham Neubig and Catherine Cui for helping out with building benchmark systems.
Thanks to everyone who created data that was used in the training of our models.

\newpage

\bibliography{main}
\bibliographystyle{acl_natbib}

\newpage

\appendix

\section{Prompts for Different Tasks}
\label{app:prompts}
\subsection{Rotten Tomatoes}
With $($review, sentiment$)$ pairs, we use the following prompts to transform them into textual data. We use \{answer\} to represent the appropriate target based on \{sentiment\}.
\begin{itemize}
    \item Multiple-choice format prompts
    \begin{enumerate}
    \item \textcolor{red}{source}: \texttt{TEXT}: \{review\} \texttt{QUERY}: What's the sentiment of this text? ``Positive", ``Negative" or ``Neutral"?\\
    \textcolor{blue}{target}: \{sentiment\}
    
    \item \textcolor{red}{source}: \texttt{TEXT}: \{review\} \texttt{QUERY}: Can you judge the sentiment of this text? The options are ``Positive", ``Negative" and ``Neutral".\\
    \textcolor{blue}{target}: \{sentiment\}
    
    \item \textcolor{red}{source}: \texttt{TEXT}: \{review\} \texttt{QUERY}: Does it seem like the reviewer who wrote this review liked the movie? ``Yes" or ``No"?\\
    \textcolor{blue}{target}: \{answer\}
    
    \item \textcolor{red}{source}: \texttt{TEXT}: \{review\} \texttt{QUERY}: Does it seem like the reviewer who wrote this review disliked the movie? ``Yes" or ``No"?\\
    \textcolor{blue}{target}: \{answer\}
    
    \item \textcolor{red}{source}: \texttt{TEXT}: \{review\} \texttt{QUERY}: Assign the correct sentiment to this review. Please choose from ``Positive", ``Negative", ``Neutral".\\
    \textcolor{blue}{target}: \{sentiment\}
    
    \item \textcolor{red}{source}: \texttt{TEXT}: \{review\} \texttt{QUERY}: Does the reviewer like this movie based on the review text? ``Yes" or ``No"?\\
    \textcolor{blue}{target}: \{answer\}
    
    \item \textcolor{red}{source}: \texttt{TEXT}: \{review\} \texttt{QUERY}: Based on the review, the reviewer likes this movie. ``True" or ``False"?\\
    \textcolor{blue}{target}: \{answer\}
    
    \item \textcolor{red}{source}: \texttt{TEXT}: \{review\} \texttt{QUERY}: Judge the sentiment of the text. You can choose from ``Positive", ``Negative", ``Neutral".\\
    \textcolor{blue}{target}: \{sentiment\}
    
    \item \textcolor{red}{source}: \texttt{TEXT}: \{review\} \texttt{QUERY}: Can you tell the sentiment of the text? ``Positive", ``Negative" or ``Neutral"?\\
    \textcolor{blue}{target}: \{sentiment\}
    
    \item \textcolor{red}{source}: \texttt{TEXT}: \{review\} \texttt{QUERY}: Is the review ``positive" or ``negative" or ``neutral"?\\
    \textcolor{blue}{target}: \{sentiment\}
    
    \item \textcolor{red}{source}: \texttt{TEXT}: \{review\} \texttt{QUERY}: Is the sentiment of this text positive? ``Yes" or ``No"?\\
    \textcolor{blue}{target}: \{answer\}
    
    \item \textcolor{red}{source}: \texttt{TEXT}: \{review\} \texttt{QUERY}: Is the sentiment of this text neutral? ``Yes" or ``No"?\\
    \textcolor{blue}{target}: \{answer\}
    
    \item \textcolor{red}{source}: \texttt{TEXT}: \{review\} \texttt{QUERY}: Is the sentiment of this text negative? ``Yes" or ``No"?\\
    \textcolor{blue}{target}: \{answer\}
    
    \item \textcolor{red}{source}: \texttt{TEXT}: \{review\} \texttt{QUERY}: Is it likely that the person who wrote this review will recommend this movie to others? ``Yes" or ``No"?\\
    \textcolor{blue}{target}: \{answer\}
    
    \item \textcolor{red}{source}: \texttt{TEXT}: \{review\} \texttt{QUERY}: Would the person who wrote this review advise others not to see this movie? ``Yes" or ``No"?\\
    \textcolor{blue}{target}: \{answer\}
    
    \item \textcolor{red}{source}: \texttt{TEXT}: \{review\} \texttt{QUERY}: There is positive sentiment within this text. ``True" or ``False"?\\
    \textcolor{blue}{target}: \{answer\}
    
    \item \textcolor{red}{source}: \texttt{TEXT}: \{review\} \texttt{QUERY}: There is negative sentiment within this text. ``True" or ``False"?\\
    \textcolor{blue}{target}: \{answer\}
    
    \item \textcolor{red}{source}: \texttt{TEXT}: \{review\} \texttt{QUERY}: The sentiment of this text is neutral. ``True" or ``False"?\\
    \textcolor{blue}{target}: \{answer\}
    
    \item \textcolor{red}{source}: \texttt{TEXT}: \{review\} \texttt{QUERY}: Please identify the correct sentiment of this text. You may choose from ``Positive", ``Negative", ``Neutral".\\
    \textcolor{blue}{target}: \{sentiment\}
    
    \item \textcolor{red}{source}: \texttt{TEXT}: \{review\} \texttt{QUERY}: Given the options: ``Positive", ``Negative", ``Neutral", which one best describes the previous text?\\
    \textcolor{blue}{target}: \{sentiment\}
    \end{enumerate}

    \item Generation format prompts
    \begin{enumerate}
        \item \textcolor{red}{source}: \texttt{TEXT}: \{review\} \texttt{QUERY}: What's the sentiment of this text?\\
        \textcolor{blue}{target}: \{sentiment\}
        
        \item \textcolor{red}{source}: \texttt{TEXT}: \{review\} \texttt{QUERY}: Can you judge the sentiment of this text?\\
        \textcolor{blue}{target}: \{sentiment\}
        
        \item \textcolor{red}{source}: \texttt{TEXT}: \{review\} \texttt{QUERY}: Does it seem like the reviewer who wrote this review liked the movie?\\
        \textcolor{blue}{target}: \{answer\}
        
        \item \textcolor{red}{source}: \texttt{TEXT}: \{review\} \texttt{QUERY}: Does it seem like the reviewer who wrote this review disliked the movie?\\
        \textcolor{blue}{target}: \{answer\}
        
        \item \textcolor{red}{source}: \texttt{TEXT}: \{review\} \texttt{QUERY}: Assign the correct sentiment to this review.\\
        \textcolor{blue}{target}: \{sentiment\}
        
        \item \textcolor{red}{source}: \texttt{TEXT}: \{review\} \texttt{QUERY}: Does the reviewer like this movie based on the review text?\\
        \textcolor{blue}{target}: \{answer\}
        
        \item \textcolor{red}{source}: \texttt{TEXT}: \{review\} \texttt{QUERY}: According to the review, does the reviewer like this movie?\\
        \textcolor{blue}{target}: \{answer\}
        
        \item \textcolor{red}{source}: \texttt{TEXT}: \{review\} \texttt{QUERY}: Judge the sentiment of the text.\\
        \textcolor{blue}{target}: \{sentiment\}
        
        \item \textcolor{red}{source}: \texttt{TEXT}: \{review\} \texttt{QUERY}: Can you tell the sentiment of the text?\\
        \textcolor{blue}{target}: \{sentiment\}
        
        \item \textcolor{red}{source}: \texttt{TEXT}: \{review\} \texttt{QUERY}: Can you decide the sentiment polarity of the review?\\
        \textcolor{blue}{target}: \{sentiment\}
        
        \item \textcolor{red}{source}: \texttt{TEXT}: \{review\} \texttt{QUERY}: Is the sentiment of this text positive?\\
        \textcolor{blue}{target}: \{answer\}
        
        \item \textcolor{red}{source}: \texttt{TEXT}: \{review\} \texttt{QUERY}: Is the sentiment of this text neutral?\\
        \textcolor{blue}{target}: \{answer\}
        
        \item \textcolor{red}{source}: \texttt{TEXT}: \{review\} \texttt{QUERY}: Is the sentiment of this text negative?\\
        \textcolor{blue}{target}: \{answer\}
        
        \item \textcolor{red}{source}: \texttt{TEXT}: \{review\} \texttt{QUERY}: Is it likely that the person who wrote this review will recommend this movie to others? \\
        \textcolor{blue}{target}: \{answer\}
        
        \item \textcolor{red}{source}: \texttt{TEXT}: \{review\} \texttt{QUERY}: Would the person who wrote this review advise others not to see this movie?\\
        \textcolor{blue}{target}: \{answer\}
        
        \item \textcolor{red}{source}: \texttt{TEXT}: \{review\} \texttt{QUERY}: Is there positive sentiment within this text?\\
        \textcolor{blue}{target}: \{answer\}
        
        \item \textcolor{red}{source}: \texttt{TEXT}: \{review\} \texttt{QUERY}: Is there negative sentiment within this text?\\
        \textcolor{blue}{target}: \{answer\}
        
        \item \textcolor{red}{source}: \texttt{TEXT}: \{review\} \texttt{QUERY}: Is this a text with neutral sentiment?\\
        \textcolor{blue}{target}: \{answer\}
        
        \item \textcolor{red}{source}: \texttt{TEXT}: \{review\} \texttt{QUERY}: Please identify the correct sentiment of this text.\\
        \textcolor{blue}{target}: \{sentiment\}
        
        \item \textcolor{red}{source}: \texttt{TEXT}: \{review\} \texttt{QUERY}: What sentiment best describes the previous text?\\
        \textcolor{blue}{target}: \{sentiment\}
        
    \end{enumerate}
\end{itemize}
\subsection{Daily Mail}
\paragraph{Category}
With $($summary, category$)$ pairs, we use the following prompts to transform them into textual data. Here, we use \{choices\_with\_or\} (e.g. ``option1", ``option2", or ``option3") and \{choices\_without\_or\} (e.g. ``option1", ``option2", ``option3") to refer to all the available options. We use \{other\_category\} to refer to a random wrong category for the text we aim to classify, and \{answer\} to represent the appropriate target based on \{category\}.

\begin{itemize}
    \item Multiple-choice format prompts
    \begin{enumerate}
        \item \textcolor{red}{source}: \texttt{TEXT}: \{summary\} \texttt{QUERY}: What's this text about? \{choices\_with\_or\}?\\
        \textcolor{blue}{target}: \{category\}
        
        \item \textcolor{red}{source}: \texttt{TEXT}: \{summary\} \texttt{QUERY}: Classify this text. You may choose from \{choices\_without\_or\}.\\
        \textcolor{blue}{target}: \{category\}
        
        \item \textcolor{red}{source}: \texttt{TEXT}: \{summary\} \texttt{QUERY}: How would you categorize this text? \{choices\_with\_or\}?\\
        \textcolor{blue}{target}: \{category\}
        
        \item \textcolor{red}{source}: \texttt{TEXT}: \{summary\} \texttt{QUERY}: Is this text about ``\{other\_category\}"? ``Yes" or ``No"?\\
        \textcolor{blue}{target}: \{answer\}
        
        \item \textcolor{red}{source}: \texttt{TEXT}: \{summary\} \texttt{QUERY}: Is this text about ``\{category\}"? ``Yes" or ``No"?\\
        \textcolor{blue}{target}: \{answer\}
        
        \item \textcolor{red}{source}: \texttt{TEXT}: \{summary\} \texttt{QUERY}: Can you choose an appropriate class from the following list for this text? \{choices\_without\_or\}.\\
        \textcolor{blue}{target}: \{category\}
        
        \item \textcolor{red}{source}: \texttt{TEXT}: \{summary\} \texttt{QUERY}: Given a list of categories: \{choices\_without\_or\}, what category does the paragraph belong to?\\
        \textcolor{blue}{target}: \{category\}
        
        \item \textcolor{red}{source}: \texttt{TEXT}: \{summary\} \texttt{QUERY}: Is this paragraph related to ``\{other\_ctegory\}"? ``Yes" or ``No"?\\
        \textcolor{blue}{target}: \{answer\}

        \item \textcolor{red}{source}: \texttt{TEXT}: \{summary\} \texttt{QUERY}: Is this paragraph related to ``\{category\}"? ``Yes" or ``No"?\\
        \textcolor{blue}{target}: \{answer\}
        
        \item \textcolor{red}{source}: \texttt{TEXT}: \{summary\} \texttt{QUERY}: Pick one category for the previous text. The options are \{choices\_without\_or\}.\\
        \textcolor{blue}{target}: \{category\}
        
        \item \textcolor{red}{source}: \texttt{TEXT}: \{summary\} \texttt{QUERY}: Given a choice of categories: \{choices\_with\_or\}, the text refers to which one?\\
        \textcolor{blue}{target}: \{category\}
        
        \item \textcolor{red}{source}: \texttt{TEXT}: \{summary\} \texttt{QUERY}: What is the topic of the text? \{choices\_with\_or\}?\\
        \textcolor{blue}{target}: \{category\}
        
        \item \textcolor{red}{source}: \texttt{TEXT}: \{summary\} \texttt{QUERY}: The topic of this text is ``\{other\_category\}". ``True" or ``False"?\\
        \textcolor{blue}{target}: \{answer\}
        
        \item \textcolor{red}{source}: \texttt{TEXT}: \{summary\} \texttt{QUERY}: The topic of this text is ``\{category\}". ``True" or ``False"?\\
        \textcolor{blue}{target}: \{answer\}
        
        \item \textcolor{red}{source}: \texttt{TEXT}: \{summary\} \texttt{QUERY}: Can you identify the category of this text? \{choices\_with\_or\}?\\
        \textcolor{blue}{target}: \{category\}
        
        \item \textcolor{red}{source}: \texttt{TEXT}: \{summary\} \texttt{QUERY}: Select a class from the following that best describes the text: \{choices\_without\_or\}.\\
        \textcolor{blue}{target}: \{category\}
        
        \item \textcolor{red}{source}: \texttt{TEXT}: \{summary\} \texttt{QUERY}: Is this a piece of text regarding \{choices\_with\_or\}?\\
        \textcolor{blue}{target}: \{category\}
        
        \item \textcolor{red}{source}: \texttt{TEXT}: \{text\} \texttt{QUERY}: What category best describes this paragraph? \{choices\_with\_or\}?\\
        \textcolor{blue}{target}: \{category\}
        
        \item \textcolor{red}{source}: \texttt{TEXT}: \{summary\} \texttt{QUERY}: Please classify this text into one of the following: \{choices\_without\_or\}.\\
        \textcolor{blue}{target}: \{category\}
        
        \item \textcolor{red}{source}: \texttt{TEXT}: \{summary\} \texttt{QUERY}: What's the main topic of this paragraph? \{choices\_with\_or\}?\\
        \textcolor{blue}{target}: \{category\}
        
    \end{enumerate}
    \item Generation format prompts
    \begin{enumerate}
        \item \textcolor{red}{source}: \texttt{TEXT}: \{summary\} \texttt{QUERY}: What's this text about?\\
        \textcolor{blue}{target}: \{category\}
        
        \item \textcolor{red}{source}: \texttt{TEXT}: \{summary\} \texttt{QUERY}: Classify this text.\\
        \textcolor{blue}{target}: \{category\}
        
        \item \textcolor{red}{source}: \texttt{TEXT}: \{summary\} \texttt{QUERY}: How would you categorize this text?\\
        \textcolor{blue}{target}: \{category\}
        
        \item \textcolor{red}{source}: \texttt{TEXT}: \{summary\} \texttt{QUERY}: Is this text about ``\{other\_category\}"?\\
        \textcolor{blue}{target}: \{answer\}
        
        \item \textcolor{red}{source}: \texttt{TEXT}: \{summary\} \texttt{QUERY}: Is this text about ``\{category\}"?\\
        \textcolor{blue}{target}: \{answer\}
        
        \item \textcolor{red}{source}: \texttt{TEXT}: \{summary\} \texttt{QUERY}: Can you find an appropriate class for this text?\\
        \textcolor{blue}{target}: \{category\}
        
        \item \textcolor{red}{source}: \texttt{TEXT}: \{summary\} \texttt{QUERY}: What category does the paragraph belong to?\\
        \textcolor{blue}{target}: \{category\}
        
        \item \textcolor{red}{source}: \texttt{TEXT}: \{summary\} \texttt{QUERY}: Is this paragraph related to ``\{other\_category\}"?\\
        \textcolor{blue}{target}: \{answer\}
        
        \item \textcolor{red}{source}: \texttt{TEXT}: \{summary\} \texttt{QUERY}: Is this paragraph related to ``\{category\}"?\\
        \textcolor{blue}{target}: \{answer\}
        
        \item \textcolor{red}{source}: \texttt{TEXT}: \{summary\} \texttt{QUERY}: Pick one category for the previous text.\\
        \textcolor{blue}{target}: \{category\}
        
        \item \textcolor{red}{source}: \texttt{TEXT}: \{summary\} \texttt{QUERY}: The text refers to which category?\\
        \textcolor{blue}{target}: \{category\}
        
        \item \textcolor{red}{source}: \texttt{TEXT}: \{summary\} \texttt{QUERY}: What is the topic of the text?\\
        \textcolor{blue}{target}: \{category\}
        
        \item \textcolor{red}{source}: \texttt{TEXT}: \{summary\} \texttt{QUERY}: Is the topic of this text ``\{other\_category\}"?\\
        \textcolor{blue}{target}: \{answer\}
        
        \item \textcolor{red}{source}: \texttt{TEXT}: \{summary\} \texttt{QUERY}: Is the topic of this text ``\{category\}"?\\
        \textcolor{blue}{target}: \{answer\}
        
        \item \textcolor{red}{source}: \texttt{TEXT}: \{summary\} \texttt{QUERY}: Can you identify the category of this text?\\
        \textcolor{blue}{target}: \{category\}
        
        \item \textcolor{red}{source}: \texttt{TEXT}: \{summary\} \texttt{QUERY}: Think of a class that best fits the text.\\
        \textcolor{blue}{target}: \{category\}
        
        \item \textcolor{red}{source}: \texttt{TEXT}: \{summary\} \texttt{QUERY}: Can you classify this piece of text?\\
        \textcolor{blue}{target}: \{category\}
        
        \item \textcolor{red}{source}: \texttt{TEXT}: \{summary\} \texttt{QUERY}: What category best describes this paragraph?\\
        \textcolor{blue}{target}: \{category\}
        
        \item \textcolor{red}{source}: \texttt{TEXT}: \{summary\} \texttt{QUERY}: Please classify this text.\\
        \textcolor{blue}{target}: \{category\}
        
        \item \textcolor{red}{source}: \texttt{TEXT}: \{summary\} \texttt{QUERY}: What's the main topic of this paragraph?\\
        \textcolor{blue}{target}: \{category\}
    \end{enumerate}
\end{itemize}
\paragraph{Summary}
With $($text, summary$)$ pairs, we use the following prompts to transform them into textual data. Specifically, for some prompts, we add some hints for the final length with the hope that the model can learn to generate summaries with appropriate length. We use \{length\} to represent the length of the target summary.
\begin{itemize}
    \item Generation format prompts
    \begin{enumerate}
        \item \textcolor{red}{source}: \texttt{TEXT}: \{text\} \texttt{QUERY}: Can you summarize the previous text?\\
        \textcolor{blue}{target}: \{summary\}
        
        \item \textcolor{red}{source}: \texttt{TEXT}: \{text\} \texttt{QUERY}: Generate a summary for the text.\\
        \textcolor{blue}{target}: \{summary\}
        
        \item \textcolor{red}{source}: \texttt{TEXT}: \{text\} \texttt{QUERY}: Summarize the preceding text in your own words.\\
        \textcolor{blue}{target}: \{summary\}
        
        \item \textcolor{red}{source}: \texttt{TEXT}: \{text\} \texttt{QUERY}: What are the main points one should remember from this text?\\
        \textcolor{blue}{target}: \{summary\}
        
        \item \textcolor{red}{source}: \texttt{TEXT}: \{text\} \texttt{QUERY}: Can you generate a \{length\}-word summary for the previous text?\\
        \textcolor{blue}{target}: \{summary\}
        
        \item \textcolor{red}{source}: \texttt{TEXT}: \{text\} \texttt{QUERY}: In a few sentences, what does the previous paragraph say?\\
        \textcolor{blue}{target}: \{summary\}
        
        \item \textcolor{red}{source}: \texttt{TEXT}: \{text\} \texttt{QUERY}: How would you summarize the key points of the text?\\
        \textcolor{blue}{target}: \{summary\}
        
        \item \textcolor{red}{source}: \texttt{TEXT}: \{text\} \texttt{QUERY}: Condense the text down to the essentials.\\
        \textcolor{blue}{target}: \{summary\}
        
        \item \textcolor{red}{source}: \texttt{TEXT}: \{text\} \texttt{QUERY}: In around \{length\} words, summarize the article.\\
        \textcolor{blue}{target}: \{summary\}
        
        \item \textcolor{red}{source}: \texttt{TEXT}: \{text\} \texttt{QUERY}: In around \{length\} words, briefly describe what the previous paragraph talks about.\\
        \textcolor{blue}{target}: \{summary\}
        
        \item \textcolor{red}{source}: \texttt{TEXT}: \{text\} \texttt{QUERY}: In around \{length\} words, summarize the core of the above text.\\
        \textcolor{blue}{target}: \{summary\}
        
        \item \textcolor{red}{source}: \texttt{TEXT}: \{text\} \texttt{QUERY}: In around \{length\} words, write a TLDR (Too Long Didn''t Read) summary for the above text.\\
        \textcolor{blue}{target}: \{summary\}
        
        \item \textcolor{red}{source}: \texttt{TEXT}: \{text\} \texttt{QUERY}: Can you summarize the previous text in about \{length\} words?\\
        \textcolor{blue}{target}: \{summary\}
        
        \item \textcolor{red}{source}: \texttt{TEXT}: \{text\} \texttt{QUERY}: Can you conclude the previous article in about \{length\} words?\\
        \textcolor{blue}{target}: \{summary\}
        
        \item \textcolor{red}{source}: \texttt{TEXT}: \{text\} \texttt{QUERY}: Please write a summary of about \{length\} words based on the previous article.\\
        \textcolor{blue}{target}: \{summary\}
        
        \item \textcolor{red}{source}: \texttt{TEXT}: \{text\} \texttt{QUERY}: Based on the previous text, can you write a \{length\}-word summary?\\
        \textcolor{blue}{target}: \{summary\}
        
        \item \textcolor{red}{source}: \texttt{TEXT}: \{text\} \texttt{QUERY}: What is the previous article about? Please summarize in about \{length\} words.\\
        \textcolor{blue}{target}: \{summary\}

        \item \textcolor{red}{source}: \texttt{TEXT}: \{text\} \texttt{QUERY}: Can you list the key points of the text?\\
        \textcolor{blue}{target}: \{summary\}
        
        \item \textcolor{red}{source}: \texttt{TEXT}: \{text\} \texttt{QUERY}: What can be a short description of the text?\\
        \textcolor{blue}{target}: \{summary\}
        
        \item \textcolor{red}{source}: \texttt{TEXT}: \{text\} \texttt{QUERY}: Can you express the main content of the text?\\
        \textcolor{blue}{target}: \{summary\}
        
    \end{enumerate}
\end{itemize}

With $($text, title$)$ pairs, we use both generation format prompts and multiple-choice format prompts, as shown below. Here, we use \{choices\_with\_or\} (e.g. ``option1", ``option2", or ``option3") and \{choices\_without\_or\} (e.g. ``option1", ``option2", ``option3") to refer to all the available options. We use \{other\_title\} to refer to the title of a random article.
\begin{itemize}
    \item Multiple-choice formats
    \begin{enumerate}
        \item \textcolor{red}{source}: \texttt{TEXT}: \{text\} \texttt{QUERY}: Which of the following titles can summarize this text? \{choices\_with\_or\}?\\
        \textcolor{blue}{target}: \{title\}
        
        \item \textcolor{red}{source}: \texttt{TEXT}: \{text\} \texttt{QUERY}: Select a headline for the previous text. The options are \{choices\_without\_or\}.\\
        \textcolor{blue}{target}: \{title\}
        
        \item \textcolor{red}{source}: \texttt{TEXT}: \{text\} \texttt{QUERY}: Given a list of titles: \{choices\_without\_or\}, which one can express the main idea of the text?\\
        \textcolor{blue}{target}: \{title\}
        
        \item \textcolor{red}{source}: \texttt{TEXT}: \{text\} \texttt{QUERY}: Given a list of headings: \{choices\_without\_or\}, which of these could be used as a heading for the above text?\\
        \textcolor{blue}{target}: \{title\}
        
        \item \textcolor{red}{source}: \texttt{TEXT}: \{text\} \texttt{QUERY}: Which of the following headings matches the above text? \{choices\_with\_or\}?\\
        \textcolor{blue}{target}: \{title\}
        
        \item \textcolor{red}{source}: \texttt{TEXT}: \{text\} \texttt{QUERY}: Which of the following titles covers the essence of the above text? \{choices\_with\_or\}?\\
        \textcolor{blue}{target}: \{title\}
        
        \item \textcolor{red}{source}: \texttt{TEXT}: \{text\} \texttt{QUERY}: How would you summarize the text? You may choose from \{choices\_without\_or\}.\\
        \textcolor{blue}{target}: \{title\}
        
        \item \textcolor{red}{source}: \texttt{TEXT}: \{text\} \texttt{QUERY}: Choose a title for the above text from: \{choices\_without\_or\}. \\
        \textcolor{blue}{target}: \{title\}
        
        \item \textcolor{red}{source}: \texttt{TEXT}: \{text\} \texttt{QUERY}: Select an appropriate heading for the article. The options are: \{choices\_without\_or\}.\\
        \textcolor{blue}{target}: \{title\}
        
        \item \textcolor{red}{source}: \texttt{TEXT}: \{text\} \texttt{QUERY}: Which of the following is the heading of the article? \{choices\_with\_or\}?\\
        \textcolor{blue}{target}: \{title\}
        
        \item \textcolor{red}{source}: \texttt{TEXT}: \{text\} \texttt{QUERY}: ``\{title\}" can summarize the previous text. ``True" or ``False"?\\
        \textcolor{blue}{target}: True
        
        \item \textcolor{red}{source}: \texttt{TEXT}: \{text\} \texttt{QUERY}: ``\{other\_title\}" can summarize the previous text. ``True" or ``False"?\\
        \textcolor{blue}{target}: False
        
        \item \textcolor{red}{source}: \texttt{TEXT}: \{text\} \texttt{QUERY}: Does ``\{title\}" summarize the core of the above text? ``Yes" or ``No"?\\
        \textcolor{blue}{target}: Yes
        
        \item \textcolor{red}{source}: \texttt{TEXT}: \{text\} \texttt{QUERY}: Does ``\{other\_title\}" summarize the core of the above text? ``Yes" or ``No"?\\
        \textcolor{blue}{target}: No
        
        \item \textcolor{red}{source}: \texttt{TEXT}: \{text\} \texttt{QUERY}: Here are some titles: \{choices\_without\_or\}, which one is compatible with the previous text?\\
        \textcolor{blue}{target}: \{title\}
        
        \item \textcolor{red}{source}: \texttt{TEXT}: \{text\} \texttt{QUERY}: Can you choose an appropriate title of the preceding text from the following list: \{choices\_without\_or\}?\\
        \textcolor{blue}{target}: \{title\}
        
        \item \textcolor{red}{source}: \texttt{TEXT}: \{text\} \texttt{QUERY}: Can you pick the suitable title of the previous article from the following options: \{choices\_without\_or\}?\\
        \textcolor{blue}{target}: \{title\}
        
        \item \textcolor{red}{source}: \texttt{TEXT}: \{text\} \texttt{QUERY}: Given options: \{choices\_without\_or\}, which of these is an appropriate title of the preceding text?\\
        \textcolor{blue}{target}: \{title\}
        
        \item \textcolor{red}{source}: \texttt{TEXT}: \{text\} \texttt{QUERY}: Summarize the preceding text with one of the following options: \{choices\_without\_or\}.\\
        \textcolor{blue}{target}: \{title\}
        
        \item \textcolor{red}{source}: \texttt{TEXT}: \{text\} \texttt{QUERY}: Which of the following could be the title of the preceding article? \{choices\_with\_or\}?\\
        \textcolor{blue}{target}: \{title\}
    \end{enumerate}
    \item Generation format prompts
    \begin{enumerate}
        \item \textcolor{red}{source}: \texttt{TEXT}: \{text\} \texttt{QUERY}: Can you summarize this text in around \{length\} words?\\
        \textcolor{blue}{target}: \{title\}
        
        \item \textcolor{red}{source}: \texttt{TEXT}: \{text\} \texttt{QUERY}: In around \{length\} words, summarize the article.\\
        \textcolor{blue}{target}: \{title\}
        
        \item \textcolor{red}{source}: \texttt{TEXT}: \{text\} \texttt{QUERY}: In around \{length\} words, summarize the core of the above text.\\
        \textcolor{blue}{target}: \{title\}
        
        \item \textcolor{red}{source}: \texttt{TEXT}: \{text\} \texttt{QUERY}: In around \{length\} words, write a TLDR (Too Long Didn''t Read) summary for the above text.\\
        \textcolor{blue}{target}: \{title\}
        
        \item \textcolor{red}{source}: \texttt{TEXT}: \{text\} \texttt{QUERY}: Write a headline for the previous text.\\
        \textcolor{blue}{target}: \{title\}
        
        \item \textcolor{red}{source}: \texttt{TEXT}: \{text\} \texttt{QUERY}: What could be the title of the previous text?\\
        \textcolor{blue}{target}: \{title\}
        
        \item \textcolor{red}{source}: \texttt{TEXT}: \{text\} \texttt{QUERY}: What can be an appropriate headline for the text?\\
        \textcolor{blue}{target}: \{title\}
        
        \item \textcolor{red}{source}: \texttt{TEXT}: \{text\} \texttt{QUERY}: Summarize the core of the above text in around \{length\} words.\\
        \textcolor{blue}{target}: \{title\}
        
        \item \textcolor{red}{source}: \texttt{TEXT}: \{text\} \texttt{QUERY}: Generate a title for this article.\\
        \textcolor{blue}{target}: \{title\}
        
        \item \textcolor{red}{source}: \texttt{TEXT}: \{text\} \texttt{QUERY}: Can ``\{title\}" summarize the previous text?\\
        \textcolor{blue}{target}: Yes
        
        \item \textcolor{red}{source}: \texttt{TEXT}: \{text\} \texttt{QUERY}: Can ``\{other\_titles\}" summarize the previous text?\\
        \textcolor{blue}{target}: No
        
        \item \textcolor{red}{source}: \texttt{TEXT}: \{text\} \texttt{QUERY}: Does ``\{title\}" summarize the core of the above text?\\
        \textcolor{blue}{target}: Yes
        
        \item \textcolor{red}{source}: \texttt{TEXT}: \{text\} \texttt{QUERY}: Does ``\{other\_title\}" summarize the core of the above text?\\
        \textcolor{blue}{target}: No
        
        \item \textcolor{red}{source}: \texttt{TEXT}: \{text\} \texttt{QUERY}: In around \{length\} words, what does the previous article say?\\
        \textcolor{blue}{target}: \{title\}
        
        \item \textcolor{red}{source}: \texttt{TEXT}: \{text\} \texttt{QUERY}: In around \{length\} words, briefly describe what the previous article talks about.\\
        \textcolor{blue}{target}: \{title\}
        
        \item \textcolor{red}{source}: \texttt{TEXT}: \{text\} \texttt{QUERY}: What can be a short description of the text?\\
        \textcolor{blue}{target}: \{title\}
        
        \item \textcolor{red}{source}: \texttt{TEXT}: \{text\} \texttt{QUERY}: Please write a summary of about \{length\} words based on the previous article.\\
        \textcolor{blue}{target}: \{title\}
        
        \item \textcolor{red}{source}: \texttt{TEXT}: \{text\} \texttt{QUERY}: Condense the text down to the essentials.\\
        \textcolor{blue}{target}: \{title\}
        
        \item \textcolor{red}{source}: \texttt{TEXT}: \{text\} \texttt{QUERY}: How would you summarize the text?\\
        \textcolor{blue}{target}: \{title\}
        
        \item \textcolor{red}{source}: \texttt{TEXT}: {text} \texttt{QUERY}: Can you express the main idea of the text?\\
        \textcolor{blue}{target}: \{title\}
    \end{enumerate}
\end{itemize}

\paragraph{Sentence Expansion}
We use generation format prompts for sentence expansion signals. We use \{length\} as a clue in some prompts to guide the model to generate appropriate content. For $($summary, text$)$ pairs, we use the following prompts:
\begin{itemize}
    \item Generation format prompts
    \begin{enumerate}
        \item \textcolor{red}{source}: \texttt{TEXT}: \{summary\} \texttt{QUERY}: What details would you include in a storyline to make it more engaging and informative?\\
        \textcolor{blue}{target}: \{text\}
        
        \item \textcolor{red}{source}: \texttt{TEXT}: \{summary\} \texttt{QUERY}: Write a news article based on this outline.\\
        \textcolor{blue}{target}: \{text\}
        
        \item \textcolor{red}{source}: \texttt{TEXT}: \{summary\} \texttt{QUERY}: Please expand the previous text to a more informative news article.\\
        \textcolor{blue}{target}: \{text\}
        
        \item \textcolor{red}{source}: \texttt{TEXT}: \{summary\} \texttt{QUERY}: Can you add some details to the previous text?\\
        \textcolor{blue}{target}: \{text\}
        
        \item \textcolor{red}{source}: \texttt{TEXT}: \{summary\} \texttt{QUERY}: Can you write an article that uses the previous text as a summary?\\
        \textcolor{blue}{target}: \{text\}
        
        \item \textcolor{red}{source}: \texttt{TEXT}: \{summary\} \texttt{QUERY}: Write an article of around \{length\} words based on the previous key points.\\
        \textcolor{blue}{target}: \{text\}
        
        \item \textcolor{red}{source}: \texttt{TEXT}: \{summary\} \texttt{QUERY}: Write a \{length\}-word article with the previous text as the main content.\\
        \textcolor{blue}{target}: \{text\}
        
        \item \textcolor{red}{source}: \texttt{TEXT}: \{summary\} \texttt{QUERY}: Given the previous outline, write an article.\\
        \textcolor{blue}{target}: \{text\}
        
        \item \textcolor{red}{source}: \texttt{TEXT}: \{summary\} \texttt{QUERY}: Can you expand the previous outline into a passage of about \{length\} words?\\
        \textcolor{blue}{target}: \{text\}
        
        \item \textcolor{red}{source}: \texttt{TEXT}: \{summary\} \texttt{QUERY}: Please write a text of approximately \{length\} words using the previous text as an outline for your story.\\
        \textcolor{blue}{target}: \{text\}
        
        \item \textcolor{red}{source}: \texttt{TEXT}: \{summary\} \texttt{QUERY}: Expand the previous outline into a story of around \{length\} words.\\
        \textcolor{blue}{target}: \{text\}
        
        \item \textcolor{red}{source}: \texttt{TEXT}: \{summary\} \texttt{QUERY}: Given the previous text as an outline, write a \{length\}-word article.\\
        \textcolor{blue}{target}: \{text\}
        
        \item \textcolor{red}{source}: \texttt{TEXT}: \{summary\} \texttt{QUERY}: Please use about \{length\} words to enrich the preceding text into a story.\\
        \textcolor{blue}{target}: \{text\}
        
        \item \textcolor{red}{source}: \texttt{TEXT}: \{summary\} \texttt{QUERY}: Enrich the previous text with more details to make it a news article of about \{length\} words.\\
        \textcolor{blue}{target}: \{text\}
        
        \item \textcolor{red}{source}: \texttt{TEXT}: \{summary\} \texttt{QUERY}: Given the previous outline, generate a text of around \{length\} words.\\
        \textcolor{blue}{target}: \{text\}
        
        \item \textcolor{red}{source}: \texttt{TEXT}: \{summary\} \texttt{QUERY}: Write a passage of about \{length\} words with the previous text as the core content.\\
        \textcolor{blue}{target}: \{text\}
        
        \item \textcolor{red}{source}: \texttt{TEXT}: \{summary\} \texttt{QUERY}: Write a story of about \{length\} words using the given key points as the core content.\\
        \textcolor{blue}{target}: \{text\}
        
        \item \textcolor{red}{source}: \texttt{TEXT}: \{summary\} \texttt{QUERY}: Please expand the given outline into a detailed and informative story.\\
        \textcolor{blue}{target}: \{text\}
        
        \item \textcolor{red}{source}: \texttt{TEXT}: \{summary\} \texttt{QUERY}: Write a detail-rich story with the previous text as the core content.\\
        \textcolor{blue}{target}: \{text\}
        
        \item \textcolor{red}{source}: \texttt{TEXT}: \{summary\} \texttt{QUERY}: Based on the previous overview, write a long article.\\
        \textcolor{blue}{target}: \{text\}
    \end{enumerate}
\end{itemize}

For $($title, summary$)$ pairs, we use the following prompts:
\begin{itemize}
    \item Generation format prompts
    \begin{enumerate}
        \item \textcolor{red}{source}: \texttt{TEXT}: \{title\} \texttt{QUERY}: Based on the given title, can you develop a story outline?\\
        \textcolor{blue}{target}: \{summary\}
        
        \item \textcolor{red}{source}: \texttt{TEXT}: \{title\} \texttt{QUERY}: Based on the given title, can you outline a story?\\
        \textcolor{blue}{target}: \{summary\}
        
        \item \textcolor{red}{source}: \texttt{TEXT}: \{title\} \texttt{QUERY}: Expand the given text into a story line.\\
        \textcolor{blue}{target}: \{summary\}
        
        \item \textcolor{red}{source}: \texttt{TEXT}: \{title\} \texttt{QUERY}: Can you write something relevant around this title?\\
        \textcolor{blue}{target}: \{summary\}
        
        \item \textcolor{red}{source}: \texttt{TEXT}: \{title\} \texttt{QUERY}: Using the given text as a headline, please write a \{length\}-word outline.\\
        \textcolor{blue}{target}: \{summary\}
        
        \item \textcolor{red}{source}: \texttt{TEXT}: \{title\} \texttt{QUERY}: Write a paragraph of about \{length\} words with the previous text as the core content.\\
        \textcolor{blue}{target}: \{summary\}
        
        \item \textcolor{red}{source}: \texttt{TEXT}: \{title\} \texttt{QUERY}: Please use about \{length\} words to enrich the preceding text into a story.\\
        \textcolor{blue}{target}: \{summary\}
        
        \item \textcolor{red}{source}: \texttt{TEXT}: \{title\} \texttt{QUERY}: Can you write an article that uses the previous text as a title?\\
        \textcolor{blue}{target}: \{summary\}
        
        \item \textcolor{red}{source}: \texttt{TEXT}: \{title\} \texttt{QUERY}: Write a paragraph of around \{length\} words based on the previous text as the title.\\
        \textcolor{blue}{target}: \{summary\}
        
        \item \textcolor{red}{source}: \texttt{TEXT}: \{title\} \texttt{QUERY}: Given the previous headline, write a piece of text.\\
        \textcolor{blue}{target}: \{summary\}
        
        \item \textcolor{red}{source}: \texttt{TEXT}: \{title\} \texttt{QUERY}: Can you expand the previous title into a paragraph of about \{length\} words?\\
        \textcolor{blue}{target}: \{summary\}
        
        \item \textcolor{red}{source}: \texttt{TEXT}: \{title\} \texttt{QUERY}: Please write a text of approximately \{length\} words using the previous text as the title.\\
        \textcolor{blue}{target}: \{summary\}
        
        \item \textcolor{red}{source}: \texttt{TEXT}: \{title\} \texttt{QUERY}: Expand the previous text into a storyline of around \{length\} words.\\
        \textcolor{blue}{target}: \{summary\}
        
        \item \textcolor{red}{source}: \texttt{TEXT}: \{title\} \texttt{QUERY}: Given the previous text as a title, write a \{length\}-word paragraph.\\
        \textcolor{blue}{target}: \{summary\}
        
        \item \textcolor{red}{source}: \texttt{TEXT}: \{title\} \texttt{QUERY}: Based on the previous heading, generate a storyline.\\
        \textcolor{blue}{target}: \{summary\}
        
        \item \textcolor{red}{source}: \texttt{TEXT}: \{title\} \texttt{QUERY}: Can you add some details to the previous text to make it an outline?\\
        \textcolor{blue}{target}: \{summary\}
        
        \item \textcolor{red}{source}: \texttt{TEXT}: \{title\} \texttt{QUERY}: Enrich the previous text with more details to make it about \{length\} words.\\
        \textcolor{blue}{target}: \{summary\}
        
        \item \textcolor{red}{source}: \texttt{TEXT}: \{title\} \texttt{QUERY}: Given the previous headline, generate a text of around \{length\} words.\\
        \textcolor{blue}{target}: \{summary\}
        
        \item \textcolor{red}{source}: \texttt{TEXT}: \{title\} \texttt{QUERY}: Please expand the given headline into a short story.\\
        \textcolor{blue}{target}: \{summary\}
        
        \item \textcolor{red}{source}: \texttt{TEXT}: \{title\} \texttt{QUERY}: Write a story with the previous text as the core content.\\
        \textcolor{blue}{target}: \{summary\}
    \end{enumerate}
\end{itemize}

For $($title, text$)$ pairs, we use the following prompts:
\begin{itemize}
    \item Generation format prompts
    \begin{enumerate}
        \item \textcolor{red}{source}: \texttt{TEXT}: \{title\} \texttt{QUERY}: Given the above title, can you write a news article?\\
        \textcolor{blue}{target}: \{text\}
        
        \item \textcolor{red}{source}: \texttt{TEXT}: \{title\} \texttt{QUERY}: Write a news article of about \{length\} words based on the given text as the title.\\
        \textcolor{blue}{target}: \{text\}
        
        \item \textcolor{red}{source}: \texttt{TEXT}: \{title\} \texttt{QUERY}: Using the given text as a headline, can you write a \{length\}-word article?\\
        \textcolor{blue}{target}: \{text\}
        
        \item \textcolor{red}{source}: \texttt{TEXT}: \{title\} \texttt{QUERY}: Can you develop a detailed story of around \{length\} words based on the given title?\\
        \textcolor{blue}{target}: \{text\}
        
        \item \textcolor{red}{source}: \texttt{TEXT}: \{title\} \texttt{QUERY}: Given the previous title, can you write a story?\\
        \textcolor{blue}{target}: \{text\}
        
        \item \textcolor{red}{source}: \texttt{TEXT}: \{title\} \texttt{QUERY}: Can you expand the above headline into a news article of around \{length\} words?\\
        \textcolor{blue}{target}: \{text\}
        
        \item \textcolor{red}{source}: \texttt{TEXT}: \{title\} \texttt{QUERY}: Please use about \{length\} words to enrich the preceding text into a story.\\
        \textcolor{blue}{target}: \{text\}
        
        \item \textcolor{red}{source}: \texttt{TEXT}: \{title\} \texttt{QUERY}: Can you write an article that uses the previous text as a title?\\
        \textcolor{blue}{target}: \{text\}
        
        \item \textcolor{red}{source}: \texttt{TEXT}: \{title\} \texttt{QUERY}: Write an article of around \{length\} words based on the previous text as the title.\\
        \textcolor{blue}{target}: \{text\}
        
        \item \textcolor{red}{source}: \texttt{TEXT}: \{title\} \texttt{QUERY}: Please write a text of approximately \{length\} words using the previous text as the title.\\
        \textcolor{blue}{target}: \{text\}
        
        \item \textcolor{red}{source}: \texttt{TEXT}: \{title\} \texttt{QUERY}: Expand the previous text into an article of around \{length\} words.\\
        \textcolor{blue}{target}: \{text\}
        
        \item \textcolor{red}{source}: \texttt{TEXT}: \{title\} \texttt{QUERY}: Given the previous text as a title, write a \{length\}-word article.\\
        \textcolor{blue}{target}: \{text\}
        
        \item \textcolor{red}{source}: \texttt{TEXT}: \{title\} \texttt{QUERY}: Based on the previous heading, generate a story.\\
        \textcolor{blue}{target}: \{text\}
        
        \item \textcolor{red}{source}: \texttt{TEXT}: \{title\} \texttt{QUERY}: Given the previous headline, generate a text of around \{length\} words.\\
        \textcolor{blue}{target}: \{text\}
        
        \item \textcolor{red}{source}: \texttt{TEXT}: \{title\} \texttt{QUERY}: Please expand the given headline into a long story.\\
        \textcolor{blue}{target}: \{text\}
        
        \item \textcolor{red}{source}: \texttt{TEXT}: \{title\} \texttt{QUERY}: Write a story with the previous text as the core content.\\
        \textcolor{blue}{target}: \{text\}
        
        \item \textcolor{red}{source}: \texttt{TEXT}: \{title\} \texttt{QUERY}: Write a news article based on this title.\\
        \textcolor{blue}{target}: \{text\}
        
        \item \textcolor{red}{source}: \texttt{TEXT}: \{title\} \texttt{QUERY}: Given the previous title, write a detailed and informative article.\\
        \textcolor{blue}{target}: \{text\}
        
        \item \textcolor{red}{source}: \texttt{TEXT}: \{title\} \texttt{QUERY}: Given the previous title, write a news story related to it.\\
        \textcolor{blue}{target}: \{text\}
        
        \item \textcolor{red}{source}: \texttt{TEXT}: \{title\} \texttt{QUERY}: Can you add some details to the previous text to make it more detailed?\\
        \textcolor{blue}{target}: \{text\}
    \end{enumerate}
\end{itemize}

\paragraph{Temporal Information}
With $($text, one\_bullet\_point, another\_bullet\_point$)$ triples, we can construct prompts that ask for information about the temporal order between two events. We use \{first\_bp\} to represent the bullet point within the two bullet points that occurred first, and \{last\_bp\} to represent the bullet point that occurred later.

With $($text, first\_bullet\_point, second bullet point$)$ triples,  we can design prompts that require inference about the immediate subsequent event given a prior event. Here, we use \{choices\_with\_or\} (e.g. ``option1", ``option2", or ``option3") and \{choices\_without\_or\} (e.g. ``option1", ``option2", ``option3") to refer to all the bullets within the summary of the article.

\begin{itemize}
    \item Multiple-choice format prompts
    \begin{enumerate}
        \item \textcolor{red}{source}: \texttt{TEXT}: \{text\} \texttt{QUERY}: According to the order of events in the article, which of the following events occurred first? ``\{one\_bullet\_point\}" or ``\{another\_bullet\_point\}"?\\
        \textcolor{blue}{target}: \{first\_bp\}
        
        \item \textcolor{red}{source}: \texttt{TEXT}: \{text\} \texttt{QUERY}: Which of the following events should come first in the summary of the previous article? ``\{one\_bullet\_point\}" or ``\{another\_bullet\_point\}"?\\
        \textcolor{blue}{target}: \{first\_bp\}
        
        \item \textcolor{red}{source}: \texttt{TEXT}: \{text\} \texttt{QUERY}: In summarizing the previous text, which of the following events should be placed first? ``\{one\_bullet\_point\}" or ``\{another\_bullet\_point\}"?\\
        \textcolor{blue}{target}: \{first\_bp\}
        
        \item \textcolor{red}{source}: \texttt{TEXT}: \{text\} \texttt{QUERY}: According to the previous text, which of the following events occurred later? ``\{one\_bullet\_point\}" or ``\{another\_bullet\_point\}"?\\
        \textcolor{blue}{target}: \{last\_bp\}
        
        \item \textcolor{red}{source}: \texttt{TEXT}:\{text\} \texttt{QUERY}: Based on the text, ``\{first\_bp\}" happened before ``\{last\_bp\}". ``True" or ``False"?\\
        \textcolor{blue}{target}: True
        
        \item \textcolor{red}{source}: \texttt{TEXT}: \{text\} \texttt{QUERY}: Based on the text, ``\{last\_bp\}" happened before ``\{first\_bp\}". ``True" or ``False"?\\
        \textcolor{blue}{target}: False
        
        \item \textcolor{red}{source}: \texttt{TEXT}: \{text\} \texttt{QUERY}: Given the two events: ``\{one\_bullet\_point\}" and ``\{another\_bullet\_point\}", which one happened first?\\
        \textcolor{blue}{target}: \{first\_bp\}
        
        \item \textcolor{red}{source}: \texttt{TEXT}: \{text\} \texttt{QUERY}: Which of the following key points would you like to put first if you were to summarize the whole text? ``\{one\_bullet\_point\}" or ``\{another\_bullet\_point\}"?\\
        \textcolor{blue}{target}: \{first\_bp\}
        
        \item \textcolor{red}{source}: \texttt{TEXT}: \{text\} \texttt{QUERY}: When summarizing this text, would you mention ``\{first\_bp\}" before ``\{last\_bp\}"? ``Yes" or ``No"?\\
        \textcolor{blue}{target}: Yes
        
        \item \textcolor{red}{source}: \texttt{TEXT}: \{text\} \texttt{QUERY}: When summarizing this text, would you mention ``\{last\_bp\}" before ``\{first\_bp\}"? ``Yes" or ``No"?\\
        \textcolor{blue}{target}: No
        
        \item \textcolor{red}{source}: 
        \texttt{TEXT}: \{text\} \texttt{QUERY}: In the article, which of the following is mentioned first? ``\{one\_bullet\_point\}" or ``\{another\_bullet\_point\}"?\\
        \textcolor{blue}{target}: \{first\_bp\}
        
        \item \textcolor{red}{source}: \texttt{TEXT}: \{text\} \texttt{QUERY}: According to the timeline of the article, which of the following events occurred first? ``\{one\_bullet\_point\}" or ``\{another\_bullet\_point\}"?\\
        \textcolor{blue}{target}: \{first\_bp\}
        
        \item \textcolor{red}{source}: \texttt{TEXT}: \{text\} \texttt{QUERY}: Which of the following two sentences should come first in the summary of the article? ``\{one\_bullet\_point\}" or ``\{another\_bullet\_point\}"?\\
        \textcolor{blue}{target}: \{first\_bp\}
        
        \item \textcolor{red}{source}: \texttt{TEXT}: \{text\} \texttt{QUERY}: Given two events: ``\{one\_bullet\_point\}" and ``\{another\_bullet\_point\}", which one happened first according to the timeline of the article?\\
        \textcolor{blue}{target}: \{first\_bp\}
        
        \item \textcolor{red}{source}: \texttt{TEXT}: \{text\} \texttt{QUERY}: In the article, which event is mentioned first? ``\{one\_bullet\_point\}" or ``\{another\_bullet\_point\}"?\\
        \textcolor{blue}{target}: \{first\_bp\}
        
        \item \textcolor{red}{source}: \texttt{TEXT}: \{text\} \texttt{QUERY}: According to the timeline of the article, pick the one that happened first: ``\{one\_bullet\_point\}", ``\{another\_bullet\_point\}".\\
        \textcolor{blue}{target}: \{first\_bp\}
        
        \item \textcolor{red}{source}: \texttt{TEXT}: \{text\} \texttt{QUERY}: To summarize this article, what should you write first, ``\{one\_bullet\_point\}" or ``\{another\_bullet\_point\}" in a temporal order?\\
        \textcolor{blue}{target}: \{first\_bp\}
        
        \item \textcolor{red}{source}: \texttt{TEXT}: \{text\} \texttt{QUERY}: To summarize this text, which one should be mentioned first? ``\{one\_bullet\_point\}" or ``\{another\_bullet\_point\}"?\\
        \textcolor{blue}{target}: \{first\_bp\}
        
        \item \textcolor{red}{source}: \texttt{TEXT}: \{text\} \texttt{QUERY}: In the article, which event should come right after ``\{first\_bullet\_point\}"? \{choices\_with\_or\}?\\
        \textcolor{blue}{target}: \{second\_bullet\_point\}
        
        \item \textcolor{red}{source}: \texttt{TEXT}: \{text\} \texttt{QUERY}: When summarizing the main points of this article, what would you write after writing ``\{first\_bullet\_point\}"? \{choices\_with\_or\}?\\
        \textcolor{blue}{target}: \{second\_bullet\_point\}
        
        \item \textcolor{red}{source}: \texttt{TEXT}: \{text\} \texttt{QUERY}: Given the text and its first key point ``\{first\_bullet\_point\}", what would be a second key point? \{choices\_with\_or\}?\\
        \textcolor{blue}{target}: \{second\_bullet\_point\}
        
        \item \textcolor{red}{source}: \texttt{TEXT}: \{text\} \texttt{QUERY}: Given the events: \{choices\_without\_or\}, which one came right after ``\{first\_bullet\_point\}" according to the text?\\
        \textcolor{blue}{target}: \{second\_bullet\_point\}
        
        \item \textcolor{red}{source}: \texttt{TEXT}: \{text\} \texttt{QUERY}: Given the events: \{choices\_without\_or\}, which one happened right after ``\{first\_bullet\_point\}"?\\
        \textcolor{blue}{target}: \{second\_bullet\_point\}
        
        \item \textcolor{red}{source}: \texttt{TEXT}: \{text\} \texttt{QUERY}: According to the timeline of the article, choose the one that should follow ``\{first\_bullet\_point\}": \{choices\_without\_or\}.\\
        \textcolor{blue}{target}: \{second\_bullet\_point\}
        
        \item \textcolor{red}{source}: \texttt{TEXT}: \{text\} \texttt{QUERY}: To summarize this text, what should be mentioned right after ``\{first\_bullet\_point\}"? \{choices\_with\_or\}?\\
        \textcolor{blue}{target}: \{second\_bullet\_point\}
        
        \item \textcolor{red}{source}: \texttt{TEXT}: \{text\} \texttt{QUERY}: Given the events: \{choices\_without\_or\}, choose the one that follows ``\{first\_bullet\_point\}" according to the previous text.\\
        \textcolor{blue}{target}: \{second\_bullet\_point\}
        
        \item \textcolor{red}{source}: \texttt{TEXT}: \{text\} \texttt{QUERY}: To write a summary for the text, what will you write after you write ``\{first\_bullet\_point\}"? The options are \{choices\_without\_or\}.\\
        \textcolor{blue}{target}: \{second\_bullet\_point\}
        
        \item \textcolor{red}{source}: \texttt{TEXT}: \{text\} \texttt{QUERY}: One of the sentences in the summary of the text is ``\{first\_bullet\_point\}", what could be the next sentence? \{choices\_with\_or\}?\\
        \textcolor{blue}{target}: \{second\_bullet\_point\}
        
        \item \textcolor{red}{source}: \texttt{TEXT}: \{text\} \texttt{QUERY}: In the article, what is mentioned right after ``\{first\_bullet\_point\}"? \{choices\_with\_or\}?\\
        \textcolor{blue}{target}: \{second\_bullet\_point\}
        
    \end{enumerate}
    \item Generation format prompts
    \begin{enumerate}
        \item \textcolor{red}{source}: \texttt{TEXT}: \{text\} \texttt{QUERY}: In the article, what event should come right after ``\{first\_bullet\_point\}"?\\
        \textcolor{blue}{target}: \{second\_bullet\_point\}
        
        \item \textcolor{red}{source}: \texttt{TEXT}: \{text\} \texttt{QUERY}: When summarizing the main points of this article, what would you write after writing ``\{first\_bullet\_point\}"?\\
        \textcolor{blue}{target}: \{second\_bullet\_point\}
        
        \item \textcolor{red}{source}: \texttt{TEXT}: \{text\} \texttt{QUERY}: Given the text and its first key point ``\{first\_bullet\_point\}", what would be a second key point?\\
        \textcolor{blue}{target}: \{second\_bullet\_point\}
        
        \item \textcolor{red}{source}: \texttt{TEXT}: \{text\} \texttt{QUERY}: What event came right after ``\{first\_bullet\_point\}" according to the text?\\
        \textcolor{blue}{target}: \{second\_bullet\_point\}
        
        \item \textcolor{red}{source}: \texttt{TEXT}: \{text\} \texttt{QUERY}: Based on the text, what happened right after ``\{first\_bullet\_point\}"?\\
        \textcolor{blue}{target}: \{second\_bullet\_point\}
        
        \item \textcolor{red}{source}: \texttt{TEXT}: \{text\} \texttt{QUERY}: According to the timeline of the article, write the event that should follow ``\{first\_bullet\_point\}".\\
        \textcolor{blue}{target}: \{second\_bullet\_point\}
        
        \item \textcolor{red}{source}: \texttt{TEXT}: \{text\} \texttt{QUERY}: To summarize this text, what should be mentioned right after ``\{first\_bullet\_point\}"?\\
        \textcolor{blue}{target}: \{second\_bullet\_point\}
        
        \item \textcolor{red}{source}: \texttt{TEXT}: \{text\} \texttt{QUERY}: Write the key point that follows ``\{first\_bullet\_point\}" according to the previous text.\\
        \textcolor{blue}{target}: \{second\_bullet\_point\}
        
        \item \textcolor{red}{source}: \texttt{TEXT}: \{text\} \texttt{QUERY}: To write a summary for the text, what will you write after you write ``\{first\_bullet\_point\}"?\\
        \textcolor{blue}{target}: \{second\_bullet\_point\}
        
        \item \textcolor{red}{source}: \texttt{TEXT}: \{text\} \texttt{QUERY}: One of the sentences in the summary of the text is ``\{first\_bullet\_point\}", what could be the next sentence?\\
        \textcolor{blue}{target}: \{second\_bullet\_point\}
        
        \item \textcolor{red}{source}: \texttt{TEXT}: \{text\} \texttt{QUERY}: In the article, what is mentioned right after ``\{first\_bullet\_point\}"?\\
        \textcolor{blue}{target}: \{second\_bullet\_point\}
    \end{enumerate}
\end{itemize}

\subsection{Wikidata}
\paragraph{Relation}
With $($text, subject, property, object$)$ quads, we construct prompts that ask for the property between a subject and an object. Here, we use \{objects\_with\_or\} (e.g. ``option1", ``option2", or ``option3") and \{objects\_without\_or\} (e.g. ``option1", ``option2", ``option3") to refer to some (3 to 9) available objects (include the correct one). We use \{other\_object\} to refer to one random object that is different from \{object\}. We use similar notations: \{subjects\_with\_or\}, \{subjects\_without\_or\}, \{other\_subject\}, \{properties\_with\_or\}, \{properties\_without\_or\} and \{other\_property\}.

\begin{itemize}
    \item Multiple-choice format prompts
    \begin{enumerate}
        \item \textcolor{red}{source}: \texttt{TEXT}: \{text\} \texttt{QUERY}: Given the subject ``\{subject\}" and relation ``\{property\}", what's the correct object? \{objects\_with\_or\}?\\
        \textcolor{blue}{target}: \{object\}
        
        \item \textcolor{red}{source}: \texttt{TEXT}: \{text\} \texttt{QUERY}: ``\{subject\}" can form a ``\{property\}" relationship with which of the following entities? \{objects\_with\_or\}?\\
        \textcolor{blue}{target}: \{object\}
        
        \item \textcolor{red}{source}: \texttt{TEXT}: \{text\} \texttt{QUERY}: Here is a list of entities: \{objects\_without\_or\}. With which of these can ``\{subject\}" form a ``\{property\}" relationship?\\
        \textcolor{blue}{target}: \{object\}
        
        \item \textcolor{red}{source}: \texttt{TEXT}: \{text\} \texttt{QUERY}: Is ``\{subject\}" and ``\{object\}" in a ``\{property\}" relationship? ``Yes" or ``No"?\\
        \textcolor{blue}{target}: Yes
        
        \item \textcolor{red}{source}: \texttt{TEXT}: \{text\} \texttt{QUERY}: Is ``\{subject\}" and ``\{other\_object\}" in a ``\{property\}" relationship? ``Yes" or ``No"?\\
        \textcolor{blue}{target}: No
        
        \item \textcolor{red}{source}: \texttt{TEXT}: \{text\} \texttt{QUERY}: Given the subject ``\{subject\}" and relation ``\{property\}", can you select the correct object from the following: \{objects\_without\_or\}?\\
        \textcolor{blue}{target}: \{object\}
        
        \item \textcolor{red}{source}: \texttt{TEXT}: \{text\} \texttt{QUERY}: Given the subject ``\{subject\}" and relation ``\{property\}", pick the correct object from the following: \{objects\_without\_or\}.\\
        \textcolor{blue}{target}: \{object\}
        
        \item \textcolor{red}{source}: \texttt{TEXT}: \{text\} \texttt{QUERY}: With which of the following can ``\{subject\}" form a ``\{property\}" relation? \{objects\_with\_or\}?\\
        \textcolor{blue}{target}: \{object\}
        
        \item \textcolor{red}{source}: \texttt{TEXT}: \{text\} \texttt{QUERY}: Given the relation ``\{property\}" and subject ``\{subject\}", can you identify an object for this relation from \{objects\_without\_or\}?\\
        \textcolor{blue}{target}: \{object\}
        
        \item \textcolor{red}{source}: \texttt{TEXT}: \{text\} \texttt{QUERY}: ``\{subject\}" and ``\{object\}" can form a ``\{property\}" relation. ``True" or ``False"?\\
        \textcolor{blue}{target}: True
        
        \item \textcolor{red}{source}: \texttt{TEXT}: \{text\} \texttt{QUERY}: ``\{subject\}" and ``\{other\_object\}" can form a ``\{property\}" relation. ``True" or ``False"?\\
        \textcolor{blue}{target}: False

        \item \textcolor{red}{source}: \texttt{TEXT}: \{text\} \texttt{QUERY}: Given the object ``\{object\}" and relation ``\{property\}", what's the correct subject? \{subjects\_with\_or\}?\\
        \textcolor{blue}{target}: \{subject\}
        
        \item \textcolor{red}{source}: \texttt{TEXT}: \{text\} \texttt{QUERY}: Which of the following entities could constitute a ``\{property\}" relationship with ``\{object\}"? \{subjects\_with\_or\}?\\
        \textcolor{blue}{target}: \{subject\}
        
        \item \textcolor{red}{source}: \texttt{TEXT}: \{text\} \texttt{QUERY}: Here is a list of entities: \{subjects\_without\_or\}. Which one can form a ``\{property\}" relation with ``\{object\}"?\\
        \textcolor{blue}{target}: \{subject\}
        
        \item \textcolor{red}{source}: \texttt{TEXT}: \{text\} \texttt{QUERY}: Given the object ``\{object\}" and relation ``\{property\}", choose the appropriate subject from \{subjects\_without\_or\}.\\
        \textcolor{blue}{target}: \{subject\}
        
        \item \textcolor{red}{source}: \texttt{TEXT}: \{text\} \texttt{QUERY}: Which subject can form a ``\{property\}" relation with ``\{object\}"? The options are \{subjects\_without\_or\}.\\
        \textcolor{blue}{target}: \{subject\}
        
        \item \textcolor{red}{source}: \texttt{TEXT}: \{text\} \texttt{QUERY}: Which of the following entities does ``\{subject\}" have a ``\{property\}" relationship with? \{objects\_with\_or\}?\\
        \textcolor{blue}{target}: \{object\}
        
        \item \textcolor{red}{source}: \texttt{TEXT}: \{text\} \texttt{QUERY}: Given the entities: \{objects\_without\_or\}, which one is ``\{subject\}" in a ``\{property\}" relationship with?\\
        \textcolor{blue}{target}: \{object\}
        
        \item \textcolor{red}{source}: \texttt{TEXT}: \{text\} \texttt{QUERY}: Choose the subject that has a ``\{property\}" relation with the object ``\{object\}". You may choose from \{subjects\_without\_or\}.\\
        \textcolor{blue}{target}: \{subject\}
        
        \item \textcolor{red}{source}: \texttt{TEXT}: \{text\} \texttt{QUERY}: Choose the object with which ``\{subject\}" has a ``\{property\}" relation. The options are \{objects\_without\_or\}.\\
        \textcolor{blue}{target}: \{object\}
        
        \item \textcolor{red}{source}: \texttt{TEXT}: \{text\} \texttt{QUERY}: Given the subject ``\{subject\}" and object ``\{object\}", what's the relation between them? \{properties\_with\_or\}?\\
        \textcolor{blue}{target}: \{property\}
        
        \item \textcolor{red}{source}: \texttt{TEXT}: \{text\} \texttt{QUERY}: What's the relationship between the subject ``\{subject\}" and object ``\{object\}"? \{properties\_with\_or\}?\\
        \textcolor{blue}{target}: \{property\}
        
        \item \textcolor{red}{source}: \texttt{TEXT}: \{text\} \texttt{QUERY}: Here is a list of relations: \{properties\_without\_or\}. Which one applies to ``\{subject\}" and ``\{object\}"?\\
        \textcolor{blue}{target}: \{property\}
        
        \item \textcolor{red}{source}: \texttt{TEXT}: \{text\} \texttt{QUERY}: Given the subject ``\{subject\}" and object ``\{object\}", can you select the correct relationship from the following: \{properties\_without\_or\}?\\
        \textcolor{blue}{target}: \{property\}
        
        \item \textcolor{red}{source}: \texttt{TEXT}: \{text\} \texttt{QUERY}: Which of the following is the correct relation between ``\{subject\}" and ``\{object\}"? \{properties\_with\_or\}?\\
        \textcolor{blue}{target}: \{property\}
        
        \item \textcolor{red}{source}: \texttt{TEXT}: \{text\} \texttt{QUERY}: Given the subject ``\{subject\}" and object ``\{object\}", pick the appropriate relation between them. You may choose from \{properties\_without\_or\}.\\
        \textcolor{blue}{target}: \{property\}
        
        \item \textcolor{red}{source}: \texttt{TEXT}: \{text\} \texttt{QUERY}: What relationship can be formed between ``\{subject\}" and ``\{object\}"? \{properties\_with\_or\}?\\
        \textcolor{blue}{target}: \{property\}
        
        \item \textcolor{red}{source}: \texttt{TEXT}: \{text\} \texttt{QUERY}: Which of the following relations can apply to ``\{subject\}" and ``\{object\}"? \{properties\_with\_or\}?\\
        \textcolor{blue}{target}: \{property\}
        
        \item \textcolor{red}{source}: \texttt{TEXT}: \{text\} \texttt{QUERY}: What's the relation between ``\{subject\}" and ``\{object\}"? The options are \{properties\_without\_or\}.\\
        \textcolor{blue}{target}: \{property\}
        
        \item \textcolor{red}{source}: \texttt{TEXT}: \{text\} \texttt{QUERY}: Given a list of relations: \{properties\_without\_or\}, which of them is the relationship between ``\{subject\}" and ``\{object\}"?\\
        \textcolor{blue}{target}: \{property\}
        
        \item \textcolor{red}{source}: \texttt{TEXT}: \{text\} \texttt{QUERY}: The relation between ``\{subject\}" and ``\{object\}" is ``\{property\}". ``True" or ``False"?\\
        \textcolor{blue}{target}: True
        
        \item \textcolor{red}{source}: \texttt{TEXT}: \{text\} \texttt{QUERY}: The relation between ``\{subject\}" and ``\{object\}" is ``\{other\_property\}". ``True" or ``False"?\\
        \textcolor{blue}{target}: False
        
        \item \textcolor{red}{source}: \texttt{TEXT}: \{text\} \texttt{QUERY}: Identify the correct relationship between ``\{subject\}" and ``\{object\}". Constrain your answer in \{properties\_without\_or\}.\\
        \textcolor{blue}{target}: \{property\}
        
        \item \textcolor{red}{source}: \texttt{TEXT}: \{text\} \texttt{QUERY}: Here are some relation types: \{properties\_without\_or\}. Which one of them can be used to describe the relationship between ``\{subject\}" and ``\{object\}"?\\
        \textcolor{blue}{target}: \{property\}
        
        \item \textcolor{red}{source}: \texttt{TEXT}: \{text\} \texttt{QUERY}: Is there a ``\{property\}" relation between ``\{subject\}" and ``\{object\}"? ``Yes" or ``No"?\\
        \textcolor{blue}{target}: Yes
        
        \item \textcolor{red}{source}: \texttt{TEXT}: \{text\} \texttt{QUERY}: Is there a ``\{other\_property\}" relation between ``\{subject\}" and ``\{object\}"? ``Yes" or ``No"?\\
        \textcolor{blue}{target}: No
        
        \item \textcolor{red}{source}: \texttt{TEXT}: \{text\} \texttt{QUERY}: Do ``\{subject\}" and ``\{object\}" make a ``\{property\}" relation? ``Yes" or ``No"?\\
        \textcolor{blue}{target}: Yes
        
        \item \textcolor{red}{source}: \texttt{TEXT}: \{text\} \texttt{QUERY}: Do ``\{subject\}" and ``\{object\}" make a ``\{other\_property\}" relation? ``Yes" or ``No"?\\
        \textcolor{blue}{target}: No
        
        \item \textcolor{red}{source}: \texttt{TEXT}: \{text\} \texttt{QUERY}: Given the subject ``\{subject\}" and object ``\{object\}", what's their relation? \{properties\_with\_or\}?\\
        \textcolor{blue}{target}: \{property\}
        
        \item \textcolor{red}{source}: \texttt{TEXT}: \{text\} \texttt{QUERY}: Judge the relationship between ``\{subject\}" and ``\{object\}". Please choose from \{properties\_without\_or\}.\\
        \textcolor{blue}{target}: \{property\}
    \end{enumerate}
    \item Generation format prompts
    \begin{enumerate}
        \item \textcolor{red}{source}: \texttt{TEXT}: \{text\} \texttt{QUERY}: Given the subject ``\{subject\}" and relation ``\{property\}", what's the correct object?\\
        \textcolor{blue}{target}: \{object\}
        
        \item \textcolor{red}{source}: \texttt{TEXT}: \{text\} \texttt{QUERY}: ``\{subject\}" can form a ``\{property\}" relationship with what entity?\\
        \textcolor{blue}{target}: \{object\}
        
        \item \textcolor{red}{source}: \texttt{TEXT}: \{text\} \texttt{QUERY}: With what entity can ``\{subject\}" form a ``\{property\}" relationship?\\
        \textcolor{blue}{target}: \{object\}
        
        \item \textcolor{red}{source}: \texttt{TEXT}: \{text\} \texttt{QUERY}: Is ``\{subject\}" and ``\{object\}" in a ``\{property\}" relationship?\\
        \textcolor{blue}{target}: Yes
        
        \item \textcolor{red}{source}: \texttt{TEXT}: \{text\} \texttt{QUERY}: Is ``\{subject\}" and ``\{other\_object\}" in a ``\{property\}" relationship?\\
        \textcolor{blue}{target}: No
        
        \item \textcolor{red}{source}: \texttt{TEXT}: \{text\} \texttt{QUERY}: Given the subject ``\{subject\}" and relation ``\{property\}", can you find an appropriate object?\\
        \textcolor{blue}{target}: \{object\}
        
        \item \textcolor{red}{source}: \texttt{TEXT}: \{text\} \texttt{QUERY}: What can be a suitable object given the subject ``\{subject\}" and relation ``\{property\}"?\\
        \textcolor{blue}{target}: \{object\}
        
        \item \textcolor{red}{source}: \texttt{TEXT}: \{text\} \texttt{QUERY}: Given the subject ``\{subject\}" and relation ``\{property\}", generate a suitable object.\\
        \textcolor{blue}{target}: \{object\}
        
        \item \textcolor{red}{source}: \texttt{TEXT}: \{text\} \texttt{QUERY}: Given the relation ``\{property\}" and subject ``\{subject\}", what would be the object?\\
        \textcolor{blue}{target}: \{object\}
        
        \item \textcolor{red}{source}: \texttt{TEXT}: \{text\} \texttt{QUERY}: Can ``\{subject\}" and ``\{object\}" form a ``\{property\}" relation?\\
        \textcolor{blue}{target}: Yes
        
        \item \textcolor{red}{source}: \texttt{TEXT}: \{text\} \texttt{QUERY}: Can ``\{subject\}" and ``\{other\_object\}" form a ``\{property\}" relation?\\
        \textcolor{blue}{target}: No
        
        \item \textcolor{red}{source}: \texttt{TEXT}: \{text\} \texttt{QUERY}: Given the object ``\{object\}" and relation ``\{property\}", what's the correct subject?\\
        \textcolor{blue}{target}: \{subject\}
        
        \item \textcolor{red}{source}: \texttt{TEXT}: \{text\} \texttt{QUERY}: What entity could constitute a ``\{property\}" relationship with ``\{object\}"?\\
        \textcolor{blue}{target}: \{subject\}
        
        \item \textcolor{red}{source}: \texttt{TEXT}: \{text\} \texttt{QUERY}: What entity can form a ``\{property\}" relationship with ``\{object\}"?\\
        \textcolor{blue}{target}: \{subject\}
        
        \item \textcolor{red}{source}: \texttt{TEXT}: \{text\} \texttt{QUERY}: Given the object ``\{object\}" and relation ``\{property\}", generate an appropriate subject.\\
        \textcolor{blue}{target}: \{subject\}
        
        \item \textcolor{red}{source}: \texttt{TEXT}: \{text\} \texttt{QUERY}: What subject has a ``\{property\}" relationship with ``\{object\}"?\\
        \textcolor{blue}{target}: \{subject\}
        
        \item \textcolor{red}{source}: \texttt{TEXT}: \{text\} \texttt{QUERY}: What entity does ``\{subject\}" have a ``\{property\}" relationship with?\\
        \textcolor{blue}{target}: \{object\}
        
        \item \textcolor{red}{source}: \texttt{TEXT}: \{text\} \texttt{QUERY}: What entity is ``\{subject\}" in a ``\{property\}" relation with?\\
        \textcolor{blue}{target}: \{object\}
        
        \item \textcolor{red}{source}: \texttt{TEXT}: \{text\} \texttt{QUERY}: Find a subject that has a ``\{property\}" relation with the object ``\{object\}".\\
        \textcolor{blue}{target}: \{subject\}
        
        \item \textcolor{red}{source}: \texttt{TEXT}: \{text\} \texttt{QUERY}: Find an object with which ``\{subject\}" has a ``\{property\}" relation.\\
        \textcolor{blue}{target}: \{object\}
        
        \item \textcolor{red}{source}: \texttt{TEXT}: \{text\} \texttt{QUERY}: Given the subject ``\{subject\}" and object ``\{object\}", what's the relation between them?\\
        \textcolor{blue}{target}: \{property\}
        
        \item \textcolor{red}{source}: \texttt{TEXT}: \{text\} \texttt{QUERY}: What's the relation between the subject ``\{subject\}" and object ``\{object\}"?\\
        \textcolor{blue}{target}: \{property\}
        
        \item \textcolor{red}{source}: \texttt{TEXT}: \{text\} \texttt{QUERY}: What relationship applies to the subject ``\{subject\}" and object ``\{object\}"?\\
        \textcolor{blue}{target}: \{property\}
        
        \item \textcolor{red}{source}: \texttt{TEXT}: \{text\} \texttt{QUERY}: Given the subject ``\{subject\}" and object ``\{object\}", can you determine the relationship between them?\\
        \textcolor{blue}{target}: \{property\}
        
        \item \textcolor{red}{source}: \texttt{TEXT}: \{text\} \texttt{QUERY}: What is the relation between ``\{subject\}" and ``\{object\}"?\\
        \textcolor{blue}{target}: \{property\}
        
        \item \textcolor{red}{source}: \texttt{TEXT}: \{text\} \texttt{QUERY}: Given the subject ``\{subject\}" and object ``\{object\}", find an appropriate relation between them.\\
        \textcolor{blue}{target}: \{property\}
        
        \item \textcolor{red}{source}: \texttt{TEXT}: \{text\} \texttt{QUERY}: What relationship can be formed between ``\{subject\}" and ``\{object\}"?\\
        \textcolor{blue}{target}: \{property\}
        
        \item \textcolor{red}{source}: \texttt{TEXT}: \{text\} \texttt{QUERY}: What kind of relationship can apply to ``\{subject\}" and ``\{object\}"?\\
        \textcolor{blue}{target}: \{property\}
        
        \item \textcolor{red}{source}: \texttt{TEXT}: \{text\} \texttt{QUERY}: Can you tell the relation between entity ``\{subject\}" and entity ``\{object\}"?\\
        \textcolor{blue}{target}: \{property\}
        
        \item \textcolor{red}{source}: \texttt{TEXT}: \{text\} \texttt{QUERY}: What's the relation between the following two entities: ``\{subject\}" and ``\{object\}"?\\
        \textcolor{blue}{target}: \{property\}
        
        \item \textcolor{red}{source}: \texttt{TEXT}: \{text\} \texttt{QUERY}: Can ``\{property\}" be used to summarize the relationship between ``\{subject\}" and ``\{object\}"?\\
        \textcolor{blue}{target}: Yes
        
        \item \textcolor{red}{source}: \texttt{TEXT}: \{text\} \texttt{QUERY}: Can ``\{other\_property\}" be used to summarize the relationship between ``\{subject\}" and ``\{object\}"?\\
        \textcolor{blue}{target}: No
        
        \item \textcolor{red}{source}: \texttt{TEXT}: \{text\} \texttt{QUERY}: Identify the relationship between ``\{subject\}" and ``\{object\}".\\
        \textcolor{blue}{target}: \{property\}
        
        \item \textcolor{red}{source}: \texttt{TEXT}: \{text\} \texttt{QUERY}: What can be used to describe the relationship between ``\{subject\}" and ``\{object\}"?\\
        \textcolor{blue}{target}: \{property\}
        
        \item \textcolor{red}{source}: \texttt{TEXT}: \{text\} \texttt{QUERY}: Is there a ``\{property\}" relation between ``\{subject\}" and ``\{object\}"?\\
        \textcolor{blue}{target}: Yes
        
        \item \textcolor{red}{source}: \texttt{TEXT}: \{text\} \texttt{QUERY}: Is there a ``\{other\_property\}" relation between ``\{subject\}" and ``\{object\}"?\\
        \textcolor{blue}{target}: No
        
        \item \textcolor{red}{source}: \texttt{TEXT}: \{text\} \texttt{QUERY}: Do ``\{subject\}" and ``\{object\}" make a ``\{property\}" relation?\\
        \textcolor{blue}{target}: Yes
        
        \item \textcolor{red}{source}: \texttt{TEXT}: \{text\} \texttt{QUERY}: Do ``\{subject\}" and ``\{object\}" make a ``\{other\_property\}" relation?\\
        \textcolor{blue}{target}: No
        
        \item \textcolor{red}{source}: \texttt{TEXT}: \{text\} \texttt{QUERY}: Given the subject ``\{subject\}" and object ``\{object\}", what's their relation?\\
        \textcolor{blue}{target}: \{property\}
        
        \item \textcolor{red}{source}: \texttt{TEXT}: \{text\} \texttt{QUERY}: Judge the relationship between ``\{subject\}" and ``\{object\}".\\
        \textcolor{blue}{target}: \{property\}
    \end{enumerate}
    
\end{itemize}
\paragraph{Entity typing}
With $($text, entity, entity\_type$)$ triples, we design prompts that ask for the entity type of entity that appears in a piece of text. Here, we use \{entity\_types\_with\_or\} (e.g. ``option1", ``option2", or ``option3") and \{entity\_types\_without\_or\} (e.g. ``option1", ``option2", ``option3") to refer to some (3 to 9) available entity types (include the correct one). We use \{other\_entity\_type\} to represent a random entity type that is different from the true entity type of the entity.

\begin{itemize}
    \item Multiple-choice format prompts
    \begin{enumerate}
        \item \textcolor{red}{source}: \texttt{TEXT}: \{text\} \texttt{QUERY}: What's the entity type of ``\{entity\}"? \{entity\_types\_with\_or\}?\\
        \textcolor{blue}{target}: \{entity\_type\}
        
        \item \textcolor{red}{source}: \texttt{TEXT}: \{text\} \texttt{QUERY}: Can you choose an appropriate class from the following list for ``\{entity\}"? The options are \{entity\_types\_without\_or\}.\\
        \textcolor{blue}{target}: \{entity\_type\}
        
        \item \textcolor{red}{source}: \texttt{TEXT}: \{text\} \texttt{QUERY}: ``\{entity\}" is an instance of which of the following? \{entity\_types\_with\_or\}?\\
        \textcolor{blue}{target}: \{entity\_type\}
        
        \item \textcolor{red}{source}: \texttt{TEXT}: \{text\} \texttt{QUERY}: Given a list of categories: \{entity\_types\_without\_or\}, what category does ``\{entity\}" belong to?\\
        \textcolor{blue}{target}: \{entity\_type\}
        
        \item \textcolor{red}{source}: \texttt{TEXT}: \{text\} \texttt{QUERY}: Choose the correct entity type for ``\{entity\}" from the following: \{entity\_types\_without\_or\}.\\
        \textcolor{blue}{target}: \{entity\_type\}
        
        \item \textcolor{red}{source}: \texttt{TEXT}: \{text\} \texttt{QUERY}: Does ``\{entity\}" belong to ``\{entity\_type\}"? ``Yes" or ``No"?\\
        \textcolor{blue}{target}: Yes
        
        \item \textcolor{red}{source}: \texttt{TEXT}: \{text\} \texttt{QUERY}: Does ``\{entity\}" belong to ``\{other\_entity\_type\}"? ``Yes" or ``No"?\\
        \textcolor{blue}{target}: No
        
        \item \textcolor{red}{source}: \texttt{TEXT}: \{text\} \texttt{QUERY}: In the previous text, what type of entity is ``\{entity\}"? \{entity\_types\_with\_or\}?\\
        \textcolor{blue}{target}: \{entity\_type\}
        
        \item \textcolor{red}{source}: \texttt{TEXT}: \{text\} \texttt{QUERY}: Which of the following is the entity type of ``\{entity\}"? \{entity\_types\_with\_or\}?\\
        \textcolor{blue}{target}: \{entity\_type\}
        
        \item \textcolor{red}{source}: \texttt{TEXT}: \{text\} \texttt{QUERY}: In the text, ``\{entity\}" belongs to ``\{entity\_type\}". ``True" or ``False"?\\
        \textcolor{blue}{target}: True
        
        \item \textcolor{red}{source}: \texttt{TEXT}: \{text\} \texttt{QUERY}: In the text, ``\{entity\}" belongs to ``\{other\_entity\_type\}". ``True" or ``False"?\\
        \textcolor{blue}{target}: False
        
        \item \textcolor{red}{source}: \texttt{TEXT}: \{text\} \texttt{QUERY}: Pick one category for ``\{entity\}". The options are \{entity\_types\_without\_or\}.\\
        \textcolor{blue}{target}: \{entity\_type\}
        
        \item \textcolor{red}{source}: \texttt{TEXT}: \{text\} \texttt{QUERY}: What's the entity category of ``\{entity\}"? \{entity\_types\_with\_or\}?\\
        \textcolor{blue}{target}: \{entity\_type\}
        
        \item \textcolor{red}{source}: \texttt{TEXT}: \{text\} \texttt{QUERY}: What category best describes ``\{entity\}"? \{entity\_types\_with\_or\}?\\
        \textcolor{blue}{target}: \{entity\_type\}
        
        \item \textcolor{red}{source}: \texttt{TEXT}: \{text\} \texttt{QUERY}: What type of entity is ``\{entity\}"? \{entity\_types\_with\_or\}?\\
        \textcolor{blue}{target}: \{entity\_type\}
        
        \item \textcolor{red}{source}: \texttt{TEXT}: \{text\} \texttt{QUERY}: Can you identify the category of ``\{entity\}"? \{entity\_types\_with\_or\}?\\
        \textcolor{blue}{target}: \{entity\_type\}
        
        \item \textcolor{red}{source}: \texttt{TEXT}: \{text\} \texttt{QUERY}: Given a list of entity types: \{entity\_types\_without\_or\}, which one best describes ``\{entity\}"?\\
        \textcolor{blue}{target}: \{entity\_type\}
        
        \item \textcolor{red}{source}: \texttt{TEXT}: \{text\} \texttt{QUERY}: Here is a list of entity types: \{entity\_types\_without\_or\}. Which one does ``\{entity\}" belong to?\\
        \textcolor{blue}{target}: \{entity\_type\}
        
        \item \textcolor{red}{source}: \texttt{TEXT}: \{text\} \texttt{QUERY}: The entity type of ``\{entity\}" is ``\{entity\_type\}". ``True" or ``False"?\\
        \textcolor{blue}{target}: True
        
        \item \textcolor{red}{source}: \texttt{TEXT}: \{text\} \texttt{QUERY}: The entity type of ``\{entity\}" is ``\{other\_entity\_type\}". ``True" or ``False"?\\
        \textcolor{blue}{target}: False
    \end{enumerate}
    \item Generation format prompts
    \begin{enumerate}
        \item \textcolor{red}{source}: \texttt{TEXT}: \{text\} \texttt{QUERY}: What's the entity type of ``\{entity\}"?\\
        \textcolor{blue}{target}: \{entity\_type\}
        
        \item \textcolor{red}{source}: \texttt{TEXT}: \{text\} \texttt{QUERY}: Can you find an appropriate class for ``\{entity\}"?\\
        \textcolor{blue}{target}: \{entity\_type\}
        
        \item \textcolor{red}{source}: \texttt{TEXT}: \{text\} \texttt{QUERY}: ``\{entity\}" is an instance of what entity type?\\
        \textcolor{blue}{target}: \{entity\_type\}
        
        \item \textcolor{red}{source}: \texttt{TEXT}: \{text\} \texttt{QUERY}: What category does ``\{entity\}" belong to?\\
        \textcolor{blue}{target}: \{entity\_type\}
        
        \item \textcolor{red}{source}: \texttt{TEXT}: \{text\} \texttt{QUERY}: Assign a correct entity type for ``\{entity\}".\\
        \textcolor{blue}{target}: \{entity\_type\}
        
        \item \textcolor{red}{source}: \texttt{TEXT}: \{text\} \texttt{QUERY}: Does ``\{entity\}" belong to ``\{entity\_type\}"?\\
        \textcolor{blue}{target}: Yes
        
        \item \textcolor{red}{source}: \texttt{TEXT}: \{text\} \texttt{QUERY}: Does ``\{entity\}" belong to ``\{other\_entity\_type\}"?\\
        \textcolor{blue}{target}: No
        
        \item \textcolor{red}{source}: \texttt{TEXT}: \{text\} \texttt{QUERY}: In the previous text, what type of entity is ``\{entity\}"?\\
        \textcolor{blue}{target}: \{entity\_type\}
        
        \item \textcolor{red}{source}: \texttt{TEXT}: \{text\} \texttt{QUERY}: ``\{entity\}" is an instance of what entity type?\\
        \textcolor{blue}{target}: \{entity\_type\}
        
        \item \textcolor{red}{source}: \texttt{TEXT}: \{text\} \texttt{QUERY}: In the text, is ``\{entity\}" an instance of ``\{entity\_type\}"?\\
        \textcolor{blue}{target}: Yes
        
        \item \textcolor{red}{source}: \texttt{TEXT}: \{text\} \texttt{QUERY}: In the text, is ``\{entity\}" an instance of ``\{other\_entity\_type\}"?\\
        \textcolor{blue}{target}: No
        
        \item \textcolor{red}{source}: \texttt{TEXT}: \{text\} \texttt{QUERY}: Pick one category for ``\{entity\}".\\
        \textcolor{blue}{target}: \{entity\_type\}
        
        \item \textcolor{red}{source}: \texttt{TEXT}: \{text\} \texttt{QUERY}: What's the entity category of ``\{entity\}"?\\
        \textcolor{blue}{target}: \{entity\_type\}
        
        \item \textcolor{red}{source}: \texttt{TEXT}: \{text\} \texttt{QUERY}: What category best describes ``\{entity\}"?\\
        \textcolor{blue}{target}: \{entity\_type\}
        
        \item \textcolor{red}{source}: \texttt{TEXT}: \{text\} \texttt{QUERY}: What type of entity is ``\{entity\}"?\\
        \textcolor{blue}{target}: \{entity\_type\}
        
        \item \textcolor{red}{source}: \texttt{TEXT}: \{text\} \texttt{QUERY}: Can you identify the category of ``\{entity\}"?\\
        \textcolor{blue}{target}: \{entity\_type\}
        
        \item \textcolor{red}{source}: \texttt{TEXT}: \{text\} \texttt{QUERY}: What entity type best describes ``\{entity\}"?\\
        \textcolor{blue}{target}: \{entity\_type\}
        
        \item \textcolor{red}{source}: \texttt{TEXT}: \{text\} \texttt{QUERY}: What entity type does ``\{entity\}" belong to?\\
        \textcolor{blue}{target}: \{entity\_type\}
        
        \item \textcolor{red}{source}: \texttt{TEXT}: \{text\} \texttt{QUERY}: Is the entity type of ``\{entity\}" ``\{entity\_type\}"?\\
        \textcolor{blue}{target}: Yes
        
        \item \textcolor{red}{source}: \texttt{TEXT}: \{text\} \texttt{QUERY}: Is the entity type of ``\{entity\}" ``\{other\_entity\_type\}"?\\
        \textcolor{blue}{target}: No
    \end{enumerate}
\end{itemize}

\subsection{wikiHow}
\paragraph{Text category}
With $($title\_description, category$)$ pairs, we construct prompts that ask for the category of a given piece of text. Here, we use \{choices\_with\_or\} (e.g. ``option1", ``option2", or ``option3") and \{choices\_without\_or\} (e.g. ``option1", ``option2", ``option3") to refer to some (3 to 9) available options. We use \{other\_category\} to refer to a random wrong category for the text we aim to classify.
\begin{itemize}
    \item Multiple-choice format prompts
    \begin{enumerate}
        \item \textcolor{red}{source}: \texttt{TEXT}: \{title\_description\} \texttt{QUERY}: What's this text about? \{choices\_with\_or\}?\\
        \textcolor{blue}{target}: \{category\}
        
        \item \textcolor{red}{source}: \texttt{TEXT}: \{title\_description\} \texttt{QUERY}: Classify this text. You may choose from \{choices\_without\_or\}.\\
        \textcolor{blue}{target}: \{category\}
        
        \item \textcolor{red}{source}: \texttt{TEXT}: \{title\_description\} \texttt{QUERY}: How would you categorize this text? \{choices\_with\_or\}?\\
        \textcolor{blue}{target}: \{category\}
        
        \item \textcolor{red}{source}: \texttt{TEXT}: \{title\_description\} \texttt{QUERY}: Is this text about ``\{other\_category\}"? ``Yes" or ``No"?\\
        \textcolor{blue}{target}: No
        
        \item \textcolor{red}{source}: \texttt{TEXT}: \{title\_description\} \texttt{QUERY}: Is this text about ``\{category\}"? ``Yes" or ``No"? \\
        \textcolor{blue}{target}:Yes
        
        \item \textcolor{red}{source}: \texttt{TEXT}: \{title\_description\} \texttt{QUERY}: Can you choose an appropriate class from the following list for this text? \{choices\_without\_or\}.\\
        \textcolor{blue}{target}: \{category\}
        
        \item \textcolor{red}{source}: \texttt{TEXT}: \{title\_description\} \texttt{QUERY}: Given a list of categories: \{choices\_without\_or\}, what category does the paragraph belong to?\\
        \textcolor{blue}{target}: \{category\}
        
        \item \textcolor{red}{source}: \texttt{TEXT}: \{title\_description\} \texttt{QUERY}: Is this paragraph related to ``\{other\_category\}"? ``Yes" or ``No"?\\
        \textcolor{blue}{target}: No
        
        \item \textcolor{red}{source}: \texttt{TEXT}: \{title\_description\} \texttt{QUERY}: Is this paragraph related to ``\{category\}"? ``Yes" or ``No"?\\
        \textcolor{blue}{target}: Yes
        
        \item \textcolor{red}{source}: \texttt{TEXT}: \{title\_description\} \texttt{QUERY}: Pick one category for the previous text. The options are \{choices\_without\_or\}.\\
        \textcolor{blue}{target}: \{category\}
        
        \item \textcolor{red}{source}: \texttt{TEXT}: \{title\_description\} \texttt{QUERY}: Given a choice of categories: \{choices\_with\_or\}, the text refers to which one?\\
        \textcolor{blue}{target}: \{category\}
        
        \item \textcolor{red}{source}: \texttt{TEXT}: \{title\_description\} \texttt{QUERY}: What is the topic of the text? \{choices\_with\_or\}?\\
        \textcolor{blue}{target}: \{category\}
        
        \item \textcolor{red}{source}: \texttt{TEXT}: \{title\_description\} \texttt{QUERY}: The topic of this text is ``\{other\_category\}". ``True" or ``False"?\\
        \textcolor{blue}{target}: False
        
        \item \textcolor{red}{source}: \texttt{TEXT}: \{title\_description\} \texttt{QUERY}: The topic of this text is ``\{category\}". ``True" or ``False"?\\
        \textcolor{blue}{target}: True
        
        \item \textcolor{red}{source}: \texttt{TEXT}: \{title\_description\} \texttt{QUERY}: Can you identify the category of this text? \{choices\_with\_or\}?\\
        \textcolor{blue}{target}: \{category\}
        
        \item \textcolor{red}{source}: \texttt{TEXT}: \{title\_description\} \texttt{QUERY}: Select a class from the following that best describes the text: \{choices\_without\_or\}.\\
        \textcolor{blue}{target}: \{category\}
        
        \item \textcolor{red}{source}: \texttt{TEXT}: \{title\_description\} \texttt{QUERY}: Is this a piece of text regarding \{choices\_with\_or\}?\\
        \textcolor{blue}{target}: \{category\}
        
        \item \textcolor{red}{source}: \texttt{TEXT}: \{title\_description\} \texttt{QUERY}: What category best describes this paragraph? \{choices\_with\_or\}?\\
        \textcolor{blue}{target}: \{category\}
        
        \item \textcolor{red}{source}: \texttt{TEXT}: \{title\_description\} \texttt{QUERY}: Please classify this text into one of the following: \{choices\_without\_or\}.\\
        \textcolor{blue}{target}: \{category\}
        
        \item \textcolor{red}{source}: \texttt{TEXT}: \{title\_description\} \texttt{QUERY}: What's the main topic of this paragraph? \{choices\_with\_or\}?\\
        \textcolor{blue}{target}: \{category\}
    \end{enumerate}
    \item Generation format prompts
    \begin{enumerate}
        \item \textcolor{red}{source}: \texttt{TEXT}: \{title\_description\} \texttt{QUERY}: What's this text about?\\
        \textcolor{blue}{target}: \{category\}
        
        \item \textcolor{red}{source}: \texttt{TEXT}: \{title\_description\} \texttt{QUERY}: Classify this text.\\
        \textcolor{blue}{target}: \{category\}
        
        \item \textcolor{red}{source}: \texttt{TEXT}: \{title\_description\} \texttt{QUERY}: How would you categorize this text?\\
        \textcolor{blue}{target}: \{category\}
        
        \item \textcolor{red}{source}: \texttt{TEXT}: \{title\_description\} \texttt{QUERY}: Is this text about ``\{other\_category\}"?\\
        \textcolor{blue}{target}: No
        
        \item \textcolor{red}{source}: \texttt{TEXT}: \{title\_description\} \texttt{QUERY}: Is this text about ``\{category\}"?\\
        \textcolor{blue}{target}: Yes
        
        \item \textcolor{red}{source}: \texttt{TEXT}: \{title\_description\} \texttt{QUERY}: Can you find an appropriate class for this text?\\
        \textcolor{blue}{target}: \{category\}
        
        \item \textcolor{red}{source}: \texttt{TEXT}: \{title\_description\} \texttt{QUERY}: What category does the paragraph belong to?\\
        \textcolor{blue}{target}: \{category\}
        
        \item \textcolor{red}{source}: \texttt{TEXT}: \{title\_description\} \texttt{QUERY}: Is this paragraph related to ``\{other\_category\}"?\\
        \textcolor{blue}{target}: No
        
        \item \textcolor{red}{source}: \texttt{TEXT}: \{title\_description\} \texttt{QUERY}: Is this paragraph related to ``\{category\}"?\\
        \textcolor{blue}{target}: Yes
        
        \item \textcolor{red}{source}: \texttt{TEXT}: \{title\_description\} \texttt{QUERY}: Pick one category for the previous text.\\
        \textcolor{blue}{target}: \{category\}
        
        \item \textcolor{red}{source}: \texttt{TEXT}: \{title\_description\} \texttt{QUERY}: The text refers to which category?\\
        \textcolor{blue}{target}: \{category\}
        
        \item \textcolor{red}{source}: \texttt{TEXT}: \{title\_description\} \texttt{QUERY}: What is the topic of the text?\\
        \textcolor{blue}{target}: \{category\}
        
        \item \textcolor{red}{source}: \texttt{TEXT}: \{title\_description\} \texttt{QUERY}: Is the topic of this text ``\{other\_category\}"?\\
        \textcolor{blue}{target}: No
        
        \item \textcolor{red}{source}: \texttt{TEXT}: \{title\_description\} \texttt{QUERY}: Is the topic of this text ``\{category\}"?\\
        \textcolor{blue}{target}: Yes
        
        \item \textcolor{red}{source}: \texttt{TEXT}: \{title\_description\} \texttt{QUERY}: Can you identify the category of this text?\\
        \textcolor{blue}{target}: \{category\}
        
        \item \textcolor{red}{source}: \texttt{TEXT}: \{title\_description\} \texttt{QUERY}: Think of a class that best fits the text.\\
        \textcolor{blue}{target}: \{category\}
        
        \item \textcolor{red}{source}: \texttt{TEXT}: \{title\_description\} \texttt{QUERY}: Can you classify this piece of text?\\
        \textcolor{blue}{target}: \{category\}
        
        \item \textcolor{red}{source}: \texttt{TEXT}: \{title\_description\} \texttt{QUERY}: What category best describes this paragraph?\\
        \textcolor{blue}{target}: \{category\}
        
        \item \textcolor{red}{source}: \texttt{TEXT}: \{title\_description\} \texttt{QUERY}: Please classify this text.\\
        \textcolor{blue}{target}: \{category\}
        
        \item \textcolor{red}{source}: \texttt{TEXT}: \{title\_description\} \texttt{QUERY}: What's the main topic of this paragraph?\\
        \textcolor{blue}{target}: \{category\}
    \end{enumerate}
\end{itemize}

\paragraph{Category Hierarchy} With $($low\_category, high\_category$)$ pairs, we construct prompts that ask for the hierarchical relationships between categories. Here, we use \{high\_categories\_with\_or\} (e.g. ``option1", ``option2", or ``option3") and \{high\_categories\_without\_or\} (e.g. ``option1", ``option2", ``option3") to refer to some (3 to 9) available higher-level categories (contain the appropriate one). We use \{other\_high\_category\} to refer to a random wrong high-level category for the low-level category.

\begin{itemize}
    \item Multiple-choice format prompts
    \begin{enumerate}
        \item \textcolor{red}{source}: \texttt{QUERY}: Which of the following contains ``\{low\_category\}"? \{high\_categories\_with\_or\}?\\
        \textcolor{blue}{target}: \{high\_category\}
        
        \item \textcolor{red}{source}: \texttt{QUERY}: Which of the following phrases is most relevant to ``\{low\_category\}"? \{high\_categories\_with\_or\}?\\
        \textcolor{blue}{target}: \{high\_category\}
        
        \item \textcolor{red}{source}: \texttt{QUERY}: ``\{low\_category\}" is a subclass of which of the following? \{high\_categories\_with\_or\}?\\
        \textcolor{blue}{target}: \{high\_category\}
        
        \item \textcolor{red}{source}: \texttt{QUERY}: Given a list of categories: \{high\_categories\_without\_or\}. Which one is the superclass of ``\{low\_category\}"?\\
        \textcolor{blue}{target}: \{high\_category\}
        
        \item \textcolor{red}{source}: \texttt{QUERY}: If a piece of text belongs to ``\{low\_category\}", it also belongs to which one of the following? \{high\_categories\_with\_or\}?\\
        \textcolor{blue}{target}: \{high\_category\}
        
        \item \textcolor{red}{source}: \texttt{QUERY}: Which one of the following can be a superclass of ``\{low\_category\}"? \{high\_categories\_with\_or\}?\\
        \textcolor{blue}{target}: \{high\_category\}
        
        \item \textcolor{red}{source}: \texttt{QUERY}: Given the concept ``\{low\_category\}", can you identify a similar concept? You may constrain your answers to \{high\_categories\_without\_or\}.\\
        \textcolor{blue}{target}: \{high\_category\}
        
        \item \textcolor{red}{source}: \texttt{QUERY}: Here is a list of categories: \{high\_categories\_without\_or\}. Which one is the most similar to ``\{low\_category\}"?\\
        \textcolor{blue}{target}: \{high\_category\}
        
        \item \textcolor{red}{source}: \texttt{QUERY}: Which one of the following is the superclass of ``\{low\_category\}"? \{high\_categories\_with\_or\}?\\
        \textcolor{blue}{target}: \{high\_category\}
        
        \item \textcolor{red}{source}: \texttt{QUERY}: Can you identify a superclass of ``\{low\_category\}". You may choose from \{high\_categories\_without\_or\}.\\
        \textcolor{blue}{target}: \{high\_category\}
        
        \item \textcolor{red}{source}: \texttt{QUERY}: Given a list of concepts: \{high\_categories\_without\_or\}. Which one is the closest to ``\{low\_category\}"?\\
        \textcolor{blue}{target}: \{high\_category\}
        
        \item \textcolor{red}{source}: \texttt{QUERY}: Given categories: \{high\_categories\_without\_or\}. ``\{low\_category\}" is the subclass of which one of the above?\\
        \textcolor{blue}{target}: \{high\_category\}
        
        \item \textcolor{red}{source}: \texttt{QUERY}: Please choose one category from the following that is most related to ``\{low\_category\}": \{high\_categories\_without\_or\}.\\
        \textcolor{blue}{target}: \{high\_category\}
        
        \item \textcolor{red}{source}: \texttt{QUERY}: Can you identify a similar concept for ``\{low\_category\}" from the following: \{high\_categories\_without\_or\}?\\
        \textcolor{blue}{target}: \{high\_category\}
        
        \item \textcolor{red}{source}: \texttt{QUERY}: Given concepts: \{high\_categories\_without\_or\}. You may choose one that includes ``\{low\_category\}".\\
        \textcolor{blue}{target}: \{high\_category\}
        
        \item \textcolor{red}{source}: \texttt{QUERY}: Given concepts: \{high\_categories\_without\_or\}, choose one that has superclass-subclass relation with ``\{low\_category\}".\\
        \textcolor{blue}{target}: \{high\_category\}
        
        \item \textcolor{red}{source}: \texttt{QUERY}: ``\{low\_category\}" is most relevant to which one of the following? \{high\_categories\_with\_or\}?\\
        \textcolor{blue}{target}: \{high\_category\}
        
        \item \textcolor{red}{source}: \texttt{QUERY}: Which one of the following phrases is most similar to ``\{low\_category\}"? \{high\_categories\_with\_or\}?\\
        \textcolor{blue}{target}: \{high\_category\}
        
        \item \textcolor{red}{source}: \texttt{QUERY}: ``\{low\_category\}" is a subclass of ``\{other\_high\_category\}"? ``True" or ``False"?\\
        \textcolor{blue}{target}: False
        
        \item \textcolor{red}{source}: \texttt{QUERY}: ``\{low\_category\}" is a subclass of ``\{high\_category\}"? ``True" or ``False"?\\
        \textcolor{blue}{target}: True
    \end{enumerate}
    \item Generation format prompts
    \begin{enumerate}
        \item \textcolor{red}{source}: \texttt{QUERY}: Can you find a superclass of ``\{low\_category\}"?\\
        \textcolor{blue}{target}: \{high\_category\}
        
        \item \textcolor{red}{source}: \texttt{QUERY}: What can be a superclass of ``\{low\_category\}"?\\
        \textcolor{blue}{target}: \{high\_category\}
        
        \item \textcolor{red}{source}: \texttt{QUERY}: ``\{low\_category\}" can be a subclass of which category?\\
        \textcolor{blue}{target}: \{high\_category\}
        
        \item \textcolor{red}{source}: \texttt{QUERY}: Can you identify a category that includes ``\{low\_category\}"?\\
        \textcolor{blue}{target}: \{high\_category\}
        
        \item \textcolor{red}{source}: \texttt{QUERY}: Find a concept that has superclass-subclass relation with ``\{low\_category\}"?\\
        \textcolor{blue}{target}: \{high\_category\}
        
        \item \textcolor{red}{source}: \texttt{QUERY}: Is ``\{low\_category\}" a subclass of ``\{high\_category\}"?\\
        \textcolor{blue}{target}: Yes
        
        \item \textcolor{red}{source}: \texttt{QUERY}: Is ``\{low\_category\}" a subclass of ``\{other\_high\_category\}"?\\
        \textcolor{blue}{target}: No
        
        \item \textcolor{red}{source}: \texttt{QUERY}: Is ``\{other\_high\_category\}" a superclass of ``\{low\_category\}"?\\
        \textcolor{blue}{target}: No
        
        \item \textcolor{red}{source}: \texttt{QUERY}: Is ``\{high\_category\}" a superclass of ``\{low\_category\}"?\\
        \textcolor{blue}{target}: Yes
        
        \item \textcolor{red}{source}: \texttt{QUERY}: ``\{low\_category\}" is typically included by which category?\\
        \textcolor{blue}{target}: \{high\_category\}
        
        \item \textcolor{red}{source}: \texttt{QUERY}: What can be a subclass of ``\{high\_category\}"?\\
        \textcolor{blue}{target}: \{low\_category\}
        
        \item \textcolor{red}{source}: \texttt{QUERY}: Can ``\{low\_category\}" and ``\{high\_category\}" form a subclass-superclass relation?\\
        \textcolor{blue}{target}: Yes
        
        \item \textcolor{red}{source}: \texttt{QUERY}: Can ``\{low\_category\}" and ``\{other\_high\_category\}" form a subclass-superclass relation?\\
        \textcolor{blue}{target}: No
        
        \item \textcolor{red}{source}: \texttt{QUERY}: Does ``\{high\_category\}" and ``\{low\_category\}" form a superclass-subclass relation?\\
        \textcolor{blue}{target}: Yes
        
        \item \textcolor{red}{source}: \texttt{QUERY}: Does ``\{other\_high\_category\}" and ``\{low\_category\}" form a superclass-subclass relation?\\
        \textcolor{blue}{target}: No
        
        \item \textcolor{red}{source}: \texttt{QUERY}: Think of a topic that is a subclass of a ``\{high\_category\}" topic.\\
        \textcolor{blue}{target}: \{low\_category\}
        
        \item \textcolor{red}{source}: \texttt{QUERY}: Please write a sub-category of a ``\{high\_category\}" topic.\\
        \textcolor{blue}{target}: \{low\_category\}
        
        \item \textcolor{red}{source}: \texttt{QUERY}: Please write a category of topics included in the ``\{high\_category\}" category.\\
        \textcolor{blue}{target}: \{low\_category\}
        
        \item \textcolor{red}{source}: \texttt{QUERY}: Write a more fine-grained sub-category within the ``\{high\_category\}" topic.\\
        \textcolor{blue}{target}: \{low\_category\}
        
        \item \textcolor{red}{source}: \texttt{QUERY}: Write a coarse-grained category that includes a ``\{low\_category\}" category.\\
        \textcolor{blue}{target}: \{high\_category\}
    \end{enumerate}
\end{itemize}
\paragraph{Goal-step relation}
With $($goal, step\_headline$)$ pairs, we construct prompts to ask for the steps needed to complete a goal. Here, we use \{step\_headlines\_with\_or\} (e.g. ``option1", ``option2", or ``option3") and \{step\_headlines\_without\_or\} (e.g. ``option1", ``option2", ``option3") to refer to some (3 to 9) steps (contain the appropriate one). For goals, we use similar notations: \{goals\_with\_or\} and \{goals\_without\_or\}.
\begin{itemize}
    \item Multiple-choice format prompts
    \begin{enumerate}
        \item \textcolor{red}{source}: \texttt{QUERY}: To accomplish the goal ``\{goal\}", what method would you take? \{step\_headlines\_with\_or\_\}?\\
        \textcolor{blue}{target}: \{step\_headline\}
        
        \item \textcolor{red}{source}: \texttt{QUERY}: What will you do to achieve the goal ``\{goal\}"? Choose one from the following: \{step\_headlines\_without\_or\}.\\
        \textcolor{blue}{target}: \{step\_headline\}
        
        \item \textcolor{red}{source}: \texttt{QUERY}: How to \{goal\}? \{step\_headlines\_with\_or\}?\\
        \textcolor{blue}{target}: \{step\_headline\}
        
        \item \textcolor{red}{source}: \texttt{QUERY}: In order to \{goal\}, what step would you take? \{step\_headlines\_with\_or\}?\\
        \textcolor{blue}{target}: \{step\_headline\}
        
        \item \textcolor{red}{source}: \texttt{QUERY}: Choose one of the following actions to \{goal\}: \{step\_headlines\_without\_or\}.\\
        \textcolor{blue}{target}: \{step\_headline\}
        
        \item \textcolor{red}{source}: \texttt{QUERY}: Which of the following methods are likely to \{goal\}? \{step\_headlines\_with\_or\}?\\
        \textcolor{blue}{target}: \{step\_headline\}
        
        \item \textcolor{red}{source}: \texttt{QUERY}: Given the goal ``\{goal\}", how will you achieve it? \{step\_headlines\_with\_or\}?\\
        \textcolor{blue}{target}: \{step\_headline\}
        
        \item \textcolor{red}{source}: \texttt{QUERY}: Your goal is to \{goal\}, identify the best method from the following list: \{step\_headlines\_without\_or\}.\\
        \textcolor{blue}{target}: \{step\_headline\}
        
        \item \textcolor{red}{source}: \texttt{QUERY}: Find the best approach to \{goal\} from these: \{step\_headlines\_without\_or\}.\\
        \textcolor{blue}{target}: \{step\_headline\}
        
        \item \textcolor{red}{source}: \texttt{QUERY}: To \{goal\}, which one of the following is the most reasonable way? \{step\_headlines\_with\_or\}?\\
        \textcolor{blue}{target}: \{step\_headline\}
        
        \item \textcolor{red}{source}: \texttt{QUERY}: Given that you want to \{goal\}, what will you do? \{step\_headlines\_with\_or\}?\\
        \textcolor{blue}{target}: \{step\_headline\}
        
        \item \textcolor{red}{source}: \texttt{QUERY}: How to \{goal\}? You may select from: \{step\_headlines\_without\_or\}.\\
        \textcolor{blue}{target}: \{step\_headline\}
        
        \item \textcolor{red}{source}: \texttt{QUERY}: You have a goal ``\{goal\}", to make it come true, which of the following you may try? \{step\_headlines\_with\_or\}?\\
        \textcolor{blue}{target}: \{step\_headline\}
        
        \item \textcolor{red}{source}: \texttt{QUERY}: Given the following actions: \{step\_headlines\_without\_or\}. Which one could serve for the goal ``\{goal\}"?\\
        \textcolor{blue}{target}: \{step\_headline\}
        
        \item \textcolor{red}{source}: \texttt{QUERY}: Here is an action list: \{step\_headlines\_without\_or\}. Can you choose the one that is probable to \{goal\}?\\
        \textcolor{blue}{target}: \{step\_headline\}
        
        \item \textcolor{red}{source}: \texttt{QUERY}: Find a way to \{goal\}, you can select from \{step\_headlines\_without\_or\}.\\
        \textcolor{blue}{target}: \{step\_headline\}
        
        \item \textcolor{red}{source}: \texttt{QUERY}: Given the goal ``\{goal\}", what could be a reasonable action to take? \{step\_headlines\_with\_or\}?\\
        \textcolor{blue}{target}: \{step\_headline\}
        
        \item \textcolor{red}{source}: \texttt{QUERY}: Your goal is to \{goal\}, to make it possible, choose one step you may take from the following: \{step\_headlines\_without\_or\}.\\
        \textcolor{blue}{target}: \{step\_headline\}
        
        \item \textcolor{red}{source}: \texttt{QUERY}: With the goal ``\{goal\}", what will you do next? \{step\_headlines\_with\_or\}?\\
        \textcolor{blue}{target}: \{step\_headline\}
        
        \item \textcolor{red}{source}: \texttt{QUERY}: How to \{goal\}? You may constrain your answer to one of the following: \{step\_headlines\_without\_or\}.\\
        \textcolor{blue}{target}: \{step\_headline\}

        \item \textcolor{red}{source}: \texttt{QUERY}: Given the action ``\{step\_headline\}", what could be the likely goal? \{goals\_with\_or\}?\\
        \textcolor{blue}{target}: \{goal\}
        
        \item \textcolor{red}{source}: \texttt{QUERY}: If your friend is planning to \{step\_headline\}, which one of the following is likely to be his intent? \{goals\_with\_or\}?\\
        \textcolor{blue}{target}: \{goal\}
        
        \item \textcolor{red}{source}: \texttt{QUERY}: If you \{step\_headline\}, the most probable result is? \{goals\_with\_or\}?\\
        \textcolor{blue}{target}: \{goal\}
        
        \item \textcolor{red}{source}: \texttt{QUERY}: You \{step\_headline\}, what could be the result? Choose from the following: \{goals\_without\_or\}.\\
        \textcolor{blue}{target}: \{goal\}
        
        \item \textcolor{red}{source}: \texttt{QUERY}: The method ``\{step\_headline\}" is typically associated with which intent? Select one of the following: \{goals\_without\_or\}.\\
        \textcolor{blue}{target}: \{goal\}
        
        \item \textcolor{red}{source}: \texttt{QUERY}: What is the usual purpose if you begin to \{step\_headline\}? \{goals\_with\_or\}?\\
        \textcolor{blue}{target}: \{goal\}
        
        \item \textcolor{red}{source}: \texttt{QUERY}: What would be the possible intent if someone begin to \{step\_headline\}? \{goals\_with\_or\}?\\
        \textcolor{blue}{target}: \{goal\}
        
        \item \textcolor{red}{source}: \texttt{QUERY}: What is the most likely outcome if you \{step\_headline\}? \{goals\_with\_or\}?\\
        \textcolor{blue}{target}: \{goal\}
        
        \item \textcolor{red}{source}: \texttt{QUERY}: What do you want to achieve if you \{step\_headline\}? \{goals\_with\_or\}?\\
        \textcolor{blue}{target}: \{goal\}
        
        \item \textcolor{red}{source}: \texttt{QUERY}: Given the step ``\{step\_headline\}", select the goal it serves from: \{goals\_without\_or\}.\\
        \textcolor{blue}{target}: \{goal\}
        
        \item \textcolor{red}{source}: \texttt{QUERY}: Find the most probable intent associated with the action ``\{step\_headline\}". You may constrain your answer to one of the following: \{goals\_without\_or\}.\\
        \textcolor{blue}{target}: \{goal\}
        
        \item \textcolor{red}{source}: \texttt{QUERY}: ``\{step\_headline\}" is one of the procedures to? \{goals\_with\_or\}?\\
        \textcolor{blue}{target}: \{goal\}
        
        \item \textcolor{red}{source}: \texttt{QUERY}: ``\{step\_headline\}" is a step of which of the following? \{goals\_with\_or\}?\\
        \textcolor{blue}{target}: \{goal\}
        
        \item \textcolor{red}{source}: \texttt{QUERY}: What is the ultimate goal if you \{step\_headline\}? Select one from: \{goals\_without\_or\}.\\
        \textcolor{blue}{target}: \{goal\}
        
        \item \textcolor{red}{source}: \texttt{QUERY}: What is the purpose to \{step\_headline\}? Choose the most appropriate one: \{goals\_without\_or\}.\\
        \textcolor{blue}{target}: \{goal\}
        
        \item \textcolor{red}{source}: \texttt{QUERY}: Here is a goal list: \{goals\_without\_or\}. Can you choose the right goal for the action ``\{step\_headline\}"?\\
        \textcolor{blue}{target}: \{goal\}
        
        \item \textcolor{red}{source}: \texttt{QUERY}: Which of the following goals will you \{step\_headline\} to achieve? \{goals\_with\_or\}?\\
        \textcolor{blue}{target}: \{goal\}
        
        \item \textcolor{red}{source}: \texttt{QUERY}: To accomplish which of the following purposes you will \{step\_headline\}? \{goals\_with\_or\}?\\
        \textcolor{blue}{target}: \{goal\}
        
        \item \textcolor{red}{source}: \texttt{QUERY}: Choose the most probable purpose if you \{step\_headline\}. The options are \{goals\_without\_or\}.\\
        \textcolor{blue}{target}: \{goal\}
        
        \item \textcolor{red}{source}: \texttt{QUERY}: Which one of the following is likely to be your intent if you \{step\_headline\}? \{goals\_with\_or\}?\\
        \textcolor{blue}{target}: \{goal\}
    \end{enumerate}
    \item Generation format prompts
    \begin{enumerate}
        \item \textcolor{red}{source}: \texttt{QUERY}: To accomplish the goal ``\{goal\}", what method would you take?\\
        \textcolor{blue}{target}: \{step\_headline\}
        
        \item \textcolor{red}{source}: \texttt{QUERY}: What will you do to achieve the goal ``\{goal\}"?\\
        \textcolor{blue}{target}: \{step\_headline\}
        
        \item \textcolor{red}{source}: \texttt{QUERY}: How to \{goal\}?\\
        \textcolor{blue}{target}: \{step\_headline\}
        
        \item \textcolor{red}{source}: \texttt{QUERY}: In order to \{goal\}, what step would you take?\\
        \textcolor{blue}{target}: \{step\_headline\}
        
        \item \textcolor{red}{source}: \texttt{QUERY}: Think of one action to \{goal\}.\\
        \textcolor{blue}{target}: \{step\_headline\}
        
        \item \textcolor{red}{source}: \texttt{QUERY}: What can you do to \{goal\}?\\
        \textcolor{blue}{target}: \{step\_headline\}
        
        \item \textcolor{red}{source}: \texttt{QUERY}: Given the goal ``\{goal\}", how will you achieve it?\\
        \textcolor{blue}{target}: \{step\_headline\}
        
        \item \textcolor{red}{source}: \texttt{QUERY}: Your goal is to \{goal\}, think of a way to achieve it.\\
        \textcolor{blue}{target}: \{step\_headline\}
        
        \item \textcolor{red}{source}: \texttt{QUERY}: Find an appropriate approach to \{goal\}.\\
        \textcolor{blue}{target}: \{step\_headline\}
        
        \item \textcolor{red}{source}: \texttt{QUERY}: To \{goal\}, what could be a reasonable way?\\
        \textcolor{blue}{target}: \{step\_headline\}
        
        \item \textcolor{red}{source}: \texttt{QUERY}: Given that you want to \{goal\}, what will you do?\\
        \textcolor{blue}{target}: \{step\_headline\}
        
        \item \textcolor{red}{source}: \texttt{QUERY}: Do you know how to \{goal\}?\\
        \textcolor{blue}{target}: \{step\_headline\}
        
        \item \textcolor{red}{source}: \texttt{QUERY}: You have a goal ``\{goal\}", to make it come true, which method will you try?\\
        \textcolor{blue}{target}: \{step\_headline\}
        
        \item \textcolor{red}{source}: \texttt{QUERY}: What behavior can serve for the goal ``\{goal\}"?\\
        \textcolor{blue}{target}: \{step\_headline\}
        
        \item \textcolor{red}{source}: \texttt{QUERY}: Can you tell me a method that is probable to \{goal\}?\\
        \textcolor{blue}{target}: \{step\_headline\}
        
        \item \textcolor{red}{source}: \texttt{QUERY}: Find a way to \{goal\}.\\
        \textcolor{blue}{target}: \{step\_headline\}
        
        \item \textcolor{red}{source}: \texttt{QUERY}: Given the goal ``\{goal\}", what could be a reasonable action to take?\\
        \textcolor{blue}{target}: \{step\_headline\}
        
        \item \textcolor{red}{source}: \texttt{QUERY}: Your goal is to \{goal\}, to make it possible, what do you plan to do?\\
        \textcolor{blue}{target}: \{step\_headline\}
        
        \item \textcolor{red}{source}: \texttt{QUERY}: With the goal ``\{goal\}", what will you do next?\\
        \textcolor{blue}{target}: \{step\_headline\}
        
        \item \textcolor{red}{source}: \texttt{QUERY}: What's your plan if you want to \{goal\}?\\
        \textcolor{blue}{target}: \{step\_headline\}
        
        \item \textcolor{red}{source}: \texttt{QUERY}: Given the action ``\{step\_headline\}", what could be the likely goal?\\
        \textcolor{blue}{target}: \{goal\}
        
        \item \textcolor{red}{source}: \texttt{QUERY}: If your friend is planning to \{step\_headline\}, what is likely to be his intent?\\
        \textcolor{blue}{target}: \{goal\}
        
        \item \textcolor{red}{source}: \texttt{QUERY}: If you \{step\_headline\}, the most probable result is?\\
        \textcolor{blue}{target}: \{goal\}
        
        \item \textcolor{red}{source}: \texttt{QUERY}: You \{step\_headline\}, what could be the result?\\
        \textcolor{blue}{target}: \{goal\}
        
        \item \textcolor{red}{source}: \texttt{QUERY}: The method ``\{step\_headline\}" is typically associated with what intent?\\
        \textcolor{blue}{target}: \{goal\}
        
        \item \textcolor{red}{source}: \texttt{QUERY}: What is the usual purpose if you begin to \{step\_headline\}?\\
        \textcolor{blue}{target}: \{goal\}
        
        \item \textcolor{red}{source}: \texttt{QUERY}: What would be the possible intent if someone begin to \{step\_headline\}?\\
        \textcolor{blue}{target}: \{goal\}
        
        \item \textcolor{red}{source}: \texttt{QUERY}: What is the most likely outcome if you \{step\_headline\}?\\
        \textcolor{blue}{target}: \{goal\}
        
        \item \textcolor{red}{source}: \texttt{QUERY}: What do you want to achieve if you \{step\_headline\}?\\
        \textcolor{blue}{target}: \{goal\}
        
        \item \textcolor{red}{source}: \texttt{QUERY}: Given the step ``\{step\_headline\}", identify the goal it serves.\\
        \textcolor{blue}{target}: \{goal\}
        
        \item \textcolor{red}{source}: \texttt{QUERY}: Find the most probable intent associated with the action ``\{step\_headline\}".\\
        \textcolor{blue}{target}: \{goal\}
        
        \item \textcolor{red}{source}: \texttt{QUERY}: ``\{step\_headline\}" is one of the procedures to?\\
        \textcolor{blue}{target}: \{goal\}
        
        \item \textcolor{red}{source}: \texttt{QUERY}: ``\{step\_headline\}" is a step if you want to?\\
        \textcolor{blue}{target}: \{goal\}
        
        \item \textcolor{red}{source}: \texttt{QUERY}: What is the ultimate goal if you \{step\_headline\}?\\
        \textcolor{blue}{target}: \{goal\}
        
        \item \textcolor{red}{source}: \texttt{QUERY}: What is the purpose to \{step\_headline\}?\\
        \textcolor{blue}{target}: \{goal\}
        
        \item \textcolor{red}{source}: \texttt{QUERY}: Can you recognize the right goal for the action ``\{step\_headline\}"?\\
        \textcolor{blue}{target}: \{goal\}
        
        \item \textcolor{red}{source}: \texttt{QUERY}: What goal will you \{step\_headline\} to achieve?\\
        \textcolor{blue}{target}: \{goal\}
        
        \item \textcolor{red}{source}: \texttt{QUERY}: To accomplish what purpose you will \{step\_headline\}?\\
        \textcolor{blue}{target}: \{goal\}
        
        \item \textcolor{red}{source}: \texttt{QUERY}: Tell me the most probable purpose if you \{step\_headline\}.\\
        \textcolor{blue}{target}: \{goal\}
        
        \item \textcolor{red}{source}: \texttt{QUERY}: What is likely to be your intent if you \{step\_headline\}?\\
        \textcolor{blue}{target}: \{goal\}
    \end{enumerate}
\end{itemize}

\paragraph{Summary}
With $($step\_headline, step\_description$)$ pairs, we construct prompts that serve for the task of summarization. Here, we use \{step\_headlines\_with\_or\} (e.g. ``option1", ``option2", or ``option3") and \{step\_headlines\_without\_or\} (e.g. ``option1", ``option2", ``option3") to refer to some (3 to 9) step headlines (contain the appropriate one). We use \{other\_step\_headline\} to represent a random step headline that is different from the appropriate one.

\begin{itemize}
    \item Multiple-choice format prompts
    \begin{enumerate}
        \item \textcolor{red}{source}: \texttt{TEXT}: \{step\_description\} \texttt{QUERY}: Which of the following can summarize this text? \{step\_headlines\_with\_or\}?\\
        \textcolor{blue}{target}: \{step\_headline\}
        
        \item \textcolor{red}{source}: \texttt{TEXT}: \{step\_description\} \texttt{QUERY}: Select a headline for the previous text. The options are \{step\_headlines\_without\_or\}.\\
        \textcolor{blue}{target}: \{step\_headline\}
        
        \item \textcolor{red}{source}: \texttt{TEXT}: \{step\_description\} \texttt{QUERY}: What's the main idea of the text? \{step\_headlines\_with\_or\}?\\
        \textcolor{blue}{target}: \{step\_headline\}
        
        \item \textcolor{red}{source}: \texttt{TEXT}: \{step\_description\} \texttt{QUERY}: Given a list of titles: \{step\_headlines\_without\_or\}, which one can express the main idea of the text?\\
        \textcolor{blue}{target}: \{step\_headline\}
        
        \item \textcolor{red}{source}: \texttt{TEXT}: \{step\_description\} \texttt{QUERY}: Given a list of headings: \{step\_headlines\_without\_or\}, which of these could be used as a heading for the above text?\\
        \textcolor{blue}{target}: \{step\_headline\}
        
        \item \textcolor{red}{source}: \texttt{TEXT}: \{step\_description\} \texttt{QUERY}: Which of the following headings matches the above text? \{step\_headlines\_with\_or\}?\\
        \textcolor{blue}{target}: \{step\_headline\}
        
        \item \textcolor{red}{source}: \texttt{TEXT}: \{step\_description\} \texttt{QUERY}: Which of the following titles covers the essence of the above text? \{step\_headlines\_with\_or\}?\\
        \textcolor{blue}{target}: \{step\_headline\}
        
        \item \textcolor{red}{source}: \texttt{TEXT}: \{step\_description\} \texttt{QUERY}: How would you summarize the text? You may choose from \{step\_headlines\_without\_or\}.\\
        \textcolor{blue}{target}: \{step\_headline\}
        
        \item \textcolor{red}{source}: \texttt{TEXT}: \{step\_description\} \texttt{QUERY}: Choose a summary for the above text from: \{step\_headlines\_without\_or\}.\\
        \textcolor{blue}{target}: \{step\_headline\}
        
        \item \textcolor{red}{source}: \texttt{TEXT}: \{step\_description\} \texttt{QUERY}: Select an appropriate heading for the paragraph. The options are: \{step\_headlines\_without\_or\}.\\
        \textcolor{blue}{target}: \{step\_headline\}
        
        \item \textcolor{red}{source}: \texttt{TEXT}: \{step\_description\} \texttt{QUERY}: Which of the following is the heading of the paragraph? \{step\_headlines\_with\_or\}?\\
        \textcolor{blue}{target}: \{step\_headline\}
        
        \item \textcolor{red}{source}: \texttt{TEXT}: \{step\_description\} \texttt{QUERY}: ``\{step\_headline\}" can summarize the previous text. ``True" or ``False"?\\
        \textcolor{blue}{target}: True
        
        \item \textcolor{red}{source}: \texttt{TEXT}: \{step\_description\} \texttt{QUERY}: ``\{other\_step\_headline\}" can summarize the previous text. ``True" or ``False"?\\
        \textcolor{blue}{target}: False
        
        \item \textcolor{red}{source}: \texttt{TEXT}: \{step\_description\} \texttt{QUERY}: Does ``\{step\_headline\}" summarize the core of the above text? ``Yes" or ``No"?\\
        \textcolor{blue}{target}: Yes
        
        \item \textcolor{red}{source}: \texttt{TEXT}: \{step\_description\} \texttt{QUERY}: Does ``\{other\_step\_headline\}" summarize the core of the above text? ``Yes" or ``No"?\\
        \textcolor{blue}{target}: No
        
        \item \textcolor{red}{source}: \texttt{TEXT}: \{step\_description\} \texttt{QUERY}: Here are some headlines: \{step\_headlines\_without\_or\}, which one is compatible with the previous text?\\
        \textcolor{blue}{target}: \{step\_headline\}
        
        \item \textcolor{red}{source}: \texttt{TEXT}: \{step\_description\} \texttt{QUERY}: Can you choose an appropriate summary of the preceding text from the following list: \{step\_headlines\_without\_or\}?\\
        \textcolor{blue}{target}: \{step\_headline\}
        
        \item \textcolor{red}{source}: \texttt{TEXT}: \{step\_description\} \texttt{QUERY}: Can you pick the suitable summary of the previous article from the following options: \{step\_headlines\_without\_or\}?\\
        \textcolor{blue}{target}: \{step\_headline\}
        
        \item \textcolor{red}{source}: \texttt{TEXT}: \{step\_description\} \texttt{QUERY}: Given options: \{step\_headlines\_without\_or\}, which of these is a summary of the preceding paragraph?\\
        \textcolor{blue}{target}: \{step\_headline\}
        
        \item \textcolor{red}{source}: \texttt{TEXT}: \{step\_description\} \texttt{QUERY}: Summarize the preceding text with one of the following options: \{step\_headlines\_without\_or\}.\\
        \textcolor{blue}{target}: \{step\_headline\}
    \end{enumerate}
    \item Generation format prompts
    \begin{enumerate}
        \item \textcolor{red}{source}: \texttt{TEXT}: \{step\_description\} \texttt{QUERY}: Can you summarize this text?\\
        \textcolor{blue}{target}: \{step\_headline\}
        
        \item \textcolor{red}{source}: \texttt{TEXT}: \{step\_description\} \texttt{QUERY}: What are the main points one should remember from this text?\\
        \textcolor{blue}{target}: \{step\_headline\}
        
        \item \textcolor{red}{source}: \texttt{TEXT}: \{step\_description\} \texttt{QUERY}: Can you generate a short summary for the previous text?\\
        \textcolor{blue}{target}: \{step\_headline\}
        
        \item \textcolor{red}{source}: \texttt{TEXT}: \{step\_description\} \texttt{QUERY}: Can you summarize the main idea of the text in your own words?\\
        \textcolor{blue}{target}: \{step\_headline\}
        
        \item \textcolor{red}{source}: \texttt{TEXT}: \{step\_description\} \texttt{QUERY}: Write a headline for the previous text.\\
        \textcolor{blue}{target}: \{step\_headline\}
        
        \item \textcolor{red}{source}: \texttt{TEXT}: \{step\_description\} \texttt{QUERY}: What's the main idea of the text?\\
        \textcolor{blue}{target}: \{step\_headline\}
        
        \item \textcolor{red}{source}: \texttt{TEXT}: \{step\_description\} \texttt{QUERY}: What can be an appropriate headline for the text?\\
        \textcolor{blue}{target}: \{step\_headline\}
        
        \item \textcolor{red}{source}: \texttt{TEXT}: \{step\_description\} \texttt{QUERY}: Summarize the core of the above text.\\
        \textcolor{blue}{target}: \{step\_headline\}
        
        \item \textcolor{red}{source}: \texttt{TEXT}: \{step\_description\} \texttt{QUERY}: Generate a title for this article.\\
        \textcolor{blue}{target}: \{step\_headline\}
        
        \item \textcolor{red}{source}: \texttt{TEXT}: \{step\_description\} \texttt{QUERY}: Can ``\{step\_headline\}" summarize the previous text?\\
        \textcolor{blue}{target}: Yes
        
        \item \textcolor{red}{source}: \texttt{TEXT}: \{step\_description\} \texttt{QUERY}: Can ``\{other\_step\_headline\}" summarize the previous text?\\
        \textcolor{blue}{target}: No
        
        \item \textcolor{red}{source}: \texttt{TEXT}: \{step\_description\} \texttt{QUERY}: Does ``\{step\_headline\}" summarize the core of the above text?\\
        \textcolor{blue}{target}: Yes
        
        \item \textcolor{red}{source}: \texttt{TEXT}: \{step\_description\} \texttt{QUERY}: Does ``\{other\_step\_headline\}" summarize the core of the above text?\\
        \textcolor{blue}{target}: No
        
        \item \textcolor{red}{source}: \texttt{TEXT}: \{step\_description\} \texttt{QUERY}: In a few words, what does the previous paragraph say?\\
        \textcolor{blue}{target}: \{step\_headline\}
        
        \item \textcolor{red}{source}: \texttt{TEXT}: \{step\_description\} \texttt{QUERY}: Briefly describe what the previous paragraph talks about.\\
        \textcolor{blue}{target}: \{step\_headline\}
        
        \item \textcolor{red}{source}: \texttt{TEXT}: \{step\_description\} \texttt{QUERY}: What can be a short description of the text?\\
        \textcolor{blue}{target}: \{step\_headline\}
        
        \item \textcolor{red}{source}: \texttt{TEXT}: \{step\_description\} \texttt{QUERY}: Write a TLDR (Too Long Didn''t Read) summary for the above text.\\
        \textcolor{blue}{target}: \{step\_headline\}
        
        \item \textcolor{red}{source}: \texttt{TEXT}: \{step\_description\} \texttt{QUERY}: Condense the text down to the essentials.\\
        \textcolor{blue}{target}: \{step\_headline\}
        
        \item \textcolor{red}{source}: \texttt{TEXT}: \{step\_description\} \texttt{QUERY}: How would you summarize the key points of the text?\\
        \textcolor{blue}{target}: \{step\_headline\}
        
        \item \textcolor{red}{source}: \texttt{TEXT}: \{step\_description\} \texttt{QUERY}: Can you express the main idea of the text?\\
        \textcolor{blue}{target}: \{step\_headline\}
    \end{enumerate}
\end{itemize}

\paragraph{Sentence expansion}
With $($step\_headline, step\_description$)$ pairs, we construct prompts that require the model to expand a short text piece into a long paragraph. We use \{length\} as a clue in some prompts to guide the model to generate appropriate content. 
\begin{itemize}
    \item Generation format prompts
    \begin{enumerate}
        \item \textcolor{red}{source}: \texttt{TEXT}: \{step\_headline\} \texttt{QUERY}: What details would you include in a storyline to make it more engaging and informative?\\
        \textcolor{blue}{target}: \{step\_description\}
        
        \item \textcolor{red}{source}: \texttt{TEXT}: \{step\_headline\} \texttt{QUERY}: Write a paragraph of approximately \{length\} words with this given title.\\
        \textcolor{blue}{target}: \{step\_description\}
        
        \item \textcolor{red}{source}: \texttt{TEXT}: \{step\_headline\} \texttt{QUERY}: Please expand the above sentences into a long paragraph.\\
        \textcolor{blue}{target}: \{step\_description\}
        
        \item \textcolor{red}{source}: \texttt{TEXT}: \{step\_headline\} \texttt{QUERY}: Can you add some details to the previous text?\\
        \textcolor{blue}{target}: \{step\_description\}
        
        \item \textcolor{red}{source}: \texttt{TEXT}: \{step\_headline\} \texttt{QUERY}: Can you write a \{length\}-word article that uses the previous text as a summary?\\
        \textcolor{blue}{target}: \{step\_description\}
        
        \item \textcolor{red}{source}: \texttt{TEXT}: \{step\_headline\} \texttt{QUERY}: Write a long paragraph that is semantically similar to the above sentences but has more details.\\
        \textcolor{blue}{target}: \{step\_description\}
        
        \item \textcolor{red}{source}: \texttt{TEXT}: \{step\_headline\} \texttt{QUERY}: Based on the headline, can you write a more informative paragraph using around \{length\} words?\\
        \textcolor{blue}{target}: \{step\_description\}
        
        \item \textcolor{red}{source}: \texttt{TEXT}: \{step\_headline\} \texttt{QUERY}: Can you expand the above headline into a paragraph of about \{length\} words?\\
        \textcolor{blue}{target}: \{step\_description\}
        
        \item \textcolor{red}{source}: \texttt{TEXT}: \{step\_headline\} \texttt{QUERY}: Given the above title, what text might you write? Please write about \{length\} words.\\
        \textcolor{blue}{target}: \{step\_description\}
        
        \item \textcolor{red}{source}: \texttt{TEXT}: \{step\_headline\} \texttt{QUERY}: Given the above title, please write some details for it.\\
        \textcolor{blue}{target}: \{step\_description\}
        
        \item \textcolor{red}{source}: \texttt{TEXT}: \{step\_headline\} \texttt{QUERY}: Write a \{length\}-word paragraph with the above title as the main topic.\\
        \textcolor{blue}{target}: \{step\_description\}
        
        \item \textcolor{red}{source}: \texttt{TEXT}: \{step\_headline\} \texttt{QUERY}: Write an article with the headline above as the main point.\\
        \textcolor{blue}{target}: \{step\_description\}
        
        \item \textcolor{red}{source}: \texttt{TEXT}: \{step\_headline\} \texttt{QUERY}: How to expand the above text to make it more specific?\\
        \textcolor{blue}{target}: \{step\_description\}
        
        \item \textcolor{red}{source}: \texttt{TEXT}: \{step\_headline\} \texttt{QUERY}: Continue with this text and write a consistent paragraph.\\
        \textcolor{blue}{target}: \{step\_description\}
        
        \item \textcolor{red}{source}: \texttt{TEXT}: \{step\_headline\} \texttt{QUERY}: Write a piece of text based on the above main idea.\\
        \textcolor{blue}{target}: \{step\_description\}
        
        \item \textcolor{red}{source}: \texttt{TEXT}: \{step\_headline\} \texttt{QUERY}: Can you generate a relevant paragraph of around \{length\} words with the above text as the main topic?\\
        \textcolor{blue}{target}: \{step\_description\}
        
        \item \textcolor{red}{source}: \texttt{TEXT}: \{step\_headline\} \texttt{QUERY}: Can you add some details after this text to construct a paragraph of around \{length\} words?\\
        \textcolor{blue}{target}: \{step\_description\}
        
        \item \textcolor{red}{source}: \texttt{TEXT}: \{step\_headline\} \texttt{QUERY}: Can you write a text that is semantically similar to the previous title, but with more detail and content?\\
        \textcolor{blue}{target}: \{step\_description\}
        
        \item \textcolor{red}{source}: \texttt{TEXT}: \{step\_headline\} \texttt{QUERY}: Please use about \{length\} words to enrich the above sentence so that it contains more information?\\
        \textcolor{blue}{target}: \{step\_description\}
        
        \item \textcolor{red}{source}: \texttt{TEXT}: \{step\_headline\} \texttt{QUERY}: Your task is to detail the above headline and write a paragraph of around \{length\} words.\\
        \textcolor{blue}{target}: \{step\_description\}
    
    \end{enumerate}
\end{itemize}
\paragraph{Procedure}
With $($goal, first\_step\_headline, second\_step\_headline$)$ triples, we can construct prompts to ask about the temporal relationship of the steps to accomplish a certain goal. We use \{choice1\} and \{choice2\} to represent the random order of \{first\_step\_headline\} and \{second\_step\_headline\}. Here, we use \{step\_headlines\_with\_or\} (e.g. ``option1", ``option2", or ``option3") and \{step\_headlines\_without\_or\} (e.g. ``option1", ``option2", ``option3") to refer to all step headlines (except the first one) needed to represent a goal.

\begin{itemize}
    \item Multiple-choice format prompts
    \begin{enumerate}
        \item \textcolor{red}{source}: \texttt{QUERY}: Given the goal ``\{goal\}", which of the following steps should be executed first? ``\{choice1\}" or ``\{choice2\}"?\\
        \textcolor{blue}{target}: \{first\_step\_headline\}
        
        \item \textcolor{red}{source}: \texttt{QUERY}: Given two steps: ``\{choice1\}" and ``\{choice2\}", which one should be considered first to accomplish the goal ``\{goal\}"?\\
        \textcolor{blue}{target}: \{first\_step\_headline\}
        
        \item \textcolor{red}{source}: \texttt{QUERY}: To accomplish the goal ``\{goal\}", what will you do first? ``\{choice1\}" or ``\{choice2\}"?\\
        \textcolor{blue}{target}: \{first\_step\_headline\}
        
        \item \textcolor{red}{source}: \texttt{QUERY}: You have a goal ``\{goal\}", what would be your first step? ``\{choice1\}" or ``\{choice2\}"?\\
        \textcolor{blue}{target}: \{first\_step\_headline\}
        
        \item \textcolor{red}{source}: \texttt{QUERY}: To \{goal\}, which of the following will you do first? ``\{choice1\}" or ``\{choice2\}"?\\
        \textcolor{blue}{target}: \{first\_step\_headline\}
        
        \item \textcolor{red}{source}: \texttt{QUERY}: In order to \{goal\}, which of the following should be the first step? ``\{choice1\}" or ``\{choice1\}"?\\
        \textcolor{blue}{target}: \{first\_step\_headline\}
        
        \item \textcolor{red}{source}: \texttt{QUERY}: Here are two methods: ``\{choice1\}" and ``\{choice2\}", which one should be the first step to \{goal\}?\\
        \textcolor{blue}{target}: \{first\_step\_headline\}
        
        \item \textcolor{red}{source}: \texttt{QUERY}: For the goal ``\{goal\}", which of the following actions should be performed first? ``\{choice1\}" or ``\{choice2\}"?\\
        \textcolor{blue}{target}: \{first\_step\_headline\}
        
        \item \textcolor{red}{source}: \texttt{QUERY}: Which of the two steps, ``\{choice1\}" or ``\{choice2\}", should be done first to accomplish the goal ``\{goal\}"?\\
        \textcolor{blue}{target}: \{first\_step\_headline\}
        
        \item \textcolor{red}{source}: \texttt{QUERY}: Which of the two procedures, ``\{choice1\}" or ``\{choice2\}", should come first to \{goal\}?\\
        \textcolor{blue}{target}: \{first\_step\_headline\}
        
        \item \textcolor{red}{source}: \texttt{QUERY}: Your goal is to \{goal\}, identify the first step to do from ``\{choice1\}", ``\{choice2\}".\\
        \textcolor{blue}{target}: \{first\_step\_headline\}
        
        \item \textcolor{red}{source}: \texttt{QUERY}: Given that you want to \{goal\}, will you first ``\{choice1\}" or ``\{choice2\}"?\\
        \textcolor{blue}{target}: \{first\_step\_headline\}
        
        \item \textcolor{red}{source}: \texttt{QUERY}: Consider the two procedures in the ``\{goal\}" process: ``\{choice1\}", ``\{choice2\}", which one should come first?\\
        \textcolor{blue}{target}: \{first\_step\_headline\}
        
        \item \textcolor{red}{source}: \texttt{QUERY}: Consider the two steps in the ``\{goal\}" process: ``\{choice1\}", ``\{choice2\}", which one to take first?\\
        \textcolor{blue}{target}: \{first\_step\_headline\}
        
        \item \textcolor{red}{source}: \texttt{QUERY}: In accomplishing the goal ``\{goal\}", which should come first, ``\{choice1\}" or ``\{choice2\}" in a temporal order?\\
        \textcolor{blue}{target}: \{first\_step\_headline\}
        
        \item \textcolor{red}{source}: \texttt{QUERY}: In order to \{goal\}, what should be performed first? ``\{choice1\}" or ``\{choice2\}"?\\
        \textcolor{blue}{target}: \{first\_step\_headline\}
        
        \item \textcolor{red}{source}: \texttt{QUERY}: There is a chronological relationship between the steps of ``\{choice1\}" and ``\{choice2\}" in the ``\{goal\}" process. Which one should be executed first?\\
        \textcolor{blue}{target}: \{first\_step\_headline\}
        
        \item \textcolor{red}{source}: \texttt{QUERY}: In the implementation of your plan ``\{goal\}", you will first ``\{choice1\}" or ``\{choice2\}"?\\
        \textcolor{blue}{target}: \{first\_step\_headline\}
        
        \item \textcolor{red}{source}: \texttt{QUERY}: With the goal ``\{goal\}", will you ``\{choice1\}" or ``\{choice2\}" first?\\
        \textcolor{blue}{target}: \{first\_step\_headline\}
        
        \item \textcolor{red}{source}: \texttt{QUERY}: In order to \{goal\}, chronologically speaking, should we ``\{choice1\}" or ``\{choice2\}" first?\\
        \textcolor{blue}{target}: \{first\_step\_headline\}
        
        \item \textcolor{red}{source}: \texttt{QUERY}: You want to \{goal\}, given a step ``\{first\_step\_headline\}", which of the following should be the next step? \{step\_headlines\_with\_or\}?\\
        \textcolor{blue}{target}: \{second\_step\_headline\}
        
        \item \textcolor{red}{source}: \texttt{QUERY}: To \{goal\}, given an event ``\{first\_step\_headline\}", what should be done next? \{step\_headlines\_with\_or\}?\\
        \textcolor{blue}{target}: \{second\_step\_headline\}
        
        \item \textcolor{red}{source}: \texttt{QUERY}: In order to \{goal\}, what should you do next after you ``\{first\_step\_headline\}"? \{step\_headlines\_with\_or\}?\\
        \textcolor{blue}{target}: \{second\_step\_headline\}
        
        \item \textcolor{red}{source}: \texttt{QUERY}: Your goal is to \{goal\}, what will you do after you ``\{first\_step\_headline\}"? \{step\_headlines\_with\_or\}?\\
        \textcolor{blue}{target}: \{second\_step\_headline\}
        
        \item \textcolor{red}{source}: \texttt{QUERY}: Given the goal ``\{goal\}" and steps: \{step\_headlines\_without\_or\}, which one should come next after a previous step ``\{first\_step\_headline\}"?\\
        \textcolor{blue}{target}: \{second\_step\_headline\}
        
        \item \textcolor{red}{source}: \texttt{QUERY}: Given the goal ``\{goal\}" and steps: \{step\_headlines\_without\_or\}, which one should come after ``\{first\_step\_headline\}"?\\
        \textcolor{blue}{target}: \{secons\_step\_headline\}
        
        \item \textcolor{red}{source}: \texttt{QUERY}: With the goal ``\{goal\}" and a first step ``\{first\_step\_headline\}", which one of the following could be a second step: \{step\_headlines\_with\_or\}?\\
        \textcolor{blue}{target}: \{second\_step\_headline\}
        
        \item \textcolor{red}{source}: \texttt{QUERY}: In order to achieve the goal ``\{goal\}", you first have to ``\{first\_step\_headline\}", what is the next reasonable step to take? Should you \{step\_headlines\_with\_or\}?\\
        \textcolor{blue}{target}: \{second\_step\_headline\}
        
        \item \textcolor{red}{source}: \texttt{QUERY}: For the goal ``\{goal\}", you should first ``\{first\_step\_headline\}". Choose the appropriate next step from: \{step\_headlines\_without\_or\}.\\
        \textcolor{blue}{target}: \{second\_step\_headline\}
        
        \item \textcolor{red}{source}: \texttt{QUERY}: Which of the following steps should follow ``\{first\_step\_headline\}" in the ``\{goal\}" process: \{step\_headlines\_with\_or\}?\\
        \textcolor{blue}{target}:\{second\_step\_headline\}
        
        \item \textcolor{red}{source}: \texttt{QUERY}: Here is a procedure list: \{step\_headlines\_without\_or\}, which one should come right after ``\{first\_step\_headline\}" if you want to \{goal\}?\\
        \textcolor{blue}{target}: \{second\_step\_headline\}
        
        \item \textcolor{red}{source}: \texttt{QUERY}: Given the following procedure list: \{step\_headlines\_without\_or\}, choose the one that should follow ``\{first\_step\_headline\}" in accomplishing the goal ``\{goal\}".\\
        \textcolor{blue}{target}: \{second\_step\_headline\}
        
        \item \textcolor{red}{source}: \texttt{QUERY}: Given the goal ``\{goal\}" and one previous step ``\{first\_step\_headline\}", what to do next? \{step\_headlines\_with\_or\}?\\
        \textcolor{blue}{target}: \{second\_step\_headline\}
        
        \item \textcolor{red}{source}: \texttt{QUERY}: Given the goal ``\{goal\}", which procedure should come after ``\{first\_step\_headline\}"? \{step\_headlines\_with\_or\}?\\
        \textcolor{blue}{target}: \{second\_step\_headline\}
        
        \item \textcolor{red}{source}: \texttt{QUERY}: With the goal ``\{goal\}", you identified the first step to take: ``\{first\_step\_headline\}", what about the next one? The options are \{step\_headlines\_without\_or\}.\\
        \textcolor{blue}{target}: \{second\_step\_headline\}
        
        \item \textcolor{red}{source}: \texttt{QUERY}: With the goal ``\{goal\}", you plan to \{first\_step\_headline\} first, what could be a next step? The options are \{step\_headlines\_without\_or\}.\\
        \textcolor{blue}{target}: \{second\_step\_headline\}
        
        \item \textcolor{red}{source}: \texttt{QUERY}: To \{goal\}, which of the following should be considered right after ``\{first\_step\_headline\}"? \{step\_headlines\_with\_or\}?\\
        \textcolor{blue}{target}: \{second\_step\_headline\}
        
        \item \textcolor{red}{source}: \texttt{QUERY}: For the goal ``\{goal\}", what will you do next after you ``\{first\_step\_headline\}"? You may choose from \{step\_headlines\_without\_or\}.\\
        \textcolor{blue}{target}: \{second\_step\_headline\}
        
        \item \textcolor{red}{source}: \texttt{QUERY}: In order to \{goal\}, choose the most appropriate procedure after you ``\{first\_step\_headline\}". Your options are \{step\_headlines\_without\_or\}.\\
        \textcolor{blue}{target}: \{second\_step\_headline\}
        
        \item \textcolor{red}{source}: \texttt{QUERY}: Given the goal ``\{goal\}" and a previous step ``\{first\_step\_headline\}", pick the best next step from the following list: \{step\_headlines\_without\_or\}.\\
        \textcolor{blue}{target}: \{second\_step\_headline\}
    \end{enumerate}
    \item Generation format prompts
    \begin{enumerate}
           \item \textcolor{red}{source}: \texttt{QUERY}: You want to \{goal\}, given a step ``\{first\_step\_headline\}", what should be the next step?\\
        \textcolor{blue}{target}: \{second\_step\_headline\}
        
        \item \textcolor{red}{source}: \texttt{QUERY}: To \{goal\}, given an event ``\{first\_step\_headline\}", what should be done next?\\
        \textcolor{blue}{target}: \{second\_step\_headline\}
        
        \item \textcolor{red}{source}: \texttt{QUERY}: In order to \{goal\}, what should you do next after you ``\{first\_step\_headline\}"?\\
        \textcolor{blue}{target}: \{second\_step\_headline\}
        
        \item \textcolor{red}{source}: \texttt{QUERY}: Your goal is to \{goal\}, what will you do after you ``\{first\_step\_headline\}"?\\
        \textcolor{blue}{target}: \{second\_step\_headline\}
        
        \item \textcolor{red}{source}: \texttt{QUERY}: Given the goal ``\{goal\}", what should come next after a previous step ``\{first\_step\_headline\}"?\\
        \textcolor{blue}{target}: \{second\_step\_headline\}
        
        \item \textcolor{red}{source}: \texttt{QUERY}: Given the goal ``\{goal\}", what should come after ``\{first\_step\_headline\}"?\\
        \textcolor{blue}{target}: \{second\_step\_headline\}
        
        \item \textcolor{red}{source}: \texttt{QUERY}: With the goal ``\{goal\}" and a first step ``\{first\_step\_headline\}", What could be a second step?\\
        \textcolor{blue}{target}: \{second\_step\_headline\}
        
        \item \textcolor{red}{source}: \texttt{QUERY}: In order to achieve the goal ``\{goal\}", you first have to ``\{first\_step\_headline\}", what is the next reasonable step to take?\\
        \textcolor{blue}{target}: \{second\_step\_headline\}
        
        \item \textcolor{red}{source}: \texttt{QUERY}: For the goal ``\{goal\}", you should first ``\{first\_step\_headline\}". What should be the next step?\\
        \textcolor{blue}{target}: \{second\_step\_headline\}
        
        \item \textcolor{red}{source}: \texttt{QUERY}: What step should follow ``\{first\_step\_headline\}" in the ``\{goal\}" process?\\
        \textcolor{blue}{target}: \{second\_step\_headline\}
        
        \item \textcolor{red}{source}: \texttt{QUERY}: What step should come right after ``\{first\_step\_headline\}" if you want to \{goal\}?\\
        \textcolor{blue}{target}: \{second\_step\_headline\}
        
        \item \textcolor{red}{source}: \texttt{QUERY}: Write down the step that should follow ``\{first\_step\_headline\}" in accomplishing the goal ``\{goal\}".\\
        \textcolor{blue}{target}: \{second\_step\_headline\}
        
        \item \textcolor{red}{source}: \texttt{QUERY}: Given the goal ``\{goal\}" and one previous step ``\{first\_step\_headline\}", what to do next?\\
        \textcolor{blue}{target}: \{second\_step\_headline\}
        
        \item \textcolor{red}{source}: \texttt{QUERY}: Given the goal ``\{goal\}", what procedure should come after ``\{first\_step\_headline\}"?\\
        \textcolor{blue}{target}: \{second\_step\_headline\}
        
        \item \textcolor{red}{source}: \texttt{QUERY}: With the goal ``\{goal\}", you identified the first step to take: ``\{first\_step\_headline\}", what about the next one?\\
        \textcolor{blue}{target}: \{second\_step\_headline\}
        
        \item \textcolor{red}{source}: \texttt{QUERY}: With the goal ``\{goal\}", you plan to \{first\_step\_headline\} first, what could be a next step?\\
        \textcolor{blue}{target}: \{second\_step\_headline\}
        
        \item \textcolor{red}{source}: \texttt{QUERY}: To \{goal\}, what should be considered right after ``\{first\_step\_headline\}"?\\
        \textcolor{blue}{target}: \{second\_step\_headline\}
        
        \item \textcolor{red}{source}: \texttt{QUERY}: For the goal ``\{goal\}", what will you do next after you ``\{first\_step\_headline\}"?\\
        \textcolor{blue}{target}: \{second\_step\_headline\}
        
        \item \textcolor{red}{source}: \texttt{QUERY}: In order to \{goal\}, come up with the most appropriate procedure after you ``\{first\_step\_headline\}".\\
        \textcolor{blue}{target}: \{second\_step\_headline\}
        
        \item \textcolor{red}{source}: \texttt{QUERY}: Given the goal ``\{goal\}" and a previous step ``\{first\_step\_headline\}", what do you plan to do next?\\
        \textcolor{blue}{target}: \{second\_step\_headline\}
    \end{enumerate}
  
\end{itemize}
\paragraph{questions}
With $($title, answer, first\_related\_question$)$ triples, we design the following prompts that ask for some related questions given a question-answer pair. Here, we use \{questions\_with\_or\} (e.g. ``option1", ``option2", or ``option3") and \{questions\_without\_or\} (e.g. ``option1", ``option2", ``option3") to refer to some (3 to 9) available questions (including one related question and the rest are irrelevant questions).

\begin{itemize}
    \item Multiple-choice format prompts
    \begin{enumerate}
        \item \textcolor{red}{source}: \texttt{TEXT}: Question: \{title\}? Answer: \{answer\}. \texttt{QUERY}: What's the next question you may ask? \{questions\_with\_or\}?\\
        \textcolor{blue}{target}: \{first\_related\_question\}
        
        \item \textcolor{red}{source}: \texttt{TEXT}: Someone asked ``\{title\}" and got the following answer: \{answer\} \texttt{QUERY}: Can you suggest a relevant question from the following list: \{questions\_with\_or\}?\\
        \textcolor{blue}{target}: \{first\_related\_question\}
        
        \item \textcolor{red}{source}: \texttt{TEXT}: Here is a question ``\{title\}" and its answer ``\{answer\}" \texttt{QUERY}: which of the following would be a worthwhile question to continue to ask? \{questions\_with\_or\}?\\
        \textcolor{blue}{target}: \{first\_related\_question\}
        
        \item \textcolor{red}{source}: \texttt{TEXT}: A question on the forum is ``\{title\}", someone gave the following answer: ``\{answer\}". \texttt{QUERY}: What question is the questioner likely to ask next? \{questions\_with\_or\}?\\
        \textcolor{blue}{target}: \{first\_related\_question\}
        
        \item \textcolor{red}{source}: \texttt{TEXT}: A person views the question ``\{title\}" and sees the following answer: ``\{answer\}". \texttt{QUERY}: What would you recommend for him/her to read? Please choose one from the list below: \{questions\_without\_or\}.\\
        \textcolor{blue}{target}: \{first\_related\_question\}
        
        \item \textcolor{red}{source}: \texttt{TEXT}: After searching the question ``\{title\}", you got the following answer: ``\{answer\}". \texttt{QUERY}: what question would you probably search next? Please select from the following: \{questions\_without\_or\}.\\
        \textcolor{blue}{target}: \{first\_related\_question\}
        
        \item \textcolor{red}{source}: \texttt{TEXT}: Here is a dialogue between two people. Person 1: \{title\}. Person 2: \{answer\} \texttt{QUERY}: What question might Person 1 ask next? \{questions\_with\_or\}?\\
        \textcolor{blue}{target}: \{first\_related\_question\}
        
        \item \textcolor{red}{source}: \texttt{TEXT}: Here is a dialogue between two people. Person 1: \{title\}. Person 2: \{answer\} \texttt{QUERY}: If you were Person 1, what question would you ask next? \{questions\_with\_or\}?\\
        \textcolor{blue}{target}: \{first\_related\_question\}
        
        \item \textcolor{red}{source}: \texttt{TEXT}: You have read the answer to ``\{title\}": ``\{answer\}". \texttt{QUERY}: Now there is a list of topics: \{questions\_without\_or\}. Which one might be of interest to you?\\
        \textcolor{blue}{target}: \{first\_related\_question\}
        
        \item \textcolor{red}{source}: \texttt{TEXT}: You clicked on the ``\{title\}" page and read the answers below. The most comprehensive answer is: \{answer\}. \texttt{QUERY}: This page has links to the following topics: \{questions\_without\_or\}. Which one are you most likely to continue clicking on?\\
        \textcolor{blue}{target}: \{first\_related\_question\}
        
        \item \textcolor{red}{source}: \texttt{TEXT}: Question: \{title\}? Answer: \{answer\}. \texttt{QUERY}: Which of the following questions is relevant to the current question and answer? \{questions\_with\_or\}?\\
        \textcolor{blue}{target}: \{first\_related\_question\}
        
        \item \textcolor{red}{source}: \texttt{TEXT}: You answered the question ``\{title\}" in the forum and gave the following answer: ``\{answer\}". \texttt{QUERY}: Which one from the list are you likely to answer as well: \{questions\_with\_or\}?\\
        \textcolor{blue}{target}: \{first\_related\_question\}
        
        \item \textcolor{red}{source}: \texttt{TEXT}: Question: \{title\}? Answer: \{answer\} \texttt{QUERY}: What would be a reasonable question to continue to ask? \{questions\_with\_or\}?\\
        \textcolor{blue}{target}: \{first\_related\_question\}
        
        \item \textcolor{red}{source}: \texttt{TEXT}: Question: \{title\}? Answer: \{answer\} \texttt{QUERY}: Identify one question from the following that is most relevant to the current context: \{questions\_without\_or\}.\\
        \textcolor{blue}{target}: \{first\_related\_question\}
        
        \item \textcolor{red}{source}: \texttt{TEXT}: Question: \{title\}? Answer: \{answer\} \texttt{QUERY}: Which of the following questions can be used as a continuation of the above context? \{questions\_with\_or\}?\\
        \textcolor{blue}{target}: \{first\_related\_question\}
        
        \item \textcolor{red}{source}: \texttt{TEXT}: Question: \{title\}? Answer: \{answer\} \texttt{QUERY}: Given these questions: \{questions\_without\_or\}, can you select a relevant one based on the previous context?\\
        \textcolor{blue}{target}: \{first\_related\_question\}
        
        \item \textcolor{red}{source}: \texttt{TEXT}: One questioner asked the question ``\{title\}". A respondent answered him ``\{answer\}". \texttt{QUERY}: This questioner then asked another question. Which of the following is the most likely one? \{questions\_with\_or\}?\\
        \textcolor{blue}{target}: \{first\_related\_question\}
        
        \item \textcolor{red}{source}: \texttt{TEXT}: One person asked: \{title\}? Another person answered: \{answer\} \texttt{QUERY}: What might the first person ask next? \{questions\_with\_or\}?\\
        \textcolor{blue}{target}: \{first\_related\_question\}
        
        \item \textcolor{red}{source}: \texttt{TEXT}: One person asked: \{title\}? Another person answered: \{answer\} \texttt{QUERY}: Here is a topic list: \{questions\_without\_or\}. Which one of these might the first person be interested in as well? \{questions\_with\_or\}?\\
        \textcolor{blue}{target}: \{first\_related\_question\}
        
        \item \textcolor{red}{source}: \texttt{TEXT}: A web page contains the question: ``\{title\}" and the answer: ``\{answer\}". \texttt{QUERY}: Which of the following questions could be on this page at the same time? \{questions\_with\_or\}?\\
        \textcolor{blue}{target}: \{first\_related\_question\}
        
    \end{enumerate}
    
    \item Generation format prompts
    \begin{enumerate}
        \item \textcolor{red}{source}: \texttt{TEXT}: Question: \{title\}? Answer: \{answer\}. \texttt{QUERY}: What's the next question you may ask? \\
        \textcolor{blue}{target}: \{first\_related\_question\}
        
        \item \textcolor{red}{source}: \texttt{TEXT}: Someone asked ``\{title\}" and got the following answer: \{answer\} \texttt{QUERY}: Can you suggest a relevant question?\\
        \textcolor{blue}{target}: \{first\_related\_question\}
        
        \item \textcolor{red}{source}: \texttt{TEXT}: Here is a question ``\{title\}" and its answer ``\{answer\}" \texttt{QUERY}: What would be a worthwhile question to continue to ask?\\
        \textcolor{blue}{target}: \{first\_related\_question\}
        
        \item \textcolor{red}{source}: \texttt{TEXT}: A question on the forum is ``\{title\}", someone gave the following answer: ``\{answer\}". \texttt{QUERY}: What question is the questioner likely to ask next?\\
        \textcolor{blue}{target}: \{first\_related\_question\}
        
        \item \textcolor{red}{source}: \texttt{TEXT}: A person views the question ``\{title\}" and sees the following answer: ``\{answer\}". \texttt{QUERY}: What would you recommend for him/her to read?\\
        \textcolor{blue}{target}: \{first\_related\_question\}
        
        \item \textcolor{red}{source}: \texttt{TEXT}: After searching the question ``\{title\}", you got the following answer: ``\{answer\}". \texttt{QUERY}: What question would you probably search next?\\
        \textcolor{blue}{target}: \{first\_related\_question\}
        
        \item \textcolor{red}{source}: \texttt{TEXT}: Here is a dialogue between two people. Person 1: \{title\}. Person 2: \{answer\} \texttt{QUERY}: What question might Person 1 ask next?\\
        \textcolor{blue}{target}: \{first\_related\_question\}
        
        \item \textcolor{red}{source}: \texttt{TEXT}: Here is a dialogue between two people. Person 1: \{title\}. Person 2: \{answer\} \texttt{QUERY}: If you were Person 1, what question would you ask next?\\
        \textcolor{blue}{target}: \{first\_related\_question\}
        
        \item \textcolor{red}{source}: \texttt{TEXT}: You have read the answer to ``\{title\}": ``\{answer\}". \texttt{QUERY}: What other question might be of interest to you?\\
        \textcolor{blue}{target}: \{first\_related\_question\}
        
        \item \textcolor{red}{source}: \texttt{TEXT}: You clicked on the ``\{title\}" page and read the answers below. The most comprehensive answer is: \{answer\}. \texttt{QUERY}: What other question are you most likely to continue clicking on?\\
        \textcolor{blue}{target}: \{first\_related\_question\}
        
        \item \textcolor{red}{source}: \texttt{TEXT}: Question: \{title\}? Answer: \{answer\}. \texttt{QUERY}: Can you write down a question that is relevant to the current question and answer?\\
        \textcolor{blue}{target}: \{first\_related\_question\}
        
        \item \textcolor{red}{source}: \texttt{TEXT}: You answered the question ``\{title\}" in the forum and gave the following answer: ``\{answer\}". \texttt{QUERY}: What other question are you likely to answer as well?\\
        \textcolor{blue}{target}: \{first\_related\_question\}
        
        \item \textcolor{red}{source}: \texttt{TEXT}: Question: \{title\}? Answer: \{answer\} \texttt{QUERY}: What would be a reasonable question to continue to ask?\\
        \textcolor{blue}{target}: \{first\_related\_question\}
        
        \item \textcolor{red}{source}: \texttt{TEXT}: Question: \{title\}? Answer: \{answer\} \texttt{QUERY}: Can you think of a question that is most relevant to the current context?\\
        \textcolor{blue}{target}: \{first\_related\_question\}
        
        \item \textcolor{red}{source}: \texttt{TEXT}: Question: \{title\}? Answer: \{answer\} \texttt{QUERY}: What question can be used as a continuation of the above context?\\
        \textcolor{blue}{target}: \{first\_related\_question\}
        
        \item \textcolor{red}{source}: \texttt{TEXT}: Question: \{title\}? Answer: \{answer\} \texttt{QUERY}: Can you generate a relevant question based on the previous context?\\
        \textcolor{blue}{target}: \{first\_related\_question\}
        
        \item \textcolor{red}{source}: \texttt{TEXT}: One questioner asked the question ``\{title\}". A respondent answered him ``\{answer\}". \texttt{QUERY}: This questioner then asked another question. What could it be?\\
        \textcolor{blue}{target}: \{first\_related\_question\}
        
        \item \textcolor{red}{source}: \texttt{TEXT}: One person asked: \{title\}? Another person answered: \{answer\} \texttt{QUERY}: What might the first person ask next?\\
        \textcolor{blue}{target}: \{first\_related\_question\}
        
        \item \textcolor{red}{source}: \texttt{TEXT}: One person asked: \{title\}? Another person answered: \{answer\} \texttt{QUERY}: What other question might the first person be interested in as well?\\
        \textcolor{blue}{target}: \{first\_related\_question\}
        
        \item \textcolor{red}{source}: \texttt{TEXT}: A web page contains the question: ``\{title\}" and the answer: ``\{answer\}". \texttt{QUERY}: What question might be on this page at the same time?\\
        \textcolor{blue}{target}: \{first\_related\_question\}
    \end{enumerate}
\end{itemize}

With $($title\_description, title$)$ pairs, we design the following prompts that ask a question based on some textual descriptions. Here, we use \{questions\_with\_or\} (e.g. ``option1", ``option2", or ``option3") and \{questions\_without\_or\} (e.g. ``option1", ``option2", ``option3") to refer to some (3 to 9) available questions (including the title and other irrelevant questions). We use \{not\_related\_question\} to represent a random question that is not the \{title\} nor its related questions.
\begin{itemize}
    \item Multiple-choice format prompts
    \begin{enumerate}
        \item \textcolor{red}{source}: \texttt{TEXT}: \{title\_description\} \texttt{QUERY}: Based on this text, Which one of the following could be a relevant question? \{questions\_with\_or\}?\\
    \textcolor{blue}{target}: \{title\}
    
    \item \textcolor{red}{source}: \texttt{TEXT}: \{title\_description\} \texttt{QUERY}: After reading the paragraph, please ask a related question. You may choose from \{questions\_without\_or\}.\\
    \textcolor{blue}{target}: \{title\}
    
    \item \textcolor{red}{source}: \texttt{TEXT}: \{title\_description\} \texttt{QUERY}: Come up with a relevant question based on this text. Please constrain your answer to \{questions\_without\_or\}.\\
    \textcolor{blue}{target}: \{title\}
    
    \item \textcolor{red}{source}: \texttt{TEXT}: \{title\_description\} \texttt{QUERY}: Which of the following questions is related to the passage? \{questions\_with\_or\}?\\
    \textcolor{blue}{target}: \{title\}
    
    \item \textcolor{red}{source}: \texttt{TEXT}: \{title\_description\} \texttt{QUERY}: Based on the passage, can you choose a relevant question from the following list: \{questions\_without\_or\}?\\
    \textcolor{blue}{target}: \{title\}
    
    \item \textcolor{red}{source}: \texttt{TEXT}: \{title\_description\} \texttt{QUERY}: Here is a list of questions: \{questions\_without\_or\}. Which one of the questions is related to the paragraph?\\
    \textcolor{blue}{target}: \{title\}
    
    \item \textcolor{red}{source}: \texttt{TEXT}: \{title\_description\} \texttt{QUERY}: Which of these questions is relevant to the above paragraph: \{questions\_with\_or\}?\\
    \textcolor{blue}{target}: \{title\}
    
    \item \textcolor{red}{source}: \texttt{TEXT}: \{title\_description\} \texttt{QUERY}: After reading this text, what would be a relevant question to ask? \{questions\_with\_or\}?\\
    \textcolor{blue}{target}: \{title\}
    
    \item \textcolor{red}{source}: \texttt{TEXT}: \{title\_description\} \texttt{QUERY}: Can you ask a question based on the previous context? You may choose from \{questions\_without\_or\}.\\
    \textcolor{blue}{target}: \{title\}
    
    \item \textcolor{red}{source}: \texttt{TEXT}: \{title\_description\} \texttt{QUERY}: There are several questions here: \{questions\_without\_or\}. Which of these questions is relevant to the previous paragraph?\\
    \textcolor{blue}{target}: \{title\}
    
    \item \textcolor{red}{source}: \texttt{TEXT}: \{title\_description\} \texttt{QUERY}: After reading this passage, what question might a person ask? \{questions\_with\_or\}?\\
    \textcolor{blue}{target}: \{title\}
    
    \item \textcolor{red}{source}: \texttt{TEXT}: \{title\_description\} \texttt{QUERY}: Select a question from the list that is closely related to the context: \{questions\_without\_or\}.\\
    \textcolor{blue}{target}: \{title\}
    
    \item \textcolor{red}{source}: \texttt{TEXT}: \{title\_description\} \texttt{QUERY}: Which of the following questions might this text be a problem description for? \{questions\_with\_or\}?\\
    \textcolor{blue}{target}: \{title\}
    
    \item \textcolor{red}{source}: \texttt{TEXT}: \{title\_description\} \texttt{QUERY}: Given a question list: \{questions\_without\_or\}, which item is related to the content of the paragraph?\\
    \textcolor{blue}{target}: \{title\}
    
    \item \textcolor{red}{source}: \texttt{TEXT}: \{title\_description\} \texttt{QUERY}: Based on this text, choose a relevant question from: \{questions\_without\_or\}.\\
    \textcolor{blue}{target}: \{title\}
    
    \item \textcolor{red}{source}: \texttt{TEXT}: \{title\_description\} \texttt{QUERY}: Ask a question about this text. You may choose from: \{questions\_without\_or\}.\\
    \textcolor{blue}{target}: \{title\}
    
    \item \textcolor{red}{source}: \texttt{TEXT}: \{title\_description\} \texttt{QUERY}: Write down a question related to the paragraph. The options are: \{questions\_without\_or\}.\\
    \textcolor{blue}{target}: \{title\}
    
    \item \textcolor{red}{source}: \texttt{TEXT}: \{title\_description\} \texttt{QUERY}: Given the options: \{questions\_without\_or\}, what question you may ask about this paragraph?\\
    \textcolor{blue}{target}: \{title\}
    
    \item \textcolor{red}{source}: \texttt{TEXT}: \{title\_description\} \texttt{QUERY}: Read the above text and select the question related to the passage from: \{questions\_without\_or\}.\\
    \textcolor{blue}{target}: \{title\}
    
    \item \textcolor{red}{source}: \texttt{TEXT}: \{title\_description\} \texttt{QUERY}: Read the previous text and select a question that one could ask based on the text: \{questions\_without\_or\}.\\
    \textcolor{blue}{target}: \{title\}
    \end{enumerate}

    \item Generation format prompts
    \begin{enumerate}
        \item \textcolor{red}{source}: \texttt{TEXT}: \{title\_description\} \texttt{QUERY}: Based on this text, what could be a relevant question?\\
        \textcolor{blue}{target}: \{title\}
        
        \item \textcolor{red}{source}: \texttt{TEXT}: \{title\_description\} \texttt{QUERY}: After reading the paragraph, please ask a related question.\\
        \textcolor{blue}{target}: \{title\}
        
        \item \textcolor{red}{source}: \texttt{TEXT}: \{title\_description\} \texttt{QUERY}: Come up with a relevant question based on this text.\\
        \textcolor{blue}{target}: \{title\}
        
        \item \textcolor{red}{source}: \texttt{TEXT}: \{title\_description\} \texttt{QUERY}: Ask a question that is related to the passage.\\
        \textcolor{blue}{target}: \{title\}
        
        \item \textcolor{red}{source}: \texttt{TEXT}: \{title\_description\} \texttt{QUERY}: Based on the passage, can you think of a relevant question?\\
        \textcolor{blue}{target}: \{title\}
        
        \item \textcolor{red}{source}: \texttt{TEXT}: \{title\_description\} \texttt{QUERY}: What would be a question worth asking after reading this paragraph?\\
        \textcolor{blue}{target}: \{title\}
        
        \item \textcolor{red}{source}: \texttt{TEXT}: \{title\_description\} \texttt{QUERY}: Is ``\{title\}" a relevant question to the above text?\\
        \textcolor{blue}{target}: Yes
        
        \item \textcolor{red}{source}: \texttt{TEXT}: \{title\_description\} \texttt{QUERY}: Is ``\{not\_related\_question\}" a relevant question to the above text?\\
        \textcolor{blue}{target}: No
        
        \item \textcolor{red}{source}: \texttt{TEXT}: \{title\_description\} \texttt{QUERY}: Can you ask a question based on the previous context?\\
        \textcolor{blue}{target}: \{title\}
        
        \item \textcolor{red}{source}: \texttt{TEXT}: \{title\_description\} \texttt{QUERY}: Can the above text be used as a description of the ``\{title\}" problem?\\
        \textcolor{blue}{target}: Yes
        
        \item \textcolor{red}{source}: \texttt{TEXT}: \{title\_description\} \texttt{QUERY}: Can the above text be used as a description of the ``\{not\_related\_question\}" problem?\\
        \textcolor{blue}{target}: No
        
        \item \textcolor{red}{source}: \texttt{TEXT}: \{title\_description\} \texttt{QUERY}: After reading this passage, what question might a person ask?\\
        \textcolor{blue}{target}: \{title\}
        
        \item \textcolor{red}{source}: \texttt{TEXT}: \{title\_description\} \texttt{QUERY}: Write down a question that is closely related to the context before.\\
        \textcolor{blue}{target}: \{title\}
        
        \item \textcolor{red}{source}: \texttt{TEXT}: \{title\_description\} \texttt{QUERY}: What question might this text be a problem description for?\\
        \textcolor{blue}{target}: \{title\}
        
        \item \textcolor{red}{source}: \texttt{TEXT}: \{title\_description\} \texttt{QUERY}: Based on this text, generate a relevant question.\\
        \textcolor{blue}{target}: \{title\}
        
        \item \textcolor{red}{source}: \texttt{TEXT}: \{title\_description\} \texttt{QUERY}: Ask a question about this text.\\
        \textcolor{blue}{target}: \{title\}
        
        \item \textcolor{red}{source}: \texttt{TEXT}: \{title\_description\} \texttt{QUERY}: Write down a question related to the paragraph.\\
        \textcolor{blue}{target}: \{title\}
        
        \item \textcolor{red}{source}: \texttt{TEXT}: \{title\_description\} \texttt{QUERY}: What question you may ask about this paragraph?\\
        \textcolor{blue}{target}: \{title\}
        
        \item \textcolor{red}{source}: \texttt{TEXT}: \{title\_description\} \texttt{QUERY}: Read the above text and ask a question related to the passage.\\
        \textcolor{blue}{target}: \{title\}
        
        \item \textcolor{red}{source}: \texttt{TEXT}: \{title\_description\} \texttt{QUERY}: Read the previous text and come up with a question that one could ask.\\
        \textcolor{blue}{target}: \{title\}
    \end{enumerate}
\end{itemize}

With $($answer, title$)$ pairs, we construct prompts that ask for the question based on the answer. Here, we use \{questions\_with\_or\} (e.g. ``option1", ``option2", or ``option3") and \{questions\_without\_or\} (e.g. ``option1", ``option2", ``option3") to refer to some (3 to 9) available questions (including the title and other irrelevant questions). We use \{not\_related\_question\} to represent a random question that is not the \{title\} nor its related questions.

\begin{itemize}
    \item Multiple-choice format prompts
    \begin{enumerate}
        \item \textcolor{red}{source}: \texttt{TEXT}: \{answer\} \texttt{QUERY}: The previous text is most likely to answer which of the following questions? \{questions\_with\_or\}?\\
        \textcolor{blue}{target}: \{title\}
        
        \item \textcolor{red}{source}: \texttt{TEXT}: \{answer\} \texttt{QUERY}: The above text is an answer to a question. What might that question be? Please select from the list below: \{questions\_without\_or\}.\\
        \textcolor{blue}{target}: \{title\}
        
        \item \textcolor{red}{source}: \texttt{TEXT}: \{answer\} \texttt{QUERY}: Given these questions: \{questions\_without\_or\}, which one can be answered using the previous text?\\
        \textcolor{blue}{target}: \{title\}
        
        \item \textcolor{red}{source}: \texttt{TEXT}: \{answer\} \texttt{QUERY}: What question can be answered by the previous text? You may choose from: \{questions\_without\_or\}.\\
        \textcolor{blue}{target}: \{title\}
        
        \item \textcolor{red}{source}: \texttt{TEXT}: \{answer\} \texttt{QUERY}: Based on the previous answer, can you choose the corresponding question from: \{questions\_without\_or\}?\\
        \textcolor{blue}{target}: \{title\}
        
        \item \textcolor{red}{source}: \texttt{TEXT}: \{answer\} \texttt{QUERY}: The above text is the answer to one question, please select the question from the following list: \{questions\_without\_or\}.\\
        \textcolor{blue}{target}: \{title\}
        
        \item \textcolor{red}{source}: \texttt{TEXT}: \{answer\} \texttt{QUERY}: Which of the following questions could be answered by the above text? \{questions\_with\_or\}?\\
        \textcolor{blue}{target}: \{title\}
        
        \item \textcolor{red}{source}: \texttt{TEXT}: \{answer\} \texttt{QUERY}: Which of the following questions might you use the preceding text in answering? The options are: \{questions\_without\_or\}.\\
        \textcolor{blue}{target}: \{title\}
        
        \item \textcolor{red}{source}: \texttt{TEXT}: \{answer\} \texttt{QUERY}: If a person searches for ``\{title\}", is he/she likely to get the above text as an answer? ``Yes" or ``No"?\\
        \textcolor{blue}{target}: Yes
    
        \item \textcolor{red}{source}: \texttt{TEXT}: \{answer\} \texttt{QUERY}: If a person searches for ``\{not\_related\_question\}", is he/she likely to get the above text as an answer? ``Yes" or ``No"?\\
        \textcolor{blue}{target}: No
        
        \item \textcolor{red}{source}: \texttt{TEXT}: \{answer\} \texttt{QUERY}: The text can answer the question ``\{title\}". ``True" or ``False"?\\
        \textcolor{blue}{target}: True
        
        \item \textcolor{red}{source}: \texttt{TEXT}: \{answer\} \texttt{QUERY}: The text can answer the question ``\{not\_related\_question\}". ``True" or ``False"?\\
        \textcolor{blue}{target}: False
        
        \item \textcolor{red}{source}: \texttt{TEXT}: \{answer\} \texttt{QUERY}: Which of the following questions does the above answer address? \{questions\_with\_or\}?\\
        \textcolor{blue}{target}: \{title\}
        
        \item \textcolor{red}{source}: \texttt{TEXT}: \{answer\} \texttt{QUERY}: Which question are you most likely to get the above answer to in your search? \{questions\_with\_or\}?\\
        \textcolor{blue}{target}: \{title\}
        
        \item \textcolor{red}{source}: \texttt{TEXT}: \{answer\} \texttt{QUERY}: Which question are you most likely to get the above answer to when you ask it? \{questions\_with\_or\}?\\
        \textcolor{blue}{target}: \{title\}
        
        \item \textcolor{red}{source}: \texttt{TEXT}: \{answer\} \texttt{QUERY}: Given the questions: \{questions\_without\_or\}, which one does the previous text address?\\
        \textcolor{blue}{target}: \{title\}
        
        \item \textcolor{red}{source}: \texttt{TEXT}: \{answer\} \texttt{QUERY}: Does the above text answer the question ``\{title\}"? ``Yes" or ``No"?\\
        \textcolor{blue}{target}: Yes
        
        \item \textcolor{red}{source}: \texttt{TEXT}: \{answer\} \texttt{QUERY}: Does the above text answer the question ``\{not\_related\_question\}"? ``Yes" or ``No"?\\
        \textcolor{blue}{target}: No
        
        \item \textcolor{red}{source}: \texttt{TEXT}: \{answer\} \texttt{QUERY}: Here is a question list: \{question\_without\_or\}, select one that can be answered by the previous text.\\
        \textcolor{blue}{target}: \{title\}
        
        \item \textcolor{red}{source}: \texttt{TEXT}: \{answer\} \texttt{QUERY}: Which of the following questions is most likely to be answered using the previous text? \{questions\_with\_or\}?\\
        \textcolor{blue}{target}: \{title\}
        
    \end{enumerate}
    \item Generation format prompts
    
    \begin{enumerate}
        \item \textcolor{red}{source}: \texttt{TEXT}: \{answer\} \texttt{QUERY}: The previous text is most likely to answer what question?\\
        \textcolor{blue}{target}: \{title\}
        
        \item \textcolor{red}{source}: \texttt{TEXT}: \{answer\} \texttt{QUERY}: The above text is an answer to a question. What might that question be?\\
        \textcolor{blue}{target}: \{title\}
        
        \item \textcolor{red}{source}: \texttt{TEXT}: \{answer\} \texttt{QUERY}: What question can be answered using the previous text?\\
        \textcolor{blue}{target}: \{title\}
        
        \item \textcolor{red}{source}: \texttt{TEXT}: \{answer\} \texttt{QUERY}: Think of a question that can be answered by the previous text.\\
        \textcolor{blue}{target}: \{title\}
        
        \item \textcolor{red}{source}: \texttt{TEXT}: \{answer\} \texttt{QUERY}: Based on the previous answer, can you identify the corresponding question?\\
        \textcolor{blue}{target}: \{title\}
        
        \item \textcolor{red}{source}: \texttt{TEXT}: \{answer\} \texttt{QUERY}: The above text is the answer to one question, what question could it be?\\
        \textcolor{blue}{target}: \{title\}
        
        \item \textcolor{red}{source}: \texttt{TEXT}: \{answer\} \texttt{QUERY}: What question would you answer with the above text?\\
        \textcolor{blue}{target}: \{title\}
        
        \item \textcolor{red}{source}: \texttt{TEXT}: \{answer\} \texttt{QUERY}: What question might you use the preceding text in answering?\\
        \textcolor{blue}{target}: \{title\}
        
        \item \textcolor{red}{source}: \texttt{TEXT}: \{answer\} \texttt{QUERY}: If a person searches for ``\{title\}", is he/she likely to get the above text as an answer?\\
        \textcolor{blue}{target}: Yes
        
        \item \textcolor{red}{source}: \texttt{TEXT}: \{answer\} \texttt{QUERY}: If a person searches for ``\{not\_related\_question\}", is he/she likely to get the above text as an answer?\\
        \textcolor{blue}{target}: No
        
        \item \textcolor{red}{source}: \texttt{TEXT}: \{answer\} \texttt{QUERY}: Can the text answer the question ``\{title\}"?\\
        \textcolor{blue}{target}: Yes
        
        \item \textcolor{red}{source}: \texttt{TEXT}: \{answer\} \texttt{QUERY}: Can the text answer the question ``\{not\_related\_question\}"?\\
        \textcolor{blue}{target}: No
        
        \item \textcolor{red}{source}: \texttt{TEXT}: \{answer\} \texttt{QUERY}: What question does the above answer address?\\
        \textcolor{blue}{target}: \{title\}
        
        \item \textcolor{red}{source}: \texttt{TEXT}: \{answer\} \texttt{QUERY}: What question are you most likely to get the above answer to in your search?\\
        \textcolor{blue}{target}: \{title\}
        
        \item \textcolor{red}{source}: \texttt{TEXT}: \{answer\} \texttt{QUERY}: What question are you most likely to get the above answer to when you ask it?\\
        \textcolor{blue}{target}: \{title\}
        
        \item \textcolor{red}{source}: \texttt{TEXT}: \{answer\} \texttt{QUERY}: Write down a question that can be addressed by the text.\\
        \textcolor{blue}{target}: \{title\}
        
        \item \textcolor{red}{source}: \texttt{TEXT}: \{answer\} \texttt{QUERY}: Does the above text answer the question ``\{title\}"?\\
        \textcolor{blue}{target}: Yes
        
        \item \textcolor{red}{source}: \texttt{TEXT}: \{answer\} \texttt{QUERY}: Does the above text answer the question ``\{not\_related\_question\}"?\\
        \textcolor{blue}{target}: No
        
        \item \textcolor{red}{source}: \texttt{TEXT}: \{answer\} \texttt{QUERY}: Please generate a question that you can use the above text as an answer.\\
        \textcolor{blue}{target}: \{title\}
        
        \item \textcolor{red}{source}: \texttt{TEXT}: \{answer\} \texttt{QUERY}: Please come up with a question with the above text as an answer.\\
        \textcolor{blue}{target}: \{title\}
    \end{enumerate}
\end{itemize}

With $($title, first\_related\_question$)$ pairs, we construct prompts that ask for related questions given a question. Here, we use \{questions\_with\_or\} (e.g. ``option1", ``option2", or ``option3") and \{questions\_without\_or\} (e.g. ``option1", ``option2", ``option3") to refer to some (3 to 9) available questions (including the \{first\_related\_question\} and other irrelevant questions).

\begin{itemize}
    \item Multiple-choice format prompts
    \begin{enumerate}
        \item \textcolor{red}{source}: \texttt{QUERY}: Given the question ``\{title\}", what could be the next question to ask? \{questions\_with\_or\}?\\
        \textcolor{blue}{target}: \{first\_related\_question\}
        
        \item \textcolor{red}{source}: \texttt{QUERY}: Given a list of questions: \{questions\_without\_or\}, can you select one that is relevant to ``\{title\}"?\\
        \textcolor{blue}{target}: \{first\_related\_question\}
        
        \item \textcolor{red}{source}: \texttt{QUERY}: Can you select a related question for ``\{title\}" from the following list: \{questions\_without\_or\}?\\
        \textcolor{blue}{target}: \{first\_related\_question\}
        
        \item \textcolor{red}{source}: \texttt{QUERY}: Here is a question list: \{questions\_without\_or\}. Which one is most similar to ``\{title\}"?\\
        \textcolor{blue}{target}: \{first\_related\_question\}
        
        \item \textcolor{red}{source}: \texttt{QUERY}: Can you choose the most relevant question to ``\{title\}" from the following list: \{questions\_without\_or\}?\\
        \textcolor{blue}{target}: \{first\_related\_question\}
        
        \item \textcolor{red}{source}: \texttt{QUERY}: If you are interested in the question ``\{title\}", which of the following topics may also be of interest to you? \{questions\_with\_or\}?\\
        \textcolor{blue}{target}: \{first\_related\_question\}
        
        \item \textcolor{red}{source}: \texttt{QUERY}: If a person searches for ``\{title\}", which of the following questions is he/she likely to search for? \{questions\_with\_or\}?\\
        \textcolor{blue}{target}: \{first\_related\_question\}
        
        \item \textcolor{red}{source}: \texttt{QUERY}: Which of the following is most relevant to the question ``\{title\}"? \{questions\_with\_or\}?\\
        \textcolor{blue}{target}: \{first\_related\_question\}
        
        \item \textcolor{red}{source}: \texttt{QUERY}: Ask a question related to ``\{title\}", you can choose from the following list of questions: \{questions\_without\_or\}.\\
        \textcolor{blue}{target}: \{first\_related\_question\}
        
        \item \textcolor{red}{source}: \texttt{QUERY}: After you ask the question ``\{title\}", what other question might you ask? \{questions\_with\_or\}?\\
        \textcolor{blue}{target}: \{first\_related\_question\}
        
        \item \textcolor{red}{source}: \texttt{QUERY}: Given the following list of questions: \{questions\_without\_or\}, which of these is most relevant to ``\{title\}"?\\
        \textcolor{blue}{target}: \{first\_related\_question\}
        
        \item \textcolor{red}{source}: \texttt{QUERY}: Those who are interested in ``\{title\}" are most likely to be interested in which of the following questions as well? \{questions\_with\_or\}?\\
        \textcolor{blue}{target}: \{first\_related\_question\}
        
        \item \textcolor{red}{source}: \texttt{QUERY}: If a person searches for ``\{title\}", which of the following questions would you continue to recommend to him or her? \{questions\_with\_or\}?\\
        \textcolor{blue}{target}: \{first\_related\_question\}
        
        \item \textcolor{red}{source}: \texttt{QUERY}: People who are concerned about ``\{title\}" are more likely to be concerned about which of the following questions? \{questions\_with\_or\}?\\
        \textcolor{blue}{target}: \{first\_related\_question\}
        
        \item \textcolor{red}{source}: \texttt{QUERY}: Given the question list: \{questions\_without\_or\}. Which of the above questions is most similar to ``\{title\}"?\\
        \textcolor{blue}{target}: \{first\_related\_question\}
        
        \item \textcolor{red}{source}: \texttt{QUERY}: Identify which of the following questions is most similar to ``\{title\}": \{questions\_without\_or\}.\\
        \textcolor{blue}{target}: \{first\_related\_question\}
        
        \item \textcolor{red}{source}: \texttt{QUERY}: Can you suggest topics for people who are interested in ``\{title\}"? You may choose from: \{questions\_without\_or\}.\\
        \textcolor{blue}{target}: \{first\_related\_question\}
        
        \item \textcolor{red}{source}: \texttt{QUERY}: When a person asks ``\{title\}", which question is he or she more likely to ask? \{questions\_with\_or\}?\\
        \textcolor{blue}{target}: \{first\_related\_question\}
        
        \item \textcolor{red}{source}: \texttt{QUERY}: Which of the following questions is likely to appeal to the same group of people as ``\{title\}"? \{questions\_with\_or\}?\\
        \textcolor{blue}{target}: \{first\_related\_question\}
        
        \item \textcolor{red}{source}: \texttt{QUERY}: Which of the following questions has a similar audience to ``\{title\}"? \{questions\_with\_or\}?\\
        \textcolor{blue}{target}: \{first\_related\_question\}
    \end{enumerate}
    \item Generation format prompts
    \begin{enumerate}
        \item \textcolor{red}{source}: \texttt{QUERY}: Given the question ``\{title\}", what could be the next question to ask?\\
        \textcolor{blue}{target}: \{first\_related\_question\}
        
        \item \textcolor{red}{source}: \texttt{QUERY}: Write down a question that is relevant to ``\{title\}".\\
        \textcolor{blue}{target}: \{first\_related\_question\}
        
        \item \textcolor{red}{source}: \texttt{QUERY}: Can you think of a related question for ``\{title\}"?\\
        \textcolor{blue}{target}: \{first\_related\_question\}
        
        \item \textcolor{red}{source}: \texttt{QUERY}: Can you write a question that is similar to ``\{title\}"?\\
        \textcolor{blue}{target}: \{first\_related\_question\}
        
        \item \textcolor{red}{source}: \texttt{QUERY}: What could be a relevant question to ``\{title\}"?\\
        \textcolor{blue}{target}: \{first\_related\_question\}
        
        \item \textcolor{red}{source}: \texttt{QUERY}: If you are interested in the question ``\{title\}", what question may also be of interest to you?\\
        \textcolor{blue}{target}: \{first\_related\_question\}
        
        \item \textcolor{red}{source}: \texttt{QUERY}: If a person searches for ``\{title\}", what other question is he/she likely to search for?\\
        \textcolor{blue}{target}: \{first\_related\_question\}
        
        \item \textcolor{red}{source}: \texttt{QUERY}: Come up with a question that is most relevant to ``\{title\}".\\
        \textcolor{blue}{target}: \{first\_related\_question\}
        
        \item \textcolor{red}{source}: \texttt{QUERY}: Ask a question related to ``\{title\}".\\
        \textcolor{blue}{target}: \{first\_related\_question\}
        
        \item \textcolor{red}{source}: \texttt{QUERY}: After you ask the question ``\{title\}", what other question might you ask?\\
        \textcolor{blue}{target}: \{first\_related\_question\}
        
        \item \textcolor{red}{source}: \texttt{QUERY}: Please suggest a question related to ``\{title\}".\\
        \textcolor{blue}{target}: \{first\_related\_question\}
        
        \item \textcolor{red}{source}: \texttt{QUERY}: What questions would people interested in ``\{title\}" also be interested in?\\
        \textcolor{blue}{target}: \{first\_related\_question\}
        
        \item \textcolor{red}{source}: \texttt{QUERY}: If a person searches for ``\{title\}", what question would you continue to recommend to him or her?\\
        \textcolor{blue}{target}: \{first\_related\_question\}
        
        \item \textcolor{red}{source}: \texttt{QUERY}: People who are concerned about ``\{title\}" are likely to be concerned about what question as well?\\
        \textcolor{blue}{target}: \{first\_related\_question\}
        
        \item \textcolor{red}{source}: \texttt{QUERY}: Identify a question that is very similar to ``\{title\}".\\
        \textcolor{blue}{target}: \{first\_related\_question\}
        
        \item \textcolor{red}{source}: \texttt{QUERY}: Is there a question similar to ``\{title\}"?\\
        \textcolor{blue}{target}: \{first\_related\_question\}
        
        \item \textcolor{red}{source}: \texttt{QUERY}: Can you suggest another question for people who are interested in ``\{title\}"?\\
        \textcolor{blue}{target}: \{first\_related\_question\}
        
        \item \textcolor{red}{source}: \texttt{QUERY}: When a person asks ``\{title\}", what question is he or she more likely to ask as well?\\
        \textcolor{blue}{target}: \{first\_related\_question\}
        
        \item \textcolor{red}{source}: \texttt{QUERY}: What question is likely to appeal to the same group of people as ``\{title\}"? \texttt{QUERY}: What question is likely to appeal to the same group of people as ``\{title\}"?\\
        \textcolor{blue}{target}: \{first\_related\_question\}
        
        \item \textcolor{red}{source}: \texttt{QUERY}: What question has a similar audience to ``\{title\}"?\\
        \textcolor{blue}{target}: \{first\_related\_question\}
    \end{enumerate}
\end{itemize}

\subsection{Wikipedia}
\paragraph{Section title}
With $($section\_text, section\_title$)$ pairs, we construct prompts that ask for the appropriate title given a piece of text. Here, we use \{section\_titles\_with\_or\} (e.g. ``option1", ``option2", or ``option3") and \{section\_titles\_without\_or\} (e.g. ``option1", ``option2", ``option3") to refer to some (3 to 9) available titles (including the correct one and other section titles within the same Wikipedia article). We use \{other\_section\_title\} to refer to a random section title that is different from the target title within the same Wikipedia article.

\begin{itemize}
    \item Multiple-choice format prompts
    \begin{enumerate}
        \item \textcolor{red}{source}: \texttt{TEXT}: \{section\_text\} \texttt{QUERY}: What's this text about? \{section\_titles\_with\_or\}?\\
        \textcolor{blue}{target}: \{section\_title\}
        
        \item \textcolor{red}{source}: \texttt{TEXT}: \{section\_text\} \texttt{QUERY}: Classify this text. You may choose from \{section\_titles\_without\_or\}.\\
        \textcolor{blue}{target}: \{section\_title\}
        
        \item \textcolor{red}{source}: \texttt{TEXT}: \{section\_text\} \texttt{QUERY}: How would you categorize this text? \{section\_titles\_with\_or\}?\\
        \textcolor{blue}{target}: \{section\_title\}
        
        \item \textcolor{red}{source}: \texttt{TEXT}: \{section\_text\} \texttt{QUERY}: Is this text about ``\{other\_section\_title\}"? ``Yes" or ``No"?\\
        \textcolor{blue}{target}: No
        
        \item \textcolor{red}{source}: \texttt{TEXT}: \{section\_text\} \texttt{QUERY}: Is this text about ``\{section\_title\}"? ``Yes" or ``No"?\\
        \textcolor{blue}{target}: Yes
        
        \item \textcolor{red}{source}: \texttt{TEXT}: \{section\_text\} \texttt{QUERY}: Can you choose an appropriate class from the following list for this text? \{section\_titles\_without\_or\}.\\
        \textcolor{blue}{target}: \{section\_title\}
        
        \item \textcolor{red}{source}: \texttt{TEXT}: \{section\_text\} \texttt{QUERY}: Given a list of categories: \{section\_titles\_without\_or\}, what category does the paragraph belong to?\\
        \textcolor{blue}{target}: \{section\_title\}
        
        \item \textcolor{red}{source}: \texttt{TEXT}: \{section\_text\} \texttt{QUERY}: Is this paragraph related to ``\{other\_section\_title\}"? ``Yes" or ``No"?\\
        \textcolor{blue}{target}: No
        
        \item \textcolor{red}{source}: \texttt{TEXT}: \{section\_text\} \texttt{QUERY}: Is this paragraph related to ``\{section\_title\}"? ``Yes" or ``No"?\\
        \textcolor{blue}{target}: Yes
        
        \item \textcolor{red}{source}: \texttt{TEXT}: \{section\_text\} \texttt{QUERY}: Pick one category for the previous text. The options are \{section\_titles\_without\_or\}.\\
        \textcolor{blue}{target}: \{section\_title\}
        
        \item \textcolor{red}{source}: \texttt{TEXT}: \{section\_text\} \texttt{QUERY}: Given a choice of categories: \{section\_titles\_with\_or\}, the text refers to which one?\\
        \textcolor{blue}{target}: \{section\_title\}
        
        \item \textcolor{red}{source}: \texttt{TEXT}: \{section\_text\} \texttt{QUERY}: What is the topic of the text? \{section\_titles\_with\_or\}?\\
        \textcolor{blue}{target}: \{section\_title\}
        
        \item \textcolor{red}{source}: \texttt{TEXT}: \{section\_text\} \texttt{QUERY}: The topic of this text is ``\{other\_section\_title\}". ``True" or ``False"?\\
        \textcolor{blue}{target}: False
        
        \item \textcolor{red}{source}: \texttt{TEXT}: \{section\_text\} \texttt{QUERY}: The topic of this text is ``\{section\_title\}". ``True" or ``False"?\\
        \textcolor{blue}{target}: True
        
        \item \textcolor{red}{source}: \texttt{TEXT}: \{section\_text\} \texttt{QUERY}: Can you identify the category of this text? \{section\_titles\_with\_or\}?\\
        \textcolor{blue}{target}: \{section\_title\}
        
        \item \textcolor{red}{source}: \texttt{TEXT}: \{section\_text\} \texttt{QUERY}: Select a class from the following that best describes the text: \{section\_titles\_without\_or\}.\\
        \textcolor{blue}{target}: \{section\_title\}
        
        \item \textcolor{red}{source}: \texttt{TEXT}: \{section\_text\} \texttt{QUERY}: Is this a piece of text regarding \{section\_titles\_with\_or\}?\\
        \textcolor{blue}{target}: \{section\_title\}
        
        \item \textcolor{red}{source}: \texttt{TEXT}: \{section\_text\} \texttt{QUERY}: What category best describes this paragraph? \{section\_titles\_with\_or\}?\\
        \textcolor{blue}{target}: \{section\_title\}
        
        \item \textcolor{red}{source}: \texttt{TEXT}: \{section\_text\} \texttt{QUERY}: Please classify this text into one of the following: \{section\_titles\_without\_or\}.\\
        \textcolor{blue}{target}: \{section\_title\}
        
        \item \textcolor{red}{source}: \texttt{TEXT}: \{section\_text\} \texttt{QUERY}: What's the main topic of this paragraph? \{section\_titles\_with\_or\}?\\
        \textcolor{blue}{target}: \{section\_title\}
    \end{enumerate}
    \item Generation format prompts
    
    \begin{enumerate}
        \item \textcolor{red}{source}: \texttt{TEXT}: \{section\_text\} \texttt{QUERY}: What's this text about?\\
        \textcolor{blue}{target}: \{section\_title\}
        
        \item \textcolor{red}{source}: \texttt{TEXT}: \{section\_text\} \texttt{QUERY}: Please give a heading for this text.\\
        \textcolor{blue}{target}: \{section\_title\}
        
        \item \textcolor{red}{source}: \texttt{TEXT}: \{section\_text\} \texttt{QUERY}: Is ``\{section\_title\}" an appropriate heading for this text?\\
        \textcolor{blue}{target}: Yes
        
        \item \textcolor{red}{source}: \texttt{TEXT}: \{section\_text\} \texttt{QUERY}: Is this text about ``\{other\_section\_title\}"?\\
        \textcolor{blue}{target}: No
        
        \item \textcolor{red}{source}: \texttt{TEXT}: \{section\_text\} \texttt{QUERY}: Is this text about ``\{section\_title\}"?\\
        \textcolor{blue}{target}: Yes
        
        \item \textcolor{red}{source}: \texttt{TEXT}: \{section\_text\} \texttt{QUERY}: Can you find an appropriate title for this text?\\
        \textcolor{blue}{target}: \{section\_title\}
        
        \item \textcolor{red}{source}: \texttt{TEXT}: \{section\_text\} \texttt{QUERY}: Is ``\{other\_section\_title\}" an appropriate heading for this text?\\
        \textcolor{blue}{target}: No
        
        \item \textcolor{red}{source}: \texttt{TEXT}: \{section\_text\} \texttt{QUERY}: Is this paragraph related to ``\{other\_section\_title\}"?\\
        \textcolor{blue}{target}: No
        
        \item \textcolor{red}{source}: \texttt{TEXT}: \{section\_text\} \texttt{QUERY}: Is this paragraph related to ``\{section\_title\}"?\\
        \textcolor{blue}{target}: Yes
        
        \item \textcolor{red}{source}: \texttt{TEXT}: \{section\_text\} \texttt{QUERY}: Pick one title for the previous text.\\
        \textcolor{blue}{target}: \{section\_title\}
        
        \item \textcolor{red}{source}: \texttt{TEXT}: \{section\_text\} \texttt{QUERY}: Generate a title for this article.\\
        \textcolor{blue}{target}: \{section\_title\}
        
        \item \textcolor{red}{source}: \texttt{TEXT}: \{section\_text\} \texttt{QUERY}: What is the topic of the text?\\
        \textcolor{blue}{target}: \{section\_title\}
        
        \item \textcolor{red}{source}: \texttt{TEXT}: \{section\_text\} \texttt{QUERY}: Is the topic of this text ``\{other\_section\_title\}"?\\
        \textcolor{blue}{target}: No
        
        \item \textcolor{red}{source}: \texttt{TEXT}: \{section\_text\} \texttt{QUERY}: Is the topic of this text ``\{section\_title\}"?\\
        \textcolor{blue}{target}: Yes
        
        \item \textcolor{red}{source}: \texttt{TEXT}: \{section\_text\} \texttt{QUERY}: What can be an appropriate title for the text?\\
        \textcolor{blue}{target}: \{section\_title\}
        
        \item \textcolor{red}{source}: \texttt{TEXT}: \{section\_text\} \texttt{QUERY}: Think of a title that best fits the text.\\
        \textcolor{blue}{target}: \{section\_title\}
        
        \item \textcolor{red}{source}: \texttt{TEXT}: \{section\_text\} \texttt{QUERY}: Under what short title is this text most likely to appear?\\
        \textcolor{blue}{target}: \{section\_title\}
        
        \item \textcolor{red}{source}: \texttt{TEXT}: \{section\_text\} \texttt{QUERY}: What title best sums up the text?\\
        \textcolor{blue}{target}: \{section\_title\}
        
        \item \textcolor{red}{source}: \texttt{TEXT}: \{section\_text\} \texttt{QUERY}: Please write a heading for this text.\\
        \textcolor{blue}{target}: \{section\_title\}
    \end{enumerate}
\end{itemize}

\paragraph{Entity}
With $($paragraph, entities$)$ pairs, we can construct prompts that ask for the entities given a paragraph. Here, \{entities\} is a textual string that contains all the entities within the paragraph, listed according to their order of appearance and separated by a comma (can be an empty string as well). We use \{entity\} to represent one random entity out of \{entities\} and use \{not\_entity\} to represent a random text span within this paragraph that is not an entity. We use \{choice0\} and \{choice1\} to represent a random order of an entity and a non-entity. Besides, we use \{answer\} to represent the target based on whether there are named entities in the text.

\begin{itemize}
    \item Multiple-choice format prompts
    \begin{enumerate}
        \item \textcolor{red}{source}: \texttt{TEXT}: \{paragraph\} \texttt{QUERY}: Is ``\{entity\}" a named entity? ``Yes" or ``No"?\\
        \textcolor{blue}{target}: Yes
        
        \item \textcolor{red}{source}: \texttt{TEXT}: \{paragraph\} \texttt{QUERY}: Is ``\{not\_entity\}" a named entity? ``Yes" or ``No"?\\
        \textcolor{blue}{target}: No
        
        \item \textcolor{red}{source}: \texttt{TEXT}: \{paragraph\} \texttt{QUERY}: In the previous text, ``\{entity\}" is a named entity. ``True" or ``False"?\\
        \textcolor{blue}{target}: True
        
        \item \textcolor{red}{source}: \texttt{TEXT}: \{paragraph\} \texttt{QUERY}: In the previous text, ``\{not\_entity\}" is a named entity. ``True" or ``False"?\\
        \textcolor{blue}{target}: False
        
        \item \textcolor{red}{source}: \texttt{TEXT}: \{paragraph\} \texttt{QUERY}: There are named entities in the previous text. ``True" or ``False"?\\
        \textcolor{blue}{target}: \{answer\}
        
        \item \textcolor{red}{source}: \texttt{TEXT}: \{paragraph\} \texttt{QUERY}: There are no named entities in the preceding text. ``True" or ``False"?\\
        \textcolor{blue}{target}: \{answer\}
        
        \item \textcolor{red}{source}: \texttt{TEXT}: \{paragraph\} \texttt{QUERY}: In the text, does ``\{entity\}" refer to real-world object? ``Yes" or ``No"?\\
        \textcolor{blue}{target}: Yes
        
        \item \textcolor{red}{source}: \texttt{TEXT}: \{paragraph\} \texttt{QUERY}: In the text, does ``\{not\_entity\}" refer to real-world object? ``Yes" or ``No"?\\
        \textcolor{blue}{target}: No
        
        \item \textcolor{red}{source}: \texttt{TEXT}: \{paragraph\} \texttt{QUERY}: Are there named entities in the text? ``Yes" or ``No"?\\
        \textcolor{blue}{target}: \{answer\}
        
        \item \textcolor{red}{source}: \texttt{TEXT}: \{paragraph\} \texttt{QUERY}: Is there any named entity in the paragraph? ``Yes" or ``No"?\\
        \textcolor{blue}{target}: \{answer\}
        
        \item \textcolor{red}{source}: \texttt{TEXT}: \{paragraph\} \texttt{QUERY}: In the previous text, which of the following is a named entity? ``\{choice0\}" or ``\{choice1\}"?\\
        \textcolor{blue}{target}: \{entity\}
        
        \item \textcolor{red}{source}: \texttt{TEXT}: \{paragraph\} \texttt{QUERY}: Which one of the following is a named entity? ``\{choice0\}" or ``\{choice1\}"?\\
        \textcolor{blue}{target}: \{entity\}
        
        \item \textcolor{red}{source}: \texttt{TEXT}: \{paragraph\} \texttt{QUERY}: ``\{entity\}" is not a named entity. ``True" or ``False"?\\
        \textcolor{blue}{target}: False
        
        \item \textcolor{red}{source}: \texttt{TEXT}: \{paragraph\} \texttt{QUERY}: ``\{not\_entity\}" is not a named entity. ``True" or ``False"?\\
        \textcolor{blue}{target}: True
        
        \item \textcolor{red}{source}: \texttt{TEXT}: \{paragraph\} \texttt{QUERY}: In the previous text, does ``\{entity\}" count as a named entity? ``Yes" or ``No"?\\
        \textcolor{blue}{target}: Yes
        
        \item \textcolor{red}{source}: \texttt{TEXT}: \{paragraph\} \texttt{QUERY}: In the previous text, does ``\{not\_entity\}" count as a named entity? ``Yes" or ``No"?\\
        \textcolor{blue}{target}: No
        
        \item \textcolor{red}{source}: \texttt{TEXT}: \{paragraph\} \texttt{QUERY}: In the text, does ``\{entity\}" consider a named entity? ``Yes" or ``No"?\\
        \textcolor{blue}{target}: Yes
        
        \item \textcolor{red}{source}: \texttt{TEXT}: \{paragraph\} \texttt{QUERY}: In the text, does ``\{not\_entity\}" consider a named entity? ``Yes" or ``No"?\\
        \textcolor{blue}{target}: No
        
        \item \textcolor{red}{source}: \texttt{TEXT}: \{paragraph\} \texttt{QUERY}: Does ``\{entity\}" refer to a specific object? ``Yes" or ``No"?\\
        \textcolor{blue}{target}: Yes
        
        \item \textcolor{red}{source}: \texttt{TEXT}: \{paragraph\} \texttt{QUERY}: Does ``\{not\_entity\}" refer to a specific object? ``Yes" or ``No"?\\
        \textcolor{blue}{target}: No
        
    \end{enumerate}
    
    \item Generation format prompts
    \begin{enumerate}
        \item \textcolor{red}{source}: \texttt{TEXT}: \{paragraph\} \texttt{QUERY}: Can you find all the entities in the text?\\
        \textcolor{blue}{target}: \{entities\}
        
        \item \textcolor{red}{source}: \texttt{TEXT}: \{paragraph\} \texttt{QUERY}: List all the entities in the text.\\
        \textcolor{blue}{target}: \{entities\}
        
        \item \textcolor{red}{source}: \texttt{TEXT}: \{paragraph\} \texttt{QUERY}: What are the entities in the previous text?\\
        \textcolor{blue}{target}: \{entities\}
        
        \item \textcolor{red}{source}: \texttt{TEXT}: \{paragraph\} \texttt{QUERY}: Please enumerate all the named entities in the above paragraph.\\
        \textcolor{blue}{target}: \{entities\}
        
        \item \textcolor{red}{source}: \texttt{TEXT}: \{paragraph\} \texttt{QUERY}: Identify all the named entities in the previous paragraph.\\
        \textcolor{blue}{target}: \{entities\}
        
        \item \textcolor{red}{source}: \texttt{TEXT}: \{paragraph\} \texttt{QUERY}: What text spans in the previous text refer to specific real-world objects?\\
        \textcolor{blue}{target}: \{entities\}
        
        \item \textcolor{red}{source}: \texttt{TEXT}: \{paragraph\} \texttt{QUERY}: Please list the named entities in the paragraph in the order in which they appear.\\
        \textcolor{blue}{target}: \{entities\}
        
        \item \textcolor{red}{source}: \texttt{TEXT}: \{paragraph\} \texttt{QUERY}: Is ``\{entity\}" a named entity?\\
        \textcolor{blue}{target}: Yes
        
        \item \textcolor{red}{source}: \texttt{TEXT}: \{paragraph\} \texttt{QUERY}: Is ``\{not\_entity\}" a named entity?\\
        \textcolor{blue}{target}: No
        
        \item \textcolor{red}{source}: \texttt{TEXT}: \{paragraph\} \texttt{QUERY}: Does ``\{entity\}" refer to a real-world object?\\
        \textcolor{blue}{target}: Yes
        
        \item \textcolor{red}{source}: \texttt{TEXT}: \{paragraph\} \texttt{QUERY}: Does ``\{not\_entity\}" refer to a real-world object?\\
        \textcolor{blue}{target}: No
        
        \item \textcolor{red}{source}: \texttt{TEXT}: \{paragraph\} \texttt{QUERY}: List all text spans that refer to specific objects.\\
        \textcolor{blue}{target}: \{entities\}
        
        \item \textcolor{red}{source}: \texttt{TEXT}: \{paragraph\} \texttt{QUERY}: Write down all the named entities that appeared in the previous text.\\
        \textcolor{blue}{target}: \{entities\}
        
        \item \textcolor{red}{source}: \texttt{TEXT}: \{paragraph\} \texttt{QUERY}: Can you write out all the named entities that appear in the text?\\
        \textcolor{blue}{target}: \{entities\}
        
        \item \textcolor{red}{source}: \texttt{TEXT}: \{paragraph\} \texttt{QUERY}: Are there named entities in the text?\\
        \textcolor{blue}{target}: \{answer\}
        
        \item \textcolor{red}{source}: \texttt{TEXT}: \{paragraph\} \texttt{QUERY}: What are the named entities in the text?\\
        \textcolor{blue}{target}: \{entities\}
        
        \item \textcolor{red}{source}: \texttt{TEXT}: \{paragraph\} \texttt{QUERY}: Can you list every named entity that appears in the previous paragraph?\\
        \textcolor{blue}{target}: \{entities\}
        
        \item \textcolor{red}{source}: \texttt{TEXT}: \{paragraph\} \texttt{QUERY}: What named entities are present in the above text?\\
        \textcolor{blue}{target}: \{entities\}
        
        \item \textcolor{red}{source}: \texttt{TEXT}: \{paragraph\} \texttt{QUERY}: Is there any named entity in the paragraph?\\
        \textcolor{blue}{target}: \{answer\}
        
        \item \textcolor{red}{source}: \texttt{TEXT}: \{paragraph\} \texttt{QUERY}: Please read the text and list the named entities contained in it.\\
        \textcolor{blue}{target}: \{entities\}
        
    \end{enumerate}
\end{itemize}

\paragraph{Sentiment}
With $($text, neutral\_sentiment$)$ pairs we can construct prompts that ask for the sentiment of a given text.
\begin{itemize}
    \item Multiple-choice format prompts
    \begin{enumerate}
        \item \textcolor{red}{source}: \texttt{TEXT}: \{text\} \texttt{QUERY}: What's the sentiment of this text? ``Positive", ``Negative" or ``Neutral"?\\
        \textcolor{blue}{target}: Neutral
        
        \item \textcolor{red}{source}: \texttt{TEXT}: \{text\} \texttt{QUERY}: Can you judge the sentiment of this text? The options are ``Positive", ``Negative" and ``Neutral".\\
        \textcolor{blue}{target}: Neutral
        
        \item \textcolor{red}{source}: \texttt{TEXT}: \{text\} \texttt{QUERY}: Assign the correct sentiment to this text. Please choose from ``Positive", ``Negative", ``Neutral".\\
        \textcolor{blue}{target}: Neutral
        
        \item \textcolor{red}{source}: \texttt{TEXT}: \{text\} \texttt{QUERY}: Judge the sentiment of the text. You can choose from ``Positive", ``Negative", ``Neutral".\\
        \textcolor{blue}{target}: Neutral
        
        \item \textcolor{red}{source}: \texttt{TEXT}: \{text\} \texttt{QUERY}: Can you tell the sentiment of the text? ``Positive", ``Negative" or ``Neutral"?\\
        \textcolor{blue}{target}: Neutral
        
        \item \textcolor{red}{source}: \texttt{TEXT}: \{text\} \texttt{QUERY}: Is the text ``positive" or ``negative" or ``neutral"?\\
        \textcolor{blue}{target}: neutral
        
        \item \textcolor{red}{source}: \texttt{TEXT}: \{text\} \texttt{QUERY}: Is the sentiment of this text positive? ``Yes" or ``No"?\\
        \textcolor{blue}{target}: No
        
        \item \textcolor{red}{source}: \texttt{TEXT}: \{text\} \texttt{QUERY}: Is the sentiment of this text neutral? ``Yes" or ``No"?\\
        \textcolor{blue}{target}: Yes
        
        \item \textcolor{red}{source}: \texttt{TEXT}: \{text\} \texttt{QUERY}: Is the sentiment of this text negative? ``Yes" or ``No"?\\
        \textcolor{blue}{target}: No
        
        \item \textcolor{red}{source}: \texttt{TEXT}: \{text\} \texttt{QUERY}: There is positive sentiment within this text. ``True" or ``False"?\\
        \textcolor{blue}{target}: False
        
        \item \textcolor{red}{source}: \texttt{TEXT}: \{text\} \texttt{QUERY}: There is negative sentiment within this text. ``True" or ``False"?\\
        \textcolor{blue}{target}: False
        
        \item \textcolor{red}{source}: \texttt{TEXT}: \{text\} \texttt{QUERY}: The sentiment of this text is neutral. ``True" or ``False"?\\
        \textcolor{blue}{target}: True
        
        \item \textcolor{red}{source}: \texttt{TEXT}: \{text\} \texttt{QUERY}: Please identify the correct sentiment of this text. You may choose from ``Positive", ``Negative", ``Neutral".\\
        \textcolor{blue}{target}: Neutral
        
        \item \textcolor{red}{source}: \texttt{TEXT}: \{text\} \texttt{QUERY}: Given the options: ``Positive", ``Negative", ``Neutral", which one best describes the sentiment of the previous text?\\
        \textcolor{blue}{target}: Neutral
        
        \item \textcolor{red}{source}: \texttt{TEXT}: \{text\} \texttt{QUERY}: With what emotion did the author write this paragraph? ``Positive", ``Negative" or ``Neutral"?\\
        \textcolor{blue}{target}: Neutral
        
        \item \textcolor{red}{source}: \texttt{TEXT}: \{text\} \texttt{QUERY}: Can you decide the sentiment polarity of the paragraph? ``Positive", ``Negative" or ``Neutral"?\\
        \textcolor{blue}{target}: Neutral
        
        \item \textcolor{red}{source}: \texttt{TEXT}: \{text\} \texttt{QUERY}: Can you read the sentiment polarity of the paragraph? ``Positive", ``Negative" or ``Neutral"?\\
        \textcolor{blue}{target}: Neutral
        
        \item \textcolor{red}{source}: \texttt{TEXT}: \{text\} \texttt{QUERY}: Does the paragraph contain ``positive", ``negative" or ``neutral" sentiment?\\
        \textcolor{blue}{target}: neutral
        
        \item \textcolor{red}{source}: \texttt{TEXT}: \{text\} \texttt{QUERY}: This is a neutral text. ``True" or ``False"?\\
        \textcolor{blue}{target}: True
        
        \item \textcolor{red}{source}: \texttt{TEXT}: \{text\} \texttt{QUERY}: What sentiment best describes the previous paragraph? ``Positive", ``Negative" or ``Neutral"?\\
        \textcolor{blue}{target}: Neutral
    \end{enumerate}
    \item Generation format prompts
    \begin{enumerate}
        \item \textcolor{red}{source}: \texttt{TEXT}: \{text\} \texttt{QUERY}: What's the sentiment of this text?\\
        \textcolor{blue}{target}: Neutral
        
        \item \textcolor{red}{source}: \texttt{TEXT}: \{text\} \texttt{QUERY}: Can you judge the sentiment of this text?\\
        \textcolor{blue}{target}: Neutral
        
        \item \textcolor{red}{source}: \texttt{TEXT}: \{text\} \texttt{QUERY}: Does it seem like the author wrote this text with a positive sentiment?\\
        \textcolor{blue}{target}: No
        
        \item \textcolor{red}{source}: \texttt{TEXT}: \{text\} \texttt{QUERY}: Does it seem like the author wrote this text with a neutral sentiment?\\
        \textcolor{blue}{target}: Yes
        
        \item \textcolor{red}{source}: \texttt{TEXT}: \{text\} \texttt{QUERY}: Assign the correct sentiment to this text.\\
        \textcolor{blue}{target}: Neutral
        
        \item \textcolor{red}{source}: \texttt{TEXT}: \{text\} \texttt{QUERY}: With what emotion did the author write this paragraph?\\
        \textcolor{blue}{target}: Neutral
        
        \item \textcolor{red}{source}: \texttt{TEXT}: \{text\} \texttt{QUERY}: Can you read the sentiment polarity of the paragraph?\\
        \textcolor{blue}{target}: Neutral
        
        \item \textcolor{red}{source}: \texttt{TEXT}: \{text\} \texttt{QUERY}: Judge the sentiment of the text.\\
        \textcolor{blue}{target}: Neutral
        
        \item \textcolor{red}{source}: \texttt{TEXT}: \{text\} \texttt{QUERY}: Can you tell the sentiment of the text?\\
        \textcolor{blue}{target}: Neutral
        
        \item \textcolor{red}{source}: \texttt{TEXT}: \{text\} \texttt{QUERY}: Can you decide the sentiment polarity of the text?\\
        \textcolor{blue}{target}: Neutral
        
        \item \textcolor{red}{source}: \texttt{TEXT}: \{text\} \texttt{QUERY}: Is the sentiment of this text positive?\\
        \textcolor{blue}{target}: No
        
        \item \textcolor{red}{source}: \texttt{TEXT}: \{text\} \texttt{QUERY}: Is the sentiment of this text neutral?\\
        \textcolor{blue}{target}: Yes
        
        \item \textcolor{red}{source}: \texttt{TEXT}: \{text\} \texttt{QUERY}: Is the sentiment of this text negative?\\
        \textcolor{blue}{target}: No
        
        \item \textcolor{red}{source}: \texttt{TEXT}: \{text\} \texttt{QUERY}: Does the paragraph contain positive sentiment?\\
        \textcolor{blue}{target}: No
        
        \item \textcolor{red}{source}: \texttt{TEXT}: \{text\} \texttt{QUERY}: Does the paragraph contain negative sentiment?\\
        \textcolor{blue}{target}: No
        
        \item \textcolor{red}{source}: \texttt{TEXT}: \{text\} \texttt{QUERY}: Is there positive sentiment within this text?\\
        \textcolor{blue}{target}: No
        
        \item \textcolor{red}{source}: \texttt{TEXT}: \{text\} \texttt{QUERY}: Is there negative sentiment within this text?\\
        \textcolor{blue}{target}: No
        
        \item \textcolor{red}{source}: \texttt{TEXT}: \{text\} \texttt{QUERY}: Is this a text with neutral sentiment?\\
        \textcolor{blue}{target}: Yes
        
        \item \textcolor{red}{source}: \texttt{TEXT}: \{text\} \texttt{QUERY}: Please identify the correct sentiment of this text.\\
        \textcolor{blue}{target}: Neutral
        
        \item \textcolor{red}{source}: \texttt{TEXT}: \{text\} \texttt{QUERY}: What sentiment best describes the previous text?\\
        \textcolor{blue}{target}: Neutral
    \end{enumerate}
\end{itemize}

\subsection{WordNet}
\paragraph{Part-of-speech}
With $($word, sentence, POS$)$ triples, we construct prompts that ask for the part of speech of a given word. We use \{other\_POS\} to refer to a random POS that is different from \{POS\}.
\begin{itemize}
    \item Multiple-choice format prompts
    \begin{enumerate}
        \item \textcolor{red}{source}: \texttt{TEXT}: \{sentence\} \texttt{QUERY}: In the text, what is the part of speech of the word ``\{word\}"? ``adjective", ``adverb", ``noun" or ``verb"?\\
        \textcolor{blue}{target}: \{POS\}
        
        \item \textcolor{red}{source}: \texttt{TEXT}: \{sentence\} \texttt{QUERY}: What is the part of speech of ``\{word\}" in the previous sentence? "``djective", ``adverb", ``noun" or ``verb"?\\
        \textcolor{blue}{target}: \{POS\}
        
        \item \textcolor{red}{source}: \texttt{TEXT}: \{sentence\} \texttt{QUERY}: Given the following parts of speech: ``adjective", ``adverb", ``noun", ``verb", which one applies to the word ``\{word\}" in the sentence?\\
        \textcolor{blue}{target}: \{POS\}
        
        \item \textcolor{red}{source}: \texttt{TEXT}: \{sentence\} \texttt{QUERY}: Choose the right part of speech for ``\{word\}". The options are ``adjective", ``adverb", ``noun", ``verb".\\
        \textcolor{blue}{target}: {POS}
        
        \item \textcolor{blue}{target}: \texttt{TEXT}: \{sentence\} \texttt{QUERY}: Is ``\{word\}" a/an ``\{POS\}" in the text? ``Yes" or ``No"?\\
        \textcolor{blue}{target}: Yes
        
        \item \textcolor{red}{source}: \texttt{TEXT}: \{sentence\} \texttt{QUERY}: Is ``\{word\}" a/an ``\{other\_POS\}" in the text? ``Yes" or ``No"?\\
        \textcolor{blue}{target}: No
        
        \item \textcolor{red}{source}: \texttt{TEXT}: \{sentence\} \texttt{QUERY}: Pick the right part of speech for ``\{word\}" in the context from the following: ``adjective", ``adverb", ``noun", ``verb".\\
        \textcolor{blue}{target}: \{POS\}
        
        \item \textcolor{red}{source}: \texttt{TEXT}: \{sentence\} \texttt{QUERY}: Given four parts of speech: ``adjective", ``adverb", ``noun", ``verb", what kind of part of speech is ``\{word\}" in the text?\\
        \textcolor{blue}{target}: \{POS\}
        
        \item \textcolor{red}{source}: \texttt{TEXT}: \{sentence\} \texttt{QUERY}: In the text, ``\{word\}" is a/an ``\{POS\}". ``True" or ``False"?\\
        \textcolor{blue}{target}: True
        
        \item \textcolor{red}{source}: \texttt{TEXT}: \{sentence\} \texttt{QUERY}: In the text, ``\{word\}" is a/an ``\{other\_POS\}". ``True" or ``False"?\\
        \textcolor{blue}{target}: False
    \end{enumerate}
    \item Generation format prompts
    \begin{enumerate}
        \item \textcolor{red}{source}: \texttt{TEXT}: \{sentence\} \texttt{QUERY}: In the text, what is the part of speech of the word ``\{word\}"?\\
        \textcolor{blue}{target}: \{POS\}
        
        \item \textcolor{red}{source}: \texttt{TEXT}: \{sentence\} \texttt{QUERY}: What is the part of speech of ``\{word\}" in the previous sentence?\\
        \textcolor{blue}{target}: \{POS\}
        
        \item \textcolor{red}{source}: \texttt{TEXT}: \{sentence\} \texttt{QUERY}: What part of speech applies to the word ``\{word\}" in the sentence?\\
        \textcolor{blue}{target}: \{POS\}
        
        \item \textcolor{red}{source}: \texttt{TEXT}: \{sentence\} \texttt{QUERY}: Choose the right part of speech for ``\{word\}".\\
        \textcolor{blue}{target}: \{POS\}
        
        \item \textcolor{red}{source}: \texttt{TEXT}: \{sentence\} \texttt{QUERY}: Is ``\{word\}" a/an ``\{POS\}" in the text?\\
        \textcolor{blue}{target}: Yes
        
        \item \textcolor{red}{source}: \texttt{TEXT}: \{sentence\} \texttt{QUERY}: Is ``\{word\}" a/an ``\{other\_POS\}" in the text?\\
        \textcolor{blue}{target}: No
        
        \item \textcolor{red}{source}: \texttt{TEXT}: \{sentence\} \texttt{QUERY}: Identify the right part of speech for ``\{word\}" in the context.\\
        \textcolor{blue}{target}: \{POS\}
        
        \item \textcolor{red}{source}: \texttt{TEXT}: \{sentence\} \texttt{QUERY}: What kind of part of speech is ``\{word\}" in the text?\\
        \textcolor{blue}{target}: \{POS\}
        
        \item \textcolor{red}{source}: \texttt{TEXT}: \{sentence\} \texttt{QUERY}: In the text, is ``\{word\}" used as a/an ``\{POS\}"?\\
        \textcolor{blue}{target}: Yes
        
        \item \textcolor{red}{source}: \texttt{TEXT}: \{sentence\} \texttt{QUERY}: In the text, is ``\{word\}" used as a/an ``\{other\_POS\}\"?\\
        \textcolor{blue}{target}: No
    \end{enumerate}
\end{itemize}

\paragraph{Meaning}
With $($word, sentence, meaning$)$ triples, we construct prompts that ask for the meaning of a word within a certain context. Here, we use \{meanings\_with\_or\} (e.g. ``option1", ``option2", or ``option3") and \{meaning\_without\_or\} (e.g. ``option1", ``option2", ``option3") to refer to all the available meanings of a given word. We use \{other\_meaning\} to refer to a random meaning out of all the meanings for the given word, but is different from \{meaning\}.
\begin{itemize}
    \item Multiple-choice format prompts
    \begin{enumerate}
        \item \textcolor{red}{source}: \texttt{TEXT}: \{sentence\} \texttt{QUERY}: What does the word ``\{word\}" mean in the previous text? \{meanings\_with\_or\}?\\
        \textcolor{blue}{target}: \{meaning\}
        
        \item \textcolor{red}{source}: \texttt{TEXT}: \{sentence\} \texttt{QUERY}: What's the meaning of ``\{word\}" in the text? \{meanings\_with\_or\}?\\
        \textcolor{blue}{target}: \{meaning\}
        
        \item \textcolor{red}{source}: \texttt{TEXT}: \{sentence\} \texttt{QUERY}: Given the following meanings of the word ``\{word\}": \{meanings\_without\_or\}. Can you choose the right meaning according to the previous context?\\
        \textcolor{blue}{target}: \{meaning\}
        
        \item \textcolor{red}{source}: \texttt{TEXT}: \{sentence\} \texttt{QUERY}: How to understand the word ``\{word\}" in the previous text? You may choose from: \{meanings\_without\_or\}.\\
        \textcolor{blue}{target}: \{meaning\}
        
        \item \textcolor{red}{source}: \texttt{TEXT}: \{sentence\} \texttt{QUERY}: Can you explain the meaning of ``\{word\}" in the previous text? The options are \{meanings\_without\_or\}.\\
        \textcolor{blue}{target}: \{meaning\}
        
        \item \textcolor{red}{source}: \texttt{TEXT}: \{sentence\} \texttt{QUERY}: Given a list of meanings of ``\{word\}": 
        \{meanings\_without\_or\}, which of these does the word ``\{word\}" in the previous text correspond to?\\
        \textcolor{blue}{target}: \{meaning\}
        
        \item \textcolor{red}{source}: \texttt{TEXT}: \{sentence\} \texttt{QUERY}: Which of the following is the meaning of ``\{word\}" in the above text: \{meanings\_without\_or\}?\\
        \textcolor{blue}{target}: \{meaning\}
        
        \item \textcolor{red}{source}: \texttt{TEXT}: \{sentence\} \texttt{QUERY}: Read the above text, and select the appropriate meaning of the word ``\{word\}". The options are \{meanings\_without\_or\}.\\
        \textcolor{blue}{target}: \{meaning\}
        
        \item \textcolor{red}{source}: \texttt{TEXT}: \{sentence\} \texttt{QUERY}: In the text, does the word ``\{word\}" mean ``\{meaning\}"? ``Yes" or ``No"?\\
        \textcolor{blue}{target}: Yes
        
        \item \textcolor{red}{source}: \texttt{TEXT}: \{sentence\} \texttt{QUERY}: In the text, does the word ``\{word\}" mean ``\{other\_meaning\}"? ``Yes" or ``No"?\\
        \textcolor{blue}{target}: No
        
        \item \textcolor{red}{source}: \texttt{TEXT}: \{sentence\} \texttt{QUERY}: The word ``\{word\}" in the previous text means ``\{meaning\}". ``True" or ``False"?\\
        \textcolor{blue}{target}: True
        
        \item \textcolor{red}{source}: \texttt{TEXT}: \{sentence\} \texttt{QUERY}: The word ``\{word\}" in the previous text means ``\{other\_meaning\}". ``True" or ``False"?\\
        \textcolor{blue}{target}: False

    \end{enumerate}
    \item Generation format prompts
    \begin{enumerate}
        \item \textcolor{red}{source}: \texttt{TEXT}: \{sentence\} \texttt{QUERY}: What does the word ``\{word\}" mean in the previous text?\\
        \textcolor{blue}{target}: \{meaning\}
        
        \item \textcolor{red}{source}: \texttt{TEXT}: \{sentence\} \texttt{QUERY}: What's the meaning of ``\{word\}" in the text?\\
        \textcolor{blue}{target}: \{meaning\}
        
        \item \textcolor{red}{source}: \texttt{TEXT}: \{sentence\} \texttt{QUERY}: In the context of the above, what is the definition of the word ``\{word\}"?\\
        \textcolor{blue}{target}: \{meaning\}
        
        \item \textcolor{red}{source}: \texttt{TEXT}: \{sentence\} \texttt{QUERY}: How to understand the word ``\{word\}" in the previous text?\\
        \textcolor{blue}{target}: \{meaning\}
        
        \item \textcolor{red}{source}: \texttt{TEXT}: \{sentence\} \texttt{QUERY}: Can you explain the meaning of ``\{word\}" in the previous text?\\
        \textcolor{blue}{target}: \{meaning\}
        
        \item \textcolor{red}{source}: \texttt{TEXT}: \{sentence\} \texttt{QUERY}: Can you give a proper definition of the word ``\{word\}" in the previous sentence?\\
        \textcolor{blue}{target}: \{meaning\}
        
        \item \textcolor{red}{source}: \texttt{TEXT}: \{sentence\} \texttt{QUERY}: In the text, what does the word ``\{word\}" mean?\\
        \textcolor{blue}{target}: \{meaning\}
        
        \item \textcolor{red}{source}: \texttt{TEXT}: \{sentence\} \texttt{QUERY}: Read the above sentence, and write the appropriate meaning of the word ``\{word\}".\\
        \textcolor{blue}{target}: \{meaning\}
        
        \item \textcolor{red}{source}: \texttt{TEXT}: \{sentence\} \texttt{QUERY}: In the text, does the word ``\{word\}" mean ``\{meaning\}"?\\
        \textcolor{blue}{target}: Yes
        
        \item \textcolor{red}{source}: \texttt{TEXT}: \{sentence\} \texttt{QUERY}: In the text, does the word ``\{word\}" mean ``\{other\_meaning\}"?\\
        \textcolor{blue}{target}: No
        
        \item \textcolor{red}{source}: \texttt{TEXT}: \{sentence\} \texttt{QUERY}: Is the meaning of ``\{word\}" in the preceding text ``\{meaning\}"?\\
        \textcolor{blue}{target}: Yes
        
        \item \textcolor{red}{source}: \texttt{TEXT}: \{sentence\} \texttt{QUERY}: Is the meaning of ``\{word\}" in the preceding text ``\{other\_meaning\}\"?\\
        \textcolor{blue}{target}: No
    \end{enumerate}
\end{itemize}

\paragraph{Synonyms}
\label{synonym_prompt}
With $($word, sentence, synonym$)$ triples, we construct prompts that ask for the synonym of a word within a certain context. Here, we use \{choices\_with\_or\} (e.g. ``option1", ``option2", or ``option3") and \{choices\_without\_or\} (e.g. ``option1", ``option2", ``option3") to refer to some (3 to 9) words that containing exactly one synonym of the word. We use \{other\_word\} to refer to a word that is not a synonym of the given word we are interested in.

\begin{itemize}
    \item Multiple-choice format prompts
    \begin{enumerate}
        \item \textcolor{red}{source}: \texttt{TEXT}: \{sentence\} \texttt{QUERY}: Can you choose a synonym for the word ``\{word\}" in the preceding text from the following options: \{choices\_without\_or\}?\\
        \textcolor{blue}{target}: \{synonym\}
        
        \item \textcolor{red}{source}: \texttt{TEXT}: \{sentence\} \texttt{QUERY}: Given the following words/phrases: \{choices\_without\_or\}, which one is semantically close to ``\{word\}" in the previous text?\\
        \textcolor{blue}{target}: \{synonym\}
        
        \item \textcolor{red}{source}: \texttt{TEXT}: \{sentence\} \texttt{QUERY}: Which of the following is similar in meaning to the word ``\{word\}" in the context?\\
        \textcolor{blue}{target}: \{synonym\}
        
        \item \textcolor{red}{source}: \texttt{TEXT}: \{sentence\} \texttt{QUERY}: In the text, the word ``\{word\}" has a similar meaning to ``\{synonym\}". ``True" or ``False"?\\
        \textcolor{blue}{target}: True
        
        \item \textcolor{red}{source}: \texttt{TEXT}: \{sentence\} \texttt{QUERY}: In the text, the word ``\{word\}" has a similar meaning to ``\{other\_word\}". ``True" or ``False"?\\
        \textcolor{blue}{target}: False
        
        \item \textcolor{red}{source}: \texttt{TEXT}: \{sentence\} \texttt{QUERY}: Given the options: \{choices\_without\_or\}, choose a synonym for the word ``\{word\}" in the context.\\
        \textcolor{blue}{target}: \{synonym\}
        
        \item \textcolor{red}{source}: \texttt{TEXT}: \{sentence\} \texttt{QUERY}: Is ``\{synonym\}" a synonym for ``\{word\}" in the previous text? ``Yes" or ``No"?\\
        \textcolor{blue}{target}: Yes
        
        \item \textcolor{red}{source}: \texttt{TEXT}: \{sentence\} \texttt{QUERY}: Is ``\{other\_word\}" a synonym for ``\{word\}" in the previous text? ``Yes" or ``No"?\\
        \textcolor{blue}{target}: No
        
        \item \textcolor{red}{source}: \texttt{TEXT}: \{sentence\} \texttt{QUERY}: Which of the following is a synonym for ``\{word\}" in the preceding sentence? \{choices\_with\_or\}?\\
        \textcolor{blue}{target}: \{synonym\}
        
        \item \textcolor{red}{source}: \texttt{TEXT}: \{sentence\} \texttt{QUERY}: Pick one of the following options that is similar to the meaning of ``\{word\}" in the text: \{choices\_without\_or\}.\\
        \textcolor{blue}{target}: \{synonym\}
        
    \end{enumerate}
    \item Generation format prompts
    \begin{enumerate}
        \item \textcolor{red}{source} \texttt{TEXT}: \{sentence\} \texttt{QUERY}: Can you write a synonym for the word ``\{word\}" in the preceding text?\\
        \textcolor{blue}{target}: \{synonym\}
        
        \item \textcolor{red}{source}: \texttt{TEXT}: \{sentence\} \texttt{QUERY}: Think of a word or phrase that is semantically close to ``\{word\}" in the previous text.\\
        \textcolor{blue}{target}: \{synonym\}
        
        \item \textcolor{red}{source}: \texttt{TEXT}: \{sentence\} \texttt{QUERY}: What word or phrase is similar in meaning to the word ``\{word\}" in the context?\\
        \textcolor{blue}{target}: \{synonym\}
        
        \item \textcolor{red}{source}: \texttt{TEXT}: \{sentence\} \texttt{QUERY}: Does the word ``\{word\}" in the previous sentence mean the same as ``\{synonym\}"?\\
        \textcolor{blue}{target}: Yes
        
        \item \textcolor{red}{source}: \texttt{TEXT}: \{sentence\} \texttt{QUERY}: Does the word ``\{word\}" in the previous sentence mean the same as ``\{other\_word\}"?\\
        \textcolor{blue}{target}: No
        
        \item \textcolor{red}{source}: \texttt{TEXT}: \{sentence\} \texttt{QUERY}: Generate a synonym for the word ``\{word\}" in the context.\\
        \textcolor{blue}{target}: \{synonym\}
        
        \item \textcolor{red}{source}: \texttt{TEXT}: \{sentence\} \texttt{QUERY}: Is ``\{synonym\}" a synonym for ``\{word\}" in the previous text?\\
        \textcolor{blue}{target}: Yes
        
        \item \textcolor{red}{source}: \texttt{TEXT}: \{sentence\} \texttt{QUERY}: Is ``\{other\_word\}" a synonym for ``\{word\}" in the previous text?\\
        \textcolor{blue}{target}: No
        
        \item \textcolor{red}{source}: \texttt{TEXT}: \{sentence\} \texttt{QUERY}: Write a synonym for the word ``\{word\}" in the preceding text.\\
        \textcolor{blue}{target}: \{synonym\}
        
        \item \textcolor{red}{source}: \texttt{TEXT}: \{sentence\} \texttt{QUERY}: Find a word or phrase that is similar to the meaning of ``\{word\}" in the text.\\
        \textcolor{blue}{target}: \{synonym\}
    \end{enumerate}
\end{itemize}

\paragraph{Antonyms}
With $($word, sentence, antonym$)$ triples, we can construct similar prompts as \S\ref{synonym_prompt}-Synonyms. Here, we use \{choices\_with\_or\} (e.g. ``option1", ``option2", or ``option3") and \{choices\_without\_or\} (e.g. ``option1", ``option2", ``option3") to refer to some (3 to 9) words that containing exactly one antonym of the word. We use \{other\_word\} to refer to a word that is not an antonym of the given word we are interested in.

\begin{itemize}
    \item Multiple-choice format prompts
    \begin{enumerate}
        \item \textcolor{red}{source}: \texttt{TEXT}: \{sentence\} \texttt{QUERY}: Can you choose an antonym for the word ``\{word\}" in the preceding text from the following options: \{choices\_without\_or\}?\\
        \textcolor{blue}{target}: \{antonym\}
        
        \item \textcolor{red}{source}: \texttt{TEXT}: \{sentence\} \texttt{QUERY}: Which of the following means the opposite to the word ``\{word\}" in the text? \{choices\_with\_or\}?\\
        \textcolor{blue}{target}: \{antonym\}
        
        \item \textcolor{red}{source}: \texttt{TEXT}: \{sentence\} \texttt{QUERY}: Which of the following is the opposite of the meaning of ``\{word\}" in the text? \{choices\_with\_or\}?\\
        \textcolor{blue}{target}: \{antonym\}
        
        \item \textcolor{red}{source}: \texttt{TEXT}: \{sentence\} \texttt{QUERY}: In the text, the word ``\{word\}" has an opposite meaning to ``\{antonym\}". ``True" or ``False"?\\
        \textcolor{blue}{target}: True
        
        \item \textcolor{red}{source}: \texttt{TEXT}: \{sentence\} \texttt{QUERY}: In the text, the word ``\{word\}" has an opposite meaning to ``\{other\_word\}". ``True" or ``False"?\\
        \textcolor{blue}{target}: False
        
        \item \textcolor{red}{source}: \texttt{TEXT}: \{sentence\} \texttt{QUERY}: Given the options: \{choices\_without\_or\}, choose an antonym for the word ``\{word\}" in the context.\\
        \textcolor{blue}{target}: \{antonym\}
        
        \item \textcolor{red}{source}: \texttt{TEXT}: \{sentence\} \texttt{QUERY}: Is ``\{antonym\}" an antonym for ``\{word\}" in the previous text? ``Yes" or ``No"?\\
        \textcolor{blue}{target}: Yes
        
        \item \textcolor{red}{source}: \texttt{TEXT}: \{sentence\} \texttt{QUERY}: Is ``\{other\_word\}" an antonym for ``\{word\}" in the previous text? ``Yes" or ``No"?\\
        \textcolor{blue}{target}: No
        
        \item \textcolor{red}{source}: \texttt{TEXT}: \{sentence\} \texttt{QUERY}: Which of the following is an antonym for ``\{word\}" in the preceding sentence? \{choices\_with\_or\}?\\
        \textcolor{blue}{target}: \{antonym\}
        
        \item \textcolor{red}{source}: \texttt{TEXT}: \{sentence\} \texttt{QUERY}: Pick one of the following options that is opposite to the meaning of ``\{word\}" in the text: \{choices\_without\_or\}.\\
        \textcolor{blue}{target}: \{antonym\}
    \end{enumerate}
    \item Generation format prompts
    \begin{enumerate}
        \item \textcolor{red}{source}: \texttt{TEXT}: \{sentence\} \texttt{QUERY}: Can you write an antonym for the word ``\{word\}" in the preceding text.\\
        \textcolor{blue}{target}: \{antonym\}
        
        \item \textcolor{red}{source}: \texttt{TEXT}: \{sentence\} \texttt{QUERY}: Think of a word or phrase that means the opposite of the word ``\{word\}" in the previous sentence.\\
        \textcolor{blue}{target}: \{antonym\}
        
        \item \textcolor{red}{source}: \texttt{TEXT}: \{sentence\} \texttt{QUERY}: What word or phrase is opposite in meaning to the word ``\{word\}" in the context?\\
        \textcolor{blue}{target}: \{antonym\}
        
        \item \textcolor{red}{source}: \texttt{TEXT}: \{sentence\} \texttt{QUERY}: Does the word ``\{word\}" in the previous sentence mean the opposite of ``\{antonym\}"?\\
        \textcolor{blue}{target}: Yes
        
        \item \textcolor{red}{source}: \texttt{TEXT}: \{sentence\} \texttt{QUERY}: Does the word ``\{word\}" in the previous sentence mean the opposite of ``\{other\_word\}"?\\
        \textcolor{blue}{target}: No
        
        \item \textcolor{red}{source}: \texttt{TEXT}: \{sentence\} \texttt{QUERY}: Generate an antonym for the word ``\{word\}" in the context.\\
        \textcolor{blue}{target}: \{antonym\}
        
        \item \textcolor{red}{source}: \texttt{TEXT}: \{sentence\} \texttt{QUERY}: Is ``\{antonym\}" an antonym for ``\{word\}" in the previous text?\\
        \textcolor{blue}{target}: Yes
        
        \item \textcolor{red}{source}: \texttt{TEXT}: \{sentence\} \texttt{QUERY}: Is ``\{other\_word\}" an antonym for ``\{word\}" in the previous text?\\
        \textcolor{blue}{target}: No
        
        \item \textcolor{red}{source}: \texttt{TEXT}: \{sentence\} \texttt{QUERY}: Write an antonym for the word ``\{word\}" in the preceding text.\\
        \textcolor{blue}{target}: \{antonym\}
        
        \item \textcolor{red}{source}: \texttt{TEXT}: \{sentence\} \texttt{QUERY}: Find a word or phrase that is opposite in meaning to ``\{word\}" in the text.\\
        \textcolor{blue}{target}: \{antonym\}
    \end{enumerate}
\end{itemize}

\subsection{Question Answering}
\paragraph{ConTRoL}
ConTRoL is derived from verbal reasoning tests from which we can get $($premise, hypothesis, label$)$ triples. \{label\} can be ``entailment", ``neutral" and ``contradiction". We construct the following prompts.

\begin{itemize}
    \item Multiple-choice format prompts
    \begin{enumerate}
        \item \textcolor{red}{source}: \texttt{TEXT}: \{premise\} \texttt{QUERY}: Based on the previous text, can we infer that ``\{hypothesis\}"? ``Yes" or ``No" or ``Maybe"?\\
        \textcolor{blue}{target}: Yes (if \{label\} is ``entailment") / No (if \{label\} is ``contradiction") / Maybe (if \{label\} is ``neutral")
        
        \item \textcolor{red}{source}: \texttt{TEXT}: \{premise\} \texttt{QUERY}: Given the previous premise, is the hypothesis: ``\{hypothesis\}" ``true" or ``false" or ``undetermined"?\\
        \textcolor{blue}{target}: true (if \{label\} is ``entailment") / false (if \{label\} is ``contradiction") / undetermined (if \{label\} is ``neutral")
        
        \item \textcolor{red}{source}: \texttt{TEXT}: \{premise\} \texttt{QUERY}: Based on that information, is the claim ``\{hypothesis\}" ``true", ``false" or ``inconclusive"?\\
        \textcolor{blue}{target}: true (if \{label\} is ``entailment") / false (if \{label\} is ``contradiction") / inconclusive (if \{label\} is ``neutral")
        
        \item \textcolor{red}{source}: \texttt{TEXT}: \{premise\} \texttt{QUERY}: Given the previous text, does it imply that ``\{hypothesis\}"? ``Yes", ``No" or ``Maybe"?\\
        \textcolor{blue}{target}: Yes (if \{label\} is ``entailment") / No (if \{label\} is ``contradiction") / Maybe (if \{label\} is ``neutral")
        
        \item \textcolor{red}{source}: \texttt{TEXT}: \{premise\} \texttt{QUERY}: Based on the preceding paragraph, ``\{hypothesis\}" is ``guaranteed", ``possible" or ``impossible"?\\
        \textcolor{blue}{target}: guaranteed (if \{label\} is ``entailment") / possible (if \{label\} is ``neutral") / impossible (if \{label\} is ``contradiction")
        
        \item \textcolor{red}{source}: \texttt{TEXT}: Premise: \{premise\} Hypothesis: \{hypothesis\} \texttt{QUERY}: What's the relationship between the hypothesis and the premise? ``Entailment", ``Neutral" or ``Contradiction"?\\
        \textcolor{blue}{target}: Entailment (if \{label\} is ``entailment") / Neutral (if \{label\} is ``neutral") / Contradiction (if \{label\} is ``contradiction")
        
        \item \textcolor{red}{source}: \texttt{TEXT}: Premise: \{premise\} Hypothesis: \{hypothesis\} \texttt{QUERY}: The hypothesis is entailed by the premise. ``True" or ``False"?\\
        \textcolor{blue}{target}: True (if \{label\} is ``entailment") / False (if \{label\} is not ``entailment")
        
        \item \textcolor{red}{source}: \texttt{TEXT}: Premise: \{premise\} Hypothesis: \{hypothesis\} \texttt{QUERY}: The hypothesis is contradicted by the premise. ``True" or ``False"?\\
        \textcolor{blue}{target}: True (if \{label\} is ``contradiction") / False (if \{label\} is not ``contradiction")
        
        \item \textcolor{red}{source}: \texttt{TEXT}: \{premise\} \texttt{QUERY}: \{hypothesis\} ``True" or ``False" or ``Unknown"?\\
        \textcolor{blue}{target}: True (if \{label\} is ``entailment") / False (if \{label\} is ``contradiction") / Unknown (if \{label\} is ``neutral")
        
        \item \textcolor{red}{source}: \texttt{TEXT}: \{premise\} \texttt{QUERY}: Based on the previous paragraph, is it true that ``\{hypothesis\}"? ``Yes", ``No" or ``Maybe"?\\
        \textcolor{blue}{target}: Yes (if \{label\} is ``entailment") / No (if \{label\} is ``contradiction") / Maybe (if \{label\} is ``neutral")
    \end{enumerate}
    \item Generation format prompts
    
    \begin{enumerate}
        \item \textcolor{red}{source}: \texttt{TEXT}: \{premise\} \texttt{QUERY}: Based on the previous text, can we infer that ``\{hypothesis\}"?\\
        \textcolor{blue}{target}: Yes (if \{label\} is ``entailment") / No (if \{label\} is ``contradiction") / Maybe (if \{label\} is ``neutral")
        
        \item \textcolor{red}{source}: \texttt{TEXT}: \{premise\} \texttt{QUERY}: Given the previous premise, is the hypothesis: ``\{hypothesis\}" ``true" or ``false" or ``undetermined"?\\
        \textcolor{blue}{target}: true (if \{label\} is ``entailment") / false (if \{label\} is ``contradiction") / undetermined (if \{label\} is ``neutral")
        
        \item \textcolor{red}{source}: \texttt{TEXT}: \{premise\} \texttt{QUERY}: Based on that information, is the claim ``\{hypothesis\}" ``true", ``false" or ``inconclusive"?\\
        \textcolor{blue}{target}: true (if \{label\} is ``entailment") / false (if \{label\} is ``contradiction") / inconclusive (if \{label\} is ``neutral")
        
        \item \textcolor{red}{source}: \texttt{TEXT}: \{premise\} \texttt{QUERY}: Given the previous text, does it imply that ``\{hypothesis\}"?\\
        \textcolor{blue}{target}: Yes (if \{label\} is ``entailment") / No (if \{label\} is ``contradiction") / Maybe (if \{label\} is ``neutral")
        
        \item \textcolor{red}{source}: \texttt{TEXT}: \{premise\} \texttt{QUERY}: Based on the preceding paragraph, ``\{hypothesis\}" is ``guaranteed", ``possible" or ``impossible"?\\
        \textcolor{blue}{target}: guaranteed (if \{label\} is ``entailment") / possible (if \{label\} is ``neutral") / impossible (if \{label\} is ``contradiction")
        
        \item \textcolor{red}{source}: \texttt{TEXT}: Premise: \{premise\} Hypothesis: \{hypothesis\} \texttt{QUERY}: What's the relationship between the hypothesis and the premise?\\
        \textcolor{blue}{target}: Entailment (if \{label\} is ``entailment") / Neutral (if \{label\} is ``neutral") / Contradiction (if \{label\} is ``contradiction")
        
        \item \textcolor{red}{source}: \texttt{TEXT}: Premise: \{premise\} Hypothesis: \{hypothesis\} \texttt{QUERY}: Is the hypothesis entailed by the premise?\\
        \textcolor{blue}{target}: Yes (if \{label\} is ``entailment") / No (if \{label\} is not ``entailment")
        
        \item \textcolor{red}{source}: \texttt{TEXT}: Premise: {premise} Hypothesis: {hypothesis} \texttt{QUERY}: Is the hypothesis contradicted by the premise?\\
        \textcolor{blue}{target}: Yes (if \{label\} is ``contradiction") / No (if \{label\} is not ``contradiction")
        
        \item \textcolor{red}{source}: \texttt{TEXT}: \{premise\} \texttt{QUERY}: Based on the previous premise, should we assume that ``\{hypothesis\}" is true?\\
        \textcolor{blue}{target}: Yes (if \{label\} is ``entailment") / No (if \{label\} is ``contradiction") / Maybe (if \{label\} is ``neutral")
        
        \item \textcolor{red}{source}: \texttt{TEXT}: \{premise\} \texttt{QUERY}: Based on the previous paragraph, is it true that ``\{hypothesis\}"?\\
        \textcolor{blue}{target}: Yes (if \{label\} is ``entailment") / No (if \{label\} is ``contradiction") / Maybe (if \{label\} is ``neutral")
    \end{enumerate}
\end{itemize}
\paragraph{DREAM}
DREAM contains questions with multiple choices for each question, from which we can get $($context, question, choices, answer$)$ quads. Here, we use \{choices\_with\_or\} (e.g. ``option1", ``option2", or ``option3") to refer to all the available options. We construct the following prompts.

\begin{itemize}
    \item Multiple-choice format prompts
    \begin{enumerate}
        \item \textcolor{red}{source}: \texttt{TEXT}: \{context\} \texttt{QUERY}: \{question\} \{choices\_with\_or\}?\\
        \textcolor{blue}{target}: \{answer\}
    \end{enumerate}
    \item Generation format prompts
    \begin{enumerate}
        \item \textcolor{red}{source}: \texttt{TEXT}: \{context\} \texttt{QUERY}: \{question\}\\
        \textcolor{blue}{target}: \{answer\}
    \end{enumerate}
\end{itemize}

\paragraph{LogiQA}
LogiQA contains questions with multiple choices for each question, from which we can get $($context, question, choices, answer$)$ quads. Here, we use \{choices\_with\_or\} (e.g. ``option1", ``option2", or ``option3") to refer to all the available options. We construct the following prompts.

\begin{itemize}
    \item Multiple-choice format prompts
    \begin{enumerate}
        \item \textcolor{red}{source}: \texttt{TEXT}: \{context\} \texttt{QUERY}: \{question\} \{choices\_with\_or\}?\\
        \textcolor{blue}{target}: \{answer\}
    \end{enumerate}
    \item Generation format prompts
    \begin{enumerate}
        \item \textcolor{red}{source}: \texttt{TEXT}: \{context\} \texttt{QUERY}: \{question\}\\
        \textcolor{blue}{target}: \{answer\}
    \end{enumerate}
\end{itemize}

\paragraph{RACE \& RACE-C}
There are both multiple-choice questions and cloze questions in RACE and RACE-C, where we design different prompts for them. We still can get $($context, question, choices, answer$)$ quads. Here, we use \{choices\_with\_or\} (e.g. ``option1", ``option2", or ``option3") to refer to all the available options. For multiple-choice questions, we design the following prompts.

\begin{itemize}
    \item Multiple-choice format prompts
    \begin{enumerate}
        \item \textcolor{red}{source}: \texttt{TEXT}: \{context\} \texttt{QUERY}: \{question\} \{choices\_with\_or\}?\\
        \textcolor{blue}{target}: \{answer\}
    \end{enumerate}
\end{itemize}

For cloze questions, we replace the underscore within the question into a special mark ``\texttt{[BLANK]}" and design the following prompts.

\begin{itemize}
    \item Multiple-choice format prompts
    \begin{enumerate}
        \item \textcolor{red}{source}: \texttt{TEXT}: \{context\} \texttt{QUERY}: \{question\} What should be filled in the \texttt{[BLANK]} position? \{choices\_with\_or\}?\\
        \textcolor{blue}{target}: \{answer\}
    \end{enumerate}
\end{itemize}

\paragraph{ReClor}
ReClor contains questions with multiple choices for each question, from which we can get $($context, question, choices, answer$)$ quads. Here, we use \{choices\_with\_or\} (e.g. ``option1", ``option2", or ``option3") to refer to all the available options. We construct the following prompts.

\begin{itemize}
    \item Multiple-choice format prompts
    \begin{enumerate}
        \item \textcolor{red}{source}: \texttt{TEXT}: \{context\} \texttt{QUERY}: \{question\} \{choices\_with\_or\}?\\
        \textcolor{blue}{target}: \{answer\}
    \end{enumerate}
    \item Generation format prompts
    \begin{enumerate}
        \item \textcolor{red}{source}: \texttt{TEXT}: \{context\} \texttt{QUERY}: \{question\}\\
        \textcolor{blue}{target}: \{answer\}
    \end{enumerate}
\end{itemize}

\paragraph{TriviaQA}
With $($context, question, answer$)$ triples, we construct the following prompts.

\begin{itemize}
    \item Generation format prompts
    \begin{enumerate}
        \item \textcolor{red}{source}: \texttt{TEXT}: \{context\} \texttt{QUERY}: \{question\}\\
        \textcolor{blue}{target}: \{answer\}
    \end{enumerate}
\end{itemize}

\subsection{arXiv}
\paragraph{Category}
With $($abstract, category$)$ pairs, we construct prompts that ask the category of a paper based on its abstract part. Here, we use \{categories\_with\_or\} (e.g. ``option1", ``option2", or ``option3") and \{categories\_without\_or\} (e.g. ``option1", ``option2", ``option3") to refer to all the categories. We use \{other\_category\} to represent a random category that is different from \{category\}.

\begin{itemize}
    \item Multiple-choice format prompts
    \begin{enumerate}
        \item \textcolor{red}{source}: \texttt{TEXT}: \{abstract\} \texttt{QUERY}: What's this text about? \{categories\_with\_or\}?\\
        \textcolor{blue}{target}: \{category\}
        
        \item \textcolor{red}{source}: \texttt{TEXT}: \{abstract\} \texttt{QUERY}: Classify this paper. You may choose from \{categories\_without\_or\}.\\
        \textcolor{blue}{target}: \{category\}
        
        \item \textcolor{red}{source}: \texttt{TEXT}: \{abstract\} \texttt{QUERY}: Is this text about ``\{category\}"? ``Yes" or ``No"?\\
        \textcolor{blue}{target}: Yes
        
        \item \textcolor{red}{source}: \texttt{TEXT}: \{abstract\} \texttt{QUERY}: Is this text about ``\{other\_category\}"? ``Yes" or ``No"?\\
        \textcolor{blue}{target}: No
        
        \item \textcolor{red}{source}: \texttt{TEXT}: \{abstract\} \texttt{QUERY}: Given a list of categories: \{categories\_without\_or\}, what category does the paper belong to?\\
        \textcolor{blue}{target}: \{category\}
        
        \item \textcolor{red}{source}: \texttt{TEXT}: \{abstract\} \texttt{QUERY}: This text is related to ``\{category\}". ``True" or ``False"?\\
        \textcolor{blue}{target}: True
        
        \item \textcolor{red}{source}: \texttt{TEXT}: \{abstract\} \texttt{QUERY}: This text is related to ``\{other\_category\}". ``True" or ``False"?\\
        \textcolor{blue}{target}: False
        
        \item \textcolor{red}{source}: \texttt{TEXT}: \{abstract\} \texttt{QUERY}: Can you identify the category of this text? The options are: \{categories\_without\_or\}.\\
        \textcolor{blue}{target}: \{category\}
        
        \item \textcolor{red}{source}: \texttt{TEXT}: \{abstract\} \texttt{QUERY}: What category best describes this paper? \{categories\_with\_or\}?\\
        \textcolor{blue}{target}: \{category\}
        
        \item \textcolor{red}{source}: \texttt{TEXT}: \{abstract\} \texttt{QUERY}: Pick one category for the previous text. Please choose from the following: \{categories\_without\_or\}.\\
        \textcolor{blue}{target}: \{category\}
        
    \end{enumerate}
    \item Generation format prompts
    \begin{enumerate}
        \item \textcolor{red}{source}: \texttt{TEXT}: \{abstract\} \texttt{QUERY}: What's this text about?\\
        \textcolor{blue}{target}: \{category\}
        
        \item \textcolor{red}{source}: \texttt{TEXT}: \{abstract\} \texttt{QUERY}: Which discipline is this academic paper about?\\
        \textcolor{blue}{target}: \{category\}
        
        \item \textcolor{red}{source}: \texttt{TEXT}: \{abstract\} \texttt{QUERY}: Is this text about ``\{category\}"?\\
        \textcolor{blue}{target}: Yes
        
        \item \textcolor{red}{source}: \texttt{TEXT}: \{abstract\} \texttt{QUERY}: Is this text about ``\{other\_category\}"?\\
        \textcolor{blue}{target}: No
        
        \item \textcolor{red}{source}: \texttt{TEXT}: \{abstract\} \texttt{QUERY}: Which academic field is this article most likely to be a paper in?\\
        \textcolor{blue}{target}: \{category\}
        
        \item \textcolor{red}{source}: \texttt{TEXT}: \{abstract\} \texttt{QUERY}: Which academic field is the previous article about?\\
        \textcolor{blue}{target}: \{category\}
        
        \item \textcolor{red}{source}: \texttt{TEXT}: \{abstract\} \texttt{QUERY}: Does this article focus on ``\{category\}"?\\
        \textcolor{blue}{target}: Yes
        
        \item \textcolor{red}{source}: \texttt{TEXT}: \{abstract\} \texttt{QUERY}: Does this article focus on ``\{other\_category\}"?\\
        \textcolor{blue}{target}: No
        
        \item \textcolor{red}{source}: \texttt{TEXT}: \{abstract\} \texttt{QUERY}: Can you identify the academic discipline of this text?\\
        \textcolor{blue}{target}: \{category\}
        
        \item \textcolor{red}{source}: \texttt{TEXT}: \{abstract\} \texttt{QUERY}: What category best describes this paper?\\
        \textcolor{blue}{target}: \{category\}
    \end{enumerate}
\end{itemize}

\paragraph{Summary}
With $($abstract, title$)$ pairs, we construct prompts that ask for the title of an academic paper based on its abstract. Here, we use \{titles\_with\_or\} (e.g. ``option1", ``option2", or ``option3") and \{titles\_without\_or\} (e.g. ``option1", ``option2", ``option3") to refer to some (3 to 9) titles that contain the correct one. Those incorrect titles were chosen from papers in the same category as the target paper. We use \{other\_title\} to represent a random title that is different from \{title\} and its paper shares the same category as the target paper. We use \{length\} to represent the length of the target title.

\begin{itemize}
    \item Multiple-choice format prompts
    \begin{enumerate}
        \item \textcolor{red}{source}: \texttt{TEXT}: \{abstract\} \texttt{QUERY}: The preceding text is an abstract of a paper, what might the title of this paper be? \{titles\_with\_or\}?\\
        \textcolor{blue}{target}: \{title\}
        
        \item \textcolor{red}{source}: \texttt{TEXT}: \{abstract\} \texttt{QUERY}: Select a title for the previous text. The options are \{titles\_without\_or\}.\\
        \textcolor{blue}{target}: \{title\}
        
        \item \textcolor{red}{source}: \texttt{TEXT}: \{abstract\} \texttt{QUERY}: Given a list of titles: \{titles\_without\_or\}, which one can express the main idea of the previous text?\\
        \textcolor{blue}{target}: \{title\}
        
        \item \textcolor{red}{source}: \texttt{TEXT}: \{abstract\} \texttt{QUERY}: Select an appropriate heading for the text. The options are: \{titles\_without\_or\}.\\
        \textcolor{blue}{target}: \{title\}
        
        \item \textcolor{red}{source}: \texttt{TEXT}: \{abstract\} \texttt{QUERY}: ``\{title\}" can summarize the previous text. ``True" or ``False"?\\
        \textcolor{blue}{target}: True
    
        \item \textcolor{red}{source}: \texttt{TEXT}: \{abstract\} \texttt{QUERY}: ``\{other\_title\}" can summarize the previous text. ``True" or ``False"?\\
        \textcolor{blue}{target}: False
        
        \item \textcolor{red}{source}: \texttt{TEXT}: \{abstract\} \texttt{QUERY}: Does ``\{title\}" summarize the core of the above text? ``Yes" or ``No"?\\
        \textcolor{blue}{target}: Yes
        
        \item \textcolor{red}{source}: \texttt{TEXT}: \{abstract\} \texttt{QUERY}: Does ``\{other\_title\}" summarize the core of the above text? ``Yes" or ``No"?\\
        \textcolor{blue}{target}: No
        
        \item \textcolor{red}{source}: \texttt{TEXT}: \{abstract\} \texttt{QUERY}: Can you pick the suitable title of the previous paper from the following options: \{titles\_without\_or\}?\\
        \textcolor{blue}{target}: \{title\}
        
        \item \textcolor{red}{source}: \texttt{TEXT}: \{abstract\} \texttt{QUERY}: Which of the following could be the title of the preceding paper? \{titles\_with\_or\}?\\
        \textcolor{blue}{target}: \{title\}
    \end{enumerate}
    \item Generation format prompts
    \begin{enumerate}
        \item \textcolor{red}{source}: \texttt{TEXT}: \{abstract\} \texttt{QUERY}: What could be the title of the previous paper?\\
        \textcolor{blue}{target}: \{title\}
        
        \item \textcolor{red}{source}: \texttt{TEXT}: \{abstract\} \texttt{QUERY}: What can be an appropriate title for the previous text?\\
        \textcolor{blue}{target}: \{title\}
        
        \item \textcolor{red}{source}: \texttt{TEXT}: \{abstract\} \texttt{QUERY}: Generate a title for this article.\\
        \textcolor{blue}{target}: \{title\}
        
        \item \textcolor{red}{source}: \texttt{TEXT}: \{abstract\} \texttt{QUERY}: Can you write a title for the previous text?\\
        \textcolor{blue}{target}: \{title\}
        
        \item \textcolor{red}{source}: \texttt{TEXT}: \{abstract\} \texttt{QUERY}: In about \{length\} words, create a short title for the preceding article.\\
        \textcolor{blue}{target}: \{title\}
        
        \item \textcolor{red}{source}: \texttt{TEXT}: \{abstract\} \texttt{QUERY}: Can ``\{title\}" summarize the previous text?\\
        \textcolor{blue}{target}: Yes
        
        \item \textcolor{red}{source}: \texttt{TEXT}: \{abstract\} \texttt{QUERY}: Can ``\{other\_title\}" summarize the previous text?\\
        \textcolor{blue}{target}: No
        
        \item \textcolor{red}{source}: \texttt{TEXT}: \{abstract\} \texttt{QUERY}: Does ``\{title\}" summarize the core of the above text?\\
        \textcolor{blue}{target}: Yes
        
        \item \textcolor{red}{source}: \texttt{TEXT}: \{abstract\} \texttt{QUERY}: Does ``\{other\_title\}" summarize the core of the above text?\\
        \textcolor{blue}{target}: No
        
        \item \textcolor{red}{source}: \texttt{TEXT}: \{abstract\} \texttt{QUERY}: What can be a short description of the previous text? You may write about \{length\} words.\\
        \textcolor{blue}{target}: \{title\}
    \end{enumerate}
\end{itemize}

\subsection{Papers With Code}
\paragraph{Entity}
With $($sentence, entities$)$ pairs, we can construct prompts that ask for the entities given a sentence. Here, \{entities\} is a textual string that contains all the entities within the sentence, listed according to their order of appearance and separated by a comma (can be an empty string as well). We use \{entity\} to represent one random entity out of \{entities\} and use \{not\_entity\} to represent a random text span within this sentence that is not an entity. We use \{choice0\} and \{choice1\} to represent a random order of an entity and a non-entity.

\begin{itemize}
    \item Multiple-choice format prompts
    \begin{enumerate}
        \item \textcolor{red}{source}: \texttt{TEXT}: \{sentence\} \texttt{QUERY}: Is ``\{entity\}" an entity? ``Yes" or ``No"?\\
        \textcolor{blue}{target}: Yes
        
        \item \textcolor{red}{source}: \texttt{TEXT}: \{sentence\} \texttt{QUERY}: Is ``\{not\_entity\}" an entity? ``Yes" or ``No"?\\
        \textcolor{blue}{target}: No
        
        \item \textcolor{red}{source}: \texttt{TEXT}: \{sentence\} \texttt{QUERY}: In the previous text, ``\{entity\}" is a scientific term. ``True" or ``False"?\\
        \textcolor{blue}{target}: True
        
        \item \textcolor{red}{source}: \texttt{TEXT}: \{sentence\} \texttt{QUERY}: In the previous text, ``\{not\_entity\}" is a scientific term. ``True" or ``False"?\\
        \textcolor{blue}{target}: False
        
        \item \textcolor{red}{source}: \texttt{TEXT}: \{sentence\} \texttt{QUERY}: There are scientific entities in the previous text. ``True" or ``False"?\\
        \textcolor{blue}{target}: True (if \{entities\} is not an empty string) / False (if \{entities\} is an empty string)
        
        \item \textcolor{red}{source}: \texttt{TEXT}: \{sentence\} \texttt{QUERY}: There are no scientific entities in this text. ``True" or ``False"?\\
        \textcolor{blue}{target}: True (if \{entities\} is an empty string) / False (if \{entities\} is not an empty string)
        
        \item \textcolor{red}{source}: \texttt{TEXT}: \{sentence\} \texttt{QUERY}: ``\{entity\}" is not a scientific entity. ``True" or ``False"?\\
        \textcolor{blue}{target}: False
        
        \item \textcolor{red}{source}: \texttt{TEXT}: \{sentence\} \texttt{QUERY}: ``\{not\_entity\}" is not a scientific entity. ``True" or ``False"?\\
        \textcolor{blue}{target}: True
        
        \item \textcolor{red}{source}: \texttt{TEXT}: \{sentence\} \texttt{QUERY}: In this scientific text, does ``\{entity\}" consider an entity? ``Yes" or ``No"?\\
        \textcolor{blue}{target}: Yes
        
        \item \textcolor{red}{source}: \texttt{TEXT}: \{sentence\} \texttt{QUERY}: In this scientific text, does ``\{not\_entity\}" consider an entity? ``Yes" or ``No"?\\
        \textcolor{blue}{target}: No
    \end{enumerate}
    \item Generation format prompts
    \begin{enumerate}
        \item \textcolor{red}{source}: \texttt{TEXT}: \{sentence\} \texttt{QUERY}: Can you find all the entities in this scientific text?\\
        \textcolor{blue}{target}: \{entities\}
        
        \item \textcolor{red}{source}: \texttt{TEXT}: \{sentence\} \texttt{QUERY}: List all the scientific terms in the text.\\
        \textcolor{blue}{target}: \{entities\}
        
        \item \textcolor{red}{source}: \texttt{TEXT}: \{sentence\} \texttt{QUERY}: What are the entities in the previous text?\\
        \textcolor{blue}{target}: \{entities\}
        
        \item \textcolor{red}{source}: \texttt{TEXT}: \{sentence\} \texttt{QUERY}: Please enumerate all the scientific terms in the above text.\\
        \textcolor{blue}{target}: \{entities\}
        
        \item \textcolor{red}{source}: \texttt{TEXT}: \{sentence\} \texttt{QUERY}: Identify all the entities in the previous sentence.\\
        \textcolor{blue}{target}: \{entities\}
        
        \item \textcolor{red}{source}: \texttt{TEXT}: \{sentence\} \texttt{QUERY}: What entities are present in the above sentence?\\
        \textcolor{blue}{target}: \{entities\}
        
        \item \textcolor{red}{source}: \texttt{TEXT}: \{sentence\} \texttt{QUERY}: Please list the entities in the sentence in the order in which they appear.\\
        \textcolor{blue}{target}: \{entities\}
        
        \item \textcolor{red}{source}: \texttt{TEXT}: \{sentence\} \texttt{QUERY}: Is ``\{entity\}" a scientific entity?\\
        \textcolor{blue}{target}: Yes
        
        \item \textcolor{red}{source}: \texttt{TEXT}: \{sentence\} \texttt{QUERY}: Is ``\{not\_entity\}" a scientific entity?\\
        \textcolor{blue}{target}: No
        
        \item \textcolor{red}{source}: \texttt{TEXT}: \{sentence\} \texttt{QUERY}: Write down all the entities that appeared in the previous text.\\
        \textcolor{blue}{target}: \{entities\}
    \end{enumerate}
\end{itemize}
\paragraph{Entity Typing}
With $($sentence, entity, entity\_type$)$ triples, we design prompts that ask for the entity type of entity that appears in a piece of text. Here, we use \{entity\_types\_with\_or\} (e.g. ``option1", ``option2", or ``option3") and \{entity\_types\_without\_or\} (e.g. ``option1", ``option2", ``option3") to refer to some (3 to 9) available entity types (include the correct one). We use \{other\_entity\_type\} to represent a random entity type that is different from the true entity type of the entity.

\begin{itemize}
    \item Multiple-choice format prompts
    \begin{enumerate}
        \item \textcolor{red}{source}: \texttt{TEXT}: \{sentence\} \texttt{QUERY}: What's the entity type of ``\{entity\}"? \{entity\_types\_with\_or\}?\\
        \textcolor{blue}{target}: \{entity\_type\}
        
        \item \textcolor{red}{source}: \texttt{TEXT}: \{sentence\} \texttt{QUERY}: Can you choose an appropriate class from the following list for ``\{entity\}"? The options are \{entity\_types\_without\_or\}.\\
        \textcolor{blue}{target}: \{entity\_type\}
        
        \item \textcolor{red}{source}: \texttt{TEXT}: \{sentence\} \texttt{QUERY}: ``\{entity\}" is an instance of which of the following? \{entity\_types\_with\_or\}?\\
        \textcolor{blue}{target}: \{entity\_type\}
        
        \item \textcolor{red}{source}: \texttt{TEXT}: \{sentence\} \texttt{QUERY}: Given a list of categories: \{entity\_types\_without\_or\}, what category does ``\{entity\}" belong to?\\
        \textcolor{blue}{target}: \{entity\_type\}
        
        \item \textcolor{red}{source}: \texttt{TEXT}: \{sentence\} \texttt{QUERY}: Choose the correct entity type for ``\{entity\}" from the following: \{entity\_types\_without\_or\}.\\
        \textcolor{blue}{target}: \{entity\_type\}
        
        \item \textcolor{red}{source}: \texttt{TEXT}: \{sentence\} \texttt{QUERY}: Does ``\{entity\}" belong to ``\{entity\_type\}"? ``Yes" or ``No"?\\
        \textcolor{blue}{target}: Yes
        
        \item \textcolor{red}{source}: \texttt{TEXT}: \{sentence\} \texttt{QUERY}: Does ``\{entity\}" belong to ``\{other\_entity\_type\}"? ``Yes" or ``No"?\\
        \textcolor{blue}{target}: No
        
        \item \textcolor{red}{source}: \texttt{TEXT}: \{sentence\} \texttt{QUERY}: In the previous text, what type of entity is ``\{entity\}"? \{entity\_types\_with\_or\}?\\
        \textcolor{blue}{target}: \{entity\_type\}
        
        \item \textcolor{red}{source}: \texttt{TEXT}: \{sentence\} \texttt{QUERY}: Which of the following is the entity type of ``\{entity\}"? \{entity\_types\_with\_or\}?\\
        \textcolor{blue}{target}: \{entity\_type\}
        
        \item \textcolor{red}{source}: \texttt{TEXT}: \{sentence\} \texttt{QUERY}: In the text, ``\{entity\}" belongs to ``\{entity\_type\}". ``True" or ``False"?\\
        \textcolor{blue}{target}: True
        
        \item \textcolor{red}{source}: \texttt{TEXT}: \{sentence\} \texttt{QUERY}: In the text, ``\{entity\}" belongs to ``\{other\_entity\_type\}". ``True" or ``False"?\\
        \textcolor{blue}{target}: False
    \end{enumerate}
    \item Generation format prompts
    \begin{enumerate}
        \item \textcolor{red}{source}: \texttt{TEXT}: \{sentence\} \texttt{QUERY}: What's the entity type of ``\{entity\}"?\\
        \textcolor{blue}{target}: \{entity\_type\}
        
        \item \textcolor{red}{source}: \texttt{TEXT}: \{sentence\} \texttt{QUERY}: Can you find an appropriate class for ``\{entity\}"?\\
        \textcolor{blue}{target}: \{entity\_type\}
        
        \item \textcolor{red}{source}: \texttt{TEXT}: \{sentence\} \texttt{QUERY}: ``\{entity\}" is an instance of what entity type?\\
        \textcolor{blue}{target}: \{entity\_type\}
        
        \item \textcolor{red}{source}: \texttt{TEXT}: \{sentence\} \texttt{QUERY}: What category does ``\{entity\}" belong to?\\
        \textcolor{blue}{target}: \{entity\_type\}
        
        \item \textcolor{red}{source}: \texttt{TEXT}: \{sentence\} \texttt{QUERY}: Assign a correct entity type for ``\{entity\}".\\
        \textcolor{blue}{target}: \{entity\_type\}
        
        \item \textcolor{red}{source}: \texttt{TEXT}: \{sentence\} \texttt{QUERY}: Does ``\{entity\}" belong to ``\{entity\_type\}"?\\
        \textcolor{blue}{target}: Yes
        
        \item \textcolor{red}{source}: \texttt{TEXT}: \{sentence\} \texttt{QUERY}: Does ``\{entity\}" belong to ``\{other\_entity\_type\}"?\\
        \textcolor{blue}{target}: No
        
        \item \textcolor{red}{source}: \texttt{TEXT}: \{sentence\} \texttt{QUERY}: In the previous text, what type of entity is ``\{entity\}"?\\
        \textcolor{blue}{target}: \{entity\_type\}
        
        \item \textcolor{red}{source}: \texttt{TEXT}: \{sentence\} \texttt{QUERY}: ``\{entity\}" is an instance of what entity type?\\
        \textcolor{blue}{target}: \{entity\_type\}
        
        \item \textcolor{red}{source}: \texttt{TEXT}: \{sentence\} \texttt{QUERY}: In the text, is ``\{entity\}" an instance of ``\{entity\_type\}"?\\
        \textcolor{blue}{target}: Yes
        
        \item \textcolor{red}{source}: \texttt{TEXT}: \{sentence\} \texttt{QUERY}: In the text, is ``\{entity\}" an instance of ``\{other\_entity\_type\}"?\\
        \textcolor{blue}{target}: No
    \end{enumerate}
\end{itemize}
\paragraph{Summary}
With $($introduction, abstract$)$ pairs, we design prompts that ask a model to write the abstract of a paper based on its introduction part. We use \{length\} to represent the length of the target abstract.

\begin{itemize}
    \item Generation format prompts
    \begin{enumerate}
        \item \textcolor{red}{source}: \texttt{TEXT}: \{introduction\} \texttt{QUERY}: Can you summarize the previous text?\\
        \textcolor{blue}{target}: \{abstract\}
        
        \item \textcolor{red}{source}: \texttt{TEXT}: \{introduction\} \texttt{QUERY}: Generate a summary for the text.\\
        \textcolor{blue}{target}: \{abstract\}
        
        \item \textcolor{red}{source}: \texttt{TEXT}: \{introduction\} \texttt{QUERY}: Can you generate a \{length\}-word summary for the previous text?\\
        \textcolor{blue}{target}: \{abstract\}
        
        \item \textcolor{red}{source}: \texttt{TEXT}: \{introduction\} \texttt{QUERY}: In around \{length\} words, summarize the article.\\
        \textcolor{blue}{target}: \{abstract\}
        
        \item \textcolor{red}{source}: \texttt{TEXT}: \{introduction\} \texttt{QUERY}: In around \{length\} words, briefly describe what the previous paragraph talks about.\\
        \textcolor{blue}{target}: \{abstract\}
        
        \item \textcolor{red}{source}: \texttt{TEXT}: \{introduction\} \texttt{QUERY}: Please write a summary of about \{length\} words based on the previous article.\\
        \textcolor{blue}{target}: \{abstract\}
        
        \item \textcolor{red}{source}: \texttt{TEXT}: \{introduction\} \texttt{QUERY}: Can you express the main content of the text?\\
        \textcolor{blue}{target}: \{abstract\}
        
        \item \textcolor{red}{source}: \texttt{TEXT}: \{introduction\} \texttt{QUERY}: Condense the text down to the essentials.\\
        \textcolor{blue}{target}: \{abstract\}
        
        \item \textcolor{red}{source}: \texttt{TEXT}: \{introduction\} \texttt{QUERY}: What is the previous article about? Please summarize in about \{length\} words.\\
        \textcolor{blue}{target}: \{abstract\}
        
        \item \textcolor{red}{source}: \texttt{TEXT}: \{introduction\} \texttt{QUERY}: Summarize the preceding text in your own words.\\
        \textcolor{blue}{target}: \{abstract\}
    \end{enumerate}
\end{itemize}

\section{Training Details}
\label{app:training-details}

\paragraph{Topic classification}
Models are fine-tuned for 100,000 steps. We sampled 1,000 training data from AGNews \cite{Zhang2015CharacterlevelCN} as our first validation set, and 1,000 training data from DBPedia \cite{DBLP:conf/nips/ZhangZL15} as our second validation set. We selected the checkpoint that achieved the highest average accuracy on the two validation sets.

\paragraph{Sentiment classification}
Models are fine-tuned for 100,000 steps. We sampled 1,000 training data from Amazon \cite{DBLP:conf/nips/ZhangZL15} as our first validation set and 1,000 training data from IMDB \cite{maas-EtAl:2011:ACL-HLT2011} as our second validation set. We selected the checkpoint with the highest average accuracy on the two validation sets.

\paragraph{Information extraction}
Models are fine-tuned for 200,000 steps in the first training stage. We sampled 1,000 training data from CoNLL03 \cite{tjong-kim-sang-de-meulder-2003-introduction} as our first validation set and 1,000 training data from Wiki80 as our second validation set. When selecting the model from the first stage, we used the second validation set only and selected the checkpoint that achieved the highest accuracy. Models are fine-tuned for 200,000 steps in the second training stage. When selecting the model from the second stage, we used both the first and second validation sets. We selected the checkpoint that achieved the highest average rank on the two validation sets (i.e., we rank the performance of a checkpoint on each validation set).

\paragraph{Natural language inference}
Models are fine-tuned for 200,000 steps. We sampled 1,000 training data from MultiNLI \cite{N18-1101} as our validation set and selected the checkpoint that achieved the highest accuracy.

\paragraph{Intent detection}
Models are fine-tuned for 100,000 steps. We sampled 1,000 training data from SNIPS \cite{https://doi.org/10.48550/arxiv.1805.10190} as our first validation set and 1,000 training data from FB \cite{schuster-etal-2019-cross-lingual} as our second validation set. We selected the checkpoint that achieved the highest average accuracy on the two validation sets.

\paragraph{Fact retrieval}
Models are fine-tuned for 400,000 steps. Using Wikidata, we have constructed three types of relation signals (i.e., querying subject, object, and relation). We randomly sampled 1,000 data (that are not in the training set) for each of them to construct a validation set. We selected the checkpoint that achieved the highest accuracy on this validation set.

\paragraph{Temporal reasoning}
Models are fine-tuned for 200,000 steps. We used TRACIE \cite{ZRNKSR21} training data as our first validation set and sampled 1,000 UDS-T temporal-relation training data as our second validation set. We selected the checkpoint with the highest average accuracy on the two validation sets.

\paragraph{Word sense disambiguation}
Models are fine-tuned for 100,000 steps. We used Senseval3 \cite{snyder-palmer-2004-english} dataset as our validation set and selected the checkpoint that achieved the highest f1 score.

\paragraph{Summary}
Models are fine-tuned for 500,000 steps. We sampled 200 training samples from CNNDM \cite{NIPS2015_afdec700} as our first validation set and 200 training samples from XSUM \cite{Narayan2018DontGM} as our second validation set. We selected the checkpoint that achieved the highest average ROUGE-1 on the two validation sets.
    
\paragraph{All}
\begin{itemize}
    \item Stage 1: Models are fine-tuned for 1,000,000 steps. When selecting a model from the first training stage, we used the following validation sets mentioned above: AGNews, DBPedia, Amazon, IMDB, Wiki80, MultiNLI, SNIPS, FB, UDS-T, TRACIE. We selected the checkpoint with the highest average rank on those validation sets.
    \item Stage 2: Models are fine-tuned for 2,000,000 steps. When selecting a model from the second training stage, we used all the validation sets mentioned above in different tasks. We selected the checkpoint with the highest average rank on those validation sets.
\end{itemize}

\section{Prompts for evaluation datasets}
Please note that we use our designed prompts for T0pp as well, except that we remove the special markers (i.e., \texttt{TEXT} and \texttt{QUERY}) for T0pp prompts.
\label{app:eval-prompts}
\subsection{Topic Classification}
\paragraph{subj} With $($text, label$)$ pairs, we construct the following prompts.
\begin{enumerate}
    \item \textcolor{red}{source}: \texttt{TEXT}: \{text\} \texttt{QUERY}: Is this text ``subjective" or ``objective"?\\
    \textcolor{blue}{target}: \{label\}
    \item \textcolor{red}{source}: \texttt{TEXT}: \{text\} \texttt{QUERY}: Is this text subjective? ``Yes" or ``No"?\\
    \textcolor{blue}{target}: Yes (if \{label\} is ``subjective") / No (if \{label\} is ``objective")
    \item \textcolor{red}{source}: \texttt{TEXT}: \{text\} \texttt{QUERY}: Is this text objective? ``Yes" or ``No"?\\
    \textcolor{blue}{target}: Yes (if \{label\} is ``objective") / No (if \{label\} is ``subjective")
    \item \textcolor{red}{source}: \texttt{TEXT}: \{text\} \texttt{QUERY}: Does this text contain the author's subjectivity? ``Yes" or ``No"?\\
    \textcolor{blue}{target}: Yes (if \{label\} is ``subjective") / No (if \{label\} is ``objective")
    \item \textcolor{red}{source}: \texttt{TEXT}: \{text\} \texttt{QUERY}: Is this text an objective description? ``Yes" or ``No"?\\
    \textcolor{blue}{target}: Yes (if \{label\} is ``objective") / No (if \{label\} is ``subjective")
\end{enumerate}

\paragraph{qc \& yahoo\_answers\_topics} With $($text, label$)$ pairs, we construct the following prompts. Here, we use \{choices\_with\_or\} (e.g. ``option1", ``option2", or ``option3") and \{choices\_without\_or\} (e.g. ``option1", ``option2", ``option3") to refer to all the available options.
\begin{enumerate}
    \item \textcolor{red}{source}: \texttt{TEXT}: \{text\} \texttt{QUERY}: What's this text about? \{choices\_with\_or\}?\\
    \textcolor{blue}{target}: \{label\}
    
    \item \textcolor{red}{source}: \texttt{TEXT}: \{text\} \texttt{QUERY}: Classify this text. You may choose from \{choices\_without\_or\}.\\
    \textcolor{blue}{target}: \{label\}
    
    \item \textcolor{red}{source}: \texttt{TEXT}: \{text\} \texttt{QUERY}: Select a class from the following that best describes the text: \{choices\_without\_or\}.\\
    \textcolor{blue}{target}: \{label\}
    
    \item \textcolor{red}{source}: \texttt{TEXT}: \{text\} \texttt{QUERY}: How would you categorize this text? \{choices\_with\_or\}?\\
    \textcolor{blue}{target}: \{label\}
    
    \item \textcolor{red}{source}: \texttt{TEXT}: \{text\} \texttt{QUERY}: What is the topic of the text? \{choices\_with\_or\}?\\
    \textcolor{blue}{target}: \{label\}
    
\end{enumerate}

\paragraph{hate\_speech18} With $($text, label$)$ pairs, we construct the following prompts.
\begin{enumerate}
    \item \textcolor{red}{source}: \texttt{TEXT}: \{text\} \texttt{QUERY}: Does this text contain hate speech? ``Yes", ``No" or ``Unknown"?\\
    \textcolor{blue}{target}: \{label\}
    
    \item \textcolor{red}{source}: \texttt{TEXT}: \{text\} \texttt{QUERY}: Does this text convey the author's hatred towards something or someone? ``Yes", ``No" or ``Unknown"?\\
    \textcolor{blue}{target}: \{label\}
    
    \item \textcolor{red}{source}: \texttt{TEXT}: \{text\} \texttt{QUERY}: Is this a hateful text? ``Yes", ``No" or ``Unknown"?\\
    \textcolor{blue}{target}: \{label\}
    
    \item \textcolor{red}{source}: \texttt{TEXT}: \{text\} \texttt{QUERY}: Does this text convey hatred? ``Yes", ``No" or ``Unknown"?\\
    \textcolor{blue}{target}: \{label\}
    
    \item \textcolor{red}{source}: \texttt{TEXT}: \{text\} \texttt{QUERY}: Is there hate speech in the text? ``Yes", ``No" or ``Unknown"?\\
    \textcolor{blue}{target}: \{label\}
\end{enumerate}

\paragraph{tweet\_eval{/}emotion} With $($text, label$)$ pairs, we construct the following prompts. Here, we use \{choices\_with\_or\} (e.g. ``option1", ``option2", or ``option3") and \{choices\_without\_or\} (e.g. ``option1", ``option2", ``option3") to refer to all the available options.
\begin{enumerate}
    \item \textcolor{red}{source}: \texttt{TEXT}: \{text\} \texttt{QUERY}: What's the emotion of this text? \{choices\_with\_or\}?\\
    \textcolor{blue}{target}: \{label\}
    
    \item \textcolor{red}{source}: \texttt{TEXT}: \{text\} \texttt{QUERY}: Select an emotion from the following that best describes the text: \{choices\_without\_or\}.\\
    \textcolor{blue}{target}: \{label\}
    
    \item \textcolor{red}{source}: \texttt{TEXT}: \{text\} \texttt{QUERY}: What is the emotion expressed in this message? \{choices\_with\_or\}?\\
    \textcolor{blue}{target}: \{label\}
    
    \item \textcolor{red}{source}: \texttt{TEXT}: \{text\} \texttt{QUERY}: How will you feel about the message? \{choices\_with\_or\}?\\
    \textcolor{blue}{target}: \{label\}
    
    \item \textcolor{red}{source}: \texttt{TEXT}: \{text\} \texttt{QUERY}: What emotion does the writer express for the message? \{choices\_with\_or\}?\\
    \textcolor{blue}{target}: \{label\}
\end{enumerate}

\paragraph{tweet\_eval{/}hate} With $($text, label$)$ pairs, we construct the following prompts. 
\begin{enumerate}
    \item \textcolor{red}{source}: \texttt{TEXT}: \{text\} \texttt{QUERY}: Does this text contain hate speech? ``Yes" or ``No"?\\
    \textcolor{blue}{target}: Yes (if \{label\} is ``hate") / No (if \{label\} is ``non-hate")
    
    \item \textcolor{red}{source}: \texttt{TEXT}: \{text\} \texttt{QUERY}: Does this text convey the author's hatred towards something or someone? ``Yes" or ``No"?\\
    \textcolor{blue}{target}: Yes (if \{label\} is ``hate") / No (if \{label\} is ``non-hate")
    
    \item \textcolor{red}{source}: \texttt{TEXT}: \{text\} \texttt{QUERY}: Is this a hateful text? ``Yes" or ``No"?\\
    \textcolor{blue}{target}: Yes (if \{label\} is ``hate") / No (if \{label\} is ``non-hate")
    
    \item \textcolor{red}{source}: \texttt{TEXT}: \{text\} \texttt{QUERY}: Does this text convey hatred? ``Yes" or ``No"?\\
    \textcolor{blue}{target}: Yes (if \{label\} is ``hate") / No (if \{label\} is ``non-hate")
    
    \item \textcolor{red}{source}: \texttt{TEXT}: \{text\} \texttt{QUERY}: Is there hate speech in the text? ``Yes" or ``No"?\\
    \textcolor{blue}{target}: Yes (if \{label\} is ``hate") / No (if \{label\} is ``non-hate")
\end{enumerate}

\paragraph{tweet\_eval{/}irony} With $($text, label$)$ pairs, we construct the following prompts.
\begin{enumerate}
    \item \textcolor{red}{source}: \texttt{TEXT}: \{text\} \texttt{QUERY}: Does this text contain irony? ``Yes" or ``No"?\\
    \textcolor{blue}{target}: Yes (if \{label\} is ``irony") / No (if \{label\} is ``non-irony")
    
    \item \textcolor{red}{source}: \texttt{TEXT}: \{text\} \texttt{QUERY}: Is this an ironic text? ``Yes" or ``No"?\\
    \textcolor{blue}{target}: Yes (if \{label\} is ``irony") / No (if \{label\} is ``non-irony")
    
    \item \textcolor{red}{source}: \texttt{TEXT}: \{text\} \texttt{QUERY}: Is there irony in this text? ``Yes" or ``No"?\\
    \textcolor{blue}{target}: Yes (if \{label\} is ``irony") / No (if \{label\} is ``non-irony")
    
    \item \textcolor{red}{source}: \texttt{TEXT}: \{text\} \texttt{QUERY}: Does the author of the text express irony? ``Yes" or ``No"?\\
    \textcolor{blue}{target}: Yes (if \{label\} is ``irony") / No (if \{label\} is ``non-irony")
    
    \item \textcolor{red}{source}: \texttt{TEXT}: \{text\} \texttt{QUERY}: Is this text ``ironic" or ``not ironic"?\\
    \textcolor{blue}{target}: ironic (if \{label\} is ``irony") / not ironic (if \{label\} is ``non-irony")
\end{enumerate}

\paragraph{tweet\_eval{/}offensive} With $($text, label$)$ pairs, we construct the following prompts.
\begin{enumerate}
    \item \textcolor{red}{source}: \texttt{TEXT}: \{text\} \texttt{QUERY}: Does this text contain offensive content? ``Yes" or ``No"?\\
    \textcolor{blue}{target}: Yes (if \{label\} is ``offensive") / No (if \{label\} is ``non-offensive")
    
    \item \textcolor{red}{source}: \texttt{TEXT}: \{text\} \texttt{QUERY}: Does this text convey the author's offense towards something or someone? ``Yes" or ``No"?\\
    \textcolor{blue}{target}: Yes (if \{label\} is ``offensive") / No (if \{label\} is ``non-offensive")
    
    \item \textcolor{red}{source}: \texttt{TEXT}: \{text\} \texttt{QUERY}: Is this an offensive text? ``Yes" or ``No"?\\
    \textcolor{blue}{target}: Yes (if \{label\} is ``offensive") / No (if \{label\} is ``non-offensive")
    
    \item \textcolor{red}{source}: \texttt{TEXT}: \{text\} \texttt{QUERY}: Does this text convey offense? ``Yes" or ``No"?\\
    \textcolor{blue}{target}: Yes (if \{label\} is ``offensive") / No (if \{label\} is ``non-offensive")
    
    \item \textcolor{red}{source}: \texttt{TEXT}: \{text\} \texttt{QUERY}: Is there offensive content in the text? ``Yes" or ``No"?\\
    \textcolor{blue}{target}: Yes (if \{label\} is ``offensive") / No (if \{label\} is ``non-offensive")
\end{enumerate}

\subsection{Sentiment Classification}
\paragraph{financial\_phrasebank} With $($text, label$)$ pairs, we construct the following prompts.
\begin{enumerate}
    \item \textcolor{red}{source}: \texttt{TEXT}: \{text\} \texttt{QUERY}: Is this news positive, negative or neutral?\\
    \textcolor{blue}{target}: \{label\}
    
    \item \textcolor{red}{source}: \texttt{TEXT}: \{text\} \texttt{QUERY}: What's the sentiment of this text? ``Positive", ``Negative" or ``Neutral"?\\
    \textcolor{blue}{target}: Positive (if \{label\} is ``positive") / Negative (if \{label\} is ``negative") / Neutral (if \{label\} is ``neutral")
    
    \item \textcolor{red}{source}: \texttt{TEXT}: \{text\} \texttt{QUERY}: Can you judge the sentiment of this text? The options are ``Positive", ``Negative" and ``Neutral".\\
    \textcolor{blue}{target}: Positive (if \{label\} is ``positive") / Negative (if \{label\} is ``negative") / Neutral (if \{label\} is ``neutral")
    
    \item \textcolor{red}{source}: \texttt{TEXT}: \{text\} \texttt{QUERY}: Assign the correct sentiment to this news. Please choose from ``Positive", ``Negative", ``Neutral".\\
    \textcolor{blue}{target}: Positive (if \{label\} is ``positive") / Negative (if \{label\} is ``negative") / Neutral (if \{label\} is ``neutral")
    
    \item \textcolor{red}{source}: \texttt{TEXT}: \{text\} \texttt{QUERY}: Judge the sentiment of the text. You can choose from ``Positive", ``Negative", ``Neutral".\\
    \textcolor{blue}{target}: Positive (if \{label\} is ``positive") / Negative (if \{label\} is ``negative") / Neutral (if \{label\} is ``neutral")
\end{enumerate}

\paragraph{mr \& sst2} With $($text, label$)$ pairs, we construct the following prompts.
\begin{enumerate}
    \item \textcolor{red}{source}: \texttt{TEXT}: \{text\} \texttt{QUERY}: What's the sentiment of this text? ``Positive", ``Negative" or ``Neutral"?\\
    \textcolor{blue}{target}: Positive (if \{label\} is ``positive") / Negative (if \{label\} is ``negative") / Neutral (if \{label\} is ``neutral")
    
    \item \textcolor{red}{source}: \texttt{TEXT}: \{text\} \texttt{QUERY}: Can you judge the sentiment of this text? The options are ``Positive", ``Negative" and ``Neutral".\\
    \textcolor{blue}{target}: Positive (if \{label\} is ``positive") / Negative (if \{label\} is ``negative") / Neutral (if \{label\} is ``neutral")
    
    \item \textcolor{red}{source}: \texttt{TEXT}: \{text\} \texttt{QUERY}: Does it seem like the reviewer who wrote this review liked the movie? ``Yes" or ``No"?\\
    \textcolor{blue}{target}: Yes (if \{label\} is ``positive") / No (if \{label\} is ``negative")
    
    \item \textcolor{red}{source}: \texttt{TEXT}: \{text\} \texttt{QUERY}: Assign the correct sentiment to this review. Please choose from ``Positive", ``Negative", ``Neutral".\\
    \textcolor{blue}{target}: Positive (if \{label\} is ``positive") / Negative (if \{label\} is ``negative") / Neutral (if \{label\} is ``neutral")
    
    \item \textcolor{red}{source}: \texttt{TEXT}: \{text\} \texttt{QUERY}: Does the reviewer like this movie based on the review text? ``Yes" or ``No"?\\
    \textcolor{blue}{target}: Yes (if \{label\} is ``positive") / No (if \{label\} is ``negative")
\end{enumerate}

\subsection{Information Extraction}
\paragraph{conll03 \& OntoNotes 5.0 \& wnut17} For the named entity recognition task, we adopt a two-step strategy where we first identify all the entities contained in the text and then determine the entity type for each entity. For the first stage, with $($text, entities$)$ paris, we construct the following prompts.
\begin{enumerate}
    \item \textcolor{red}{source}: \texttt{TEXT}: \{text\} \texttt{QUERY}: Can you find all the entities in the text?\\
    \textcolor{blue}{target}: \{entities\}
    
    \item \textcolor{red}{source}: \texttt{TEXT}: \{text\} \texttt{QUERY}: List all the entities in the text.\\
    \textcolor{blue}{target}: \{entities\}
    
    \item \textcolor{red}{source}: \texttt{TEXT}: \{text\} \texttt{QUERY}: What are the entities in the previous text?\\
    \textcolor{blue}{target}: \{entities\}
    
    \item \textcolor{red}{source}: \texttt{TEXT}: \{text\} \texttt{QUERY}: Please enumerate all the named entities in the above paragraph.\\
    \textcolor{blue}{target}: \{entities\}
    
    \item \textcolor{red}{source}: \texttt{TEXT}: \{text\} \texttt{QUERY}: Identify all the named entities in the previous paragraph.\\
    \textcolor{blue}{target}: \{entities\}
\end{enumerate}

Then, with all the entities predicted by the model, we decide the fine-grained entity type for each entity. We construct the following prompts. Here, we use \{choices\_with\_or\} (e.g. ``option1", ``option2", or ``option3") and \{choices\_without\_or\} (e.g. ``option1", ``option2", ``option3") to refer to all the available options.
\begin{enumerate}
    \item \textcolor{red}{source}: \texttt{TEXT}: \{text\} \texttt{QUERY}: What's the entity type of ``\{entity\}"? \{choices\_with\_or\}?
    
    \item \textcolor{red}{source}: \texttt{TEXT}: \{text\} \texttt{QUERY}: Can you choose an appropriate class from the following list for ``\{entity\}"? The options are \{choices\_without\_or\}.
    
    \item \textcolor{red}{source}: \texttt{TEXT}: \{text\} \texttt{QUERY}: ``\{entity\}" is an instance of which of the following? \{choices\_with\_or\}?
    
    \item \textcolor{red}{source}: \texttt{TEXT}: \{text\} \texttt{QUERY}: Given a list of categories: \{choices\_without\_or\}, what category does ``\{entity\}" belong to?
    
    \item \textcolor{red}{source}: \texttt{TEXT}: \{text\} \texttt{QUERY}: Choose the correct entity type for ``\{entity\}" from the following: \{choices\_without\_or\}.
\end{enumerate}

\paragraph{semeval\_rel \& wiki80} With $($context, head, tail, relation$)$ quads, we construct the following prompts. Here, we use \{choices\_with\_or\} (e.g. ``option1", ``option2", or ``option3") and \{choices\_without\_or\} (e.g. ``option1", ``option2", ``option3") to refer to all the available options.
\begin{enumerate}
    \item \textcolor{red}{source}: \texttt{TEXT}: \{context\} \texttt{QUERY}: Given the subject ``\{head\}" and object ``\{tail\}", what's the relation between them? \{choices\_with\_or\}?\\
    \textcolor{blue}{target}: \{relation\}
    
    \item \textcolor{red}{source}: \texttt{TEXT}: \{context\} \texttt{QUERY}: What's the relationship between the subject ``\{head\}" and object ``\{tail\}"? \{choices\_with\_or\}?\\
    \textcolor{blue}{target}: \{relation\}
    
    \item \textcolor{red}{source}: \texttt{TEXT}: \{context\} \texttt{QUERY}: Here is a list of relations: \{choices\_without\_or\}. Which one applies to ``\{head\}" and ``\{tail\}"?\\
    \textcolor{blue}{target}: \{relation\}
    
    \item \textcolor{red}{source}: \texttt{TEXT}: \{context\} \texttt{QUERY}: Given the subject ``\{head\}" and object ``\{tail\}", can you select the correct relationship from the following: \{choices\_without\_or\}?\\
    \textcolor{blue}{target}: \{relation\}
    
    \item \textcolor{red}{source}: \texttt{TEXT}: \{context\} \texttt{QUERY}: Which of the following is the correct relation between ``\{head\}" and ``\{tail\}"? \{choices\_with\_or\}?\\
    \textcolor{blue}{target}: \{relation\}
\end{enumerate}

\subsection{Natural Language Inference}
With $($premise, hypothesis, label$)$ triples, we construct the following prompts.
\paragraph{anli \& cb \& multi\_nli \& sick \& snli} 
\begin{enumerate}
    \item \textcolor{red}{source}: \texttt{TEXT}: \{premise\} \texttt{QUERY}: Based on the previous text, can we infer that ``\{hypothesis\}"? ``Yes" or ``No" or ``Maybe"?\\
    \textcolor{blue}{target}: Yes (if \{label\} is ``entailment") / No (if \{label\} is ``contradiction") / Maybe (if \{label\} is ``neutral")
    
    \item \textcolor{red}{source}: \texttt{TEXT}: \{premise\} \texttt{QUERY}: \{hypothesis\} ``True" or ``False" or ``Unknown"?\\
    \textcolor{blue}{target}: True (if \{label\} is ``entailment") / False (if \{label\} is ``contradiction") / Unknown (if \{label\} is ``neutral")
    
    \item \textcolor{red}{source}: \texttt{TEXT}: \{premise\} \texttt{QUERY}: Based on that information, is the claim ``\{hypothesis\}" ``true", ``false" or ``inconclusive"?\\
    \textcolor{blue}{target}: true (if \{label\} is ``entailment") / false (if \{label\} is ``contradiction") / inconclusive (if \{label\} is ``neutral")
    
    \item \textcolor{red}{source}: \texttt{TEXT}: \{premise\} \texttt{QUERY}: Does it imply that ``\{hypothesis\}"? ``Yes", ``No" or ``Maybe"?\\
    \textcolor{blue}{target}: Yes (if \{label\} is ``entailment") / No (if \{label\} is ``contradiction") / Maybe (if \{label\} is ``neutral")
    
    \item \textcolor{red}{source}: \texttt{TEXT}: \{premise\} \texttt{QUERY}: ``\{hypothesis\}" is ``guaranteed", ``possible" or ``impossible"?\\
    \textcolor{blue}{target}: guaranteed (if \{label\} is ``entailment") / impossible (if \{label\} is ``contradiction") / possible (if \{label\} is ``neutral")
\end{enumerate}

\paragraph{rte} 
\begin{enumerate}
    \item \textcolor{red}{source}: \texttt{TEXT}: \{premise\} \texttt{QUERY}: Based on the previous text, can we infer that ``\{hypothesis\}"? ``Yes" or ``No"?\\
    \textcolor{blue}{target}: Yes (if \{label\} is ``entailment") / No (if \{label\} is ``contradiction")
    
    \item \textcolor{red}{source}: \texttt{TEXT}: \{premise\} \texttt{QUERY}: \{hypothesis\} ``True" or ``False"?\\
    \textcolor{blue}{target}:  True (if \{label\} is ``entailment") / False (if \{label\} is ``contradiction")
    
    \item \textcolor{red}{source}: \texttt{TEXT}: \{premise\} \texttt{QUERY}: Based on that information, is the claim ``\{hypothesis\}" ``true" or ``false" ?\\
    \textcolor{blue}{target}: true (if \{label\} is ``entailment") / false (if \{label\} is ``contradiction")
    
    \item \textcolor{red}{source}: \texttt{TEXT}: \{premise\} \texttt{QUERY}: Does it imply that ``\{hypothesis\}"? ``Yes" or ``No"?\\
    \textcolor{blue}{target}:  Yes (if \{label\} is ``entailment") / No (if \{label\} is ``contradiction")
    
    \item \textcolor{red}{source}: \texttt{TEXT}: \{premise\} \texttt{QUERY}: ``\{hypothesis\}" is ``guaranteed" or ``impossible"?\\
    \textcolor{blue}{target}:  guaranteed (if \{label\} is ``entailment") / impossible (if \{label\} is ``contradiction")
\end{enumerate}

\subsection{Intent Detection}
\paragraph{atis \& banking77 \& clinc150 \& fb \& hint3 \& nlued \& slurp \& snips} With $($text, label$)$ pairs, we construct the following prompts. Here, we use \{choices\_with\_or\} (e.g. ``option1", ``option2", or ``option3") and \{choices\_without\_or\} (e.g. ``option1", ``option2", ``option3") to refer to all the available options.
\begin{enumerate}
    \item \textcolor{red}{source}: \texttt{TEXT}: \{text\} \texttt{QUERY}: What's the intent of this text? \{choices\_with\_or\}?\\
    \textcolor{blue}{target}: \{label\}
    
    \item \textcolor{red}{source}: \texttt{TEXT}: \{text\} \texttt{QUERY}: Given this text, what could be the likely goal? \{choices\_with\_or\}?\\
    \textcolor{blue}{target}: \{label\}
    
    \item \textcolor{red}{source}: \texttt{TEXT}: \{text\} \texttt{QUERY}: The previous utterance is typically associated with which intent? \{choices\_with\_or\}?\\
    \textcolor{blue}{target}: \{label\}
    
    \item \textcolor{red}{source}: \texttt{QUERY}: What is the purpose of saying ``\{text\}"? Choose the most appropriate one: \{choices\_without\_or\}.\\
    \textcolor{blue}{target}: \{label\}
    
    \item \textcolor{red}{source}: \texttt{QUERY}: Choose the most probable purpose if you say ``\{text\}". The options are \{choices\_without\_or\}.\\
    \textcolor{blue}{target}: \{label\}
\end{enumerate}

\subsection{Fact Retrieval}
\paragraph{LAMA-TREx} With $($subject, relation, object$)$ triples, we construct the following prompts.
\begin{enumerate}
    \item \textcolor{red}{source}: \texttt{QUERY}: Given the subject ``\{subject\}" and relation ``\{relation\}", what's the correct object?\\
    \textcolor{blue}{target}: \{object\}
    
    \item \textcolor{red}{source}: \texttt{QUERY}: ``\{subject\}" can form a ``\{relation\}" relationship with what entity?\\
    \textcolor{blue}{target}: \{object\}
    
    \item \textcolor{red}{source}: \texttt{QUERY}: With what entity can ``\{subject\}" form a ``\{relation\}" relationship?\\
    \textcolor{blue}{target}: \{object\}
    
    \item \textcolor{red}{source}: \texttt{QUERY}: Given the subject ``\{subject\}" and relation ``\{relation\}", can you find an appropriate object?\\
    \textcolor{blue}{target}: \{object\}
    
    \item \textcolor{red}{source}: \texttt{QUERY}: What entity is ``\{subject\}" in a ``\{relation\}" relation with?\\
    \textcolor{blue}{target}: \{object\}
\end{enumerate}

\subsection{Temporal Reasoning} 
The datasets are transformed into natural language inference format datasets where each sample is in the form of $($premise, hypothesis, label$)$. Therefore, we apply the prompts we construct for the natural language inference task.
\paragraph{TRACIE \& UDS-T}
\begin{enumerate}
    \item \textcolor{red}{source}: \texttt{TEXT}: \{premise\} \texttt{QUERY}: Based on the previous text, can we infer that ``\{hypothesis\}"? ``Yes" or ``No"?\\
    \textcolor{blue}{target}: Yes (if \{label\} is ``entailment") / No (if \{label\} is ``contradiction")
    
    \item \textcolor{red}{source}: \texttt{TEXT}: \{premise\} \texttt{QUERY}: \{hypothesis\} ``True" or ``False"?\\
    \textcolor{blue}{target}:  True (if \{label\} is ``entailment") / False (if \{label\} is ``contradiction")
    
    \item \textcolor{red}{source}: \texttt{TEXT}: \{premise\} \texttt{QUERY}: Based on that information, is the claim ``\{hypothesis\}" ``true" or ``false" ?\\
    \textcolor{blue}{target}: true (if \{label\} is ``entailment") / false (if \{label\} is ``contradiction")
    
    \item \textcolor{red}{source}: \texttt{TEXT}: \{premise\} \texttt{QUERY}: Does it imply that ``\{hypothesis\}"? ``Yes" or ``No"?\\
    \textcolor{blue}{target}:  Yes (if \{label\} is ``entailment") / No (if \{label\} is ``contradiction")
    
    \item \textcolor{red}{source}: \texttt{TEXT}: \{premise\} \texttt{QUERY}: ``\{hypothesis\}" is ``guaranteed" or ``impossible"?\\
    \textcolor{blue}{target}:  guaranteed (if \{label\} is ``entailment") / impossible (if \{label\} is ``contradiction")
\end{enumerate}

\subsection{Word Sense Disambiguation}
\paragraph{semeval2007 \& semeval2013 \& semeval2015 \& senseval2} With $($context, word, label$)$ triples, we construct the following prompts. Here, we use \{choices\_with\_or\} (e.g. ``option1", ``option2", or ``option3") and \{choices\_without\_or\} (e.g. ``option1", ``option2", ``option3") to refer to all the available options.
\begin{enumerate}
    \item \textcolor{red}{source}: \texttt{TEXT}: \{context\} \texttt{QUERY}: What does the word ``\{word\}" mean in the previous text? \{choices\_with\_or\}?\\
    \textcolor{blue}{target}: \{label\}
    
    \item \textcolor{red}{source}: \texttt{TEXT}: \{context\} \texttt{QUERY}: What's the meaning of ``\{word\}" in the text? \{choices\_with\_or\}?\\
    \textcolor{blue}{target}: \{label\}
    
    \item \textcolor{red}{source}: \texttt{TEXT}: \{context\} \texttt{QUERY}: Given the following meanings of the word ``\{word\}": \{choices\_without\_or\}. Can you choose the right meaning according to the previous context?\\
    \textcolor{blue}{target}: \{label\}
    
    \item \textcolor{red}{source}: \texttt{TEXT}: \{context\} \texttt{QUERY}: How to understand the word ``\{word\}" in the previous text? You may choose from: \{choices\_without\_or\}.\\
    \textcolor{blue}{target}: \{label\}
    
    \item \textcolor{red}{source}: \texttt{TEXT}: \{context\} \texttt{QUERY}: Can you explain the meaning of ``\{word\}" in the previous text? The options are \{choices\_without\_or\}.\\
    \textcolor{blue}{target}: \{label\}
\end{enumerate}

\subsection{Summary}
\paragraph{SciTLDR \& Reddit-TIFU \& Multi-Xscience \& WikiSum \& GovReport \&BillSum \& BigPatent} With $($text, summary$)$ pairs, we construct the following prompts. Here, we use \{avg\_length\} to represent the average length of the summaries in the target dataset.
\begin{enumerate}
    \item \textcolor{red}{source}: \texttt{TEXT}: \{text\} \texttt{QUERY}: Can you summarize the previous text in around \{avg\_length\} words?\\
    \textcolor{blue}{target}: \{summary\}
    
    \item \textcolor{red}{source}: \texttt{TEXT}: \{text\} \texttt{QUERY}: Can you generate a \{avg\_length\}-word summary for the previous text?\\
    \textcolor{blue}{target}: \{summary\}
    
    \item \textcolor{red}{source}: \texttt{TEXT}: \{text\} \texttt{QUERY}: In around \{avg\_length\} words, write a TLDR (Too Long Didn't Read) summary for the above text.\\
    \textcolor{blue}{target}: \{summary\}
    
    \item \textcolor{red}{source}: \texttt{TEXT}: \{text\} \texttt{QUERY}: What is the previous article about? Please summarize in about \{avg\_length\} words.\\
    \textcolor{blue}{target}: \{summary\}
    
    \item \textcolor{red}{source}: \texttt{TEXT}: \{text\} \texttt{QUERY}: Please write a summary of about \{avg\_length\} words based on the previous article.\\
    \textcolor{blue}{target}: \{summary\}
\end{enumerate}

\section{Signal Collection for Gaokao-English}
\label{app:gaokao-english-signal}
\paragraph{Comprehension} We took the training split of those datasets, randomly selected 15,000 samples from each of them, and combined them into our training data. For validation, we took the validation split of those datasets, randomly selected 50 samples from each of them, and combined them into our validation set.

\paragraph{Language Usage} We prepared the signals for each question type within this category. The training data and validation data are a combination of the following: (i) For cloze (multiple-choice) questions, we used cloze questions from CLOTH dataset \cite{xie-etal-2018-large}. More specifically, we used the questions from high school examinations and randomly selected 40,000 training samples as our training data and 500 validation samples as our validation data. (ii) For cloze (hint) questions, we automatically constructed training and validation data. We used DailyMail as our data source and collected 20000 articles from each of the following categories: money, news, sport, health, science and technology, and travel. We considered each sentence as an individual data sample. By analyzing the cloze questions in Gaokao-English, we summarized that words with the following parts-of-speech are more likely to be chosen: NN, VBD, JJ, RB, NNS, VBN, VBG, VB, VBP, IN, DT, VBZ, JJR, CC, and WP. Therefore, we only consider words with those parts-of-speech as cloze questions. We set a cloze rate of 0.1 (e.g., if we have a sentence of ten words, we will mask out at most one word to construct a cloze question.). With 0.5 probability, we gave a derivation word of the target word as a hint, and with another 0.5 probability, we gave no hint at all. We used WordNet to find derivationally related forms of a word (e.g., poorly, poor). At last, we selected 60,000 training data (42,000 with a hint and 18,000 without a hint) and 1,000 validation data (700 with a hint and 300 without a hint). The hint vs. non-hint ratio is determined by examining the Gaokao-English questions. (iii) For reading (cloze) questions, we automatically constructed training and validation data using the following ways. We used the same data source as (ii). For each article, we set the cloze rate as 0.25 (e.g., if there are eight sentences within an article, we would mask out at most two sentences from that article). We ignored articles with fewer than eight sentences. Then we would sample non-consecutive sentences from an article and ask the model to fill those sentences correctly. At last, we got 40,000 training data and 500 validation data.

\paragraph{Writing} The training data and validation data are a combination of the following: (i) For writing (grammar) questions, we used the data from BEA 2019 Shared Task \cite{bryant-etal-2019-bea} and we selected the samples that were written by beginners since the mistakes there largely matched with the grammar errors tested by Gaokao-English. We randomly selected 300 as our training data and 100 as our validation data. (ii) For writing (essay writing) questions, we collected questions from real high school exams and Chinese graduate school entrance exams. In total, we collected 263 data for training and 28 data for validation. Notice that we did not use any data that are in our Gaokao-Benchmark.

\newpage

\section{\includegraphics[scale=0.01]{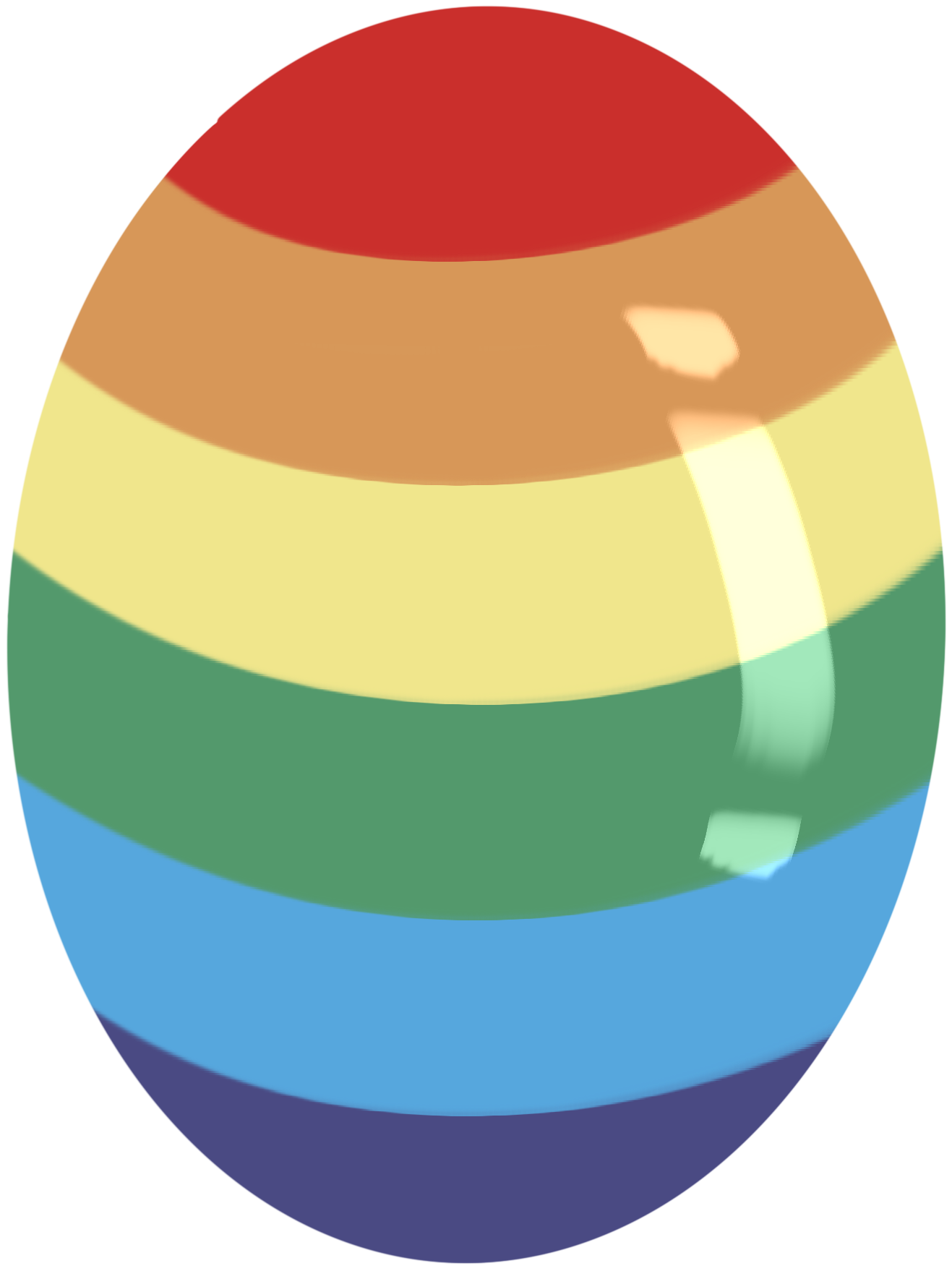} Post-credit Scene}

\newpage
\subsection{\includegraphics[scale=0.007]{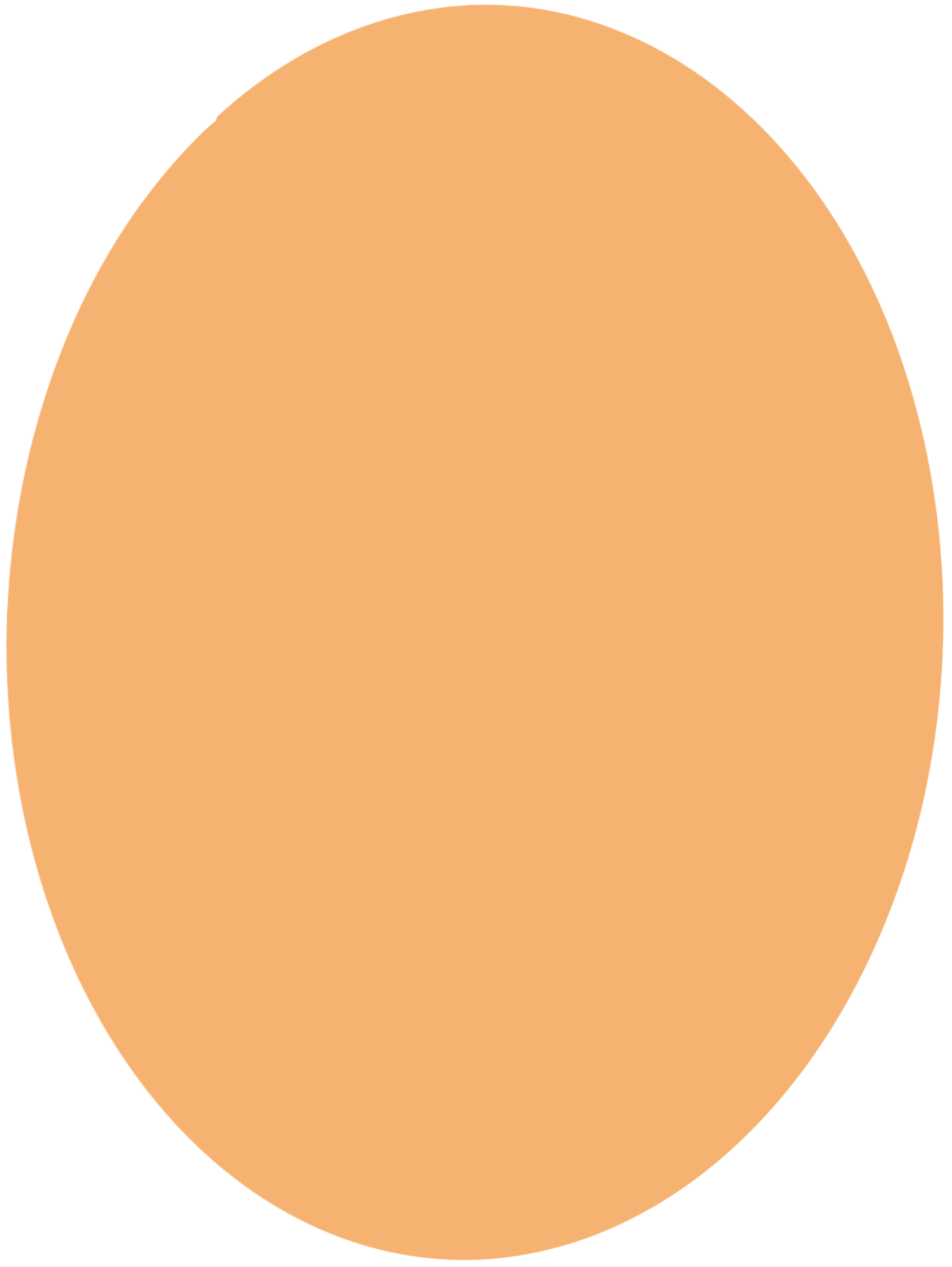} Easter Egg I}
\begin{figure*}[!th]
\centering
\includegraphics[width=15cm]{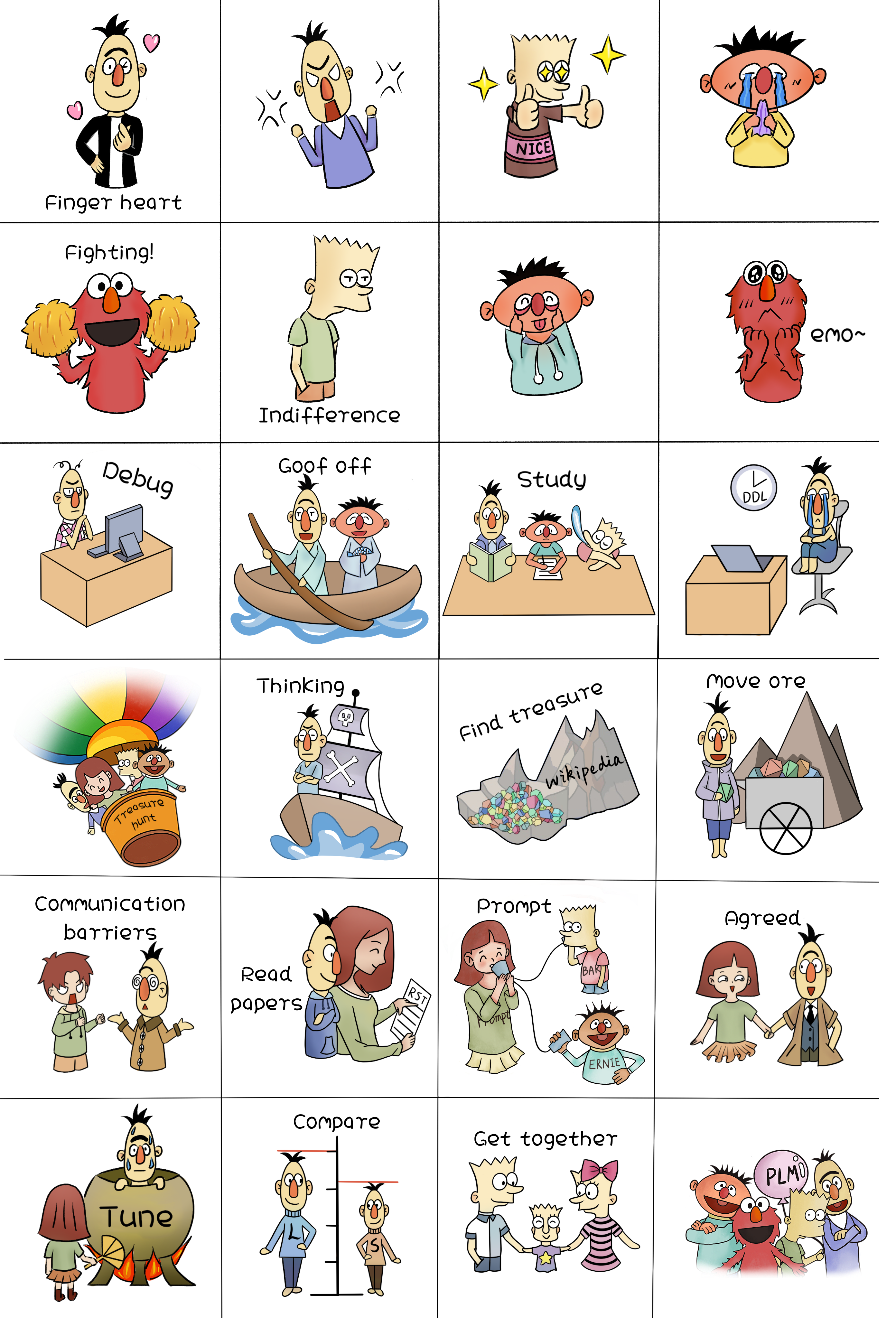}
\caption{Emojis about PLMs (the English version). More can be found at \url{http://expressai.co/peripherals/}}
\label{fig:easter_egg_1}
\end{figure*}
\clearpage

\newpage

\subsection{\includegraphics[scale=0.007]{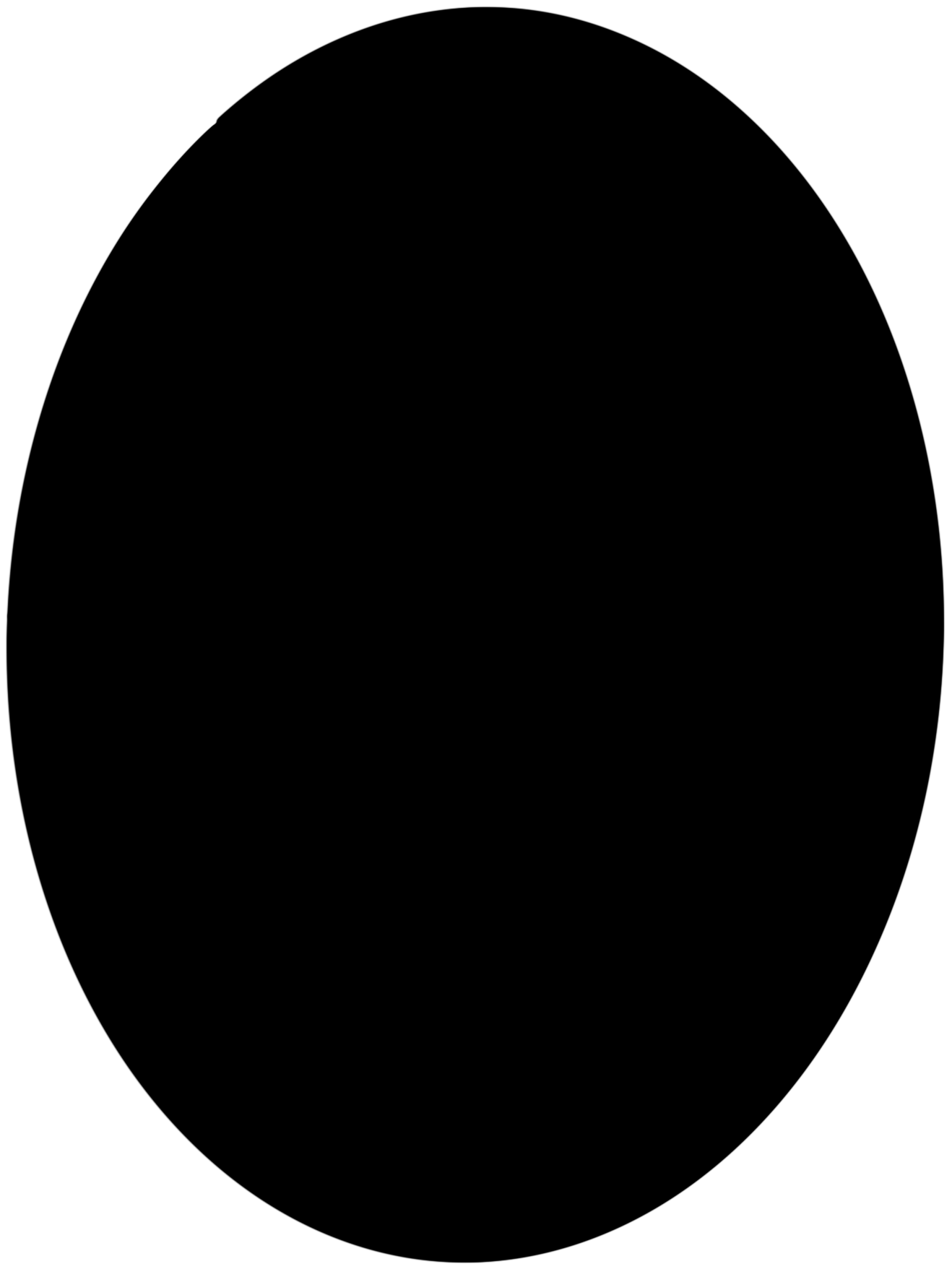} Easter Egg II}

\begin{figure*}[ht]
\centering
\includegraphics[width=16cm]{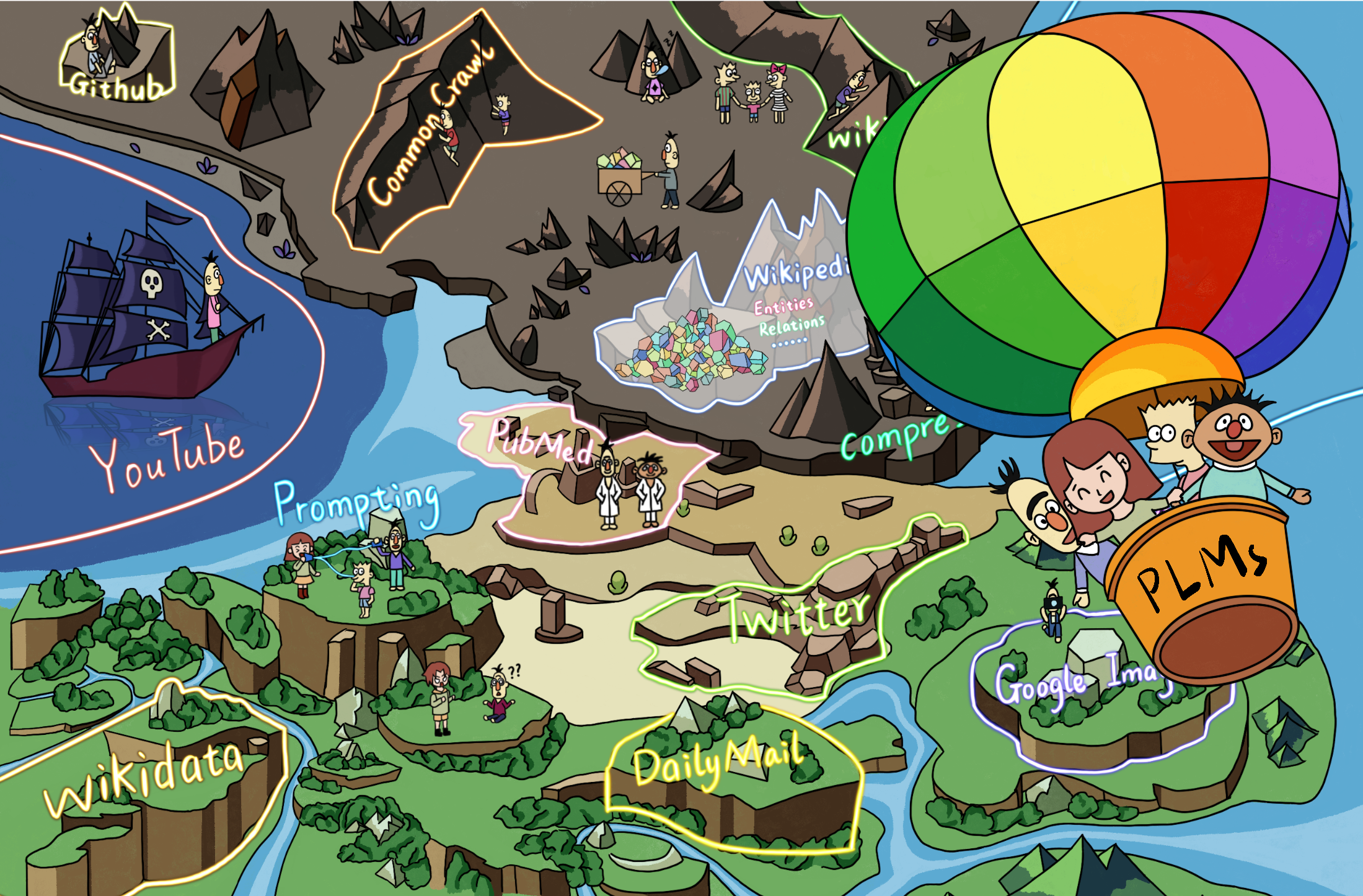}
\label{fig:easter_egg_2}
\end{figure*}
\pagecolor{black} \afterpage{\nopagecolor}
 
\textcolor{white}{\textbf{The future is all about the data; the future is all about you ...}}
 
\clearpage

\newpage

\subsection{\includegraphics[scale=0.007]{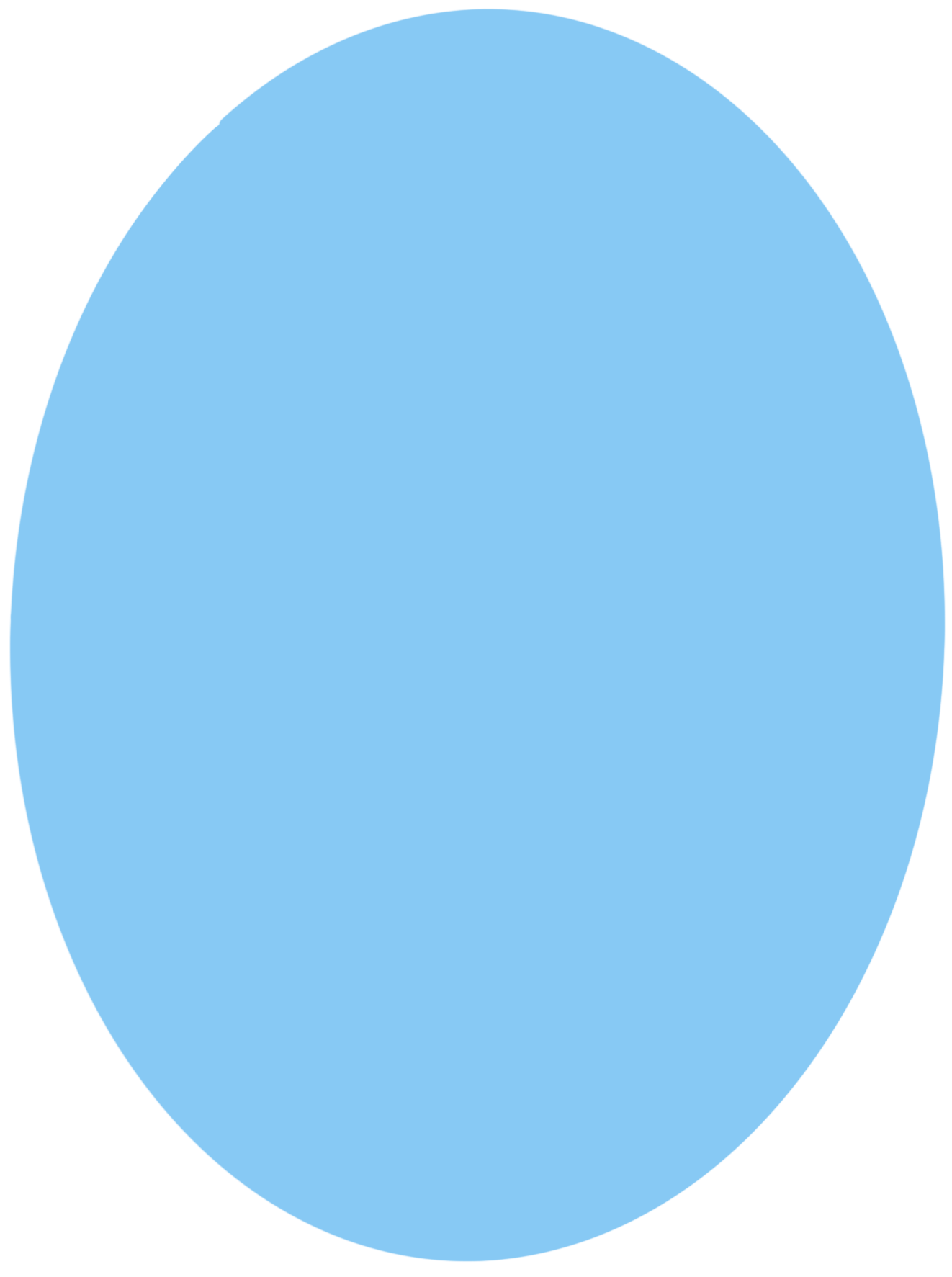} Easter Egg III} \label{easter:vi}

\paragraph{}

\paragraph{}

\paragraph{}

\paragraph{}

\paragraph{}

\paragraph{}

\begin{chapquote}[30pt]{Albert Einstein}
``Information is not knowledge''\footnote{Could pre-training be conducted over knowledge directly?}
\end{chapquote}

\end{document}